\documentclass[10pt,twocolumn,letterpaper]{article}

\usepackage[pagenumbers]{cvpr} 

\usepackage{graphicx}
\usepackage{amsmath}
\usepackage{multirow}
\usepackage{amssymb}
\usepackage{makecell}
\usepackage{tabularx}
\usepackage{booktabs}

\usepackage[accsupp]{axessibility}  

\usepackage{color}

\newcommand{\vect}[1]{\boldsymbol{\mathbf{#1}}}

\usepackage[pagebackref,breaklinks,colorlinks]{hyperref}

\usepackage[capitalize]{cleveref}
\crefname{section}{Sec.}{Secs.}
\Crefname{section}{Section}{Sections}
\Crefname{table}{Table}{Tables}
\crefname{table}{Tab.}{Tabs.}

\def\confName{CVPR}
\def\confYear{2023}

\newcommand{\papername}{Co-SLAM}
\newcommand{\methodname}{\papername\xspace}

\begin{document}

\title{\papername: Joint Coordinate and Sparse Parametric Encodings for \\Neural Real-Time SLAM
\vspace{-3mm}
}

\author{Hengyi Wang$^{\star}$ \quad Jingwen Wang$^{\star}$ \quad Lourdes Agapito\\
Department of Computer Science, University College London\\
{\tt\small \{hengyi.wang.21, jingwen.wang.17, l.agapito\}@ucl.ac.uk}
}

\twocolumn[{%
    \renewcommand\twocolumn[1][]{#1}%
    \maketitle
    \centering
    \vspace{-0.5cm}
    \includegraphics[width=0.96\linewidth]{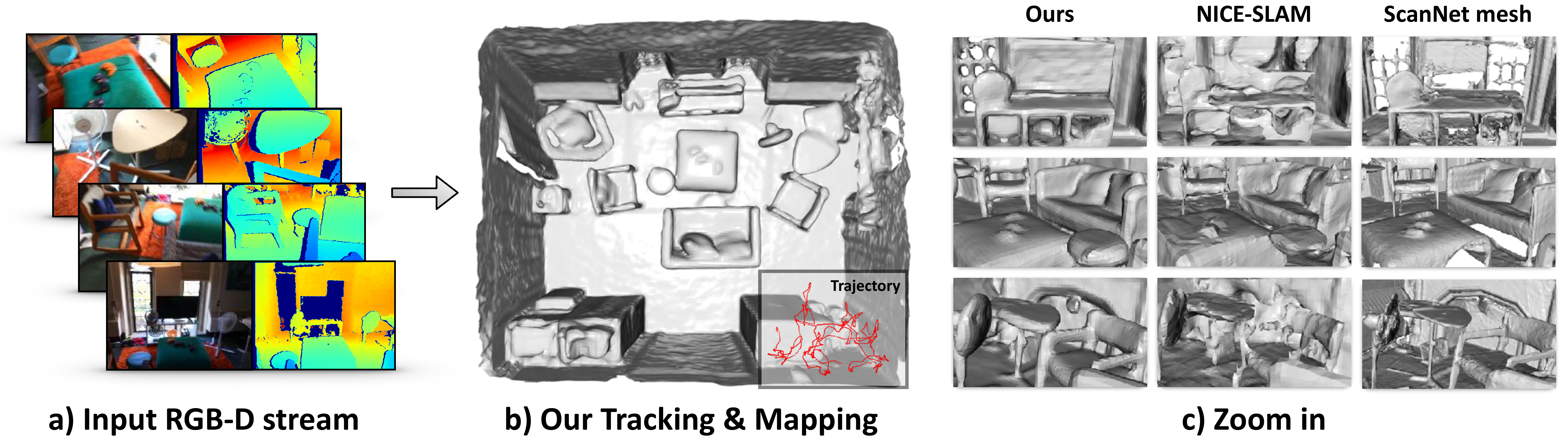}
\captionof{figure}{We present \methodname, a  neural RGB-D SLAM method that performs online tracking and mapping in real time. We propose a new hybrid representation based on a joint coordinate and sparse-parametric encoding with global bundle adjustment. Our method shows fast, high-fidelity scene reconstruction with efficient memory use and plausible hole-filling. \label{fig:teaser}}
    \vspace{4mm}
}]

\let\thefootnote\relax\footnotetext{$\star$ Indicates equal contribution.}

\begin{abstract}
We present \methodname{}, a neural RGB-D SLAM system 
based on a hybrid representation, 
that performs robust camera tracking and high-fidelity surface reconstruction in real time. \methodname{} represents the scene as a multi-resolution hash-grid 
to exploit its high convergence speed and ability to represent high-frequency local features. In addition, \methodname{} incorporates one-blob encoding, to encourage surface coherence and completion in unobserved areas. This joint parametric-coordinate encoding enables real-time and robust performance by bringing the best of both worlds: fast convergence and surface hole filling. Moreover, our ray sampling strategy allows \methodname{} to perform global bundle adjustment over all keyframes instead of requiring keyframe selection to maintain a small number of active keyframes as competing neural SLAM approaches do. Experimental results show that \methodname{} runs at $10-17$Hz and achieves state-of-the-art scene reconstruction results, and competitive tracking performance in various datasets and benchmarks (ScanNet, TUM, Replica, Synthetic RGBD). Project page: \url{https://hengyiwang.github.io/projects/CoSLAM}
\end{abstract}

\section{Introduction}

\begin{figure*}[htbp]
  \centering
  \footnotesize
  \setlength{\tabcolsep}{1.5pt}
  \newcommand{\sz}{0.21}
  \begin{tabular}{cccc}
  \footnotesize{COORDINATE}
    &\footnotesize{PARAMETRIC}
    &\footnotesize{JOINT}
    &\footnotesize{REFERENCE}\\
    \includegraphics[width=\sz\textwidth]{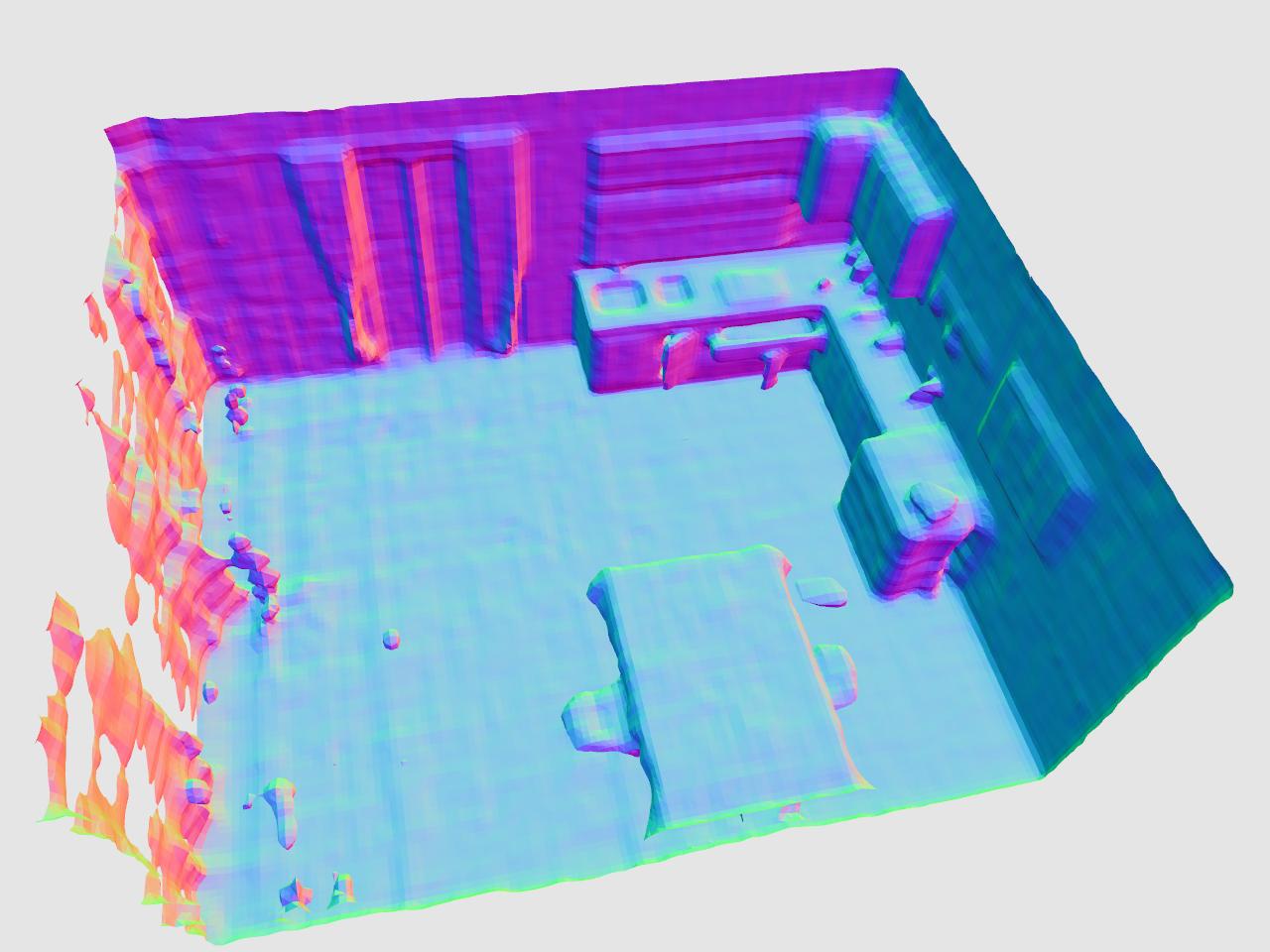}
    &\includegraphics[width=\sz\textwidth]{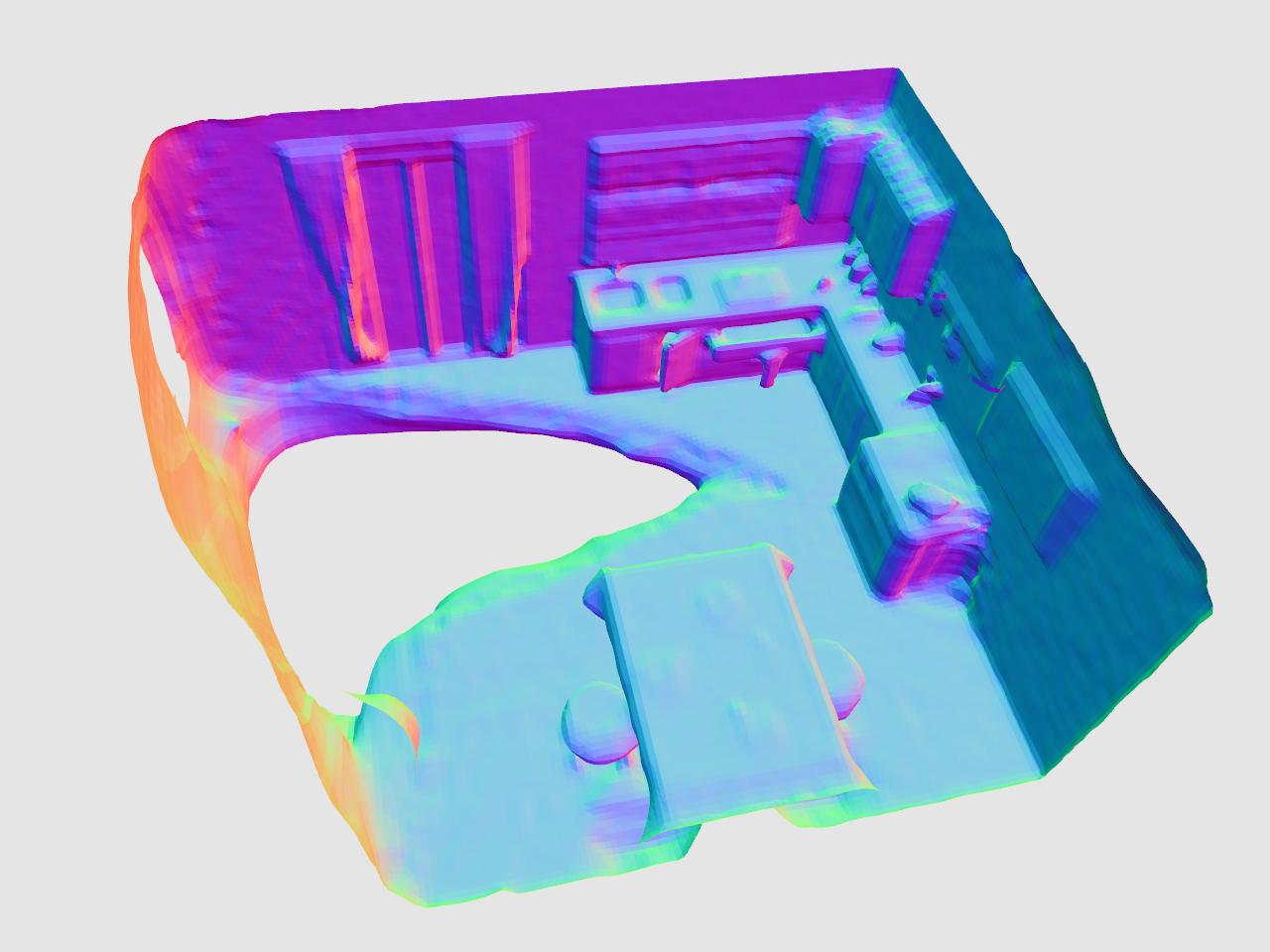}
    &\includegraphics[width=\sz\textwidth]{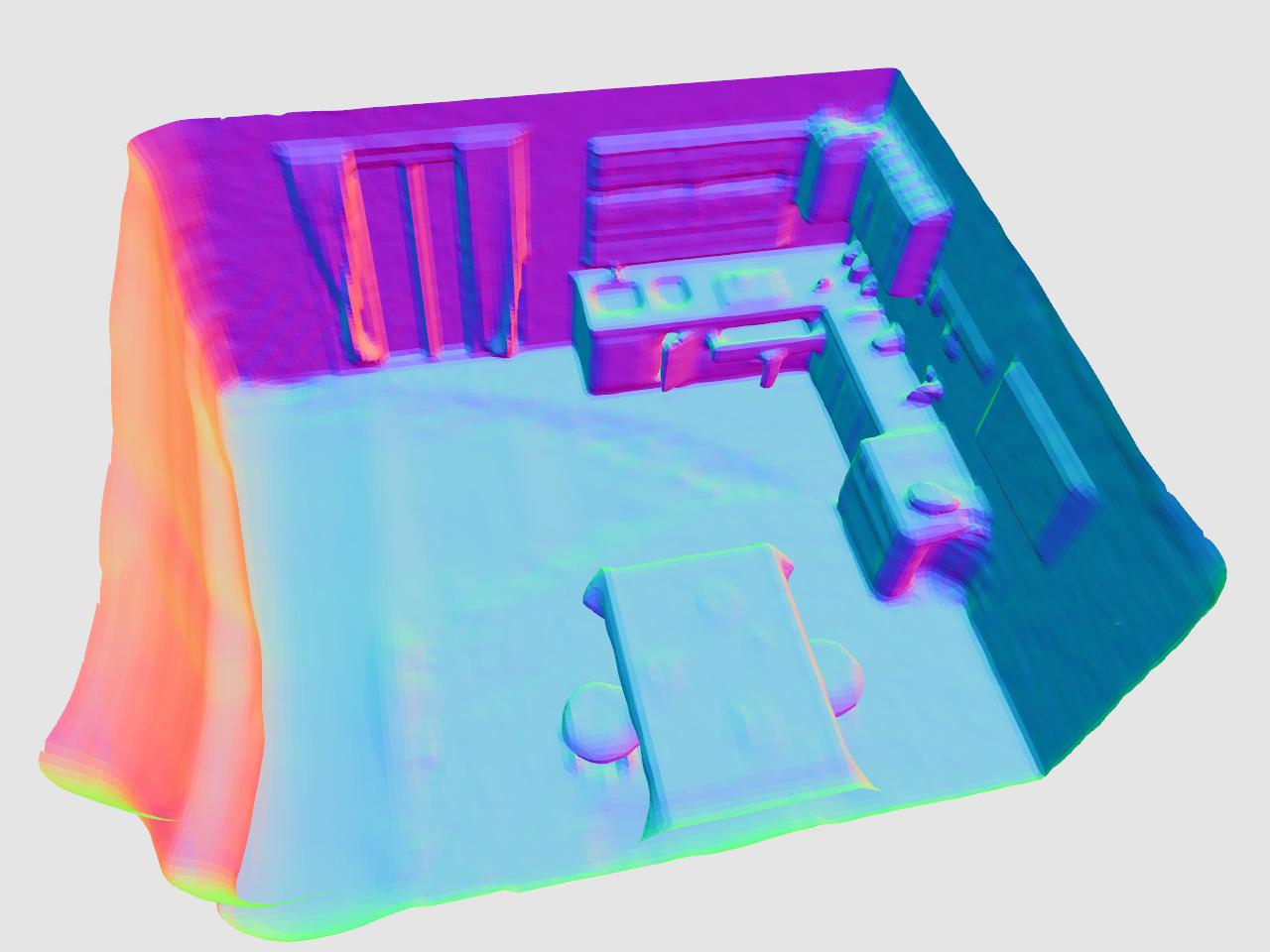}
    &\includegraphics[width=\sz\textwidth]{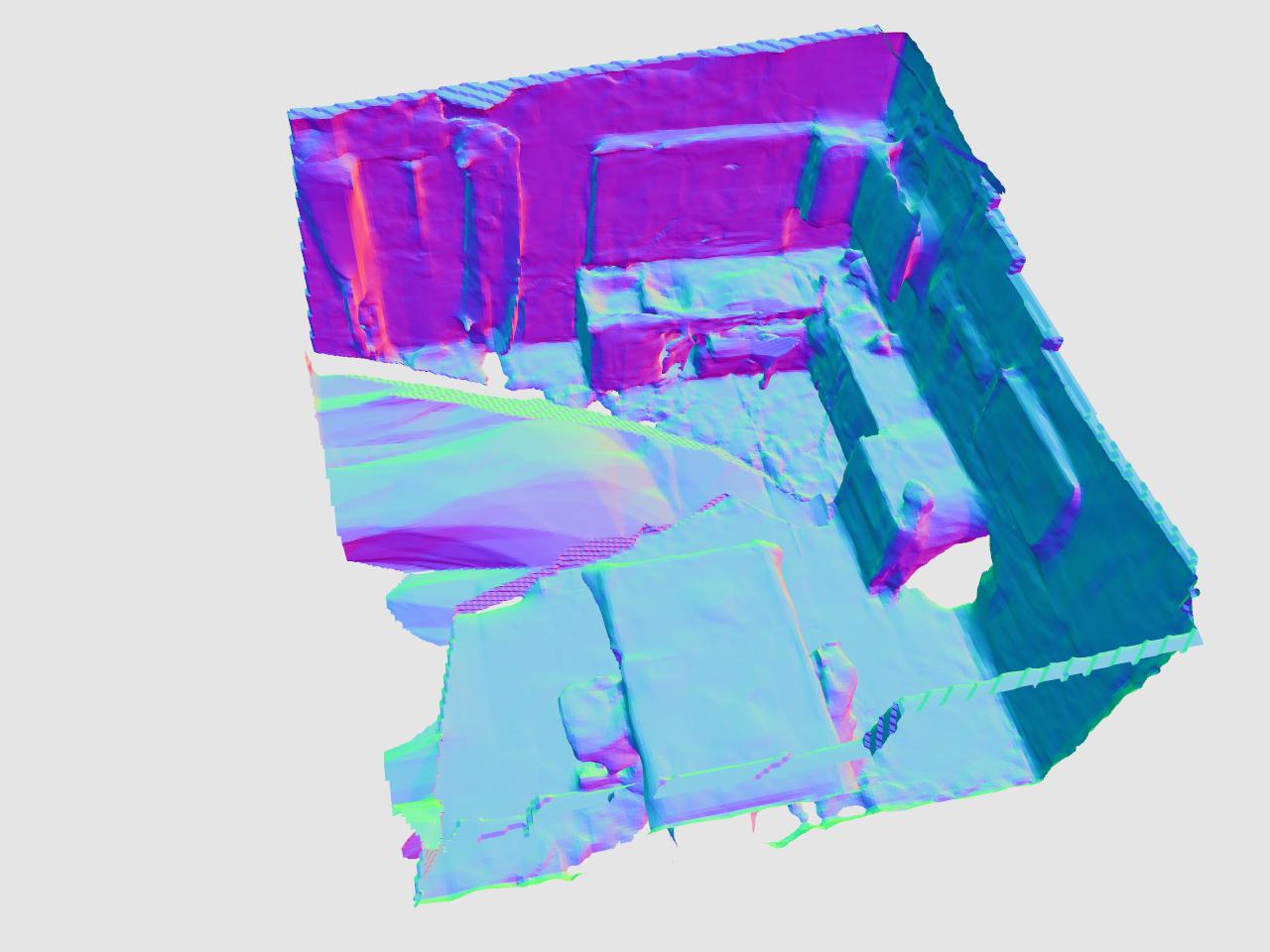}\\
    \scriptsize{Frequency}
    &\scriptsize{DenseGrid}
    &\scriptsize{DenseGrid+OneBlob}
    &\scriptsize{NICE-SLAM~\cite{zhuNiceslamNeuralImplicit2022}}\\
    \includegraphics[width=\sz\textwidth]{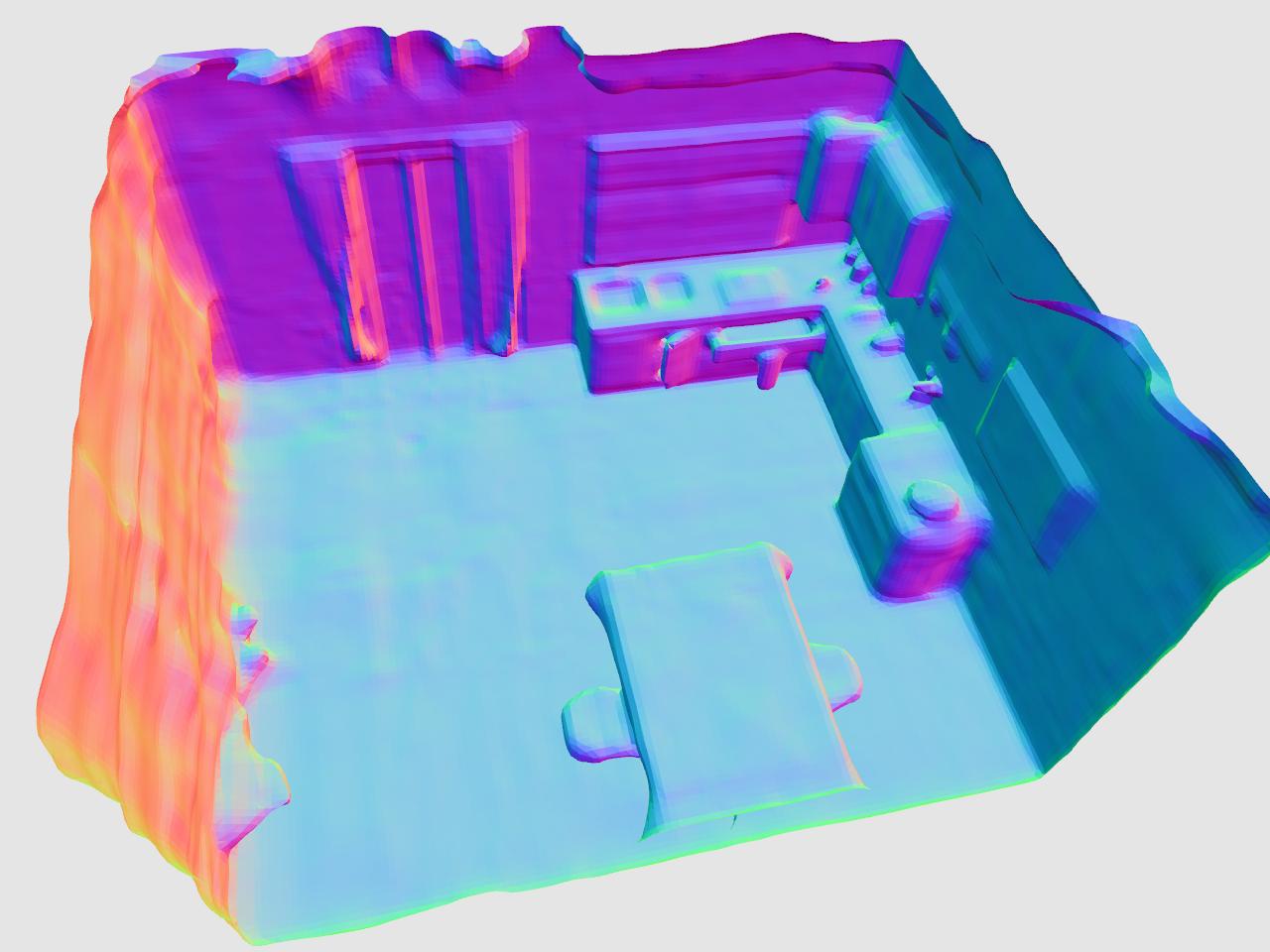}
    &\includegraphics[width=\sz\textwidth]{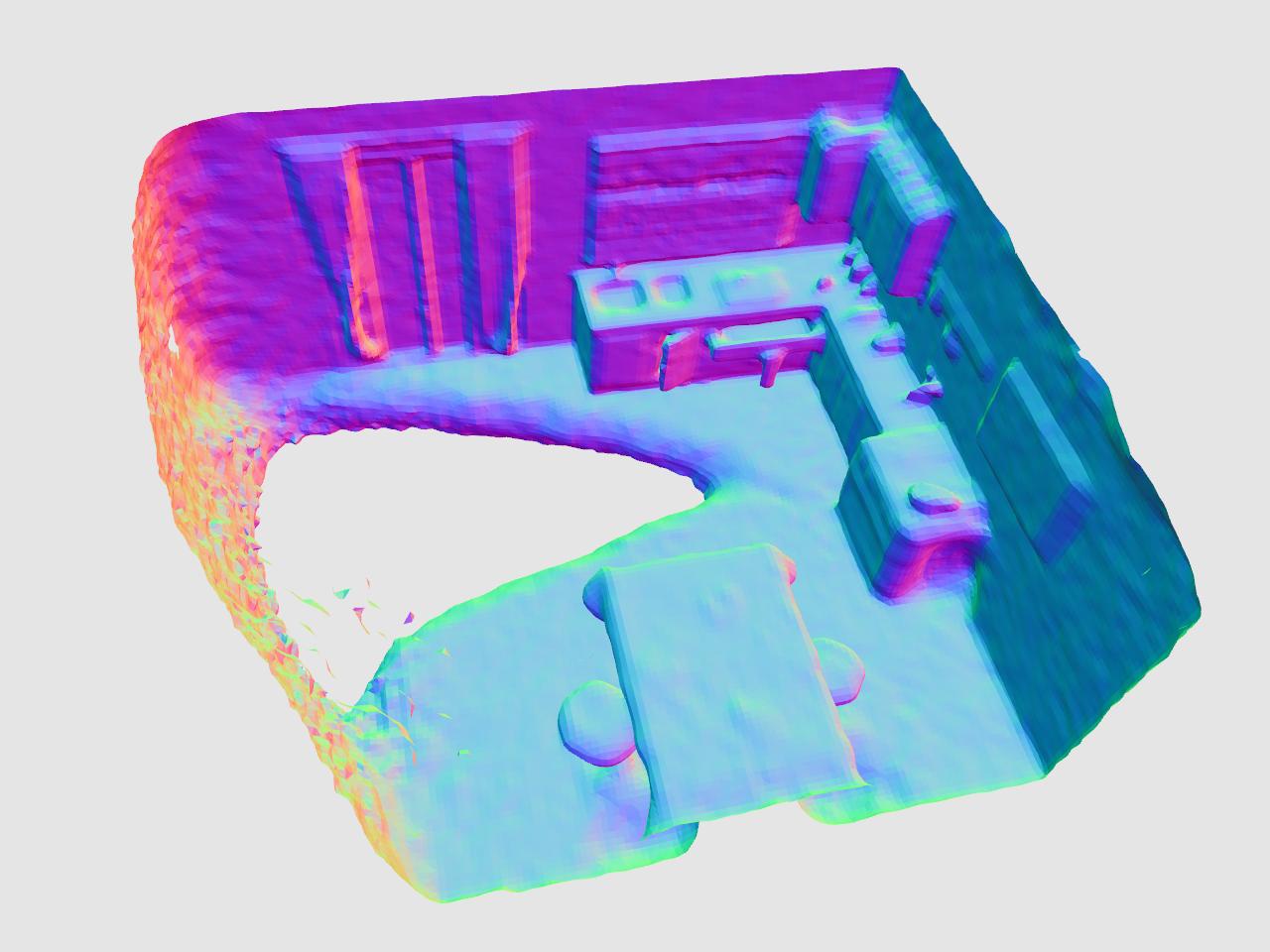}
    &\includegraphics[width=\sz\textwidth]{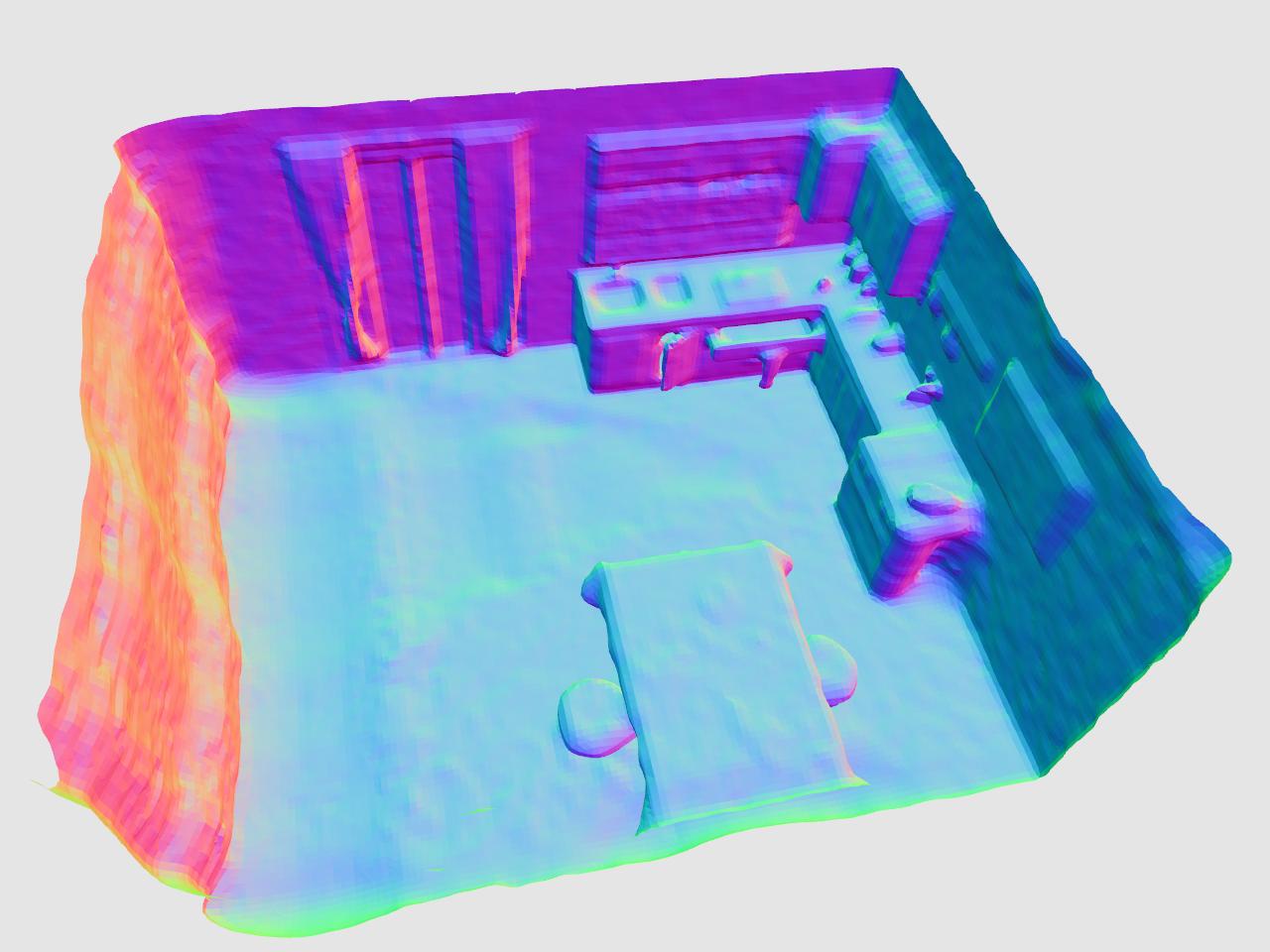}
    &\includegraphics[width=\sz\textwidth]{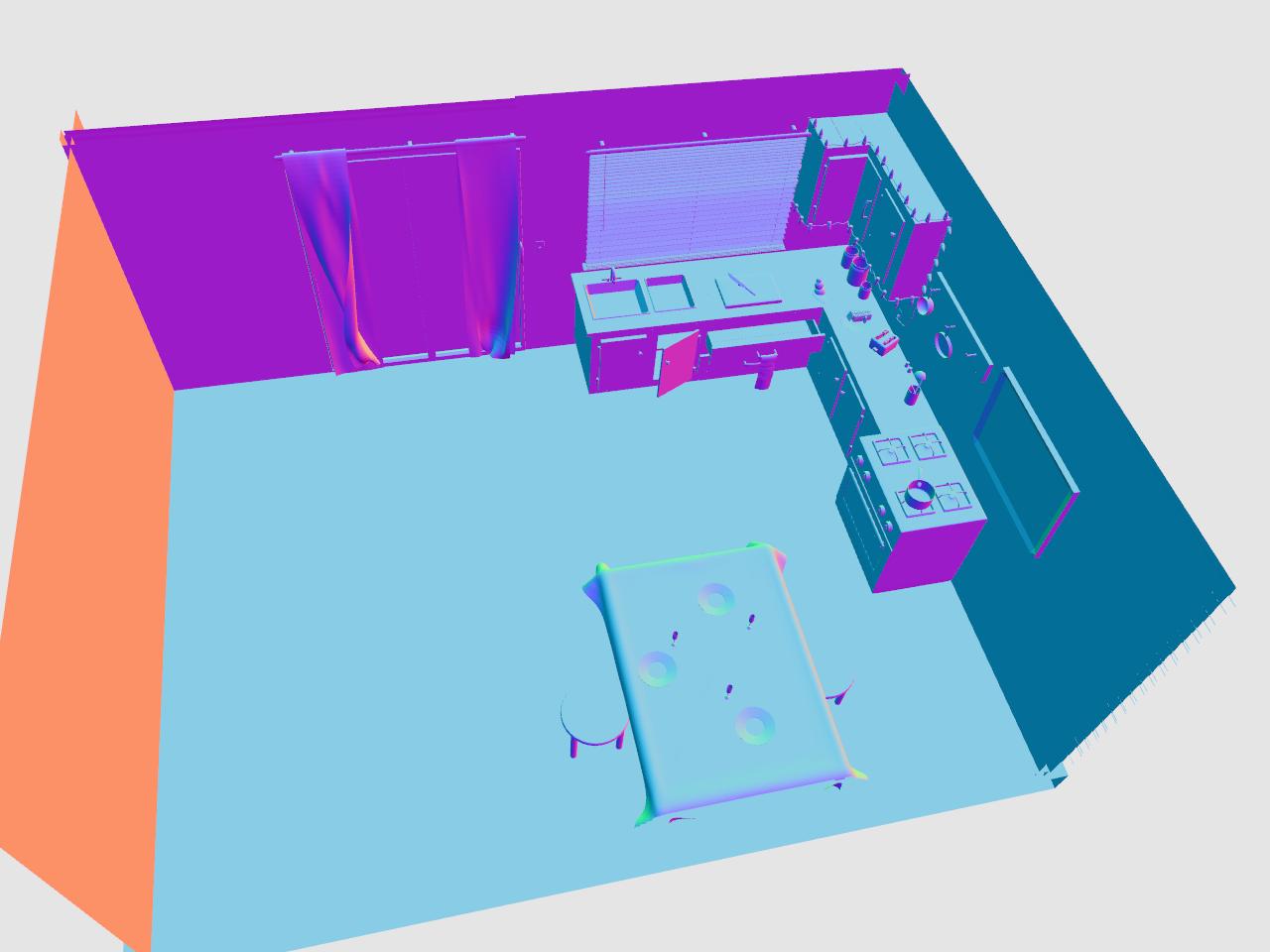}\\
    \scriptsize{OneBlob}
    &\scriptsize{HashGrid}
    &\scriptsize{HashGrid+OneBlob (Ours)}
    &\scriptsize{GT}\\
  \end{tabular} 
  \vspace{-5pt}
  \caption{Illustration of the effect of different encodings on completion. COORDINATE based encodings achieve hole filling but require long training times. PARAMETRIC encodings allow fast training, but fail to complete unobserved regions. JOINT coordinate and parametric encoding (Ours) allows smooth scene completion and fast training. NICE-SLAM~\cite{zhuNiceslamNeuralImplicit2022} uses a dense parametric encoding. }
  \label{fig:encoding_showcase}
  \vspace{-8pt}
\end{figure*}

Real-time joint camera tracking and dense surface reconstruction from RGB-D sensors has been a core  problem in computer vision and robotics for decades. Traditional SLAM solutions exist that can robustly track the position of the camera while fusing depth and/or color measurements into a single high-fidelity map. However, they rely on hand-crafted loss terms and do not exploit data-driven priors. 

Recent attention has turned to learning-based models that can exploit the ability of neural network architectures to learn smoothness and coherence priors directly from data. Coordinate-based networks have probably become the most popular representation, since they can be trained to predict the geometric and appearance properties of any point in the scene in a self-supervised way, directly from images. The most notable example, Neural Radiance Fields (NeRF)~\cite{mildenhallNerfRepresentingScenes2020a},  encodes scene density and color in the weights of a neural network. In combination with volume rendering, NeRF is trained to re-synthesize the input images and has a remarkable ability to generalize to nearby unseen views. 

Coordinate-based networks embed input point coordinates into a high dimensional space, using sinusoidal or other frequency embeddings, allowing them to capture high-frequency details that are essential for high-fidelity geometry reconstruction~\cite{azinovicNeuralRGBDSurface2022c}. Combined with the smoothness and coherence priors inherently encoded in the MLP weights, they constitute a good choice for sequential tracking and mapping~\cite{sucarIMAPImplicitMapping2021a}.
However, the weakness of MLP-based approaches is the long training times required (sometimes hours) to learn a single scene. For that reason, recent real-time capable SLAM systems built on coordinate networks with frequency embeddings such as iMAP~\cite{sucarIMAPImplicitMapping2021a} need to resort to strategies to sparsify ray sampling and reduce tracking iterations to maintain interactive operation. This comes at the cost of loss of detail in the reconstructions which are oversmoothed (Fig.~\ref{fig:replica_topdown_zoom_in}) and potential errors in camera tracking. 

Optimizable feature grids, also known as parametric embeddings, have recently become a powerful alternative scene representation to monolithic MLPs, given their ability to represent high-fidelity local features and their extremely fast convergence (orders of magnitude faster)~\cite{mueller2022instant,Yu2022MonoSDF,fridovichkeilPlenoxelsRadianceFields2022,karnewarReluFieldsLittle2022,wangGosurfNeuralFeature2022}. Recent efforts focus on sparse alternatives to these parametric embeddings such as octrees~\cite{takikawaNeuralGeometricLevel2021}, tri-plane~\cite{chanEfficientGeometryaware3D2022}, hash-grid~\cite{mueller2022instant} or sparse voxel grid~\cite{liuNeuralSparseVoxel2020,liVoxSurfVoxelbasedImplicit2022} to improve the memory efficiency of dense grids.
While these representations can be fast to train and are therefore well suited to real-time operation, they fundamentally lack the smoothness, and coherence priors inherent to MLPs and struggle with hole-filling in areas without observation. NICE-SLAM~\cite{zhuNiceslamNeuralImplicit2022} is a recent example of a multi-resolution feature grid-based SLAM method. Although it does not suffer from over-smoothness and captures local detail (as shown in Fig.~\ref{fig:encoding_showcase}) it cannot perform hole-filling which might in turn lead to drift in camera pose estimation.

Our first contribution is to design a joint coordinate and sparse grid encoding for input points that brings together the benefits of both worlds to the real-time SLAM framework. On the one hand, the smoothness and coherence priors provided by coordinate encodings (we use one-blob~\cite{mullerNeuralImportanceSampling2019} encoding), and on the other hand the optimization speed and local details of sparse feature encodings (we use hash grid~\cite{mueller2022instant}), resulting in more robust camera tracking and high-fidelity maps with better completion and hole filling.

Our second contribution relates to the bundle adjustment (BA) step in the joint optimization of the map and camera poses. So far, all neural SLAM systems~\cite{sucarIMAPImplicitMapping2021a,zhuNiceslamNeuralImplicit2022} perform BA using rays sampled from a very small subset of selected keyframes. Restricting the optimization to a very small number of viewpoints results in decreased robustness in camera tracking and increased computation due to the need for a keyframe-selection strategy. 
Instead, Co-SLAM performs global BA, sampling rays from all past keyframes, which results in an important boost in robustness and performance in pose estimation. 
In addition, we show that our BA optimization requires a fraction of the iterations of NICE-SLAM~\cite{zhuNiceslamNeuralImplicit2022} to achieve similar errors.
In practice, Co-SLAM achieves SOTA performance in camera tracking and 3D reconstruction while maintaining real time performance.  

Co-SLAM runs at 15-17Hz on Replica and Synthetic RGB-D datasets~\cite{azinovicNeuralRGBDSurface2022c}, and 12-13Hz on ScanNet~\cite{daiScannetRichlyannotated3d2017} and TUM~\cite{sturmBenchmarkEvaluationRGBD2012} scenes --- faster than NICE-SLAM (0.1-1Hz)~\cite{zhuNiceslamNeuralImplicit2022} and iMAP~\cite{sucarIMAPImplicitMapping2021a}. We perform extensive evaluations on various datasets (Replica~\cite{straubReplicaDatasetDigital2019}, Synthetic RGBD~\cite{azinovicNeuralRGBDSurface2022c}, ScanNet~\cite{daiBundlefusionRealtimeGlobally2017a}, TUM~\cite{sturmBenchmarkEvaluationRGBD2012}) where we outperform  NICE-SLAM~\cite{zhuNiceslamNeuralImplicit2022} and iMAP~\cite{sucarIMAPImplicitMapping2021a} in reconstruction and achieve better or at least on-par tracking accuracy.
\section{Related Work}

\noindent{\bf Dense Visual SLAM.} 
Taking advantage of commodity depth sensors, KinectFusion~\cite{newcombe:2011:kinfu} performs frame-to-model camera tracking via projective iterative-closest-point (ICP), and incrementally updates the scene geometry via TSDF-Fusion. 
Subsequent works focused on addressing the scalability issue by adopting more efficient data structures, such as surface elements (surfels)~\cite{whelan:2015:efusion, whelan:2016:efusion2}, VoxelHashing~\cite{niebner:2013:hashing, chen2013scalable, kahler:2015:inftam} or Octrees~\cite{zeng2013octree, Vespa:etal:RAL2018}.
While most works focus more heavily on scene reconstruction and only track per-frame poses,  BAD-SLAM~\cite{schops2019bad} proposes full direct bundle adjustment (BA) to jointly optimize keyframe (KF) poses and the dense 3D structure. 
Several recent works~\cite{yang2020d3vo, Tateno:etal:CVPR2017, koestler2021tandem, teed2021droid} leverage deep learning to improve the accuracy and robustness of traditional SLAM, and even achieve dense reconstruction with monocular SLAM. 
While these approaches introduced some learned components, the scene representation and overall pipeline still follow traditional SLAM methods.

\noindent{\bf Neural Implicit Representations.} Recently neural implicit representations~\cite{mildenhallNerfRepresentingScenes2020a} that encode 3D geometry and the appearance of a scene within the weights of a neural network have gained popularity due to their expressiveness and compactness.
Among these works, NeRF~\cite{mildenhallNerfRepresentingScenes2020a} and its variants adopt coordinate encoding~\cite{rahamanSpectralBiasNeural2019,mullerNeuralImportanceSampling2019} with MLPs and show impressive scene reconstruction using differentiable rendering. Given that coordinate encoding-based methods require lengthy training, many follow-up works~\cite{fridovichkeilPlenoxelsRadianceFields2022,karnewarReluFieldsLittle2022,sunDirectVoxelGrid2022} propose parametric encodings that increase parameter size but speed up the training. To improve the memory efficiency of parametric encoding-based methods, sparse parametric encodings, such as Octree~\cite{takikawaNeuralGeometricLevel2021}, Tri-plane~\cite{chanEfficientGeometryaware3D2022}, or sparse voxel grid~\cite{liuNeuralSparseVoxel2020,mueller2022instant,liVoxSurfVoxelbasedImplicit2022}, have been proposed. While these methods focus on novel view synthesis, others instead focus on surface reconstruction from RGB images, with implicit surface representations and differentiable renderers~\cite{wangNeusLearningNeural2021,yarivVolumeRenderingNeural2021,guoNeural3DScene2022,Yu2022MonoSDF}. Other methods~\cite{azinovicNeuralRGBDSurface2022c,sucarIMAPImplicitMapping2021a,wuVoxurfVoxelbasedEfficient2022,wangGosurfNeuralFeature2022,zhuNiceslamNeuralImplicit2022} use depth measurements as additional supervision for surface reconstruction. 

\noindent{\bf Neural Implicit SLAM.}  iMAP~\cite{sucarIMAPImplicitMapping2021a} adopts an MLP representation to perform joint tracking and mapping in quasi-real time. Smooth, plausible filling of unobserved regions is achieved thanks to the coherence priors inherent to the MLP. iMAP introduces elaborate keyframe selection and information-guided pixel sampling for speed, resulting in 10 Hz tracking and 2 Hz mapping. To reduce the computational overhead and improve scalability, NICE-SLAM~\cite{zhuNiceslamNeuralImplicit2022} adopts a multi-level feature grid for scene representation. However, as feature grids only perform local updates, they fail to achieve plausible hole-filling. With Co-SLAM we aim to address both issues. To achieve real-time performance and memory-efficiency, while maintaining high-fidelity surface reconstruction and plausible hole filling, we propose to combine the use of coordinate and sparse parametric encodings for scene representation, and perform dense global bundle adjustment using rays sampled from all keyframes.

\section{Method}

\begin{figure*}[t]
    \centering
    \includegraphics[width=0.95\textwidth]{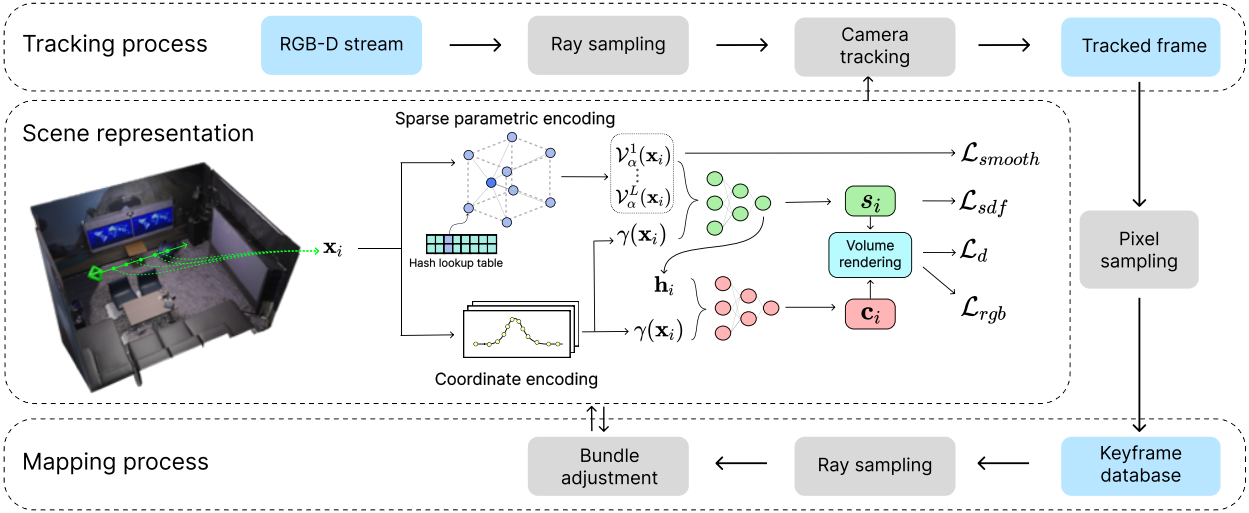}
    \vspace{-2mm}
    \caption[Overview of Co-SLAM]{\textbf{Overview of \papername.} 1) Scene representation: using our new joint coordinate+parametric encoding, input coordinates are mapped to RGB and SDF values via two shallow MLPs. 2) Tracking: optimize per-frame camera poses $\xi_t$ by minimizing losses. 3) Mapping: global bundle adjustment to jointly optimize the scene representation and camera poses taking rays sampled from all keyframes.} 
    \label{fig:overview}
    \vspace{-10pt}
\end{figure*}

Fig.~\ref{fig:overview} shows an overview of \papername. Given an input RGB-D stream $\{I_t\}_{t=1}^N$ $\{D_t\}_{t=1}^N$ with known camera intrinsics $K\in \mathbb{R}^{3 \times 3}$, we perform dense mapping and tracking by jointly optimizing camera poses $\{\xi_t\}_{t=1}^{N}$ and a neural scene representation $f_{\theta}$. Specifically, our implicit representation maps world coordinates $\vect{x}$ into color $\vect{c}$ and truncated signed distance (TSDF) $s$ values:
\begin{equation}
     f_{\theta}(\mathbf{x}) \mapsto (\vect{c}, s).
\end{equation}
Similar to most SLAM systems, the process is split into tracking and mapping. Initialization is performed by running a few training iterations on the first frame. For each subsequent frame, the camera pose is optimized first, initialized with a simple constant-speed motion model. A small fraction of pixels/rays are then sampled and copied to the global pixel-set. At each mapping iteration, global bundle adjustment is performed over a set of pixels randomly sampled from the global pixel-set, to jointly optimize the scene representation $\theta$ and all camera poses $\{\xi_t\}$.

\subsection{Joint Coordinate and Parametric Encoding}

Thanks to the coherence and smoothness priors inherent to MLPs, coordinate-based representations achieve high-fidelity scene reconstruction. However, these methods often suffer from slow convergence and catastrophic forgetting, when optimized in a sequential setting. Instead, parametric encoding based methods improve the computational efficiency, but they fall short of hole filling and smoothness.
Since both properties of speed and coherence are crucial for a real-world SLAM system, we propose a joint coordinate and parametric encoding that combines the best of both worlds: we adopt coordinate encoding for scene representation while using sparse parametric encoding to speed up training.
Specifically, we use One-blob~\cite{mullerNeuralImportanceSampling2019} encoding $\gamma(\vect{x})$ instead of embedding spatial coordinates into multiple frequency bands. As scene representation we adopt a multi-resolution hash-based feature grid~\cite{mueller2022instant} $\mathcal{V}_{\alpha}=\{\mathcal{V}_{\alpha}^{l}\}_{l=1}^{L}$. The spatial resolution of each level is set between the coarsest $R_{min}$ and the finest resolution $R_{max}$ in a progressive manner. Feature vectors $\mathcal{V}_{\alpha}(\vect{x})$ at each sampled point $\vect{x}$ are queried via trilinear interpolation. The geometry decoder outputs the predicted SDF value $s$ and a feature vector $\vect{h}$:
\begin{equation}
    f_{\tau}(\gamma(\vect{x}), \mathcal{V}_{\alpha}(\vect{x})) \mapsto (\vect{h}, s).
\end{equation}
Finally, the color MLP predicts the RGB value:
\begin{equation}
    f_{\phi}(\gamma(\vect{x}), \vect{h}) \mapsto \vect{c}.
\end{equation}
Here $\theta=\{\alpha, \phi, \tau\}$ are  learnable parameters. Injecting the One-blob encoding in the hash-based multi-resolution feature grid representation, results in fast convergence, efficient memory use, and hole filling needed for online SLAM.

\subsection{Depth and Color Rendering}

Following~\cite{sucarIMAPImplicitMapping2021a, zhuNiceslamNeuralImplicit2022}, we render depth and color by integrating the predicted values along the sampled rays. Specifically, given the camera origin $\vect{o}$ and ray direction $\vect{r}$, we uniformly sample $M$ points $\vect{x}_i = \vect{o} + d_i \vect{r}$, $i \in \{1, \dots, M\}$ with depth values $\{t_1, \dots, t_M\}$ and predicted colors $\{\vect{c}_1, \dots, \vect{c}_M\}$ and render color and depth as
\begin{equation}
\hat{\mathbf{c}}=\frac{1}{\sum_{i=1}^{M} w_{i}}\sum_{i=1}^{M} w_{i} \mathbf{c}_{i}, \quad \hat{d}=\frac{1}{\sum_{i=1}^{M} w_{i}}\sum_{i=1}^{M} w_{i} d_{i},
\end{equation}
\noindent 
where $\{w_i\}$ are the computed weights along the ray. A conversion function is needed to convert predicted SDF $s_i$ to weight $w_i$. Instead of adopting the rendering equations proposed in Neus~\cite{wangNeusLearningNeural2021, yarivVolumeRenderingNeural2021}, we follow the simple bell-shaped model  of~\cite{azinovicNeuralRGBDSurface2022c} and compute weights $w_i$ directly by multiplying the two Sigmoid functions $\sigma(\cdot)$
\begin{equation}
w_{i}=\sigma\left(\frac{s_{i}}{t r}\right) \sigma\left(-\frac{s_{i}}{t r}\right),
\end{equation}
where $tr$ is the truncation distance. 

\noindent{\bf{Depth-guided Sampling.}} 
For sampling along each ray, we observe that importance sampling does not show significant improvement while slowing down our tracking and mapping. Instead, we use depth-guided sampling: In addition to $M_c$ points uniformly sampled between $near$ and $far$ bound, for rays with a valid depth measurement, we further uniformly sample $M_f$ near-surface points within the range $[d-d_s, d+d_s]$, where $d_s$ is a small offset.

\subsection{Tracking and Bundle Adjustment}

\noindent{\textbf{Objective Functions.}}
Our tracking and bundle adjustment are performed via minimizing our objective functions with respect to learnable parameters $\theta$ and camera parameters $\xi_t$. The color and depth rendering losses are $\ell_2$ errors between the rendered results and observations:
\begin{equation}
    \mathcal{L}_{rgb} = \frac{1}{N} \sum_{n=1}^{N} (\hat{\vect{c}}_n - \vect{c}_n)^2, \mathcal{L}_d = \frac{1}{|R_d|} \sum_{r\in R_d} (\hat{d}_r - D[u, v])^2.
\end{equation}
where $R_d$ is the set of rays that have a valid depth measurement, and $u, v$ is the corresponding pixel on the image plane. 
To achieve accurate, smooth reconstructions with detailed geometry, we also apply approximate SDF and feature smoothness losses. Specifically, for samples within the truncation region, i.e. points where $|D[u, v] - d| \le tr$, we use the distance between the sampled point and its observed depth value as an approximation of the ground-truth SDF value for supervision:
\begin{equation}
    \mathcal{L}_{sdf} = \frac{1}{|R_d|} \sum_{r\in R_d} \frac{1}{|S_r^{tr}|}\sum_{p \in S_r^{tr}} \big(s_p - (D[u, v] - d)\big)^2.
\end{equation}
For points that are far from the surface ($(D[u, v] - d|) > tr$), we apply a free-space loss which forces the SDF prediction to be the truncated distance $tr$:
\begin{equation}
    \mathcal{L}_{fs} = \frac{1}{|R_d|} \sum_{r\in R_d} \frac{1}{|S_r^{fs}|}\sum_{p \in S_r^{fs}} (s_p - tr)^2.
\end{equation}
To prevent the noisy reconstructions caused by hash collisions in unobserved free-space regions we perform  additional regularization on the interpolated features $\mathcal{V}_{\alpha}(\vect{x})$:
\begin{equation}
\mathcal{L}_{smooth} = \frac{1}{|\mathcal{G}|}\sum_{\vect{x} \in \mathcal{G}} \Delta_x^2 + \Delta_y^2 + \Delta_z^2,
\end{equation}
\noindent
where $\Delta_{x, y, z} = \mathcal{V}_{\alpha}(\vect{x} + \epsilon_{x, y, z}) - \mathcal{V}_{\alpha}(\vect{x})$ denotes the feature-metric difference between adjacent sampled vertices on the hash-grid along the three dimensions. Since performing regularization on the entire feature grid is computationally infeasible for real-time mapping, we only perform it in a small random region in each iteration.

\noindent{\textbf{Camera Tracking.}} 
We track the camera-to-world transformation matrix $\vect{T}_{wc} = \exp{(\xi^{\wedge}_{t})} \in \mathbb{SE}(3)$ at each frame $t$. When a new frame comes in, we first initialize the pose of the current frame $i$ using constant speed assumption:
\begin{equation}
    \vect{T}_t = \vect{T}_{t-1} \vect{T}_{t-2}^{-1} \vect{T}_{t-1}
\end{equation}
\noindent
Then, we select $N_t$ pixels within the current frame and iteratively optimize the pose by minimizing our objective function with respect to the camera parameters $\xi_t$.

\noindent{\textbf{Bundle Adjustment.}} 
In neural SLAM, bundle adjustment usually consists of keyframe selection and joint optimization of camera poses and scene representation. Classic dense visual SLAM methods require saving keyframe (KF) images as the loss is formulated densely over all pixels. In contrast, the
advantage of neural SLAM, as first shown by iMAP~\cite{sucarIMAPImplicitMapping2021a}, is that BA can work with a sparse set of sampled rays. This is because the scene is represented as an implicit field using a neural network. However, iMAP and NICE-SLAM do not take full advantage of this - they still store full keyframe images following the classic SLAM paradigm and rely on keyframe selection (e.g. information gain, visual overlapping) to perform joint optimization on a small fraction of keyframes (usually less than 10). 

In Co-SLAM, we go further and drop the need for storing full keyframe images or keyframe selection. Instead, we only store a subset of pixels (around 5$\%$) to represent each keyframe. This allows us to insert new keyframes more frequently and maintain a much larger keyframe database. For joint optimization, we randomly sample a total number of $N_g$ rays from our global keyframe list to optimize our scene representation as well as camera poses. The joint optimization is performed in an alternating fashion. Specifically, we firstly optimize the scene representation $\theta$ for $k_m$ steps and update camera poses using the accumulated gradient on camera parameters $\{\xi_t\}$. Since each camera pose uses only $6$ parameters, this approach can improve the robustness of camera pose optimization with negligible extra computational cost on gradient accumulation.
\section{Experiments}

\subsection{Experimental Setup}

\noindent{\textbf{Datasets.}} 
We evaluate Co-SLAM on a variety of scenes from four different datasets. Following iMAP and NICE-SLAM, we quantitatively evaluate the reconstruction quality on 8 synthetic scenes from Replica~\cite{straubReplicaDatasetDigital2019}. We also evaluate on 7 synthetic scenes from NeuralRGBD~\cite{azinovicNeuralRGBDSurface2022c}, which simulates noisy depth maps.  For pose estimation, we evaluate the results on 6 scenes from ScanNet~\cite{daiScannetRichlyannotated3d2017} with their ground truth pose obtained with BundleFusion~\cite{daiBundlefusionRealtimeGlobally2017a}, and 3 scenes from TUM RGB-D dataset~\cite{sturmBenchmarkEvaluationRGBD2012} with their ground truth pose provided by a motion capture system. 

\noindent{\textbf{Metrics.}} 
We evaluate the reconstruction quality using \textit{Depth L1} (cm), \textit{Accuracy} (cm), \textit{Completion}  (cm), and \textit{Completion ratio} (\%) with a threshold of 5cm. Following NICE-SLAM~\cite{zhuNiceslamNeuralImplicit2022}, we remove the unobserved regions that are outside of any camera frustum.  In addition, we also perform an extra mesh culling that removes the noisy points within the camera frustum but outside the target scene. We observe that all methods experience a performance gain with our mesh culling strategy. Please refer to our supplementary material for more details. For evaluation of camera tracking, we adopt ATE RMSE~\cite{sturmBenchmarkEvaluationRGBD2012} (cm).

\noindent{\textbf{Baselines.}} 
We consider iMAP~\cite{sucarIMAPImplicitMapping2021a} and NICE-SLAM~\cite{zhuNiceslamNeuralImplicit2022} as our main baselines for reconstruction quality and camera tracking. For a fair comparison, we evaluate iMAP and NICE-SLAM with the same mesh culling strategy as Co-SLAM. Note that iMAP$^{\star}$ denotes the iMAP re-implementation released by the NICE-SLAM~\cite{zhuNiceslamNeuralImplicit2022} authors, which is much slower than the original implementation. To investigate the trade-off between accuracy and frame rate on real-world datasets, we report results of two versions of our method: \textit{Ours} refers to our proposed approach (which achieves real-time operation) while \textit{Ours$^\dagger$} indicates our method ran with twice as many tracking iterations.

\noindent{\textbf{Implementation Details.}}
We run Co-SLAM on a desktop PC with a 3.60GHz Intel Core i7-12700K CPU and NVIDIA RTX 3090ti GPU. For experiments with default settings (\textit{Ours}), which runs at 17 FPS on the Replica dataset, we use $N_t=1024$ pixels with $10$ iterations for tracking and $5\%$ of pixels from every $5^{th}$ frame for global bundle adjustment. We sample $M_r=32$ regular points and $M_d=11$ depth points along each camera ray, with $tr=10cm$. Please refer to supplementary materials for more specific settings on all datasets.

\begin{table*}[t]
    \centering
    \resizebox{0.98\textwidth}{!}
    {
    \begin{tabular}{l|l|cccc|ccc|c}
    \toprule
    Dataset & Method & Depth L1  (cm)$\downarrow$ & Acc. (cm)$\downarrow$ & Comp. (cm)$\downarrow$ & Comp. Ratio$\uparrow$& Tracking (ms) $\downarrow$ & Mapping (ms) $\downarrow$ & FPS $\uparrow$ & \#param. $\downarrow$  \\
    \midrule
    \multirow{4}{*}{Replica~\cite{straubReplicaDatasetDigital2019}} &
    TSDF-Fusion~\cite{curlessVolumetricMethodBuilding1996} &6.36 & \textbf{1.62}&3.94&83.93& N/A & N/A & N/A & 16.8 M\\
    &iMAP~\cite{sitzmannImplicitNeuralRepresentations2020} &4.64 & 3.62&4.93&80.51& 101 (200, 6) & 448 (1000, 10) & 9.9 & 0.26 M\\
    &NICE-SLAM~\cite{zhuNiceslamNeuralImplicit2022} & 1.90 & 2.37&2.64&91.13& 78 (200, 10) & 5470 (1000, 60) & 0.91 & 17.4 M\\
    & Ours & \textbf{1.51} & 2.10 & \textbf{2.08} & \textbf{93.44}& 58 (1024, 10) & 98 (2048, 10) & 17.4 & 0.26 M\\
    \midrule
    &
    TSDF-Fusion~\cite{curlessVolumetricMethodBuilding1996} &10.87 & \textbf{1.62} & 5.16 & 81.52 & N/A & N/A & N/A & 16.8 M \\
    Synthetic &iMAP$^{\star}$~\cite{sitzmannImplicitNeuralRepresentations2020} & 43.91 & 18.30 & 26.41 & 20.73 & 1550 (5000, 50) & 14730 (5000, 300) & 0.34 & 0.22 M \\
    RGBD~\cite{azinovicNeuralRGBDSurface2022c} & NICE-SLAM~\cite{zhuNiceslamNeuralImplicit2022} & 6.32 & 5.96 & 5.30 &77.46 & 123 (1024, 10) & 3792 (1000, 60) & 1.31 & 3.11 M \\
    & Ours&\textbf{3.02} & 2.95 & \textbf{2.96} & \textbf{86.88}& 64 (1024, 10) & 104 (2048, 10) & 15.6 & 0.26 M \\
    \bottomrule
    \end{tabular}
    }
    \vspace{-1mm}
    \caption{Reconstruction quality and run-time memory comparison on Replica~\cite{straubReplicaDatasetDigital2019} and Synthetic-RGBD~\cite{azinovicNeuralRGBDSurface2022c} with respective settings. TSDF-Fusion is reconstructed with poses estimated by Co-SLAM. Run-time is reported in \texttt{time(\#pixel, \#iter)} for a comprehensive comparison. The model size is averaged across all scenes. 
    }
    \label{tab:time_memory_performance}
    \vspace{-12pt}
\end{table*}

\begin{table}[t]
    \centering
    \resizebox{1.00\columnwidth}{!}
    {
    \begin{tabular}{l|l|ccc|cc}
    \toprule
     & Method & Track. (ms) $\downarrow$ & Map. (ms) $\downarrow$ & FPS $\uparrow$ & \#param. $\downarrow$ \\
    \midrule
    \multirow{4}{*}{\rotatebox{90}{ScanNet}}
    &
    iMAP$^{\star}$ & 30.4$\times$50 & 44.9$\times$300 & 0.37 & 0.2 M \\
    &  
    NICE-SLAM & 12.3$\times$50 & 125.3$\times$60 & 0.68 & 10.3 M \\
    & Ours$^\dagger$ & 7.8$\times$20 & 20.2$\times$10 & 6.4 & 0.8 M \\
    & Ours & 7.8$\times$10 & 20.2$\times$10 & 12.8 & 0.8 M \\
    \midrule
    \multirow{4}{*}{\rotatebox{90}{TUM}}
    &
    iMAP$^{\star}$ & 29.6$\times$200 & 44.3$\times$300 & 0.07 & 0.2 M \\
    & 
    NICE-SLAM & 47.1$\times$200 & 189.2$\times$60 & 0.08 & 101.6 M \\
    & Ours$^{\dagger}$ & 7.5$\times$20 & 19.0$\times$20 & 6.7 & 1.6 M \\
    & Ours & 7.5$\times$10 & 19.0$\times$20 & 13.3 & 1.6 M \\
    
    \bottomrule
    \end{tabular}
    }
    \vspace{-1mm}
    \caption{Run-time and memory comparison on ScanNet~\cite{daiScannetRichlyannotated3d2017} and TUM-RGBD~\cite{sturmBenchmarkEvaluationRGBD2012} with respective settings. Run-time is reported in \texttt{ms/iter $\times$ \#iter} for a detailed comparison. NICE-SLAM and iMAP$^{\star}$ run mapping at \textbf{every frame} on TUM-RGBD. Mapping happens \textbf{every 5 frames} in all other cases. The model size is averaged across all scenes. 
    }
    \label{tab:time_memory_old}
    \vspace{-5pt}
\end{table}

\begin{figure}[t]
  \centering
  \scriptsize
  \setlength{\tabcolsep}{1.5pt}
  \newcommand{\sz}{0.30}
  \begin{tabular}{lccc}
    & \tt complete\_kitchen & \tt green\_room & \tt grey\_white\_room \\
    \makecell{\rotatebox{90}{iMAP$^*$~\cite{sucarIMAPImplicitMapping2021a}}} &
    \makecell{\includegraphics[width=\sz\linewidth]{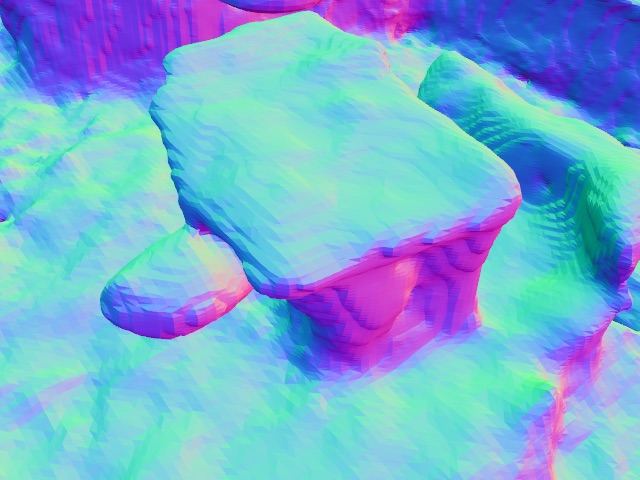}} &
    \makecell{\includegraphics[width=\sz\linewidth]{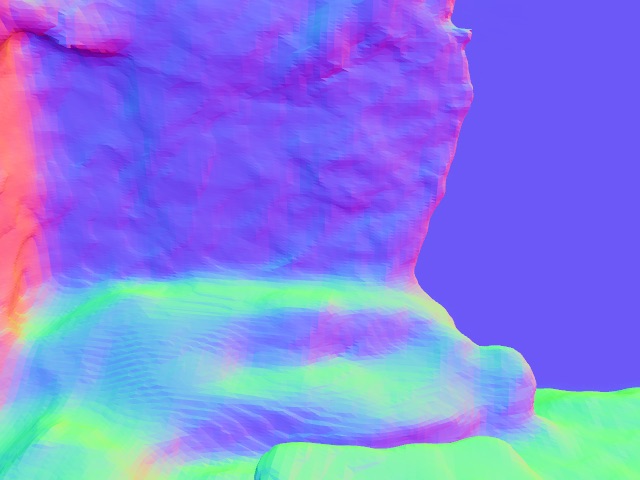}} &
    \makecell{\includegraphics[width=\sz\linewidth]{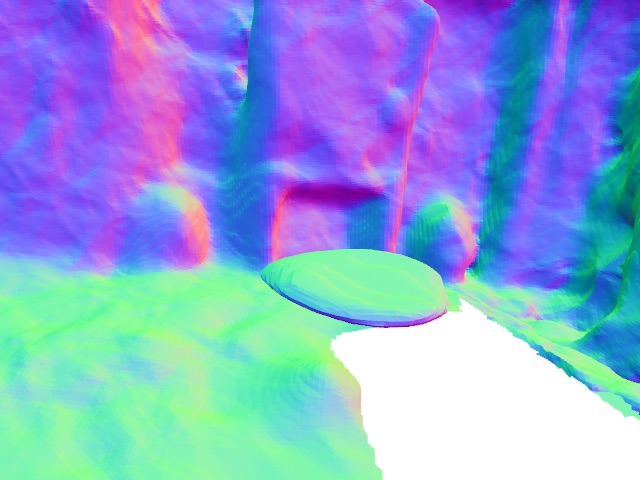}}\\
    \makecell{\rotatebox{90}{NICE-SLAM~\cite{zhuNiceslamNeuralImplicit2022}}} &
    \makecell{\includegraphics[width=\sz\linewidth]{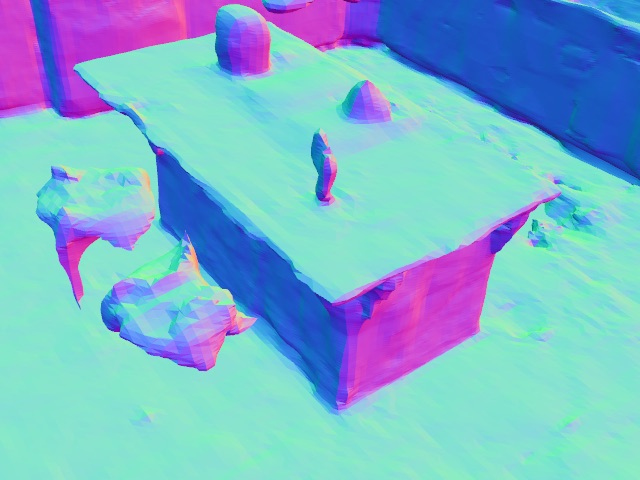}} &
    \makecell{\includegraphics[width=\sz\linewidth]{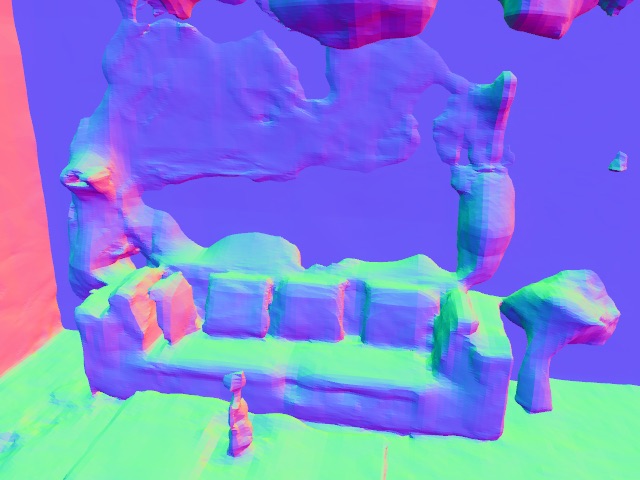}} &
    \makecell{\includegraphics[width=\sz\linewidth]{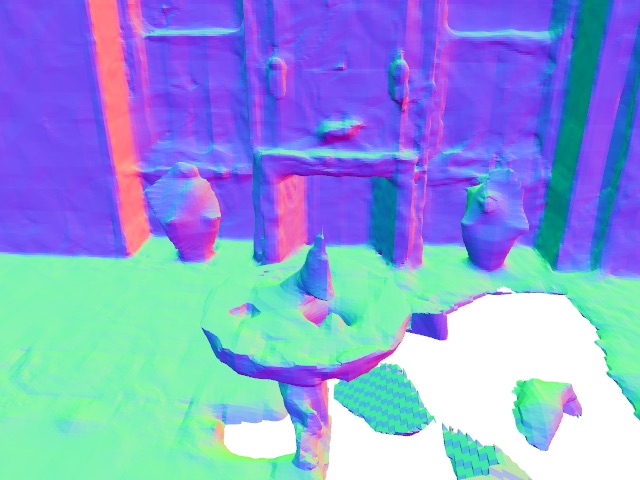}} \\
    \makecell{\rotatebox{90}{Ours}} &    
    \makecell{\includegraphics[width=\sz\linewidth]{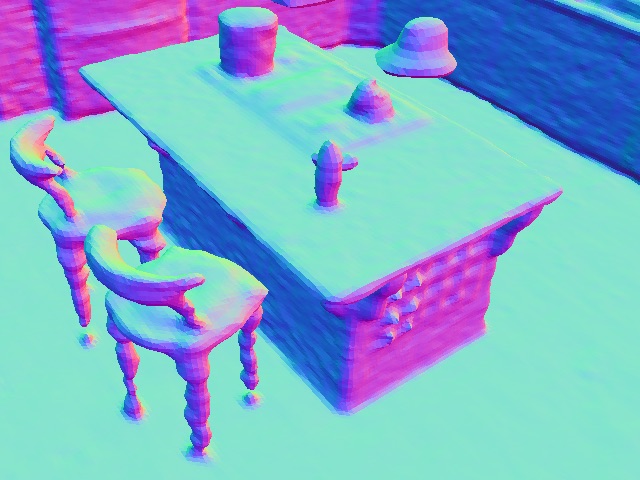}} &
    \makecell{\includegraphics[width=\sz\linewidth]{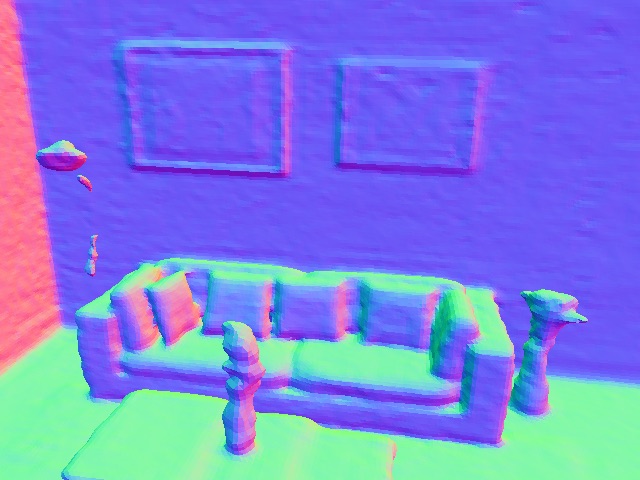}} &
    \makecell{\includegraphics[width=\sz\linewidth]{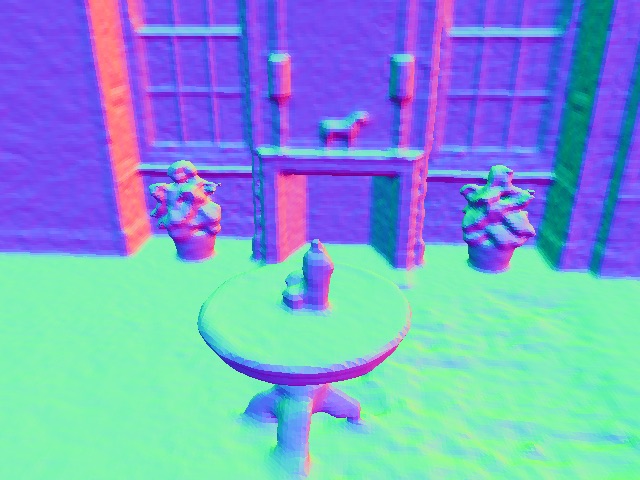}} \\
    \makecell{\rotatebox{90}{GT}} &    
    \makecell{\includegraphics[width=\sz\linewidth]{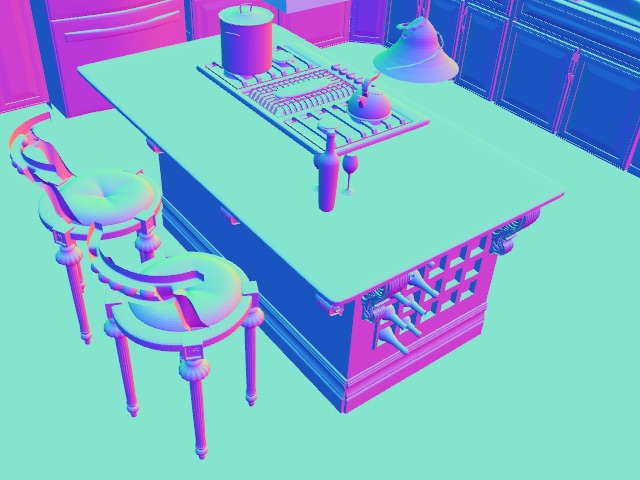}} &
    \makecell{\includegraphics[width=\sz\linewidth]{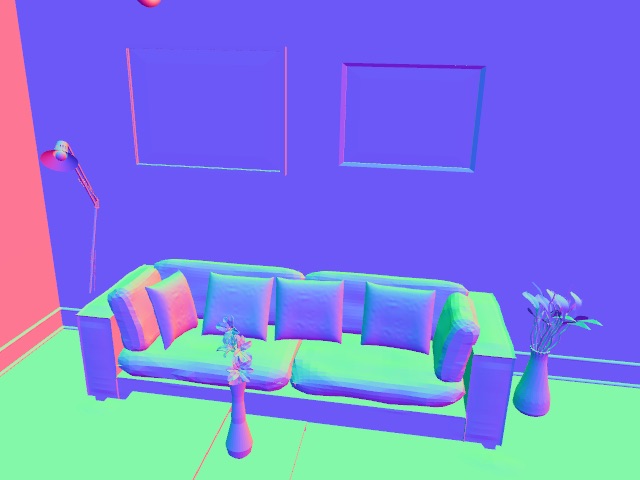}} &
    \makecell{\includegraphics[width=\sz\linewidth]{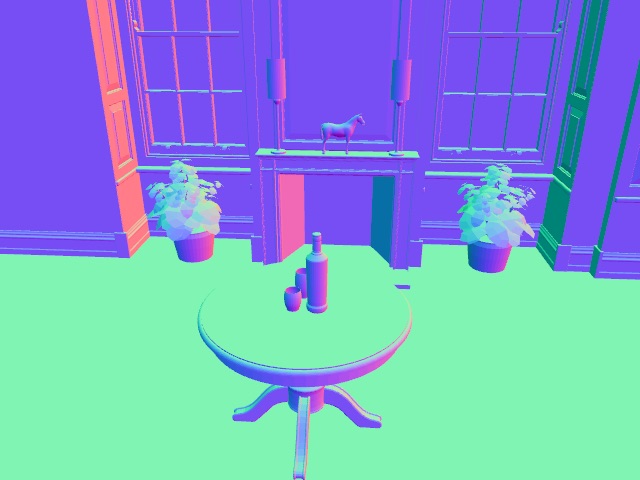}} \\
  \end{tabular}
  \vspace{-3mm}
  \caption{Reconstruction results on Synthetic RGB-D dataset~\cite{azinovicNeuralRGBDSurface2022c}. Our method can recover thin structures and achieve plausible scene completion given noisy depth measurement.
  }
  \label{fig:RGBD}
  \vspace{-12pt}
\end{figure}

\begin{figure*}[tbp]
  \centering
  \scriptsize
  \setlength{\tabcolsep}{0.5pt}
  \newcommand{\sz}{0.235}  %
  
  \begin{tabular}{lcc@{\hspace{3pt}}cc@{\hspace{3pt}}cc}
    & \multicolumn{2}{c}{\tt \footnotesize{room-0}} & \multicolumn{2}{c}{\tt \footnotesize{room-1}} & \multicolumn{2}{c}{\tt \footnotesize{office-2}} \\
    \makecell{\rotatebox{90}{iMAP~\cite{sucarIMAPImplicitMapping2021a}}}  &
    \makecell{\includegraphics[height=\sz\columnwidth]{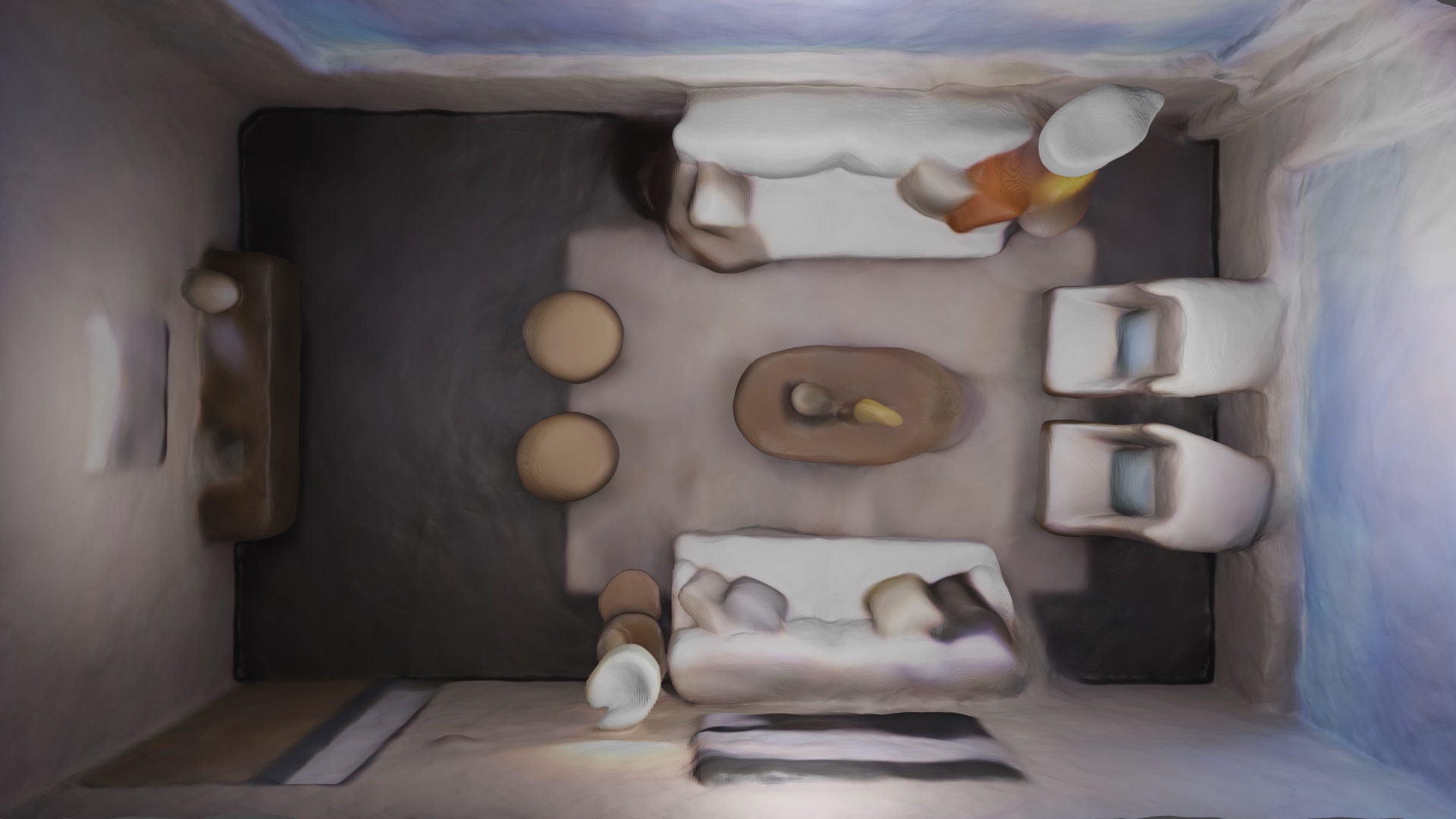}} & 
    \makecell{\includegraphics[height=\sz\columnwidth]{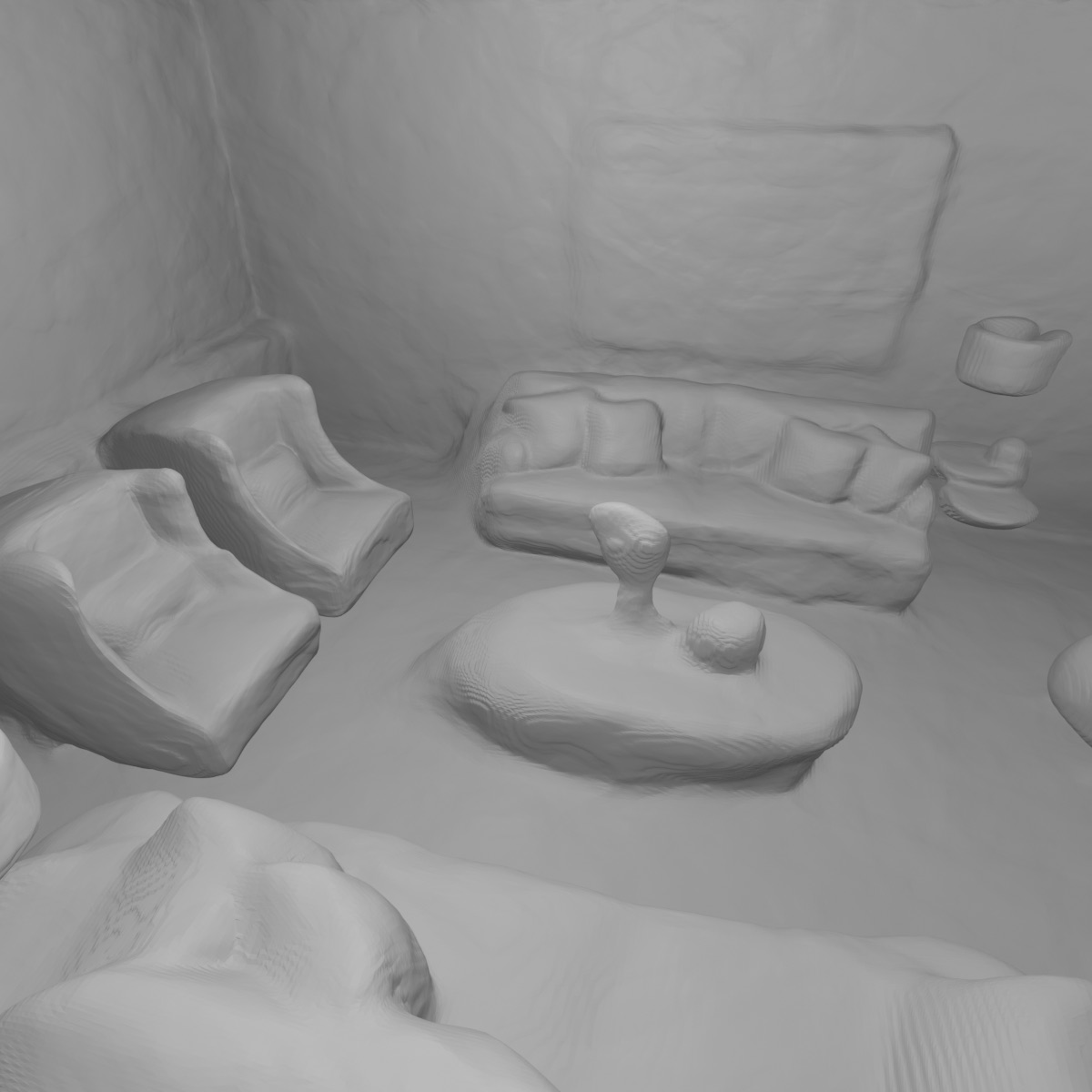}} &
    \makecell{\includegraphics[height=\sz\columnwidth]{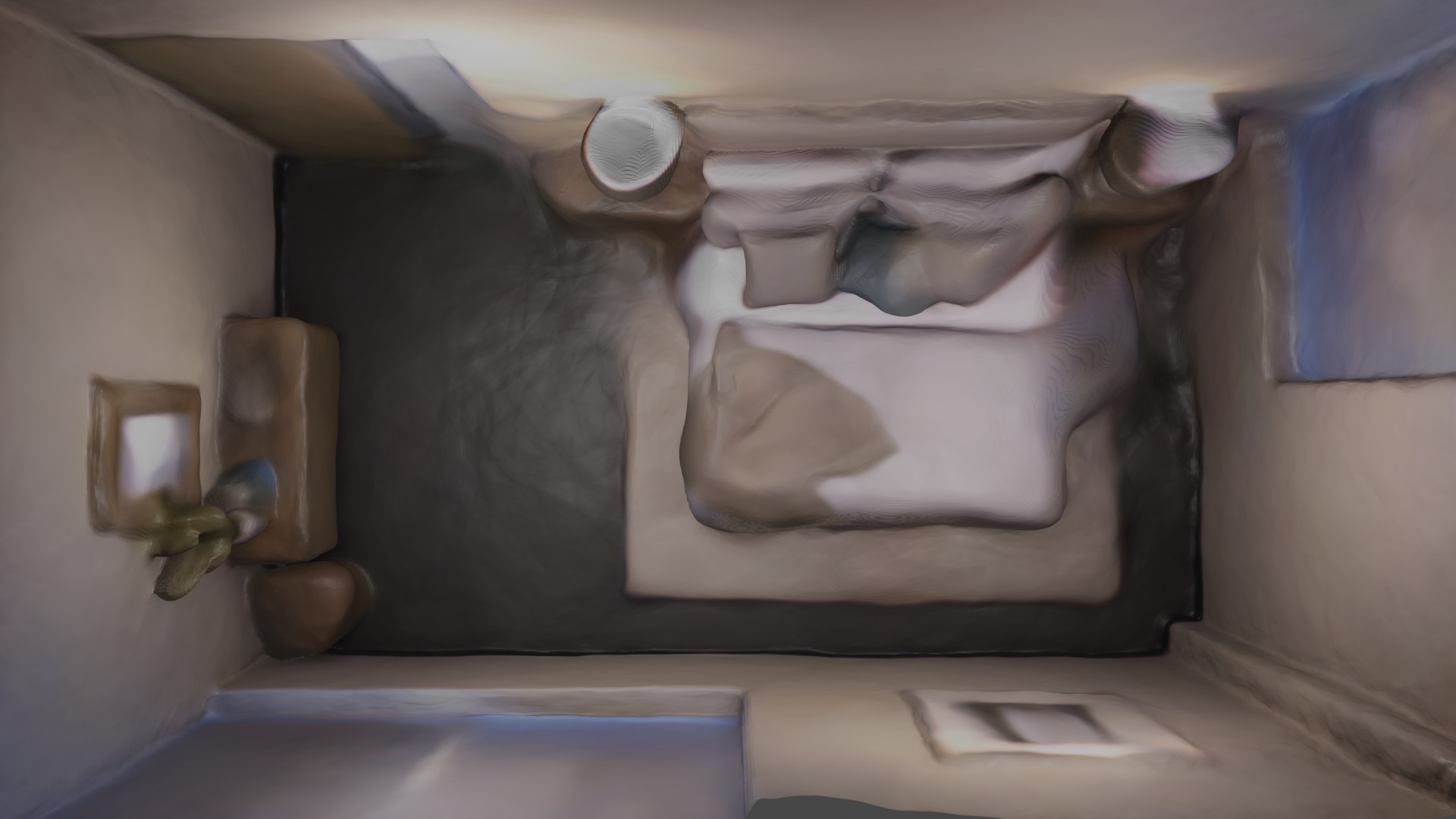}} & 
    \makecell{\includegraphics[height=\sz\columnwidth]{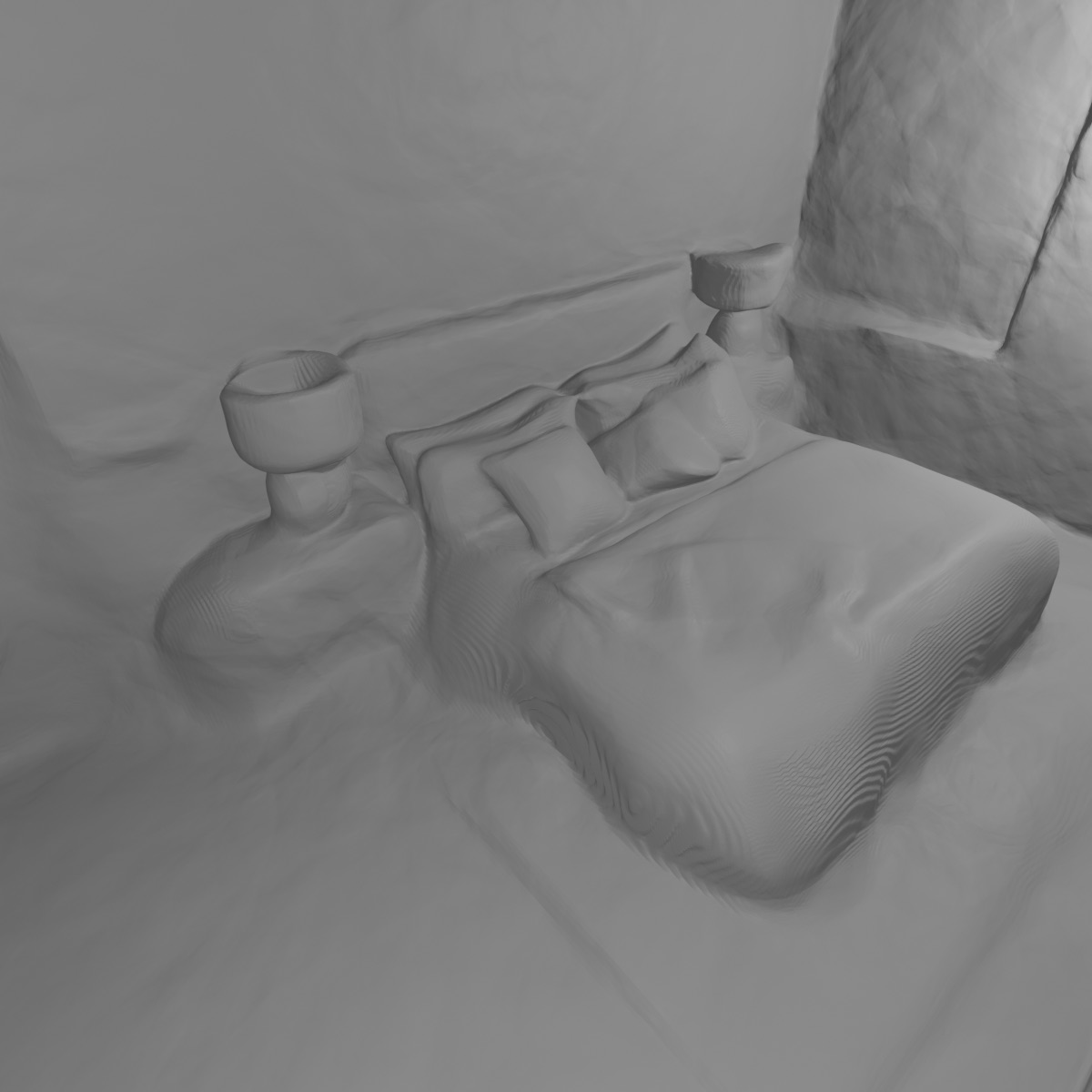}} &
    \makecell{\includegraphics[height=\sz\columnwidth]{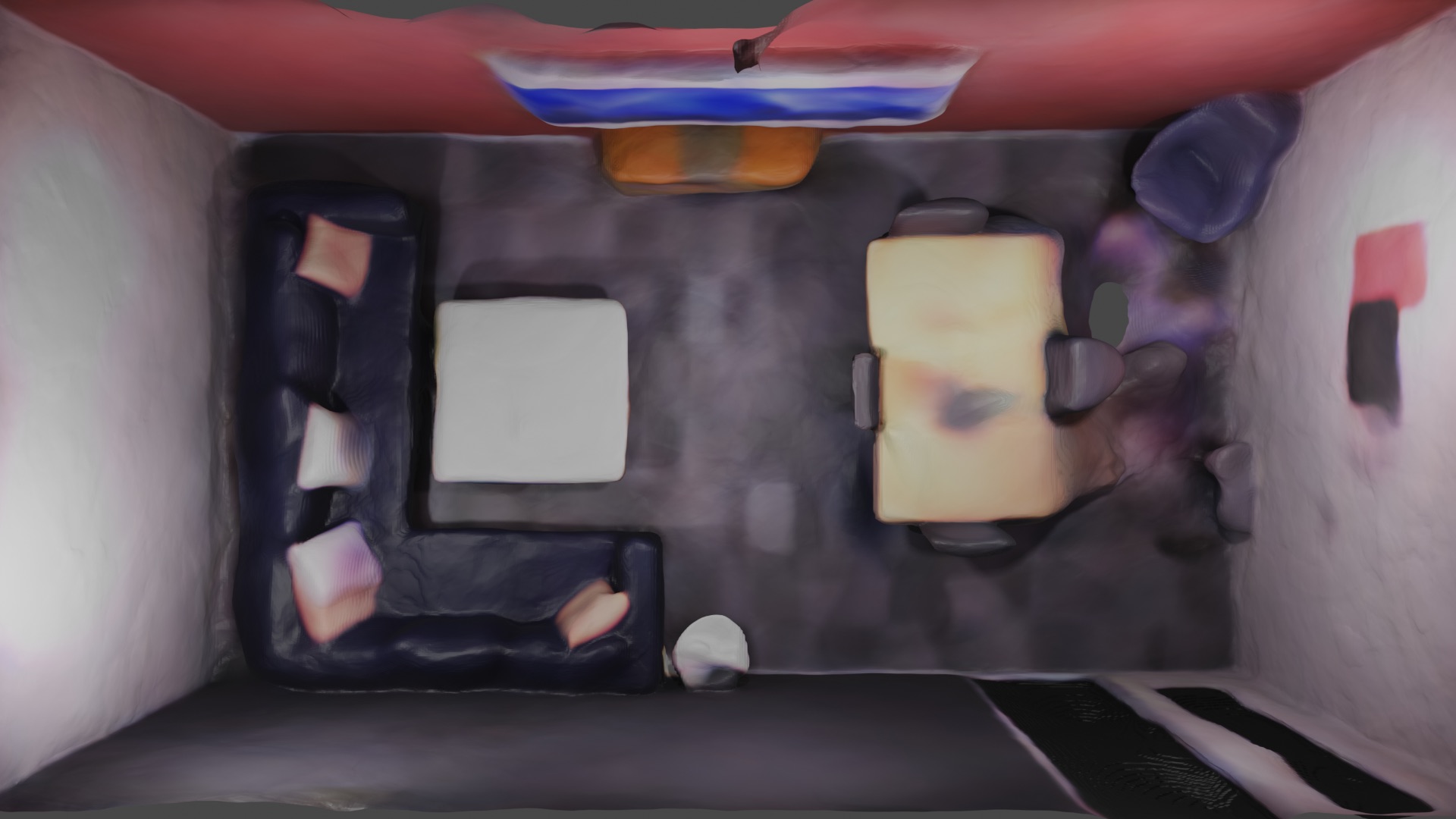}} &
    \makecell{\includegraphics[height=\sz\columnwidth]{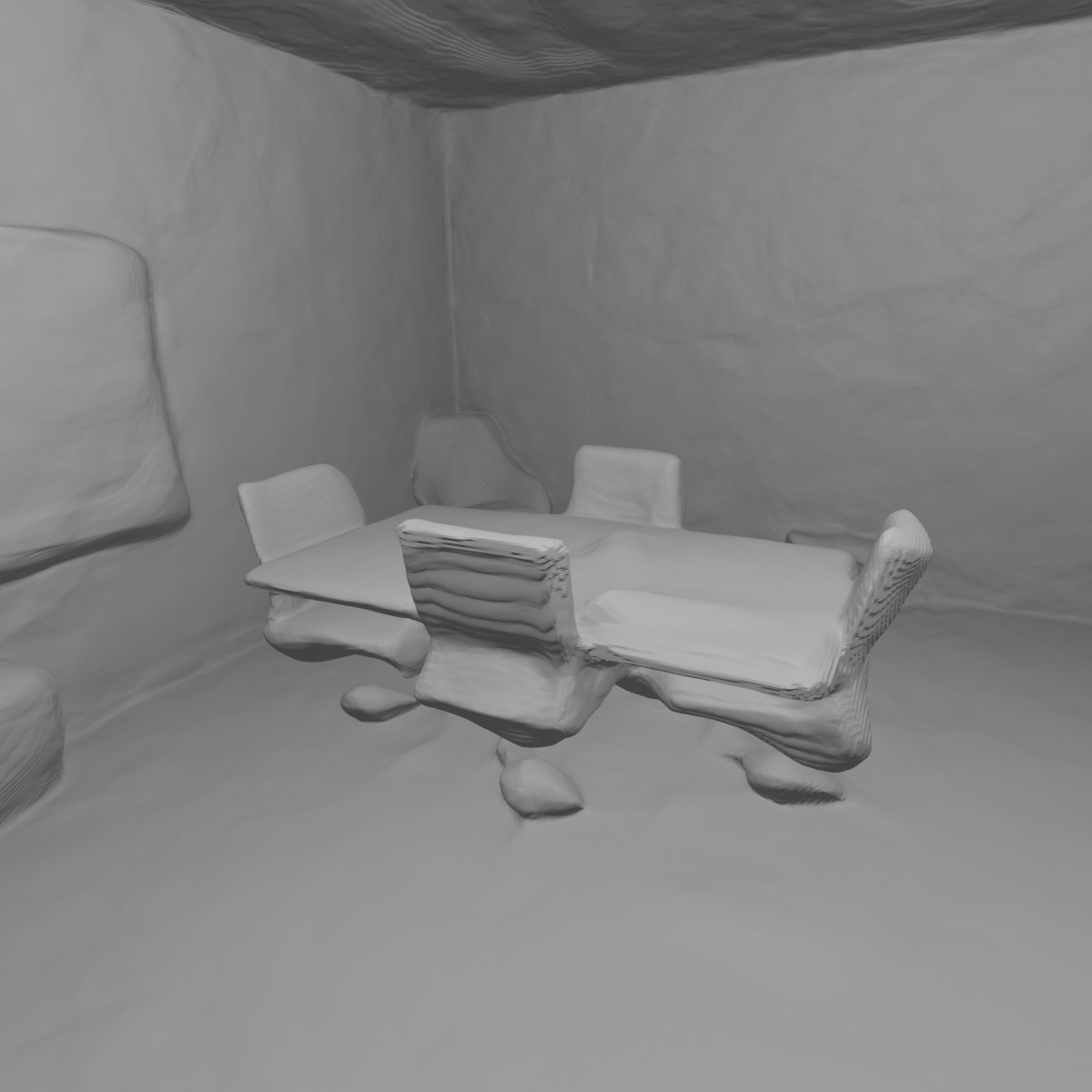}} \\
    \makecell{\rotatebox{90}{NICE-SLAM~\cite{zhuNiceslamNeuralImplicit2022}}}  &
    \makecell{\includegraphics[height=\sz\columnwidth]{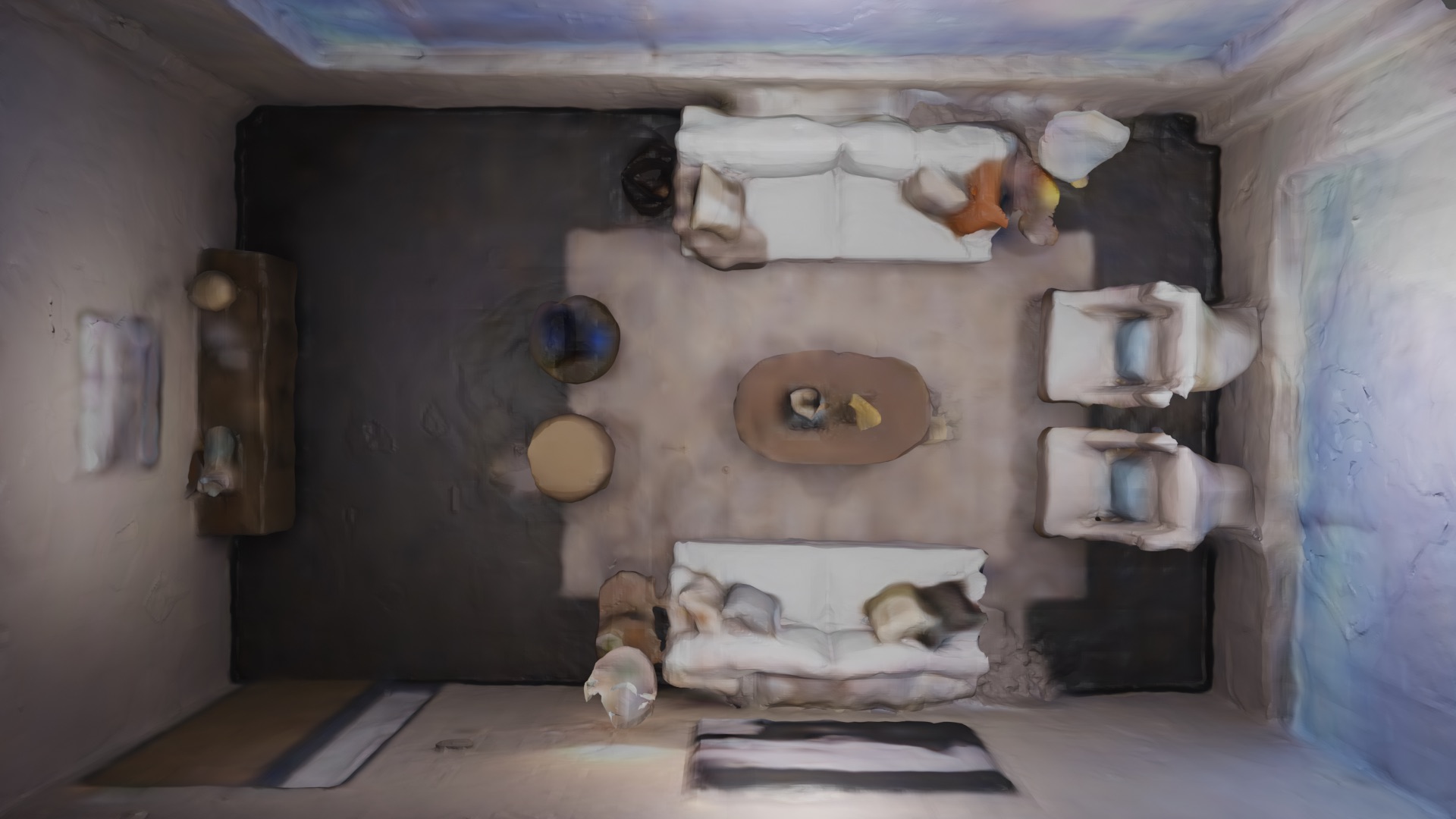}} &
    \makecell{\includegraphics[height=\sz\columnwidth]{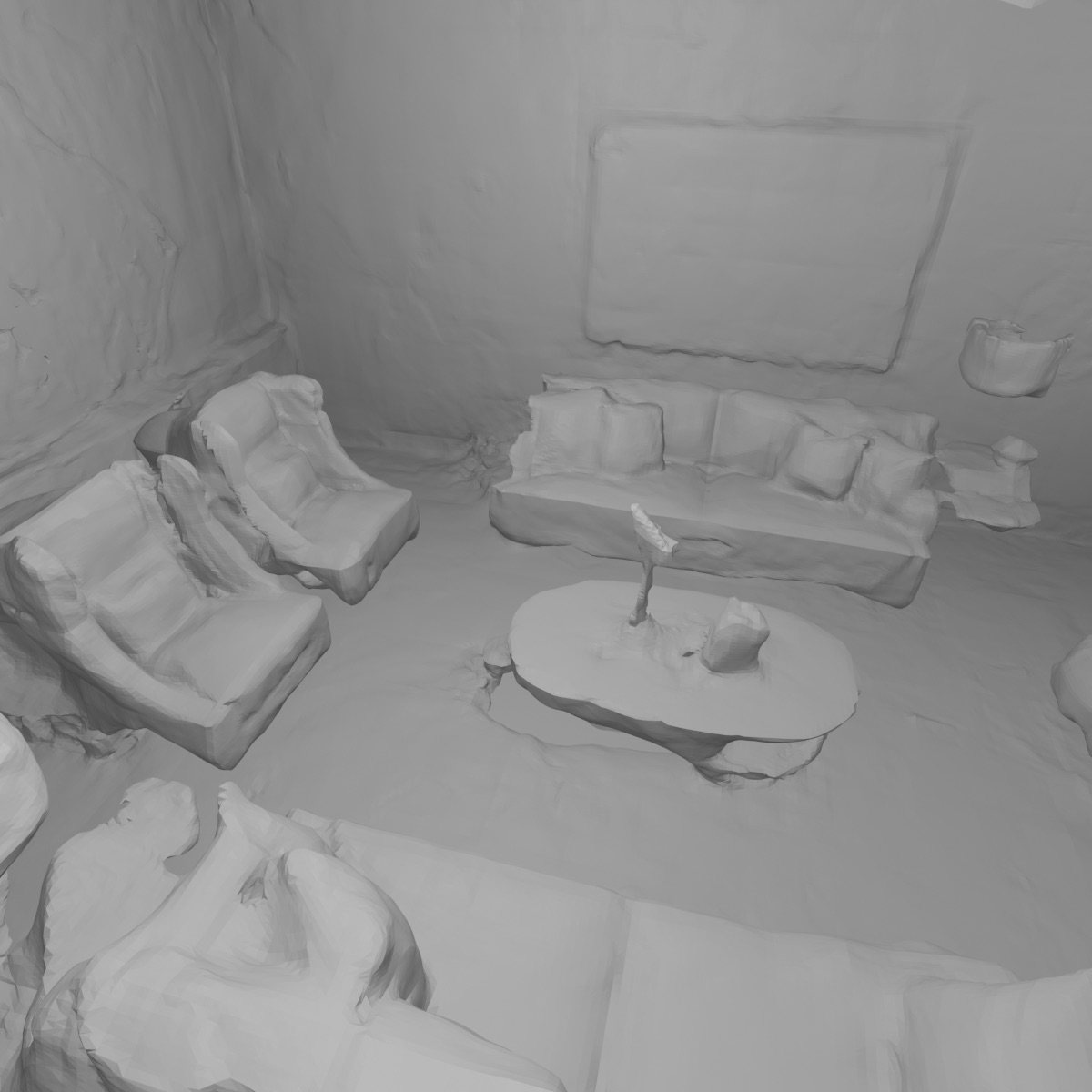}} &    
    \makecell{\includegraphics[height=\sz\columnwidth]{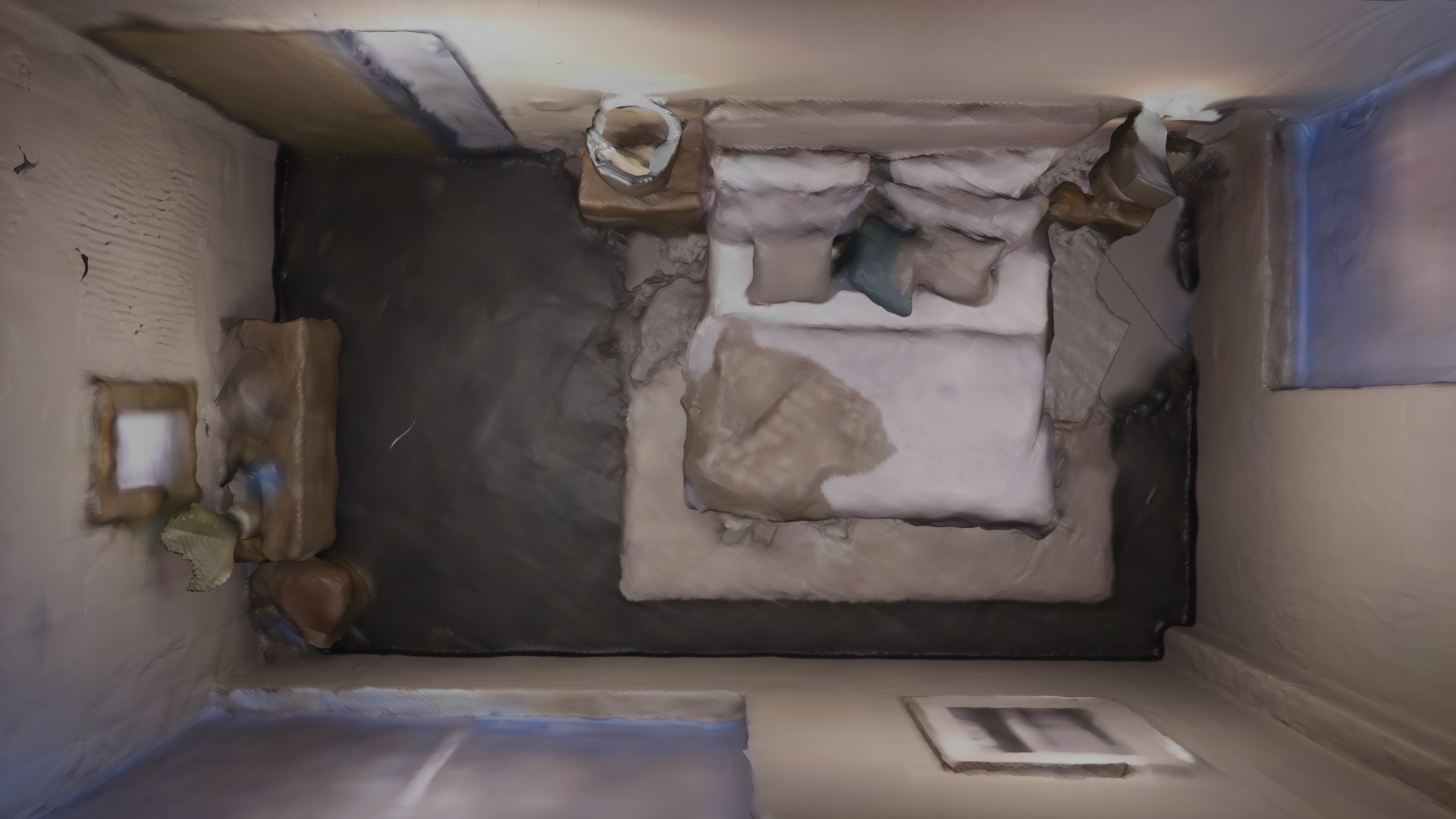}} &
    \makecell{\includegraphics[height=\sz\columnwidth]{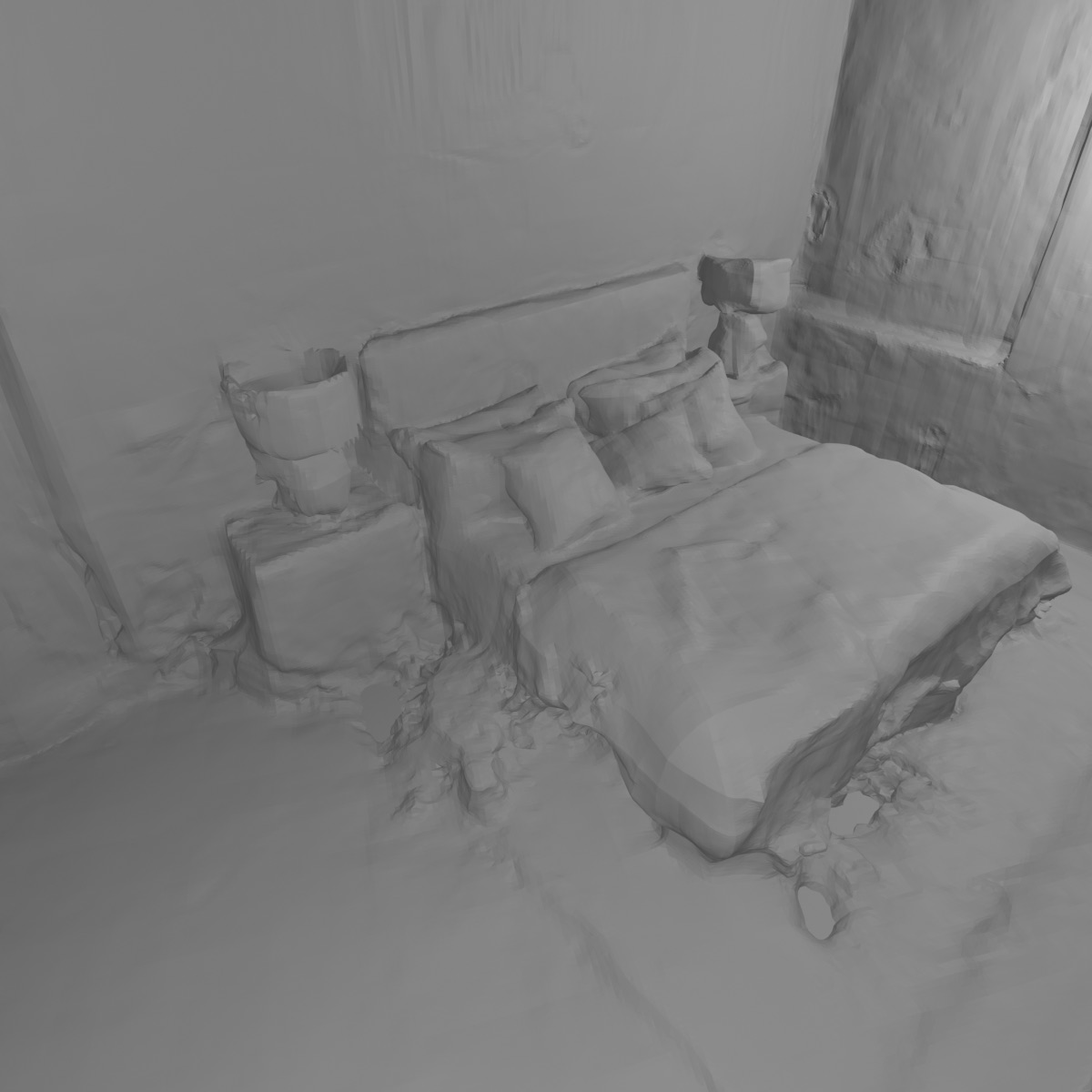}} &    
    \makecell{\includegraphics[height=\sz\columnwidth]{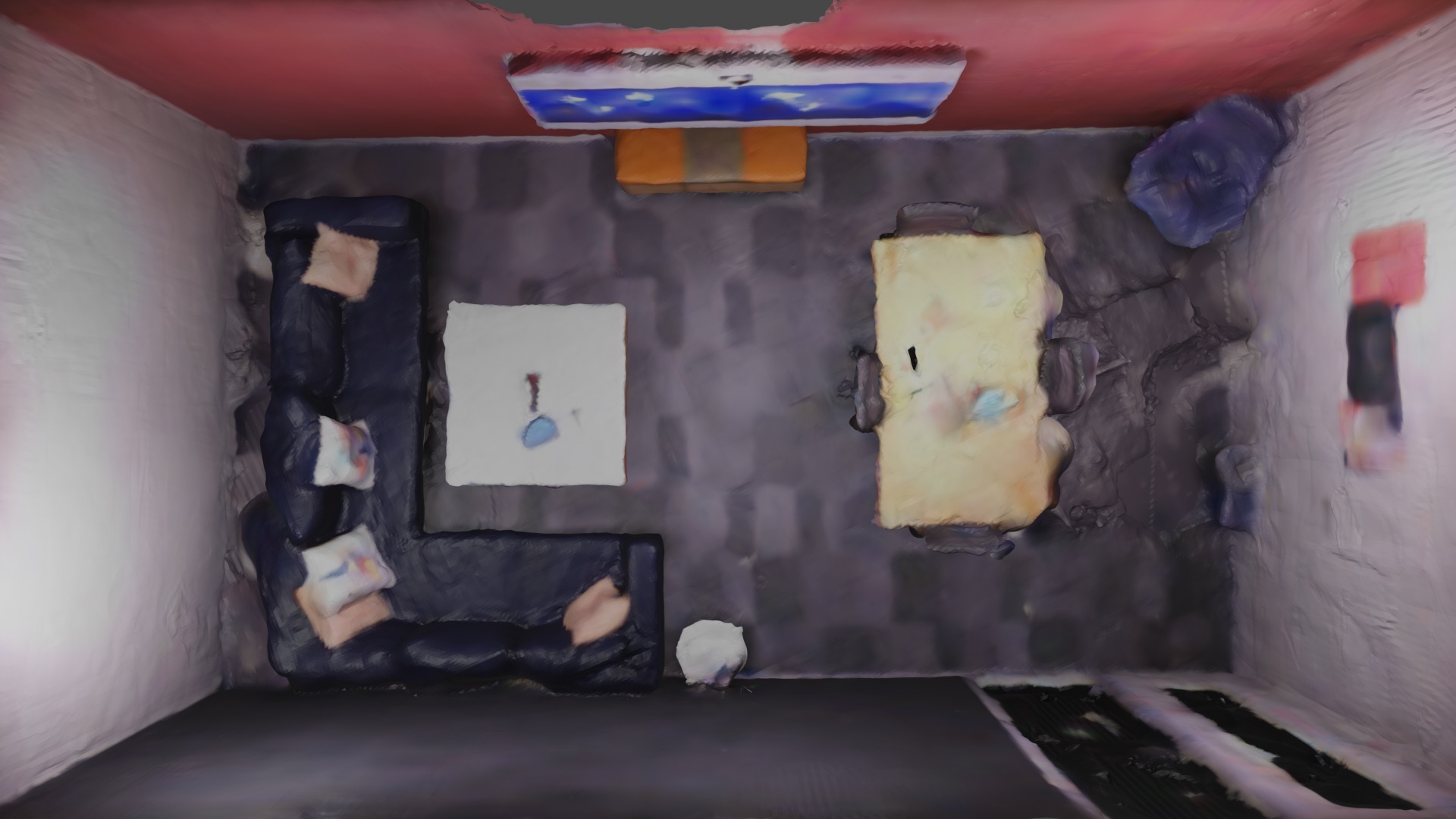}} &
    \makecell{\includegraphics[height=\sz\columnwidth]{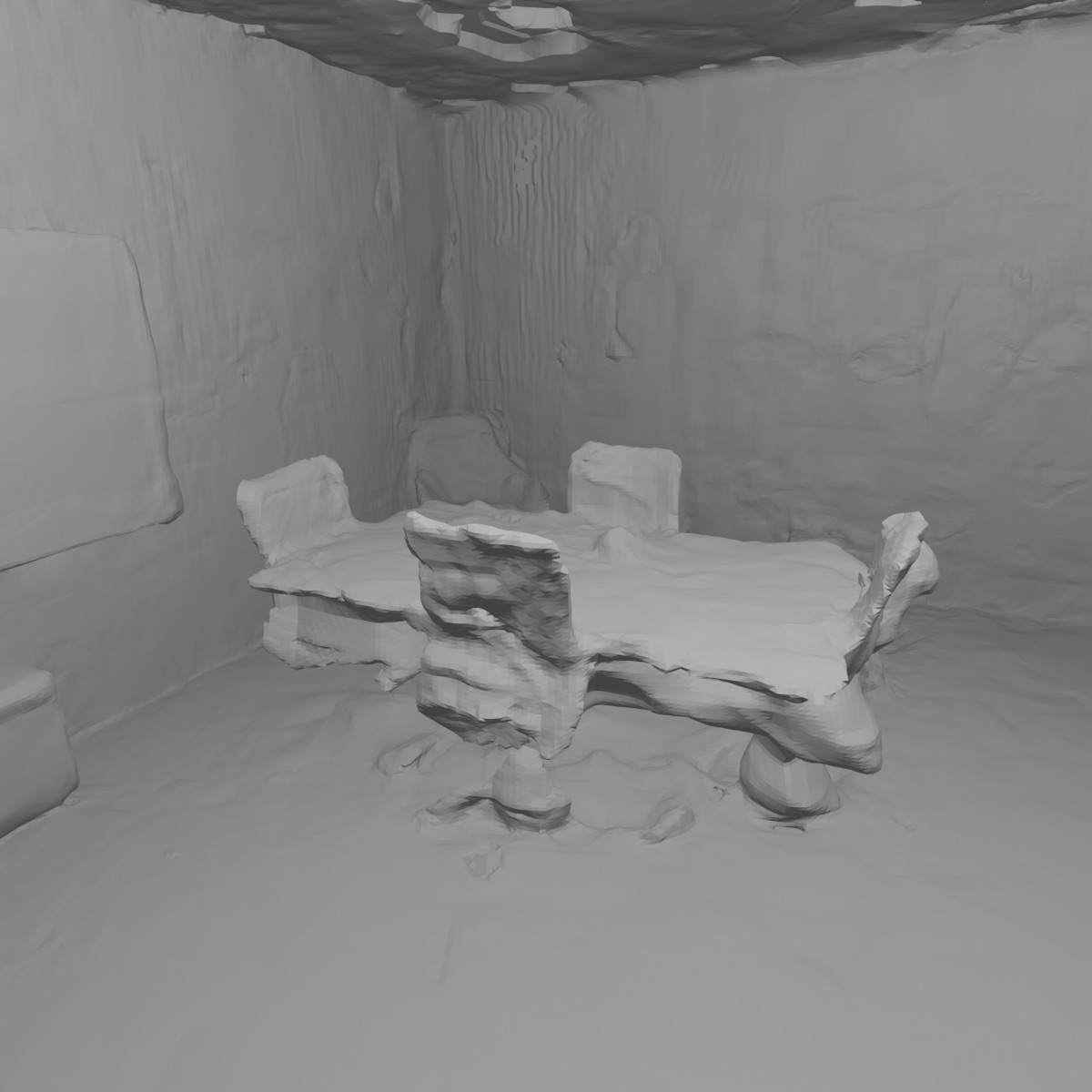}} \\
    \makecell{\rotatebox{90}{Ours}} &
    \makecell{\includegraphics[height=\sz\columnwidth]{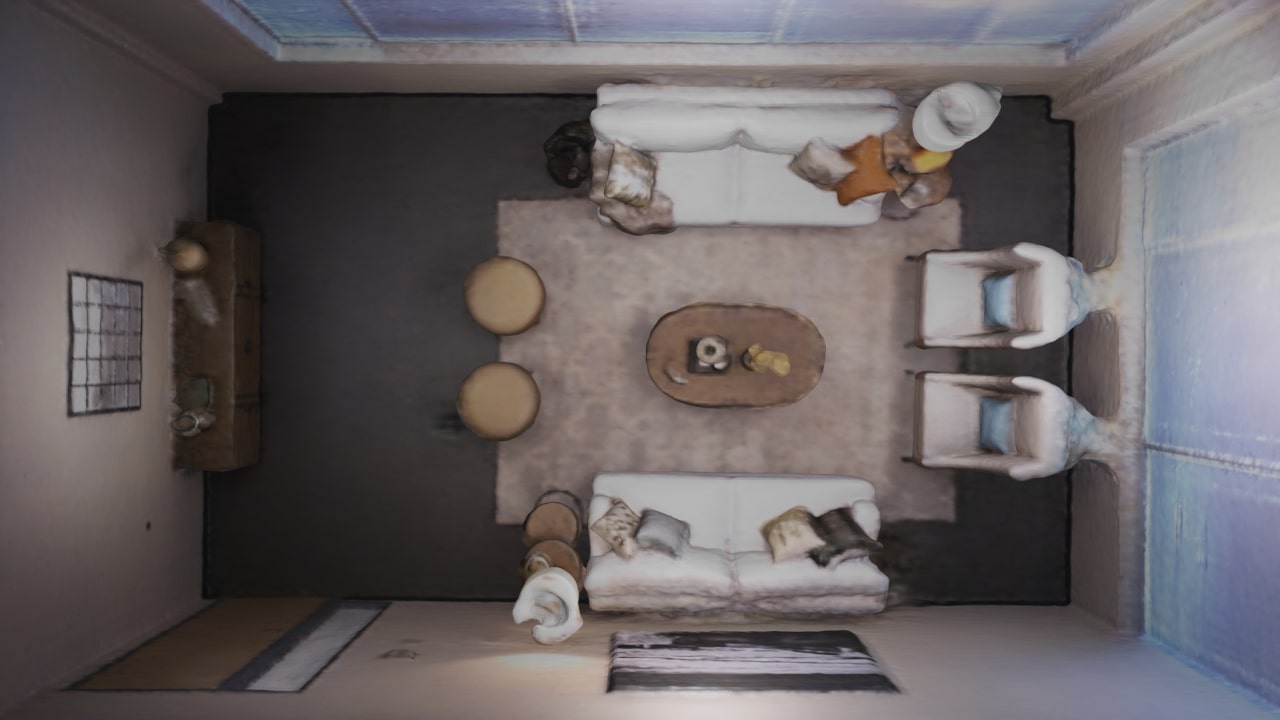}} &
    \makecell{\includegraphics[height=\sz\columnwidth]{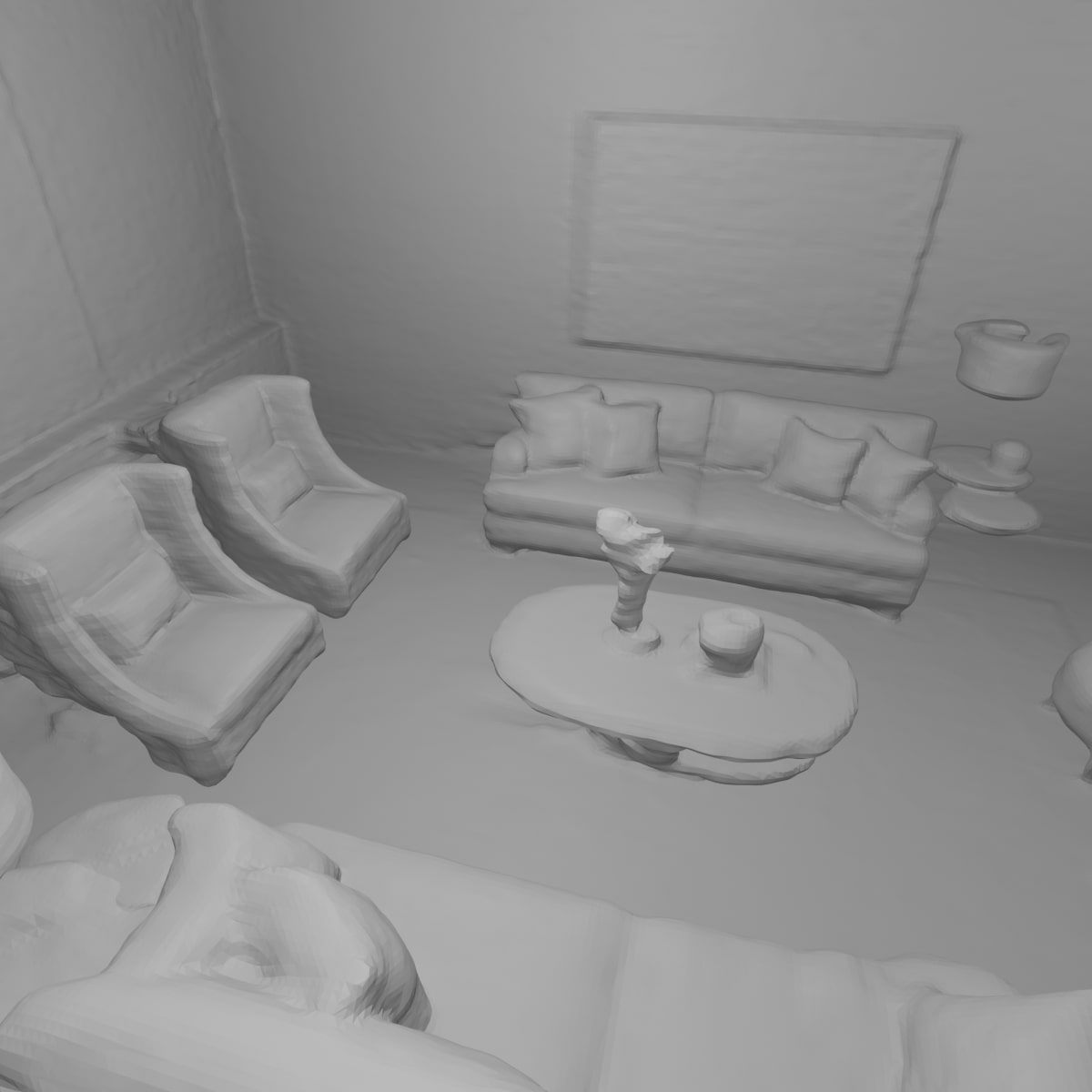}} &
    \makecell{\includegraphics[height=\sz\columnwidth]{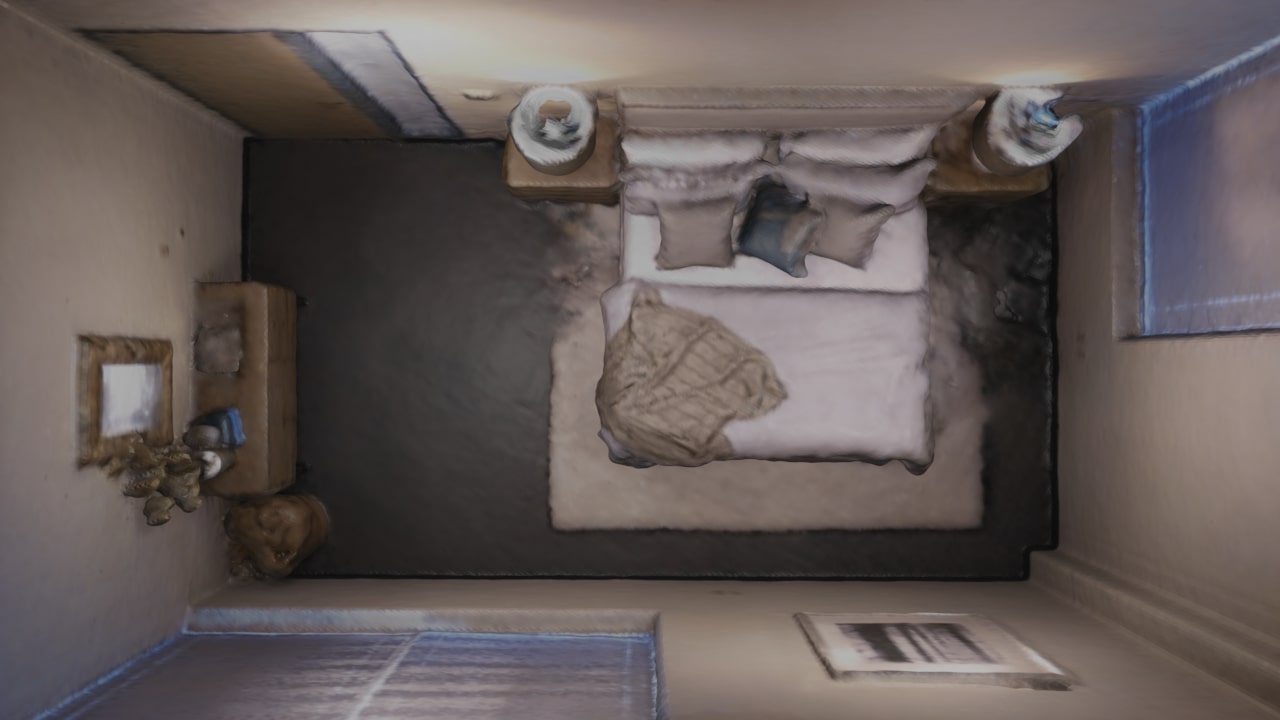}} &
    \makecell{\includegraphics[height=\sz\columnwidth]{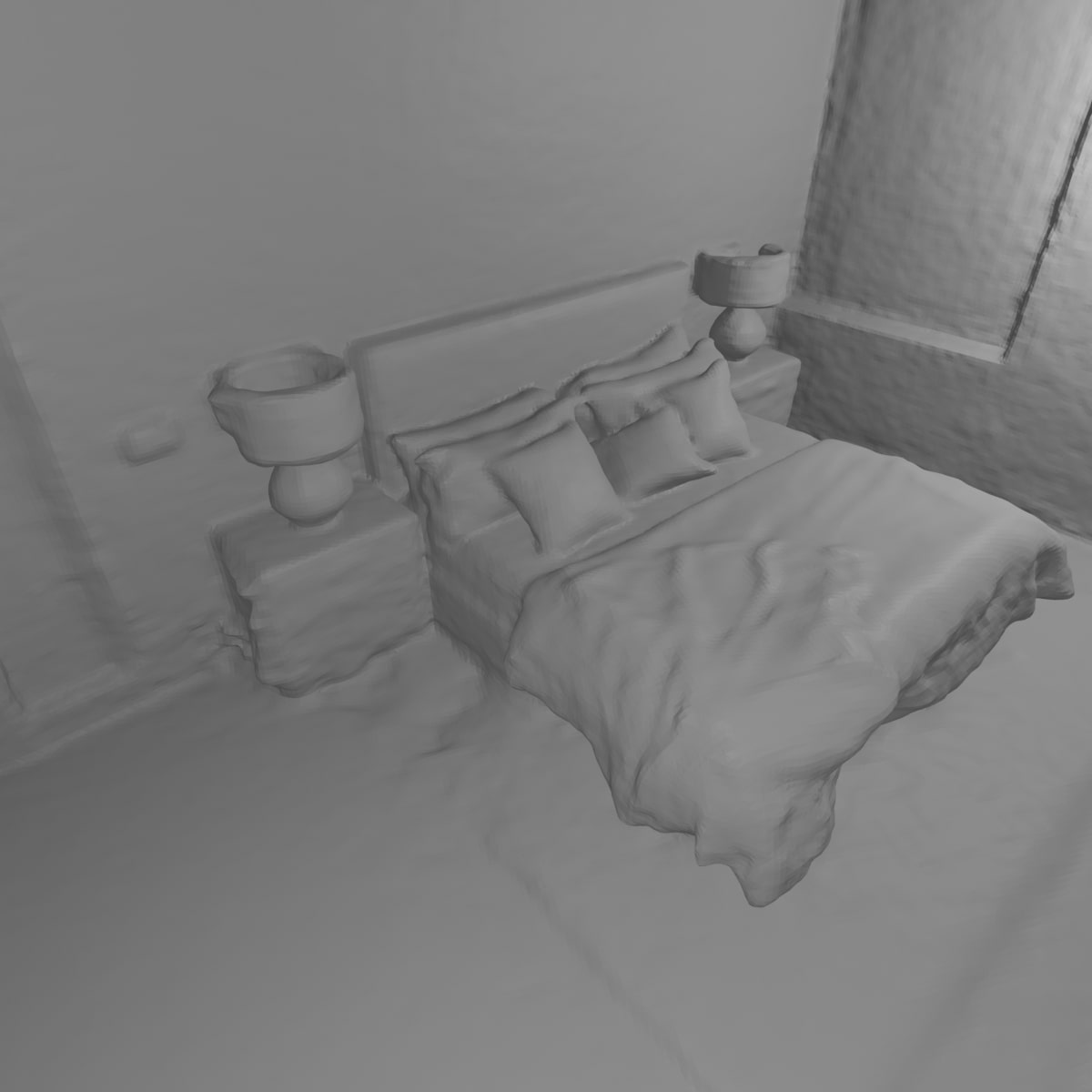}} &
    \makecell{\includegraphics[height=\sz\columnwidth]{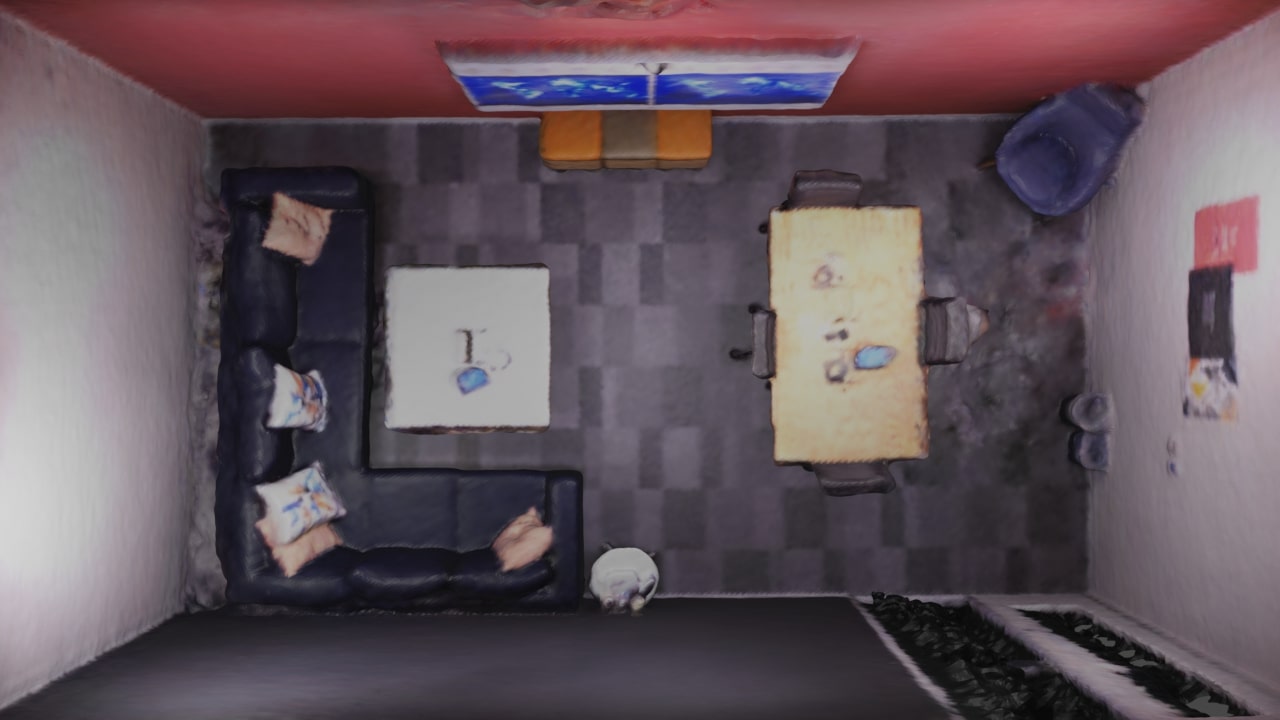}} &
    \makecell{\includegraphics[height=\sz\columnwidth]{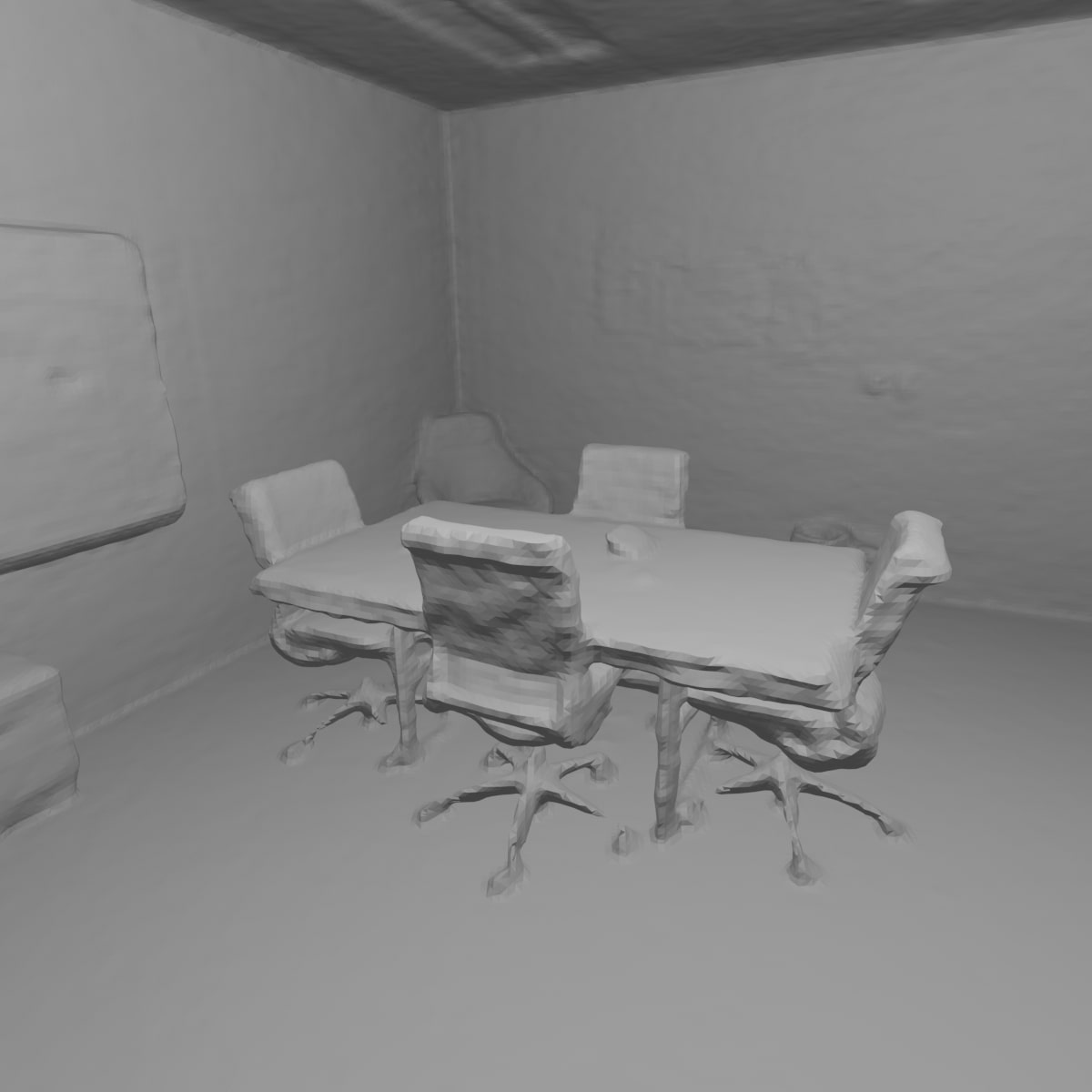}} \\
    
    \makecell{\rotatebox{90}{GT}} &
    \makecell{\includegraphics[height=\sz\columnwidth]{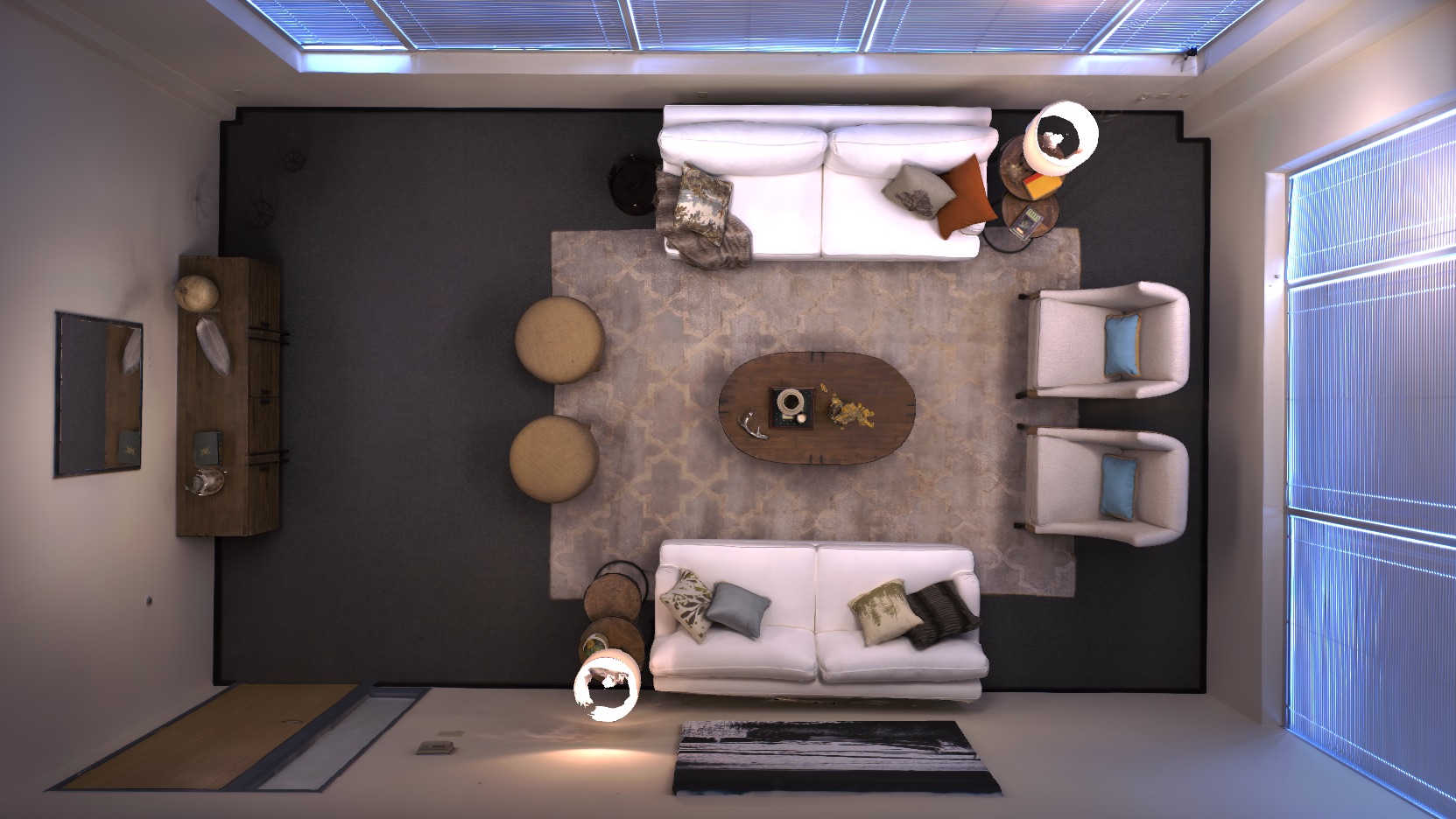}} &
    \makecell{\includegraphics[height=\sz\columnwidth]{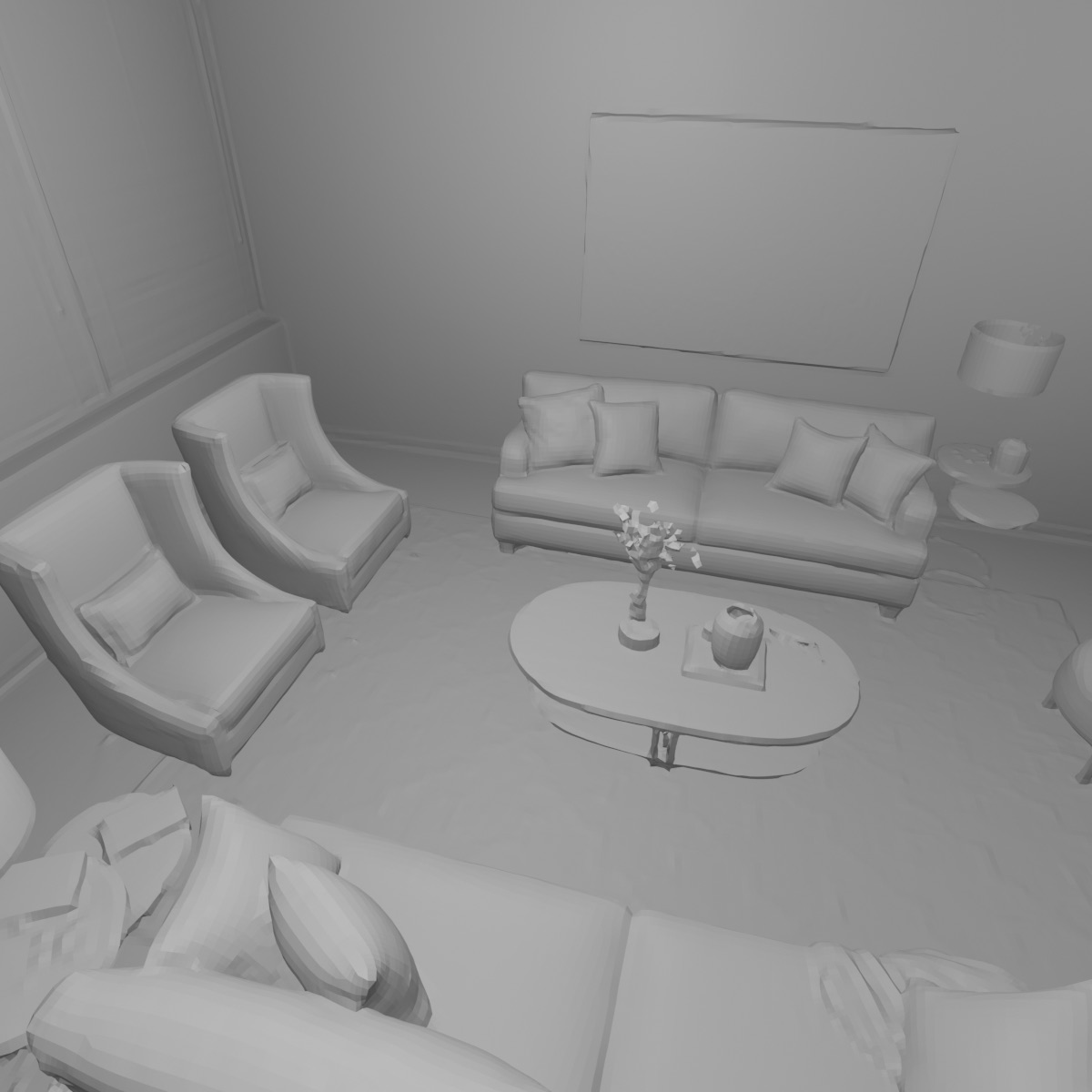}} &
    \makecell{\includegraphics[height=\sz\columnwidth]{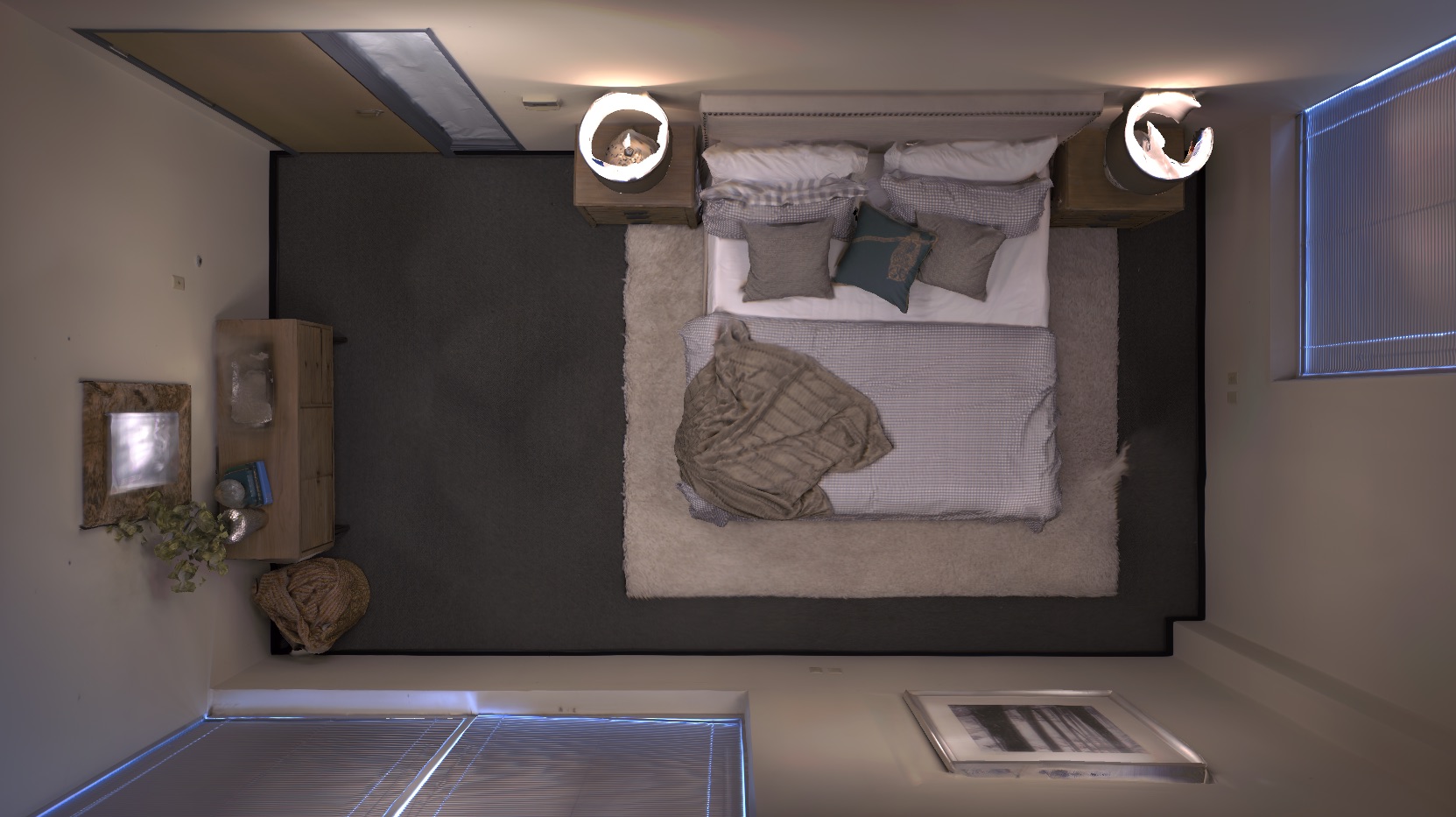}} &
    \makecell{\includegraphics[height=\sz\columnwidth]{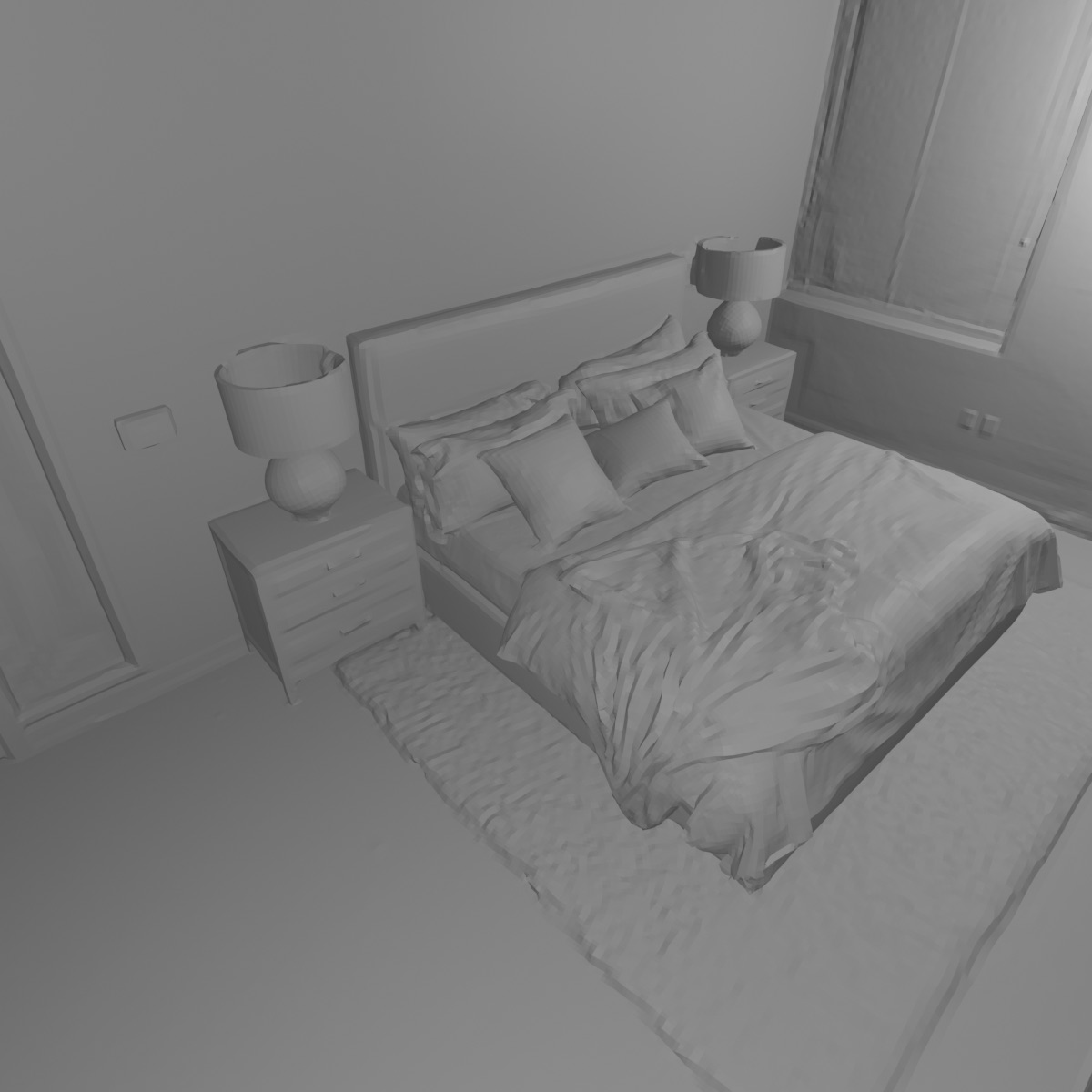}} &
    \makecell{\includegraphics[height=\sz\columnwidth]{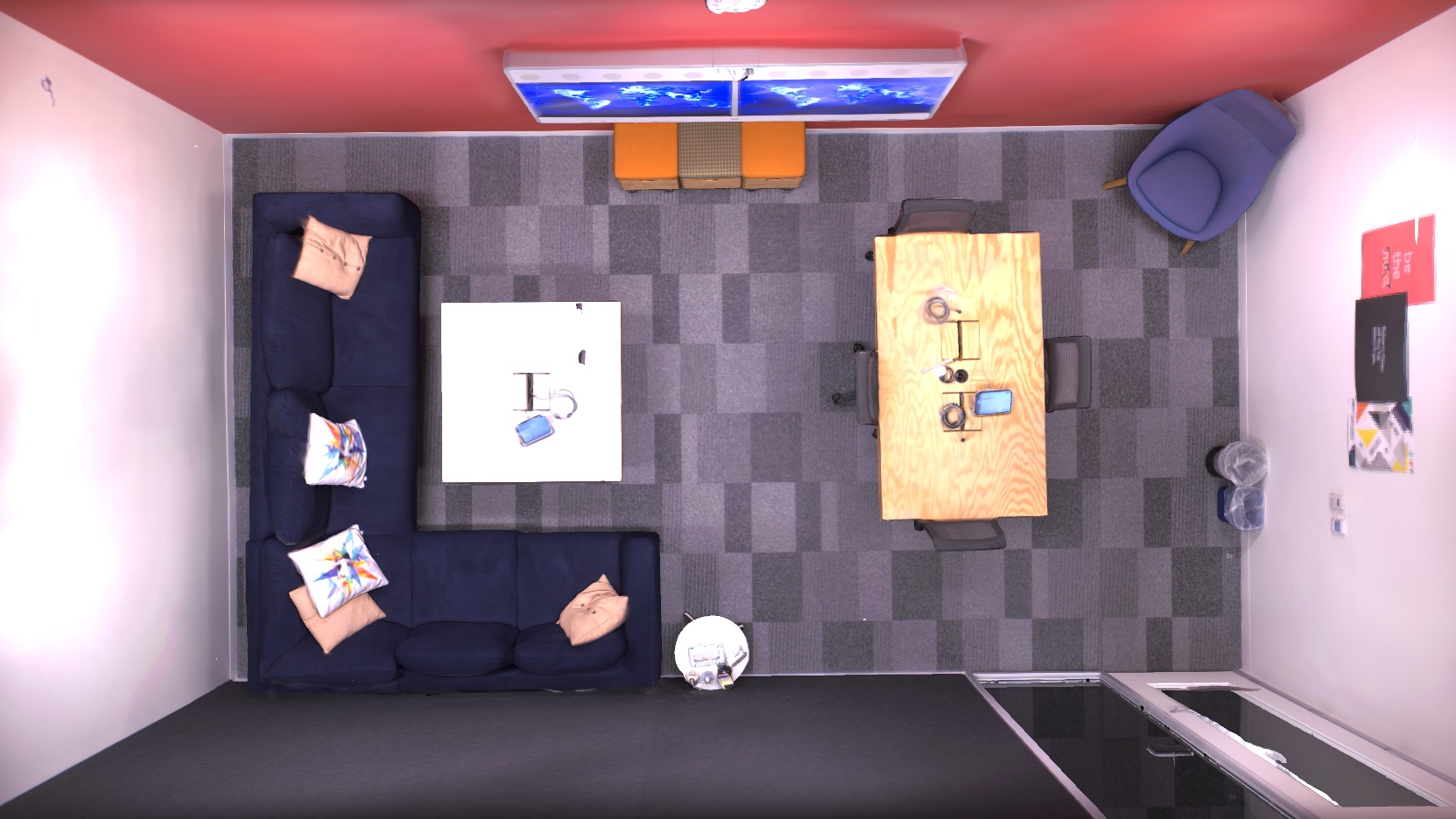}} &
    \makecell{\includegraphics[height=\sz\columnwidth]{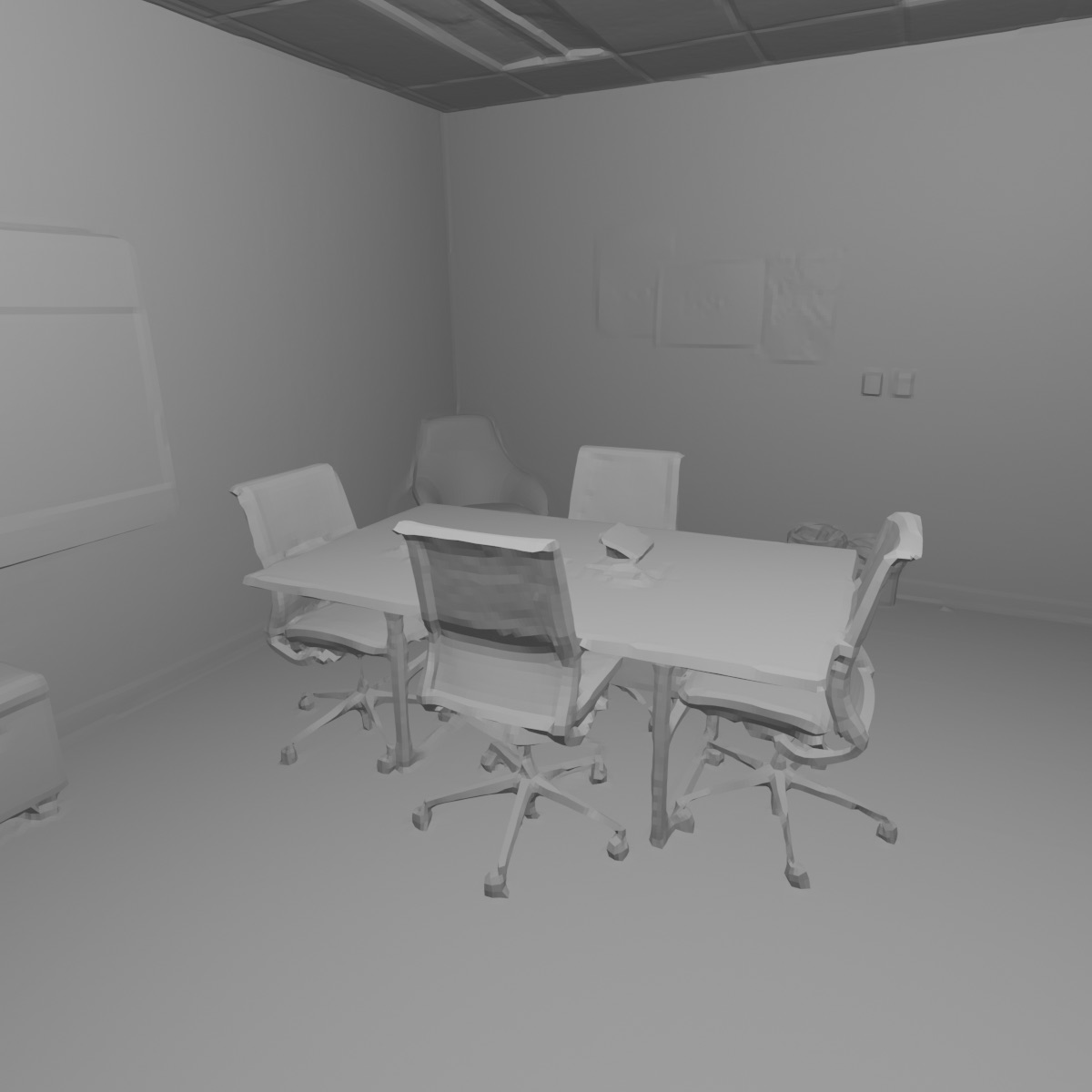}} \\
  \end{tabular}
  \vspace{-2.5mm}
  \caption{Reconstruction results on Replica~\cite{straubReplicaDatasetDigital2019} dataset. In comparison to our baselines, our methods achieve accurate and high-quality scene reconstruction and completion on various scenes.}
  \label{fig:replica_topdown_zoom_in}
  \vspace{-5pt}
\end{figure*}
\begin{figure*}[htbp]
  \centering
  \footnotesize
  \setlength{\tabcolsep}{1.5pt}
  \newcommand{\sz}{0.235}
  \begin{tabular}{lcccc}
    & iMAP$^*$~\cite{sucarIMAPImplicitMapping2021a} & NICE-SLAM~\cite{zhuNiceslamNeuralImplicit2022} & Ours & ScanNet Mesh \\
    \makecell{\rotatebox{90}{\tt scene0000}} &
    \makecell{\includegraphics[width=\sz\linewidth]{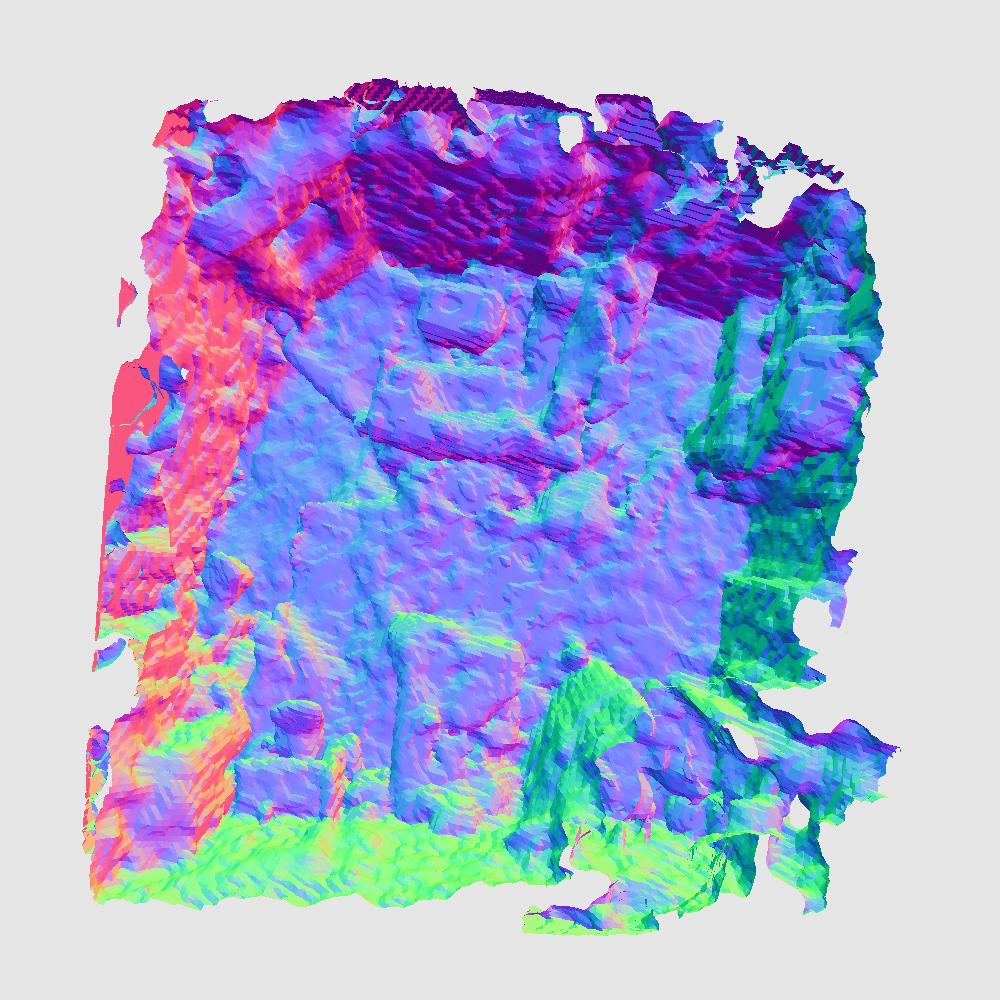}} &
    \makecell{\includegraphics[width=\sz\linewidth]{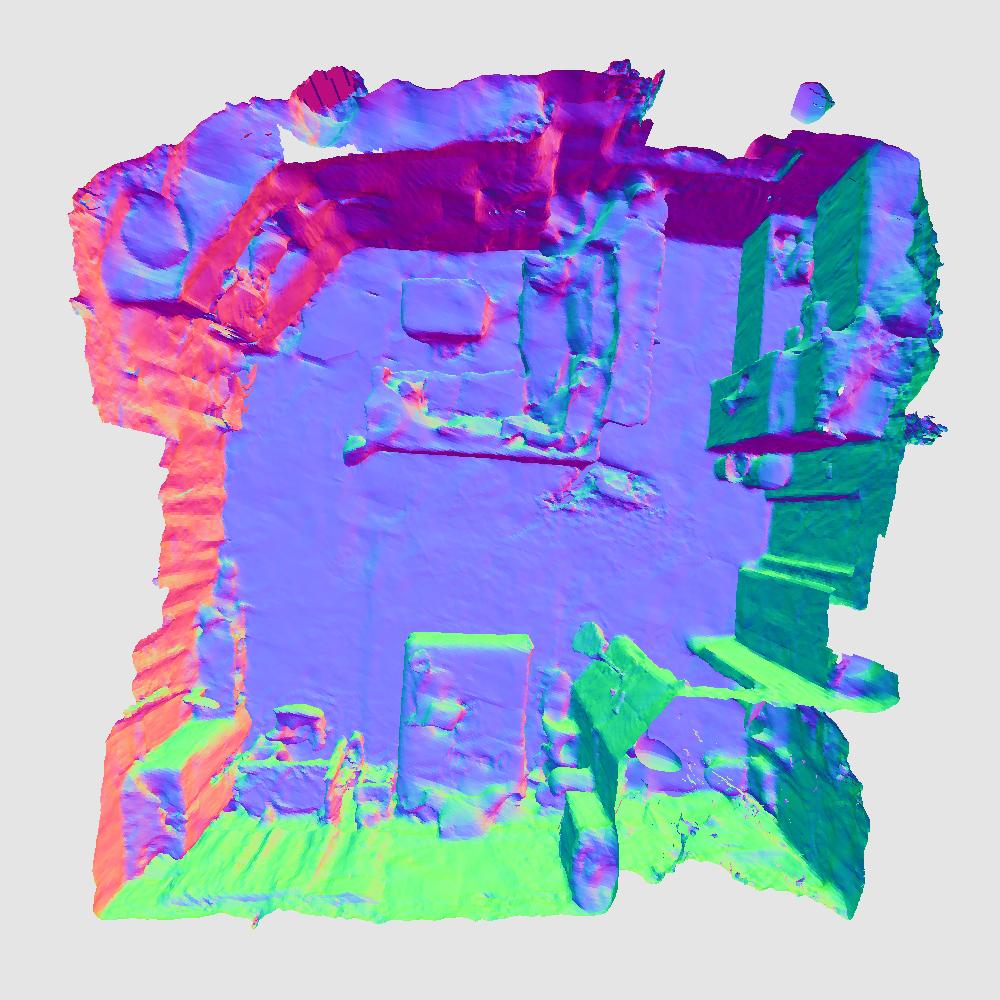}} &
    \makecell{\includegraphics[width=\sz\linewidth]{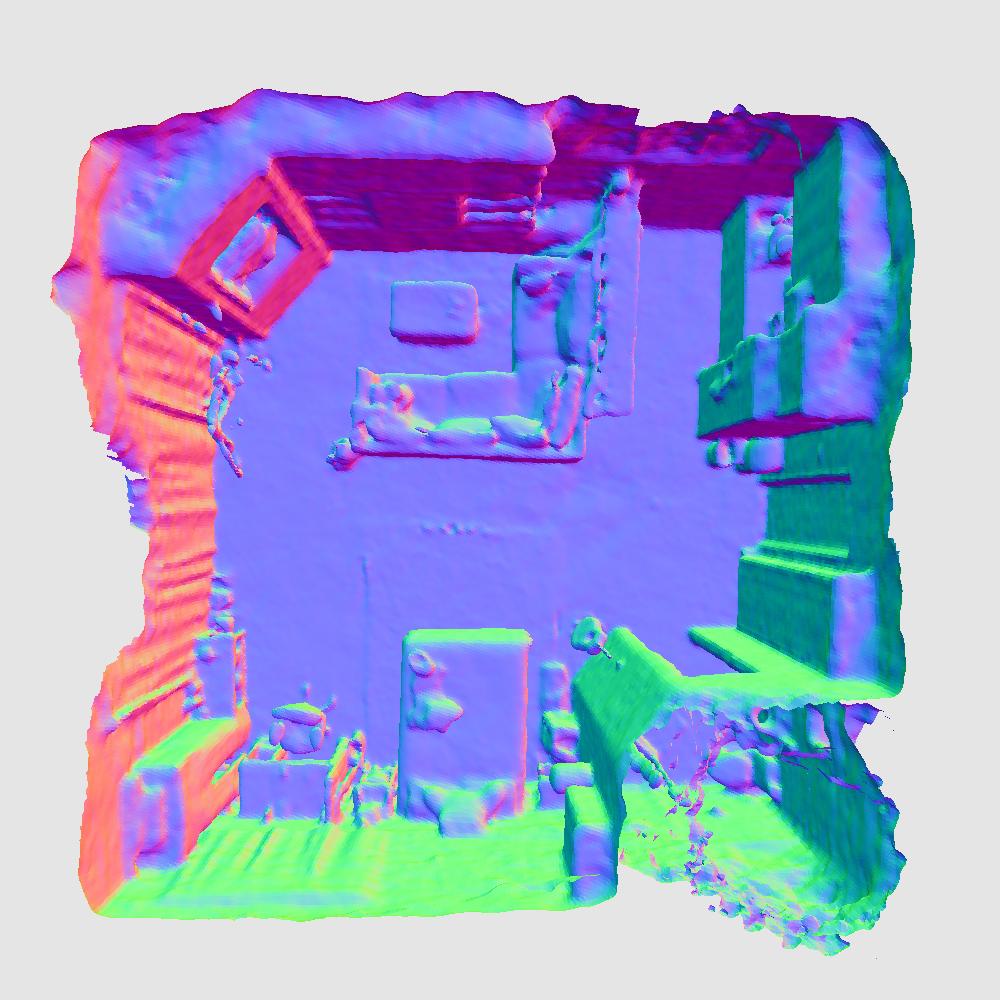}} &
    \makecell{\includegraphics[width=\sz\linewidth]{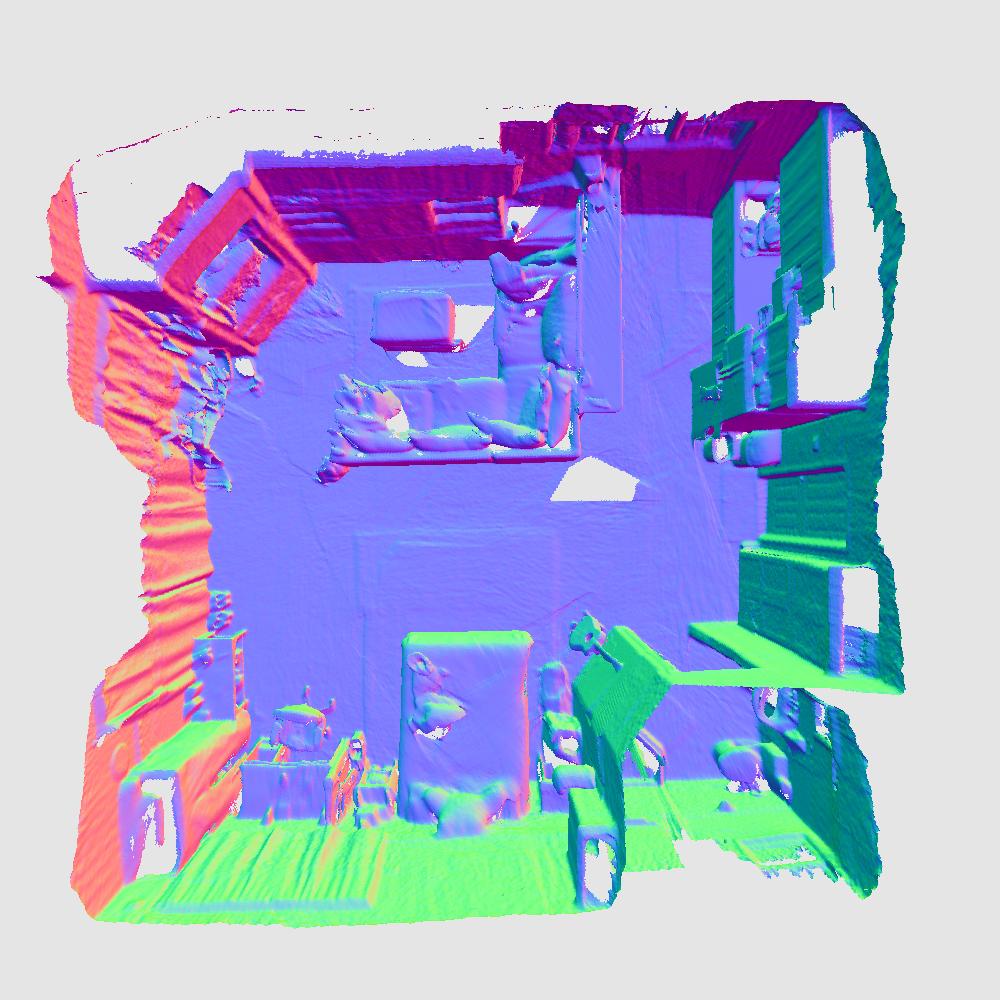}} \\
    \makecell{\rotatebox{90}{\tt scene0106}} &
    \makecell{\includegraphics[width=\sz\linewidth]{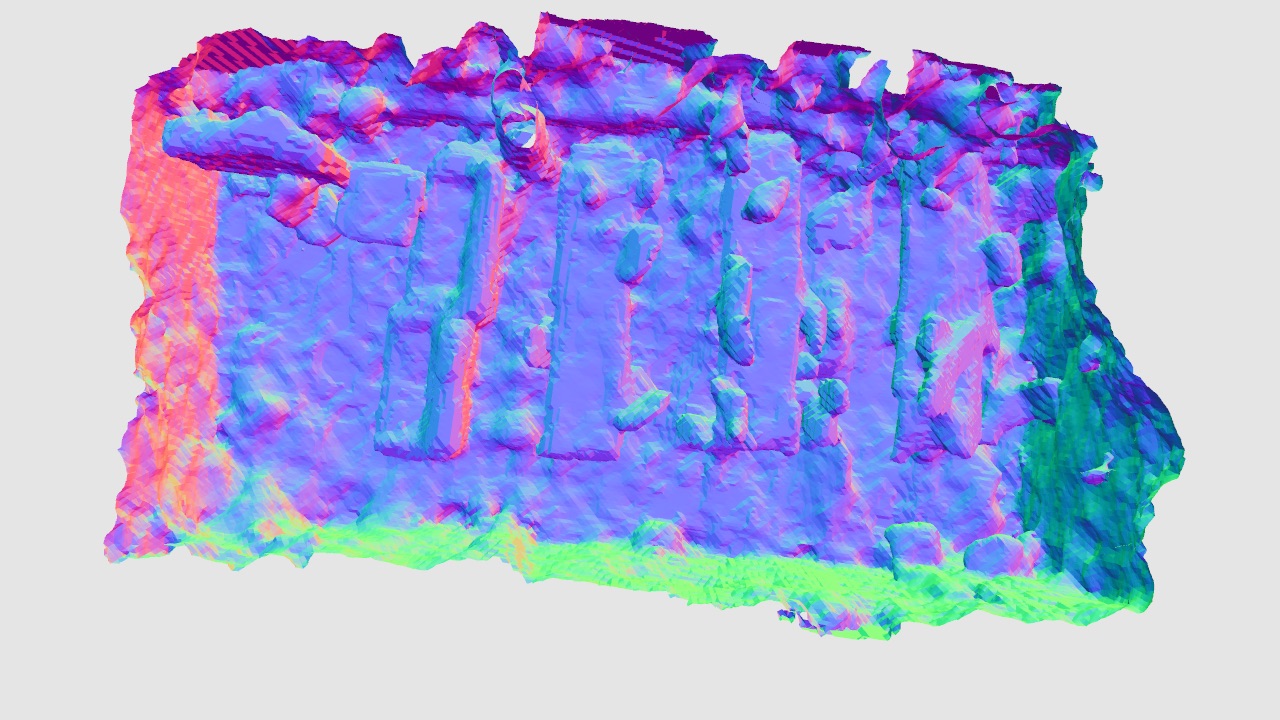}} &
    \makecell{\includegraphics[width=\sz\linewidth]{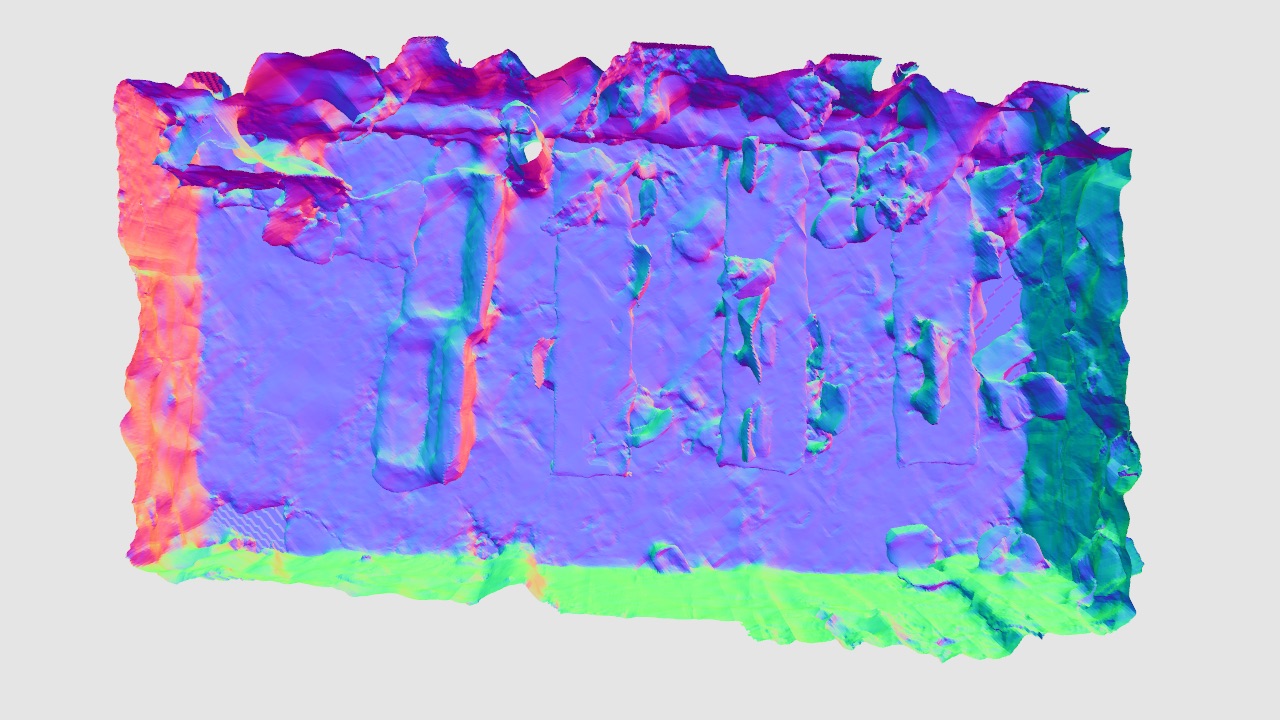}} &
    \makecell{\includegraphics[width=\sz\linewidth]{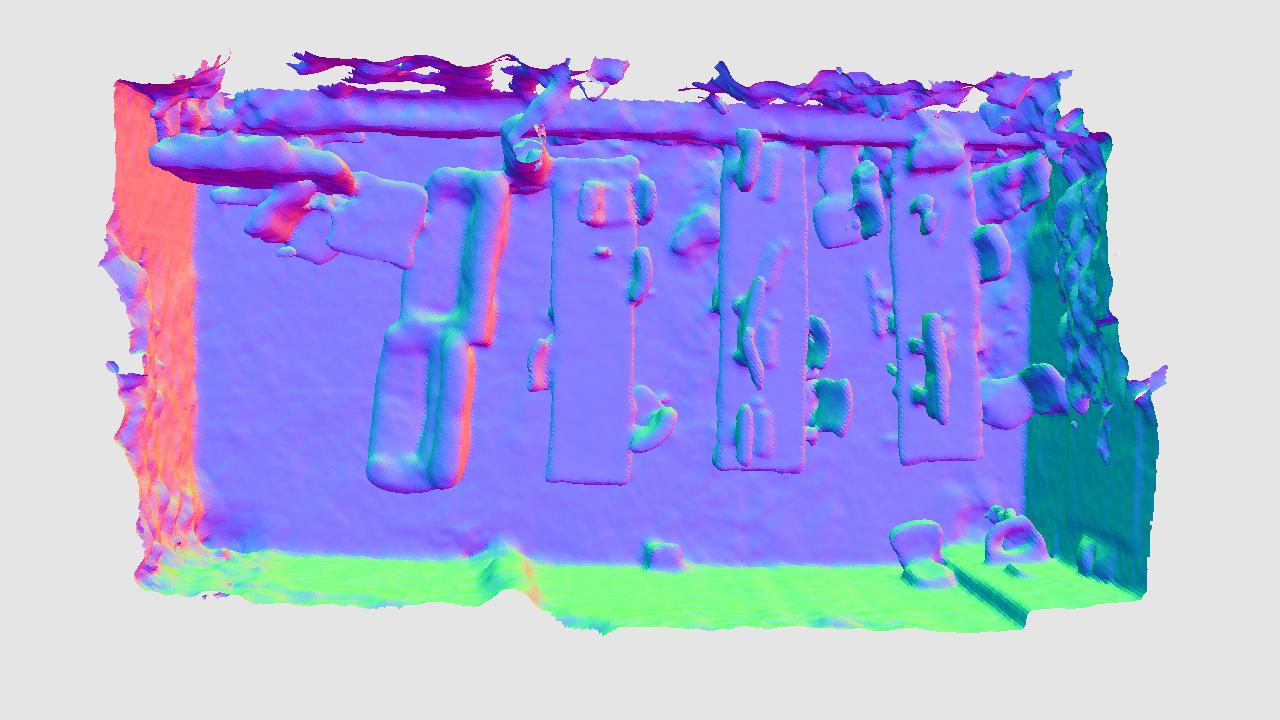}} &
    \makecell{\includegraphics[width=\sz\linewidth]{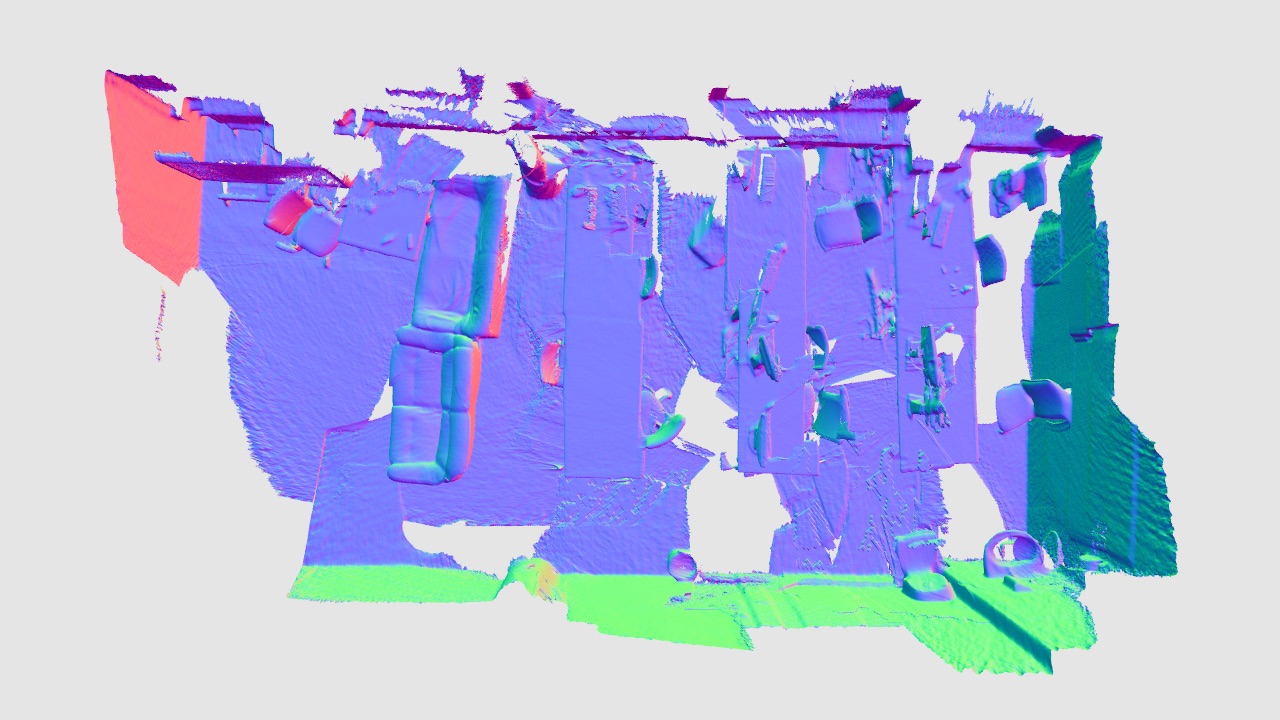}} \\
    \makecell{\rotatebox{90}{\tt scene0059}} &
    \makecell{\includegraphics[width=\sz\linewidth]{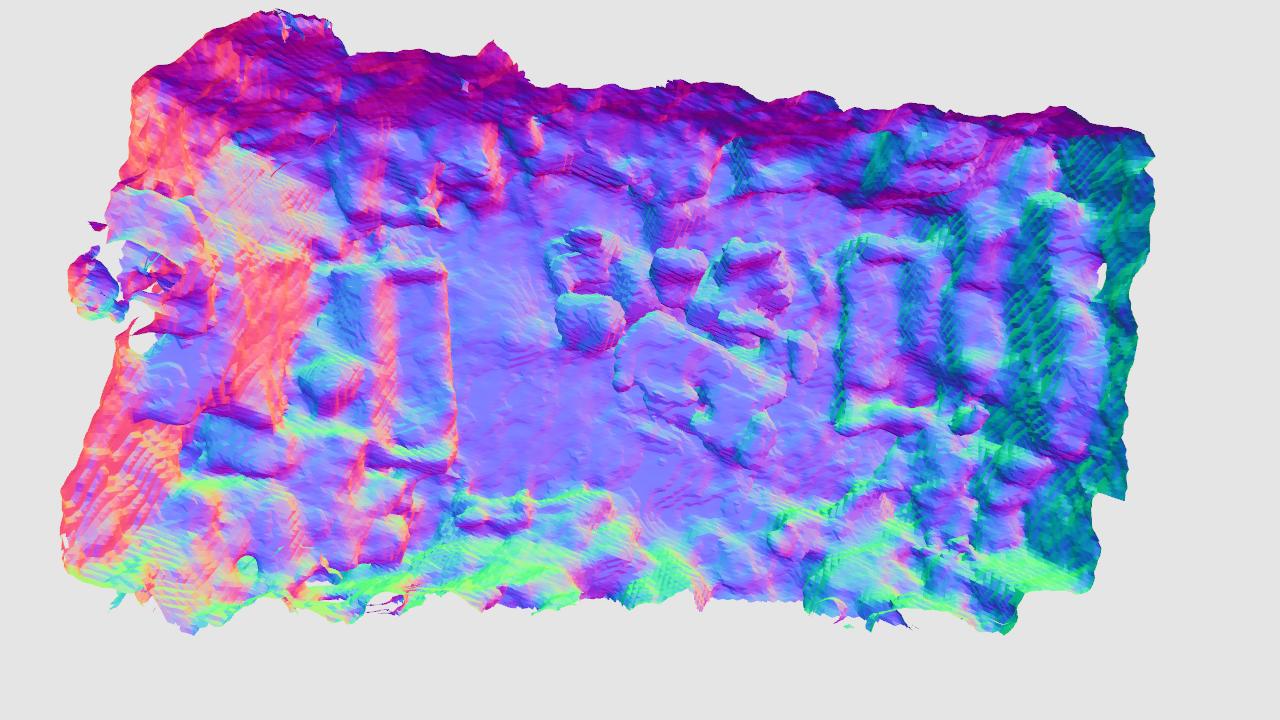}} &
    \makecell{\includegraphics[width=\sz\linewidth]{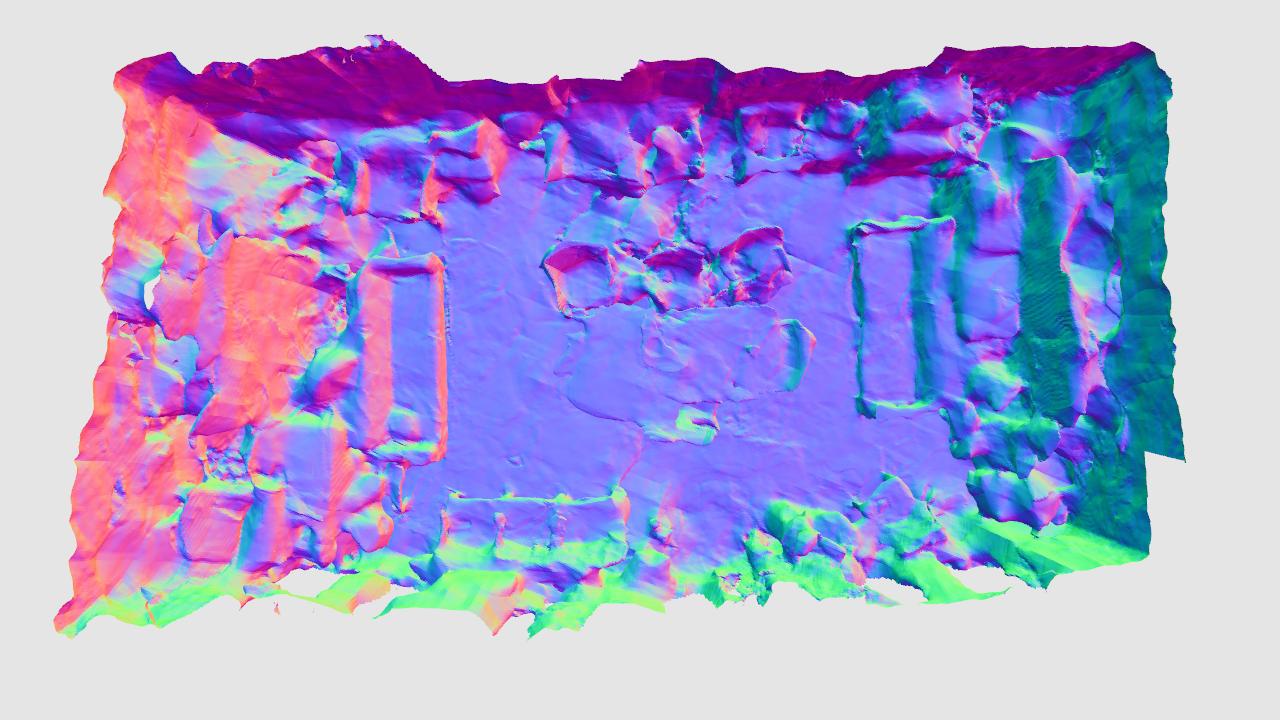}} &
    \makecell{\includegraphics[width=\sz\linewidth]{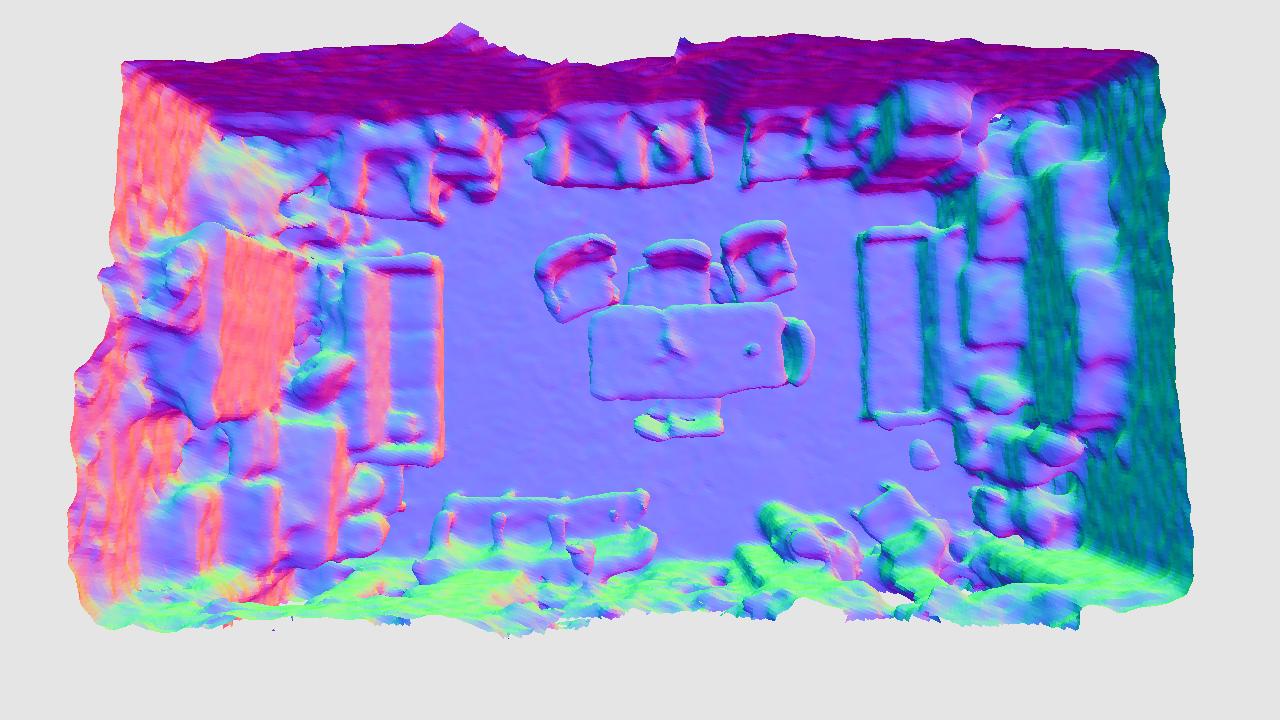}} &
    \makecell{\includegraphics[width=\sz\linewidth]{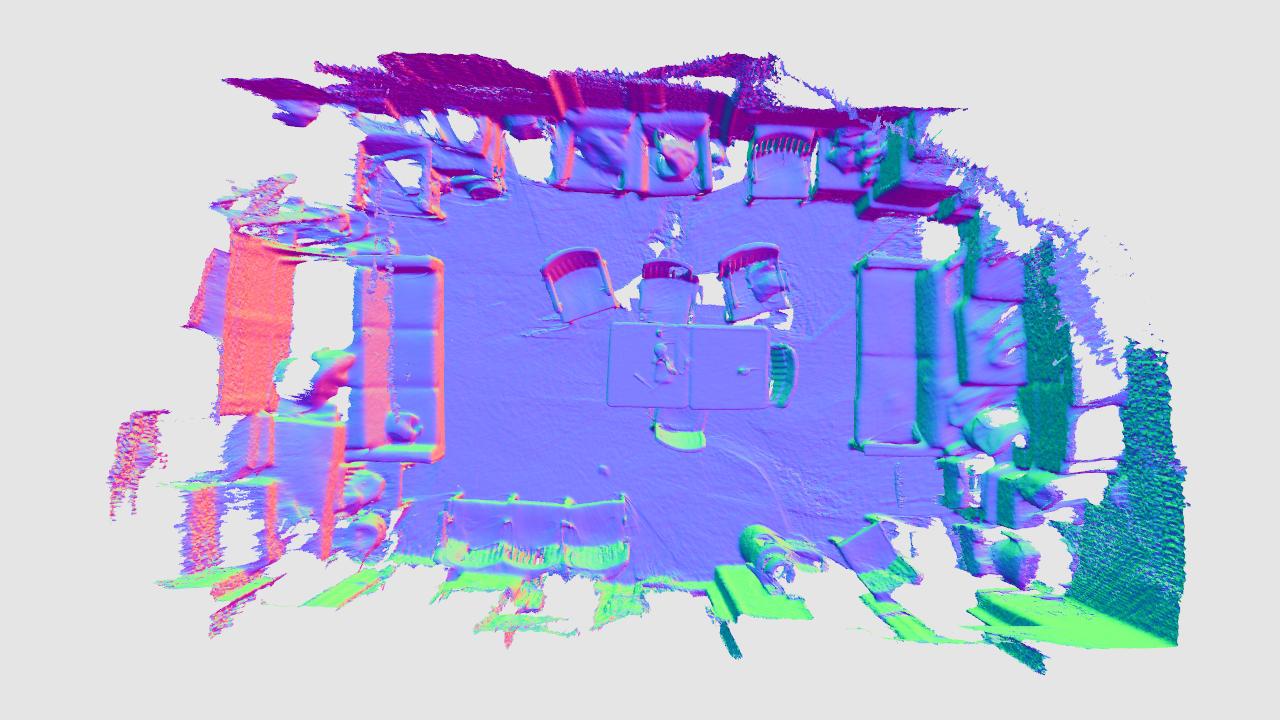}} \\
  \end{tabular} 
  \vspace{-2.5mm}
  \caption{Reconstruction results on ScanNet~\cite{daiScannetRichlyannotated3d2017}. In comparison to the previous methods, our reconstructions are smoother and contain more details thanks to our proposed joint encoding and global bundle adjustment strategy.}
  \label{fig:scannet}
  \vspace{-14pt}
\end{figure*}

\begin{table}[tp]
    \centering
 	\resizebox{1.00\columnwidth}{!}{
    \begin{tabular}{lccccccc}
        \toprule
         Scene ID & 0000 & 0059 & 0106 &  0169 &  0181 & 0207 &Avg. \\
        \midrule
        
        iMAP$^{\star}$~\cite{sucarIMAPImplicitMapping2021a} & 55.95 & 32.06 & 17.50 & 70.51 & 32.10 & 11.91 & 36.67\\
        NICE-SLAM~\cite{zhuNiceslamNeuralImplicit2022} & 8.64 & 12.25 & \textbf{8.09} & 10.28 & 12.93 & \textbf{5.59} & 9.63\\
        Ours$^{\dagger}$ & \textbf{7.13} & \textbf{11.14} & 9.36 & \textbf{5.90} & \textbf{11.81} & 7.14 & \textbf{8.75} \\
        Ours & 7.18 & 12.29 & 9.57 & 6.62 & 13.43 & 7.13 & 9.37 \\
        
        \bottomrule
    \end{tabular}
     }
     \vspace{-1mm}
    \caption{ATE RMSE (cm) results of an average of 5 runs on ScanNet. Co-SLAM achieves better or on-par performance compared to NICE-SLAM~\cite{zhuNiceslamNeuralImplicit2022} with significantly faster optimization speed. }
    \label{tab:ab_memory}
\end{table}
\begin{table}[tp]
    \centering
 	\resizebox{1.00\columnwidth}{!}{
    \begin{tabular}{lccc}
        \toprule
        & fr1/desk (cm) & fr2/xyz (cm) & fr3/office (cm) \\
        
        \midrule
        
        iMAP~\cite{sucarIMAPImplicitMapping2021a} & 4.9 & 2.0 & 5.8 \\
        iMAP$^{\star}$~\cite{sucarIMAPImplicitMapping2021a} & 7.2 & 2.1 & 9.0 \\
        NICE-SLAM~\cite{zhuNiceslamNeuralImplicit2022} & 2.7 & 1.8 & 3.0 \\
        Ours$^{\dagger}$ & \textbf{2.4} & \textbf{1.7} & \textbf{2.4} \\
        Ours & 2.7 & 1.9 & 2.6 \\

        \midrule
        
        BAD-SLAM~\cite{schopsBadSlamBundle2019a} & 1.7 & 1.1 & 1.7 \\
        Kintinuous~\cite{whelanKintinuousSpatiallyExtended2012} & 3.7 & 2.9 & 3.0 \\
        ORB-SLAM2~\cite{mur-artalOrbslam2OpensourceSlam2017} & \textbf{1.6} & \textbf{0.4} & \textbf{1.0} \\

        \bottomrule
    \end{tabular}
    }
    \vspace{-2mm}
    \caption{ATE RSME (cm) results on TUM RGB-D dataset. Our method achieves the best tracking performance among neural SLAM methods and maintains high-fidelity reconstruction.}
    \label{tab:pose_tum}
    \vspace{-12pt}
\end{table}

\subsection{Evaluation of Tracking and Reconstruction}

\noindent
\textbf{Replica dataset~\cite{straubReplicaDatasetDigital2019}.} We evaluate on the same simulated RGB-D sequences as iMAP~\cite{sucarIMAPImplicitMapping2021a}. As shown in Tab.~\ref{tab:time_memory_performance}, our method achieves higher reconstruction accuracy and faster speed. Fig.~\ref{fig:replica_topdown_zoom_in} shows the qualitative results, from which we can observe that iMAP achieves plausible completion in unobserved areas but results are over-smoothed, while NICE-SLAM maintains more reconstruction details, but results contain some artifacts (e.g. the floors beside the bed, the back of the chairs). \methodname successfully retains the advantages of both methods achieving consistent completion as well as high-fidelity reconstruction results.

\noindent
\textbf{Synthetic dataset~\cite{azinovicNeuralRGBDSurface2022c}.} We perform further experiments on the synthetic dataset from NeuralRGBD~\cite{azinovicNeuralRGBDSurface2022c}. Unlike the Replica dataset, it contains many thin structures and simulates the noise present in real depth sensor measurements. Quantitatively, our method significantly outperforms baseline methods (see Tab.~\ref{tab:time_memory_performance}) while still operating in real time (15 FPS). Fig.~\ref{fig:RGBD} shows some example qualitative results. Overall, \methodname can capture fine details (e.g. wine bottles, chair legs, etc.) and produces complete and smooth reconstructions. NICE-SLAM yields less detailed and noisier reconstructions and cannot perform hole filling, while iMAP$^*$  lost track on some occasions.

\noindent
\textbf{ScanNet dataset~\cite{daiScannetRichlyannotated3d2017}.} We evaluate the camera tracking accuracy of \methodname on 6 real-world sequences from ScanNet. The absolute trajectory error (ATE) is obtained by comparing predicted and ground-truth (generated by BundleFusion~\cite{daiBundlefusionRealtimeGlobally2017a}) trajectories. Tab.~\ref{tab:ab_memory} shows that quantitatively, our method achieves better tracking results in comparison to NICE-SLAM~\cite{zhuNiceslamNeuralImplicit2022} with fewer tracking and mapping iterations while running at $6-12$~Hz (see Tab.~\ref{tab:time_memory_old}). Fig.~\ref{fig:scannet} also shows \methodname achieves better reconstruction quality with smoother results and finer details (e.g. bike).

\noindent
\textbf{TUM dataset~\cite{sturmBenchmarkEvaluationRGBD2012}.} We further evaluate the tracking accuracy on the TUM dataset~\cite{sturmBenchmarkEvaluationRGBD2012}. As shown in Tab.~\ref{tab:pose_tum}, our method achieves competitive tracking performance at $13$ FPS. By increasing the number of tracking iterations (Ours$^{\dagger}$), our method achieves the best tracking performance among neural SLAM methods, though at the expense of the FPS dropping to $6.7$ (see Tab.~\ref{tab:time_memory_old}). Although Co-SLAM still cannot outperform classic SLAM methods, it reduces the tracking performance gap between neural and classic methods, while improving the fidelity and completeness of the reconstructions. 

\subsection{Performance Analysis}

\begin{figure}[t]
    \centering
    \includegraphics[width=0.90\columnwidth]{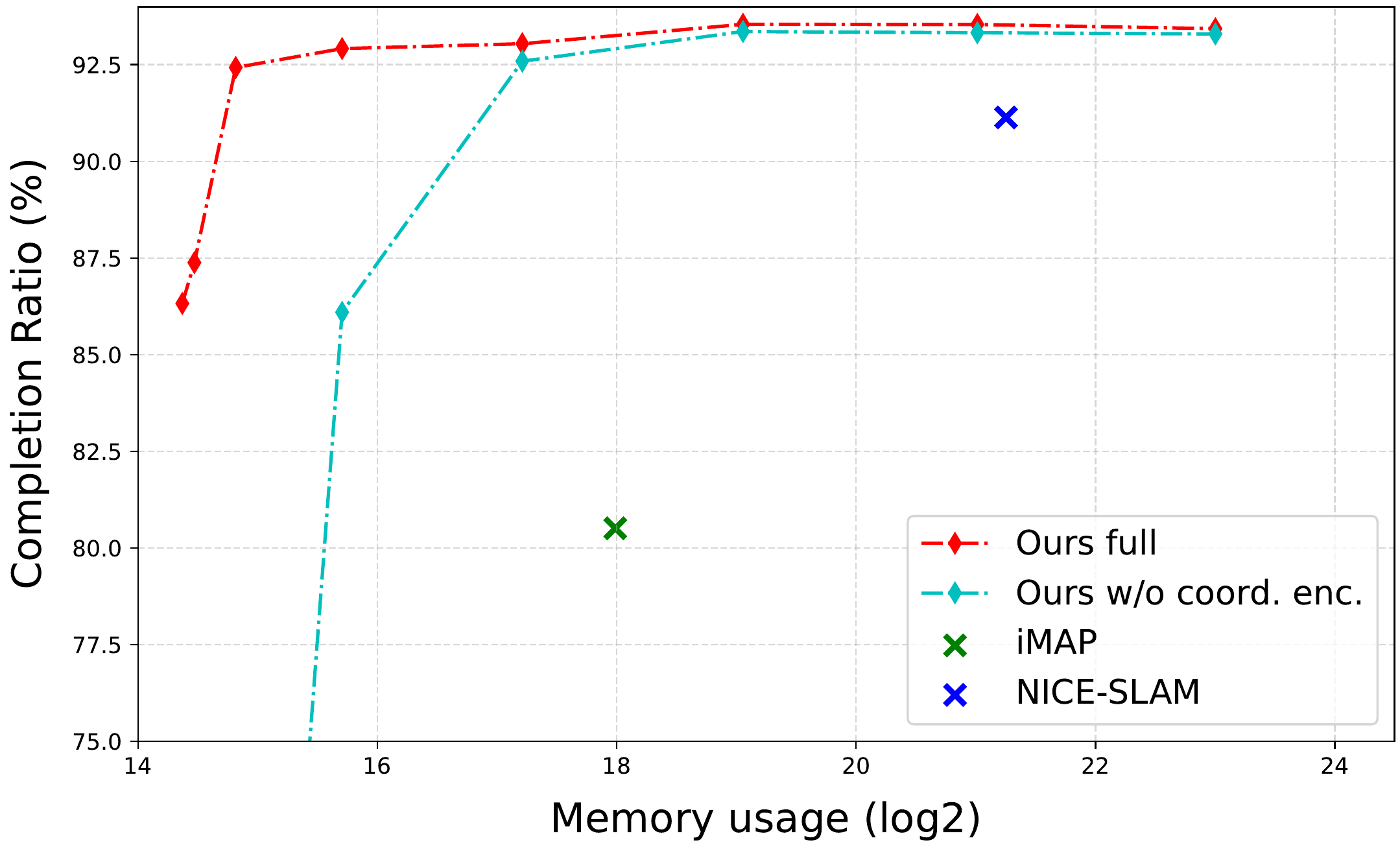}
    \vspace{-6pt}
    \caption{Completion ratio vs. model size for Co-SLAM w/ and w/o coordinate encoding. Each model corresponds to a different hash-table size. iMAP~\cite{sucarIMAPImplicitMapping2021a} and NICE-SLAM~\cite{zhuNiceslamNeuralImplicit2022} shown for reference.}
    \label{fig:memory}
    \vspace{-4pt}
\end{figure}

\noindent{\textbf{Run time and memory analysis.}}
In our default setting (Ours), \methodname can operate above $15$Hz on a desktop PC with a 3.60GHz Intel Core i7-12700K CPU and NVIDIA RTX 3090ti GPU. For more challenging scenarios such as the ScanNet and TUM datasets, Co-SLAM still achieves $5-13$Hz runtime.  
Fig.~\ref{fig:memory} shows reconstruction quality with respect to memory use. Thanks to the sparse parametric encoding, our method requires significantly less memory than NICE-SLAM~\cite{zhuNiceslamNeuralImplicit2022} while operating in real time and achieving accurate reconstruction results. Surprisingly, we found that further compressing the memory footprint (increasing the chances of hash collisions), Co-SLAM still outperforms  iMAP~\cite{sucarIMAPImplicitMapping2021a}, suggesting that our joint encoding improves the representation power of single encoding. Note that this figure is for illustration purposes, so we use the same spatial resolution throughout our hash encoding. Ideally, one can reduce the spatial resolution further to minimize hash collisions and achieve a better reconstruction quality.

\noindent{\textbf{Scene completion.}}
Fig.~\ref{fig:encoding_showcase} shows an illustration of hole filling using different encoding strategies on a small scene. Coordinate encoding-based methods achieve plausible completion at the cost of lengthy training times, while parametric encoding-based methods fail at hole-filling due to their local nature. By applying our new joint encoding, we observe that smooth hole filling can be achieved and fine structures are preserved by Co-SLAM.

\begin{table}[tp]
    \centering
    \resizebox{0.95\columnwidth}{!}{
    \begin{tabular}{lccc}
        \toprule
          & w/o hash enc. & w/o one-blob enc. & Full model\\
        \midrule
        {\bf Acc. (cm.)} $\downarrow$ & 3.69 & 2.13 & \textbf{2.10}\\
        {\bf Comp. (cm.)} $\downarrow$ & 3.43 & 2.13 & \textbf{2.08}\\
        {\bf Comp. Ratio} $\uparrow$ & 82.49 & 93.17 & \textbf{93.44}\\
        \bottomrule
    \end{tabular}
    }
    \vspace{-5pt}
    \caption{Ablation study on different encodings. Default hash-table size is $13$. Our full model with joint encoding achieves better completion and more accurate reconstructions. See also Fig.~\ref{fig:encoding_showcase}.}
    \vspace{-12pt}
    \label{tab:ablation_enc}
\end{table}
\begin{table}[t]
\centering
\resizebox{1.0\columnwidth}{!}{
\begin{tabular}{l cc ccc c c c}
\toprule
\multirow{2}{*}{Name} & \multicolumn{2}{c}{KF selection} & \multicolumn{3}{c}{\#KF} & \multirow{2}{*}{\begin{tabular}[c]{@{}c@{}}Pose \\ optim.\end{tabular}} & \multirow{2}{*}{ATE (cm)} & \multirow{2}{*}{Std. (cm)} \\
\cmidrule(lr){2-3} \cmidrule(lr){4-6}
& Local & Global & 0 & 10 & All & & & \\
\midrule
w/o BA & & & \checkmark & & & & 16.81 & 1.69\\
LBA&\checkmark& & & \checkmark & & \checkmark & 9.69 & 1.38\\

GBA-10&& \checkmark& & \checkmark & & \checkmark & 9.54 & 0.67\\
GBA&& \checkmark& &  & \checkmark& \checkmark& \textbf{8.75} & \textbf{0.33}\\
\bottomrule
\end{tabular}}
\vspace{-5pt}
\caption{Ablation of BA strategies on Co-SLAM: (LBA) BA with rays from 10 local keyframes; (GBA-10) BA with rays from 10 keyframes randomly selected from all keyframes; 
(GBA) BA with rays from all keyframes (our full method). All methods sample the same number of total rays per iteration (2048).}
\vspace{-10pt}
\label{tab:ablation_gba}
\end{table}

\subsection{Ablations}

\noindent{\textbf{Effect of joint coordinate and parametric encoding.}}
Tab.~\ref{tab:ablation_enc} illustrates a quantitative evaluation using different encodings. Our full model leads to higher accuracy and better completion than using single encodings (only one-blob or only hash-encoding). 
In addition, Fig.~\ref{fig:memory} illustrates that when compressing the size of the hash lookup table, our full model with joint coordinate and parametric encoding is more robust in comparison to using a hash-based feature grid without the coordinate encoding. 

\noindent{\textbf{Effect of global bundle adjustment.}}
Tab.~\ref{tab:ablation_gba} shows the average ATE of our SLAM method on the 6 ScanNet scenes using different BA strategies: (w/o BA) pure tracking; (LBA) BA with rays from 10 local keyframes, a similar strategy to NICE-SLAM; (GBA-10) BA using rays from only 10 keyframes randomly selected from all past keyframes; (GBA) denotes the global BA strategy of Co-SLAM.
We observe that using rays from a small (10) number of keyframes (LBA and GBA-10) leads to higher ATE errors. However, when keyframes are chosen from the entire sequence (GBA-10), instead of locally (LBA) the standard deviation is greatly reduced. Sampling rays from all keyframes (GBA) is the best overall strategy, even when all methods sample the same number of total rays (2048).
\section{Conclusion}

We presented \methodname, a dense real-time neural RGB-D SLAM system. We show that using a joint coordinate and parametric encoding with tiny MLPs as scene representation and training it with global bundle adjustment,  achieves high-fidelity mapping and accurate tracking with plausible hole filling and efficient memory use. 

\noindent
\textbf{Limitations.} Co-SLAM relies on inputs from an RGB-D sensor and is therefore sensitive to illumination changes and inaccurate depth measurements. Instead of sampling keyframe pixels randomly, an information-guided pixel sampling strategy could be helpful to further reduce the number of pixels and improve the tracking speed. 
Incorporating loop closure could lead to further improvements.

\section*{Acknowledgements}  
The research presented here has been supported by a sponsored research award from Cisco Research and the UCL Centre for Doctoral Training in Foundational AI under UKRI grant number EP/S021566/1. This project made use of time on Tier 2 HPC facility JADE2, funded by
EPSRC (EP/T022205/1).

{\small
\bibliographystyle{ieee_fullname}
\bibliography{egbib}
}

\end{document}


\title{Supplementary Material\\\papername: Joint Coordinate and Sparse Parametric Encodings for \\Neural Real-Time SLAM}  

\author{Hengyi Wang$^{\star}$ \quad Jingwen Wang$^{\star}$ \quad Lourdes Agapito\\
Department of Computer Science, University College London \\
{\tt\small \{hengyi.wang.21, jingwen.wang.17, l.agapito\}@ucl.ac.uk}
\vspace{-3mm}
}

\maketitle

\let\thefootnote\relax\footnotetext{$\star$ Indicates equal contribution.}

\section{Implementation details}

\subsection{Hyperparameters}
Here we report the detailed settings and hyper-parameters used in \methodname to achieve high-quality reconstruction and quasi-realtime performance. 

\noindent \textbf{Default setting.} For camera tracking, we select $N_t=1024$ pixels and perform $10$ iterations of tracking with $M_c=32$ regular sampling and $M_f=11$ depth-guided sampling for each camera ray. In terms of mapping and bundle adjustments, we select $N_g=2048$ pixels and use $200$ iterations for first frame mapping, $10$ iterations for bundle adjustment every $5$ frames. For scene representations, we use $L=16$ level HashGrid with from $R_{min}=16$ to $R_{max}$, where we use max voxel size $2$cm for determining $R_{max}$, and $16$ bins for OneBlob encoding of each dimension. Two $2$-layer shallow MLPs with $32$ hidden units are used to decode color and SDF. The dimension of the geometric feature $\vect{h}$ is $15$. For the training of our scene representation, we use learning rate of $1e{-3}$ for tracking and $1e{-2}$, $1e{-2}$, $1e{-3}$ for feature grid, decoder, and camera parameters during bundle adjustment. The weights of each loss are $\lambda_{rgb}=5$, $\lambda_{d}=0.1$, $\lambda_{sdf}=1000$, $\lambda_{fs}=10$, and $\lambda_{smooth}=1e{-6}$. The truncation distance $tr$ is set to $10\text{cm}$.

\noindent
\textbf{ScanNet dataset.} For ScanNet dataset, we change the voxel size to $4cm$ for the finest resolution, and increase the number of sample points to $M_c=96$, $M_f=21$. The $\lambda_{smooth}$ is increased to $1e-3$.

\noindent
\textbf{TUM dataset.}
Since scenes in TUM dataset is mostly focusing on reconstructions of tables instead of the whole room, we use 20 iterations for bundle adjustment, and set $tr$ to $5$cm. The weights of each loss are $\lambda_{rgb}=1.0$, $\lambda_{d}=0.1$, $\lambda_{sdf}=5000$, $\lambda_{fs}=10$, and $\lambda_{smooth}=1e{-8}$. The learning rate of camera parameters in tracking process is increased to $1e-2$.

\begin{figure}[tbp]
  \centering
  \scriptsize
  \setlength{\tabcolsep}{0.5pt}
  \newcommand{\sz}{0.32}  %

  \begin{tabular}{lccc}
    & frustum & frustum+occlusion & Ours \\
    \makecell{\rotatebox{90}{\tt view-1}}  &
    \makecell{\includegraphics[width=\sz\linewidth]{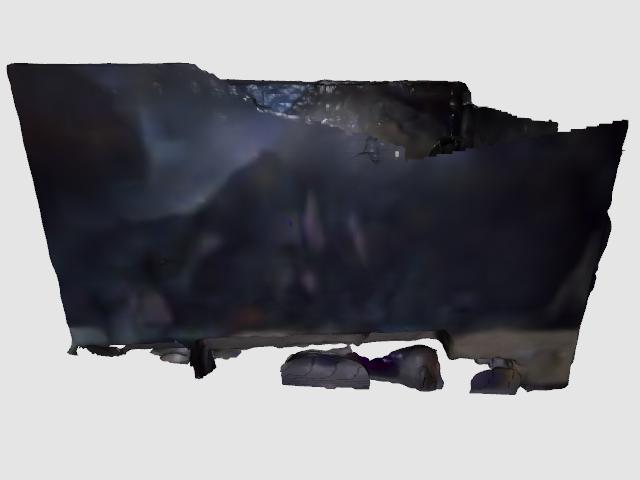}} & 
    \makecell{\includegraphics[width=\sz\linewidth]{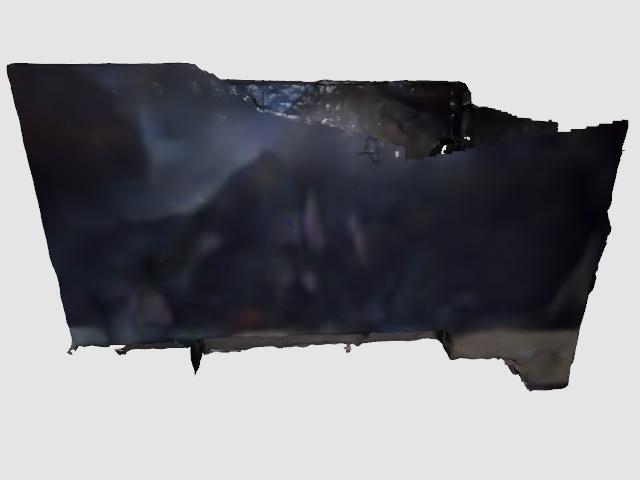}} &
    \makecell{\includegraphics[width=\sz\linewidth]{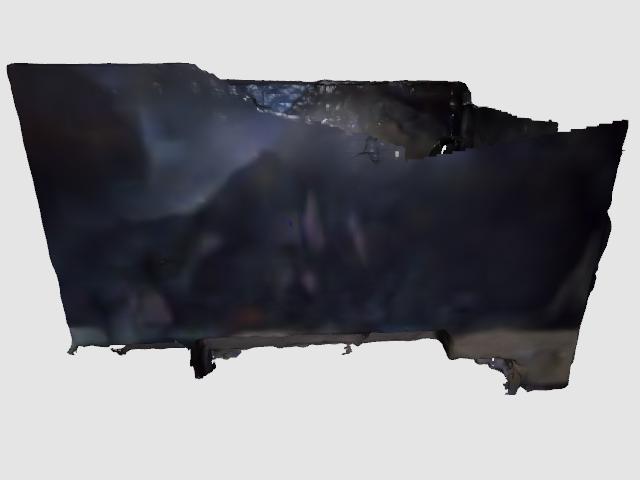}} \\
    \makecell{\rotatebox{90}{\tt view-2}}  &
    \makecell{\includegraphics[width=\sz\linewidth]{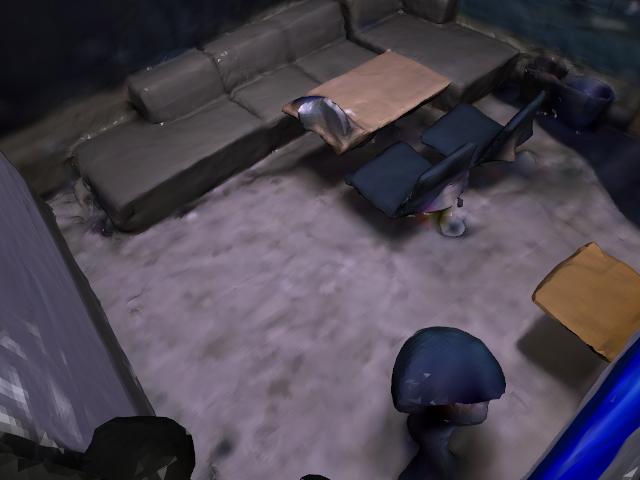}} & 
    \makecell{\includegraphics[width=\sz\linewidth]{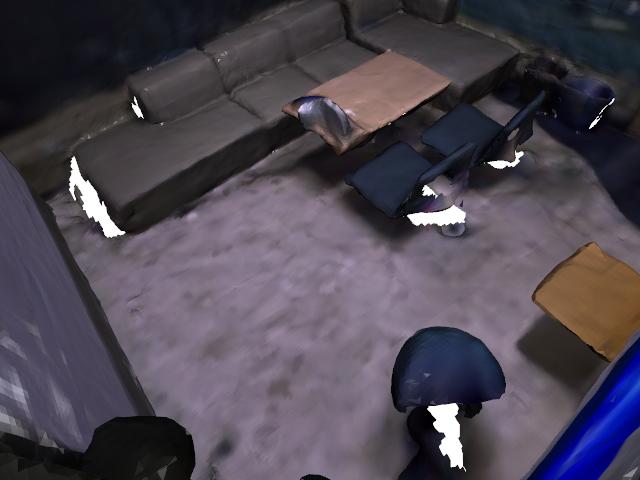}} &
    \makecell{\includegraphics[width=\sz\linewidth]{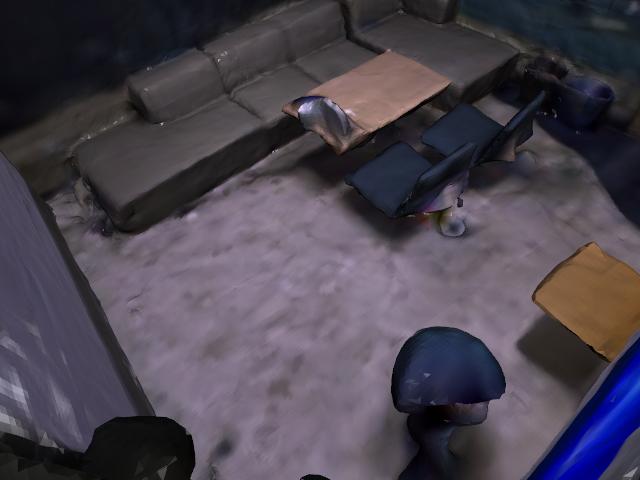}} \\
    \makecell{\rotatebox{90}{\tt view-3}}  &
    \makecell{\includegraphics[width=\sz\linewidth]{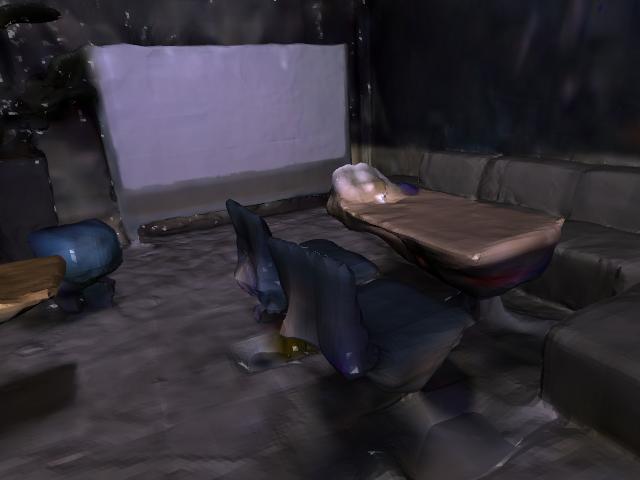}} & 
    \makecell{\includegraphics[width=\sz\linewidth]{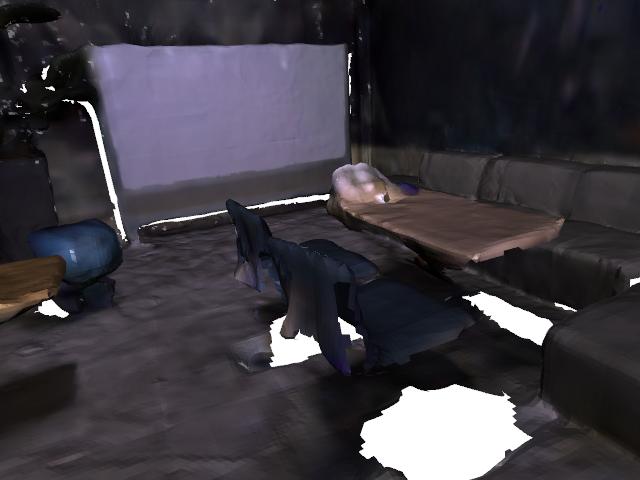}} &
    \makecell{\includegraphics[width=\sz\linewidth]{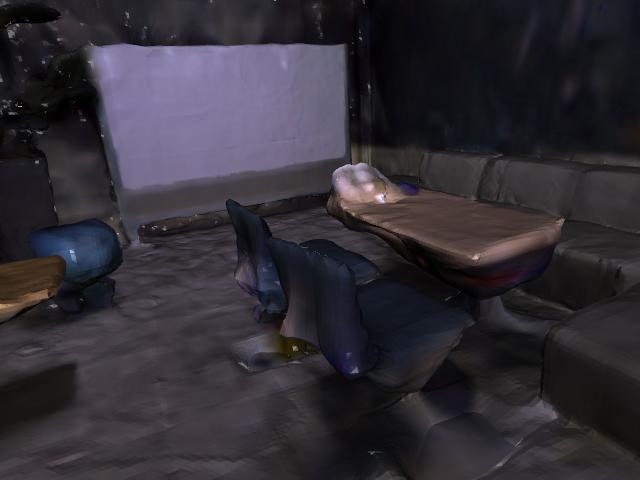}} \\
  \end{tabular}
  \caption{Visualization of different culling strategy applied on mesh reconstructed by NICE-SLAM~\cite{zhuNiceslamNeuralImplicit2022}. Culling by frustum fails to remove the artefacts outside the scene bound (\texttt{view-1}), while culling by frustum+occlusion removes occluded regions (\texttt{view-2} and \texttt{view-3}) inside the room. Our proposed method could remove unwanted artefacts outside the room but preserve the completeness inside the room.}
  \vspace{-5mm}
  \label{fig:culling_nice-slam}
\end{figure}

\subsection{Evaluation Protocol}
We have introduced a modification to the culling strategy used for the quantitative evaluation of the reconstruction accuracy, which we believe leads to a fairer comparison. In this section we describe this new culling strategy in detail and provide a justification for its use.

In the context of neural implicit reconstruction and SLAM~\cite{azinovicNeuralRGBDSurface2022c, wangGosurfNeuralFeature2022, sucarIMAPImplicitMapping2021a, zhuNiceslamNeuralImplicit2022}, due to the extrapolation ability of neural networks an extra mesh culling step is required before evaluating the reconstructed mesh. We show a demonstration in Fig.~\ref{fig:culling_nice-slam}. In previous works two different culling strategies are used: NICE-SLAM~\cite{zhuNiceslamNeuralImplicit2022} and iMAP~\cite{sucarIMAPImplicitMapping2021a} adopted a \textit{culling-by-frustum} strategy where mesh vertices outside any of the camera frustums are removed. This simple strategy works effectively well but cannot remove artifacts that are inside camera frustums but outside of scene bound, such as in \texttt{view-1}. In NeuralRGBD~\cite{azinovicNeuralRGBDSurface2022c} and GO-Surf~\cite{wangGosurfNeuralFeature2022} the \textit{frustum+occlusion} strategy is used, where in addition to the frustum criteria self-occlusion is also considered by comparing the rendered depth. While this strategy could effectively remove the artifacts in \texttt{view-1} it also removes the occluded regions as in \texttt{view-2} and \texttt{view-3}. Therefore, we propose a new culling strategy that follows the \textit{frustum+occlusion} criteria but also simulates virtual camera views that cover the occluded regions. Since we focus on the inner surface of the scene in Replica dataset~\cite{straubReplicaDatasetDigital2019}, we can remove the noisy points outside the mesh of interest to make fair comparison of different methods.

We also show evaluation of reconstruction quality using nice-slam culling strategy in Tab.~\ref{tab:recon_replica_nice}. To remove the noisy points outside the outer surface caused by hash collision, we increase the smooth weight to $\lambda_{smooth}=1e-3$ and increase the default hash lookup table size from 13 to 14.

\subsection{Evaluation Metrics} 
After mesh culling we evaluate the reconstructed mesh with a mixture of 3D (\textbf{Acc}uracy, \textbf{Comp}letion and \textbf{Comp}letion \textbf{Ratio}) and 2D (\textbf{Depth L1}) metrics.
We first uniformly sample two point clouds $P$ and $Q$ from both GT and reconstructed meshes, with $|P| = |Q| = 200000$. Then accuracy is defined as the average distance between a point on GT mesh to its nearest point on reconstructed mesh, other metrics are defined in the same fashion, see Tab.~\ref{tab:metric_defs}.

For depth L1, following~\cite{zhuNiceslamNeuralImplicit2022} we render depth from $N=1000$ virtual views of GT and reconstructed mesh. The virtual views are sampled uniformly inside a cube within the room. Views that have unobserved points will be rejected and re-sampled. Then depth L1 is defined as the average L1 difference between rendered GT depth and reconstruction depth. 

\begin{table}[tp]
    \centering
    \resizebox{0.90\columnwidth}{!}{
    \begin{tabular}{lccc}
        \toprule
          &  \textbf{iMAP}$^{\star}$~\cite{sucarIMAPImplicitMapping2021a}& \textbf{NICE-SLAM}~\cite{zhuNiceslamNeuralImplicit2022}  & \textbf{Ours}\\
        \midrule
        {\bf Depth L1} $\downarrow$  & 7.64 & 3.53 & \textbf{1.58} \\
        {\bf Acc.} $\downarrow$  & 6.95 & 2.85  & \textbf{2.15} \\
        {\bf Comp.} $\downarrow$  & 5.33 & 3.00  & \textbf{2.21} \\
        {\bf Comp. Ratio} $\uparrow$  & 66.60 & 89.33  & \textbf{92.99} \\
        \bottomrule
    \end{tabular}
    }
    \vspace{-2mm}
    \caption{Reconstruction results on Replica dataset using NICE-SLAM culling strategy. The smooth weight $\lambda_{smooth}$ is increased to $1e-3$, and hash table size is set to be 14.}
    \label{tab:recon_replica_nice}
    \vspace{-6pt}
\end{table}
\begin{table}[]
\centering
\resizebox{0.90\columnwidth}{!}{
\begin{tabular}{ll}
\hline
3D Metric & definition \\
\hline
Acc & $\sum_{p \in P} (\min_{q \in Q} \lVert p - q \rVert ) / |P|$ \\
Comp & $\sum_{q \in Q} (\min_{p \in P} \lVert p - q \rVert) / |Q|$ \\
C-$\ell_1$ & $(\text{Acc} + \text{Comp}) / 2$ \\
Comp Ratio & $\sum_{q \in Q} (\min_{p \in P} \lVert p - q \rVert < 0.05) / |Q|$ \\
\hline
\end{tabular}
}
\vspace{-5pt}
\caption{Definitions of 3D metrics used for evaluation of reconstruction quality.
}
\vspace{-12pt}
\label{tab:metric_defs}
\end{table}

\section{Additional Experimental Results}

\subsection{More Results on Synthetic Datasets}
Here we show more detailed results on all the synthetic scenes in Replica~\cite{straubReplicaDatasetDigital2019} and Synthetic RGB-D~\cite{azinovicNeuralRGBDSurface2022c}.
We show per-scene quantitative results of the Replica~\cite{straubReplicaDatasetDigital2019} and Synthetic RGB-D dataset~\cite{azinovicNeuralRGBDSurface2022c} in Tab.~\ref{tab:replica_per_scene} and Tab.~\ref{tab:synthetic_per_scene}. Our method shows consistently better results in terms of the \textit{Completion} and competitive results of \textit{Accuracy}. We also provide more qualitative comparisons on Replica dataset in Fig.~\ref{fig:Replica_office0}-\ref{fig:Replica_office4} in different scenes with different shading mode.

\begin{figure}[!tb]
  \centering
  \footnotesize
  \setlength{\tabcolsep}{1.5pt}
  \newcommand{\sz}{0.48}
  \begin{tabular}{cc}
    \includegraphics[width=\sz\linewidth]{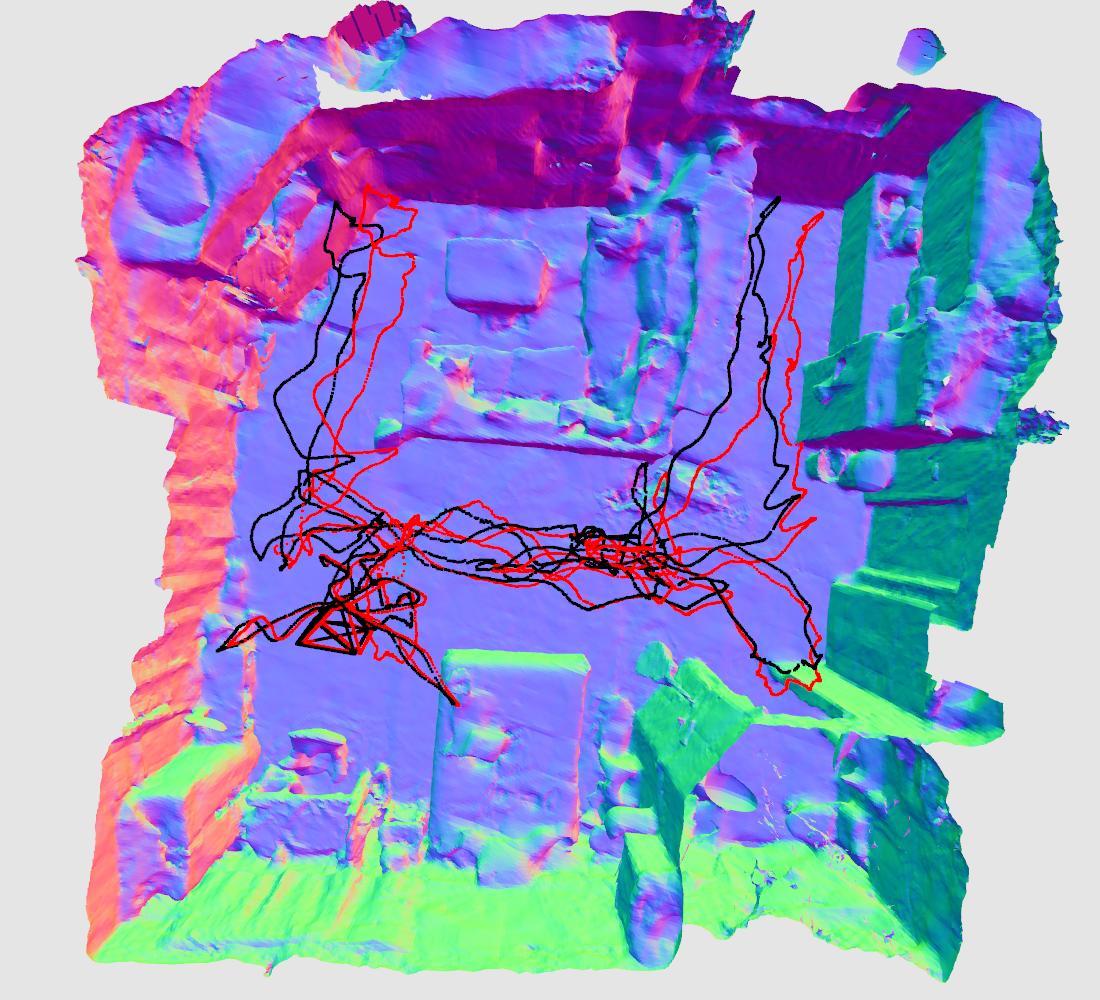} &
    \includegraphics[width=\sz\linewidth]{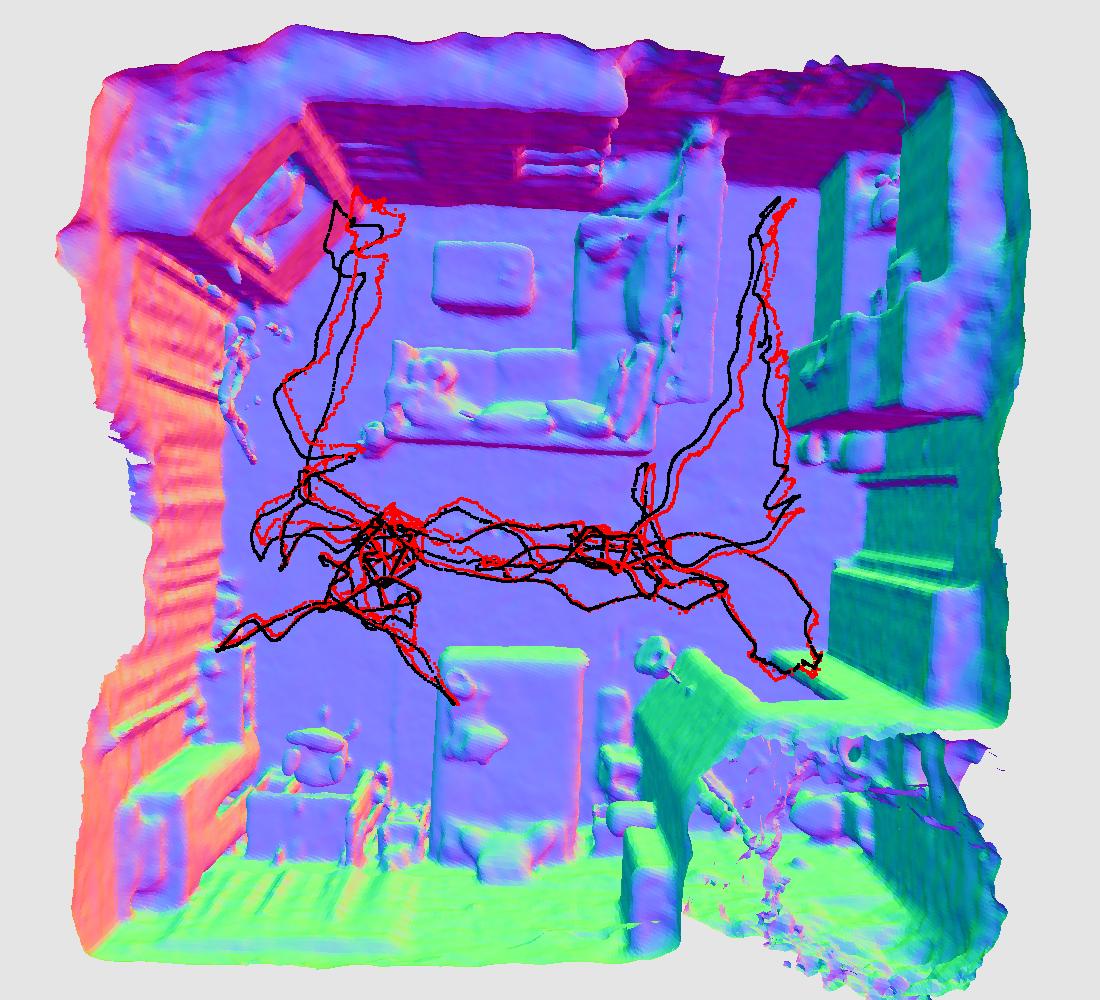} \\
    NICE-SLAM & Ours
  \end{tabular} 
  \vspace{-8pt}
  \caption{A more comprehensive comparison of raw trajectory w/o alignment. In both figures ground truth trajectory are shown in \textcolor{myblack}{black} and the estimated trajectory are shown in \textcolor{myred}{red}.}
  \label{fig:ATE}
  \vspace{-4pt}
\end{figure}
\begin{table}[tpb]
    \centering
 	\resizebox{0.99\columnwidth}{!}{
    \begin{tabular}{lccccccc}
        \toprule
         Scene ID & 0000 & 0059 & 0106 &  0169 &  0181 & 0207 &Avg. \\
        \midrule
    
        NICE-SLAM & 8.64 & 12.25 & \textbf{8.09} & 10.28 & 12.93 & \textbf{5.59} & 9.63\\
        Ours & \textbf{7.13} & \textbf{11.14} & 9.36 & \textbf{5.90} & \textbf{11.81} & 7.14 & \textbf{8.75} \\
        \hline
        NICE-SLAM & 25.24 & 25.01 & \textbf{10.40} & 30.51 & 39.98 & 12.70 & 23.97\\
        Ours & \textbf{12.94} & \textbf{19.12} & 12.12 & \textbf{19.61} & \textbf{34.41} & \textbf{9.88} & \textbf{18.01}\\
        \bottomrule
    \end{tabular}
     }
     \vspace{-5pt}
    \caption{ATE RMSE (cm) results (w/ and w/o trajectory alignment) of average of 5 runs on ScanNet.
    }
    \label{tab:ATE_no_align}
    \vspace{-3mm}
\end{table}

\subsection{More Results on Real-world Scenes}

\begin{table*}[tp]
  \centering
  \footnotesize
  \setlength{\tabcolsep}{0.3em}
    \begin{tabularx}{0.99\textwidth}{c l >{\centering\arraybackslash}X >{\centering\arraybackslash}X >{\centering\arraybackslash}X >{\centering\arraybackslash}X >{\centering\arraybackslash}X >{\centering\arraybackslash}X >{\centering\arraybackslash}X >{\centering\arraybackslash}X >{\centering\arraybackslash}X} 
      \toprule
         & & \tt{room0} & \tt{room1} & \tt{room2} & \tt{office0} & \tt{office1} & \tt{office2} & \tt{office3} & \tt{office4} & Avg. \\
          
        \midrule
        \multirow{4}{*}{\makecell{\textbf{iMAP}}}   
         & {\bf Depth L1} [cm] $\downarrow$ 
          & 5.08 & 3.44 & 5.78 & 3.79 & 3.76 & 3.97 & 5.61 & 5.71 & 4.64 \\
          & {\bf Acc.} [cm] $\downarrow$
          & 4.01 & 3.04 & 3.84 & 3.34 & 2.10 & 4.06 & 4.20 & 4.34 & 3.62\\
          & {\bf Comp.} [cm] $\downarrow$
          & 5.84 & 4.40 & 5.07 & 3.62 & 3.62 & 4.73 & 5.49 & 6.65 & 4.93\\
          & {\bf Comp. Ratio} [$<$ 5cm \%] $\uparrow$
          & 78.34 & 85.85 & 79.40 & 83.59 & 88.45 & 79.73 & 73.90 & 74.77 & 80.50 \\
      \midrule
      \multirow{4}{*}{\makecell{\textbf{NICE-SLAM}}}  
        & {\bf Depth L1} [cm] $\downarrow$ 
          & 1.79 & 1.33 & \textbf{2.20} & 1.43 & 1.58 & 2.70 & 2.10 & 2.06 & 1.90\\
          & {\bf Acc.} [cm] $\downarrow$
          &  2.44 & 2.10 & 2.17 & 1.85 & 1.56 & 3.28 & \textbf{3.01} & 2.54  & 2.37\\
          & {\bf Comp.} [cm] $\downarrow$
          & 2.60 & 2.19 & 2.73 & 1.84 & 1.82 & 3.11 & 3.16 & 3.61  & 2.63\\
          & {\bf Comp. Ratio} [$<$ 5cm \%] $\uparrow$
          &  91.81 & 93.56 & 91.48 & 94.93 & 94.11 & 88.27 & 87.68 & 87.23 & 91.13\\
     \midrule
     \multirow{4}{*}{{\makecell{\textbf{\papername}}}} 
        & {\bf Depth L1} [cm] $\downarrow$ 
          & \textbf{1.05} & \textbf{0.85} & 2.37 & \textbf{1.24} & \textbf{1.48} & \textbf{1.86} & \textbf{1.66} & \textbf{1.54} &  \textbf{1.51} \\
          & {\bf Acc. } [cm] $\downarrow$ 
          & \textbf{2.11} & \textbf{1.68} & \textbf{1.99} & \textbf{1.57} & \textbf{1.31} & \textbf{2.84} & 3.06 & \textbf{2.23} & \textbf{2.10} \\
          & {\bf Comp. } [cm] $\downarrow$ 
          & \textbf{2.02} & \textbf{1.81} & \textbf{1.96} & \textbf{1.56} & \textbf{1.59} & \textbf{2.43} & \textbf{2.72} & \textbf{2.52} & \textbf{2.08} \\
          & {\bf Comp. Ratio} [$<$ 5cm \%] $\uparrow$ 
          & \textbf{95.26} & \textbf{95.19} & \textbf{93.58} & \textbf{96.09} & \textbf{94.65} & \textbf{91.63} & \textbf{90.72} & \textbf{90.44} & \textbf{93.44}\\

    \bottomrule
    \end{tabularx}%
    \vspace{-4pt}
    \caption{Per-scene quantitative results on Replica~\cite{straubReplicaDatasetDigital2019} dataset. Our method achieves consistently better reconstruction in comparison to NICE-SLAM~\cite{zhuNiceslamNeuralImplicit2022} and iMAP~\cite{sucarIMAPImplicitMapping2021a} in most of the scenes.}
    \label{tab:replica_per_scene}
\end{table*}

\begin{table*}[tp]
  \centering
  \footnotesize
  \setlength{\tabcolsep}{0.3em}
    \begin{tabularx}{0.99\textwidth}{c l >{\centering\arraybackslash}X >{\centering\arraybackslash}X >{\centering\arraybackslash}X >{\centering\arraybackslash}X >{\centering\arraybackslash}X >{\centering\arraybackslash}X >{\centering\arraybackslash}X >{\centering\arraybackslash}X} 
      \toprule
         & & \tt{BR} & \tt{CK} &  \tt{GR} & \tt{GWR} & \tt{MA} & \tt{TG} & \tt{WR} &  Avg. \\
         
        \midrule
        \multirow{4}{*}{{\makecell{\textbf{iMAP}*}}} 
        & {\bf Depth L1} [cm] $\downarrow$ 
          & 24.03 & 63.59 & 26.22 & 21.32 & 61.29 & 29.16 & 81.71 &  47.22 \\
          & {\bf Acc. } [cm] $\downarrow$ 
          & 10.56 & 25.16 & 13.01 & 11.90 & 29.62 & 12.98 & 24.82 & 18.29 \\
          & {\bf Comp. } [cm] $\downarrow$ 
          & 11.27 & 31.09 & 19.17 & 20.39 & 49.22 & 21.07 & 32.63 & 26.41 \\
          & {\bf Comp. Ratio} [$<$ 5cm \%] $\uparrow$ 
          & 46.91 &  12.96 & 21.78 & 20.48 & 10.72 & 19.17 & 13.07 & 20.73\\
         
        \midrule
        \multirow{4}{*}{{\makecell{\textbf{NICE-SLAM}}}} 
        & {\bf Depth L1} [cm] $\downarrow$ 
          & 3.66 & 12.08 & 10.88 & 2.57 & 1.72 & 7.74 & 5.59 &  6.32 \\
          & {\bf Acc. } [cm] $\downarrow$ 
          & 3.44 & 10.92 & 5.34 & 2.63 & 6.55 & 3.57 & 9.22 & 5.95 \\
          & {\bf Comp. } [cm] $\downarrow$ 
          & 3.69 & 12.00 & 4.94 & 3.15 & 3.13 & 5.28 & 4.89 & 5.30 \\
          & {\bf Comp. Ratio} [$<$ 5cm \%] $\uparrow$ 
          & 87.69 &  55.41 & 82.78 & 87.72 & 85.04 & 72.05 & 71.56 & 77.46\\
          
        \midrule
        
        \multirow{4}{*}{{\makecell{\textbf{\papername}}}} 
        & {\bf Depth L1} [cm] $\downarrow$ 
          & \textbf{3.51} & \textbf{5.62} & \textbf{1.95} & \textbf{1.25} & \textbf{1.41} & \textbf{4.66} & \textbf{2.74} &  \textbf{3.02} \\
          & {\bf Acc. } [cm] $\downarrow$ 
          & \textbf{1.97} & \textbf{4.68} & \textbf{2.10} & \textbf{1.89} & \textbf{1.60} & \textbf{3.38} & \textbf{5.03} & \textbf{2.95} \\
          & {\bf Comp. } [cm] $\downarrow$ 
          & \textbf{1.93} & \textbf{4.94} & \textbf{2.96} & \textbf{2.16} & \textbf{2.67} & \textbf{2.74} & \textbf{3.34} & \textbf{2.96} \\
          & {\bf Comp. Ratio} [$<$ 5cm \%] $\uparrow$ 
          & \textbf{94.75} &  \textbf{68.91} & \textbf{90.80} & \textbf{95.04} & \textbf{86.98} & \textbf{86.74} & \textbf{84.94} & \textbf{86.88}\\
   
      \bottomrule
    \end{tabularx}%
    \vspace{-4pt}
    \caption{Per-scene quantitative results on Synthetic RGBD~\cite{azinovicNeuralRGBDSurface2022c} dataset. Since this dataset simulates noisy depth maps with missing depth measurement, our method surpasses NICE-SLAM~\cite{zhuNiceslamNeuralImplicit2022} by a larger margin. This indicates our method is more robust to input noise.}
    \label{tab:synthetic_per_scene}
    \vspace{-10pt}
\end{table*}

\begin{figure*}[thbp]
  \centering
  \footnotesize
  \setlength{\tabcolsep}{1.5pt}
  \newcommand{\sz}{0.235}
  \begin{tabular}{cccc}
    iMAP$^*$~\cite{sucarIMAPImplicitMapping2021a} & NICE-SLAM~\cite{zhuNiceslamNeuralImplicit2022} & Ours & ScanNet Mesh \\
    \includegraphics[width=\sz\linewidth]{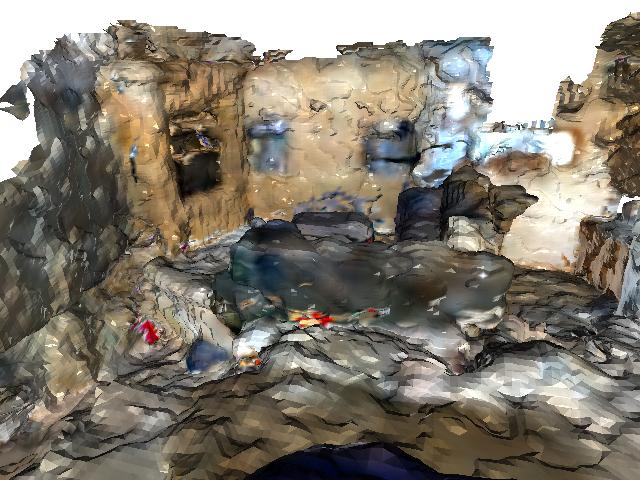} &
    \includegraphics[width=\sz\linewidth]{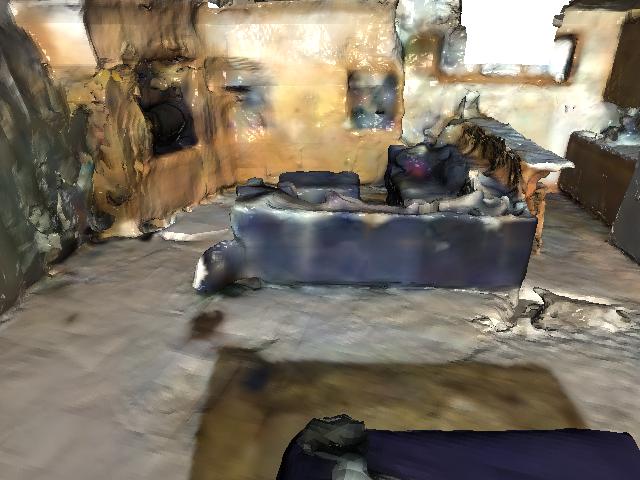} &
    \includegraphics[width=\sz\linewidth]{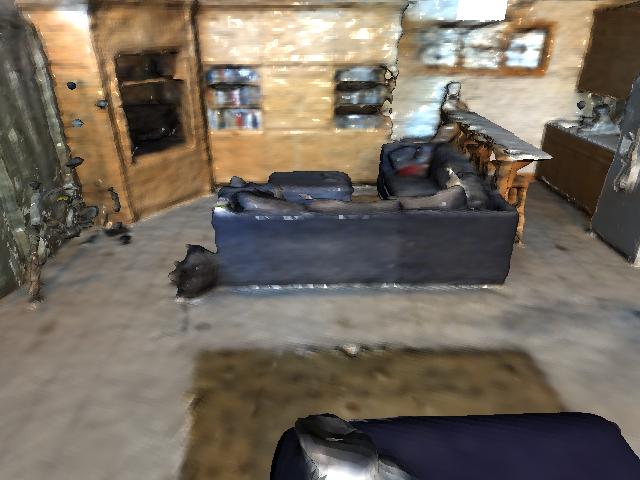} &
    \includegraphics[width=\sz\linewidth]{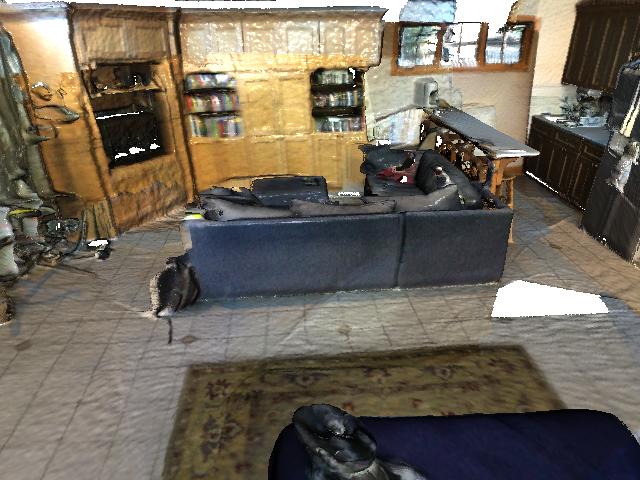} \\
    \includegraphics[width=\sz\linewidth]{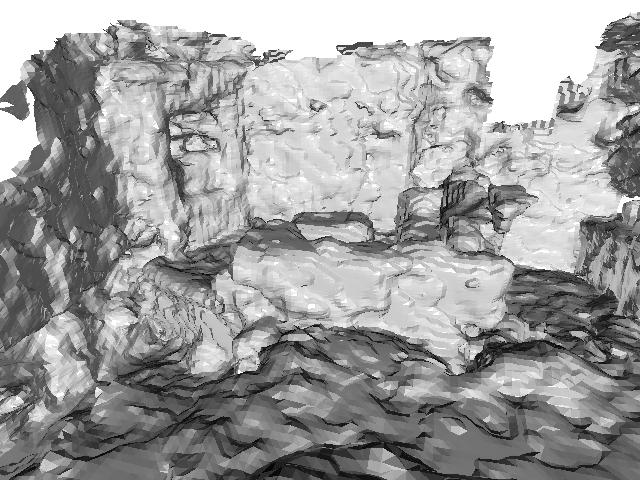} &
    \includegraphics[width=\sz\linewidth]{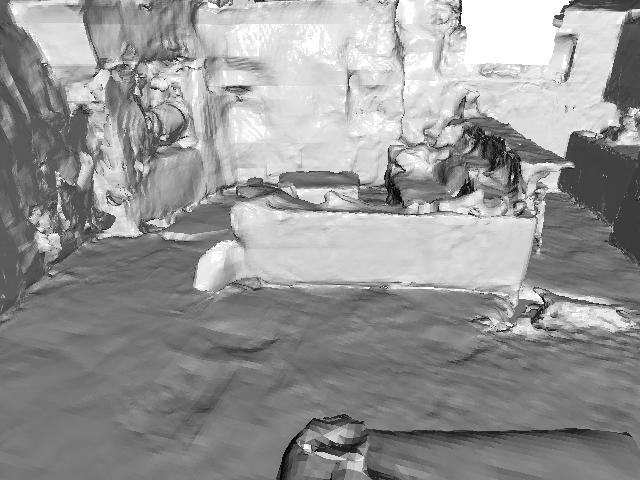} &
    \includegraphics[width=\sz\linewidth]{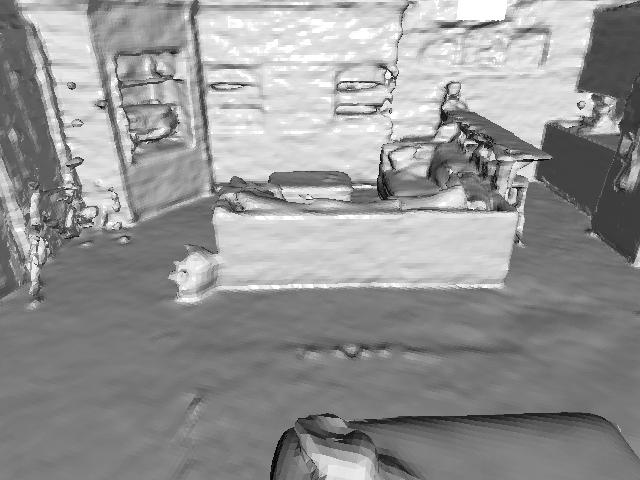} &
    \includegraphics[width=\sz\linewidth]{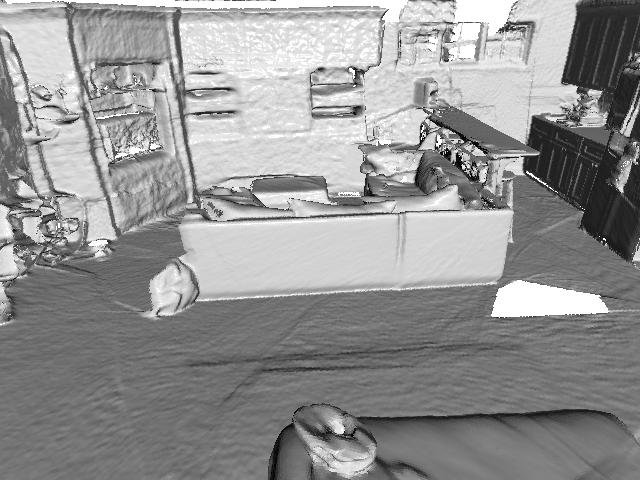} \\
    \includegraphics[width=\sz\linewidth]{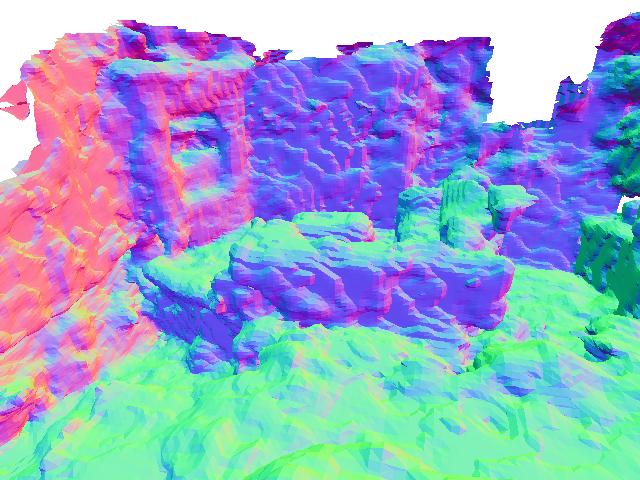} &
    \includegraphics[width=\sz\linewidth]{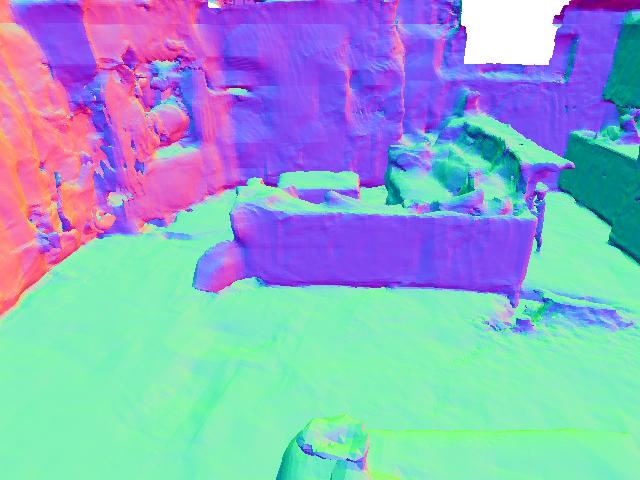} &
    \includegraphics[width=\sz\linewidth]{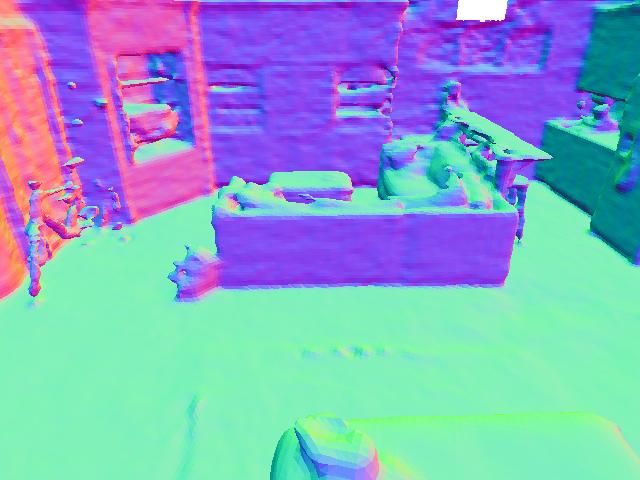} &
    \includegraphics[width=\sz\linewidth]{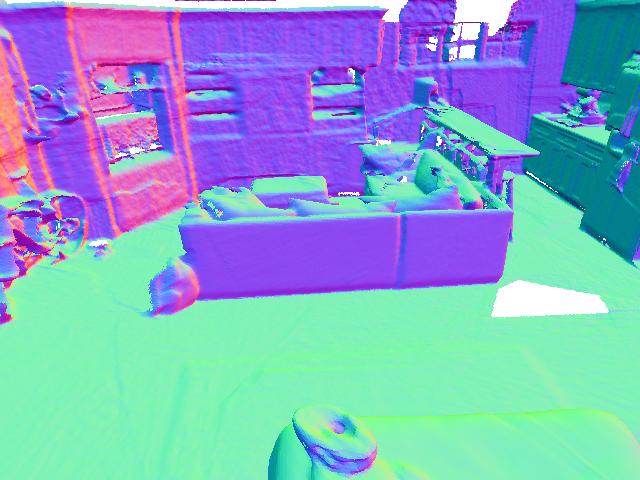} \\
  \end{tabular}
  \vspace{-5pt}
  \caption{Qualitative comparison on ScanNet \texttt{scene0000} with different shading mode. For real world scans, our method can accurately recover thin structures (e.g. bicycle) while achieve better hole fillings. Since we adopt global bundle adjustment, our scene reconstruction seems to have better coherence while reconstruction of NICE-SLAM~\cite{zhuNiceslamNeuralImplicit2022} contains some stitched effect due to the local bundle adjustment. }
  \label{fig:scannet_scene0000}
\end{figure*}

\begin{figure*}[htbp]
  \centering
  \footnotesize
  \setlength{\tabcolsep}{1.5pt}
  \newcommand{\sz}{0.235}
  \begin{tabular}{cccc}
    iMAP$^*$~\cite{sucarIMAPImplicitMapping2021a} & NICE-SLAM~\cite{zhuNiceslamNeuralImplicit2022} & Ours & ScanNet Mesh \\
    \includegraphics[width=\sz\linewidth]{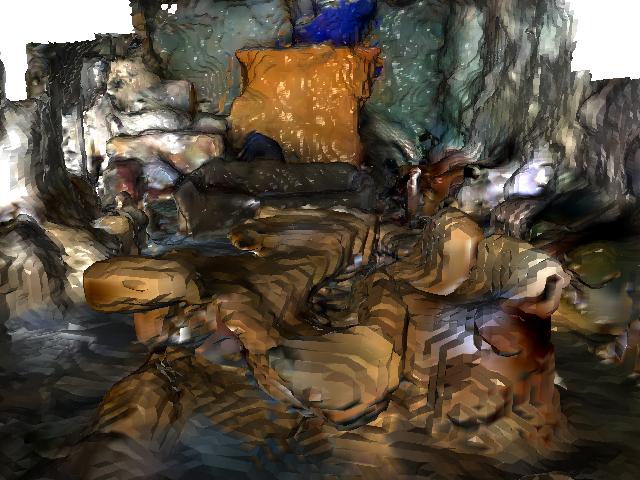} &
    \includegraphics[width=\sz\linewidth]{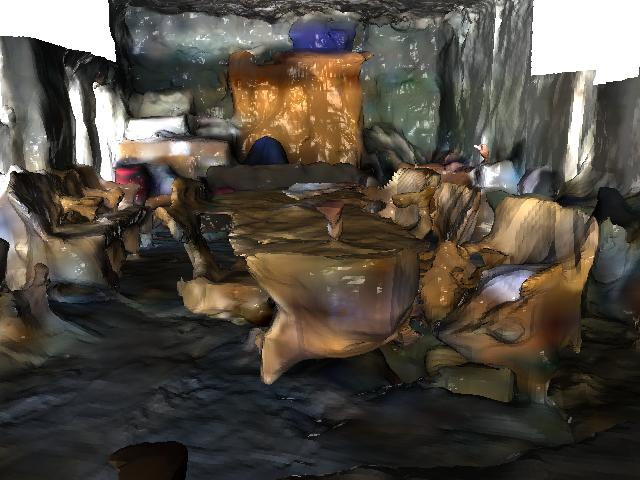} &
    \includegraphics[width=\sz\linewidth]{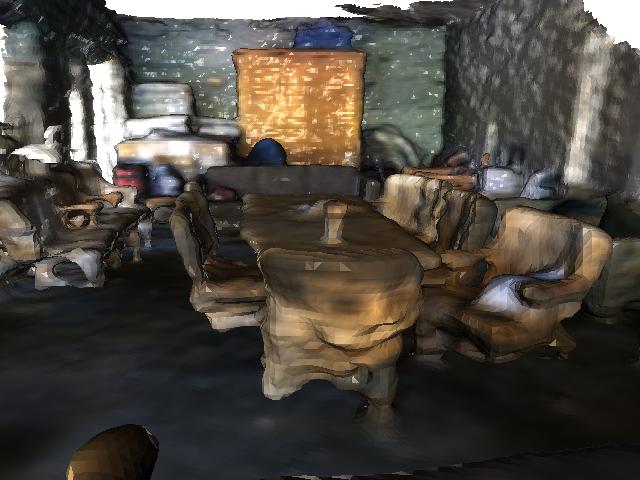} &
    \includegraphics[width=\sz\linewidth]{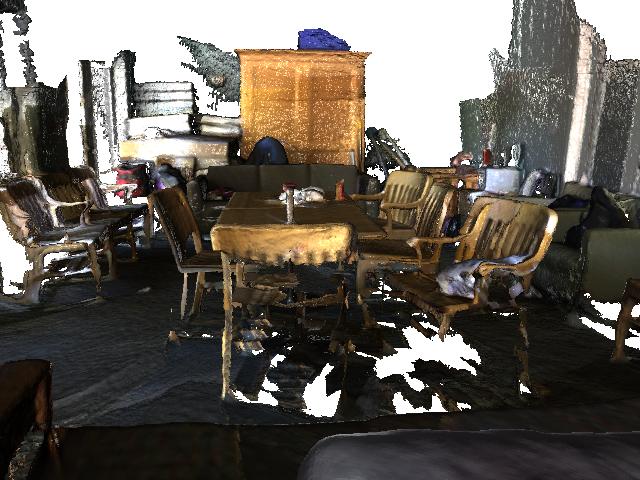} \\
    \includegraphics[width=\sz\linewidth]{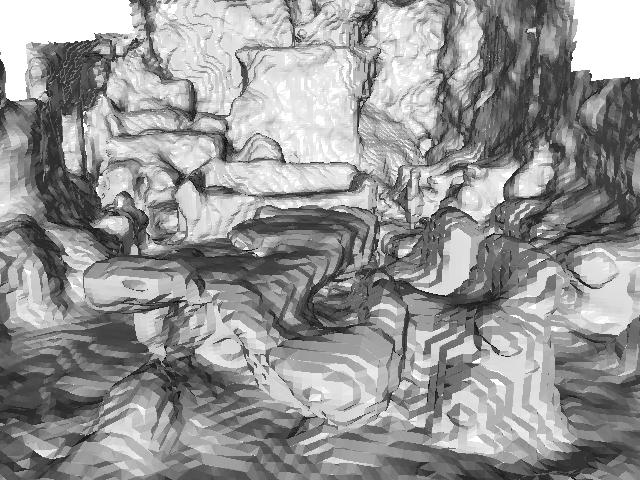} &
    \includegraphics[width=\sz\linewidth]{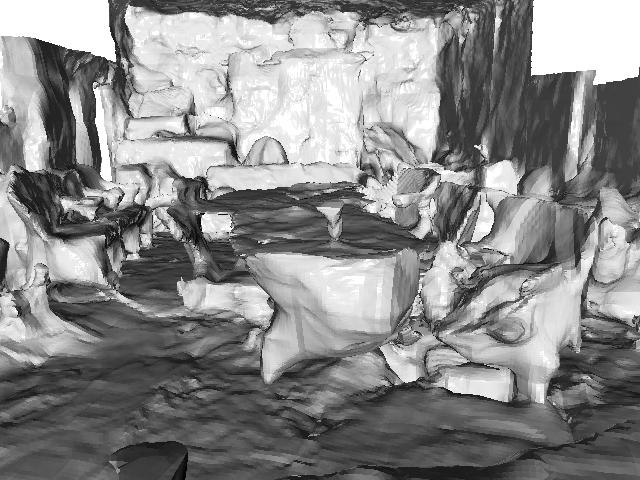} &
    \includegraphics[width=\sz\linewidth]{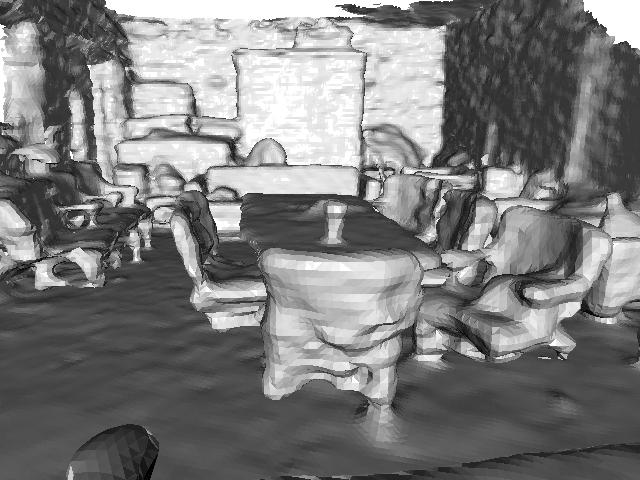} &
    \includegraphics[width=\sz\linewidth]{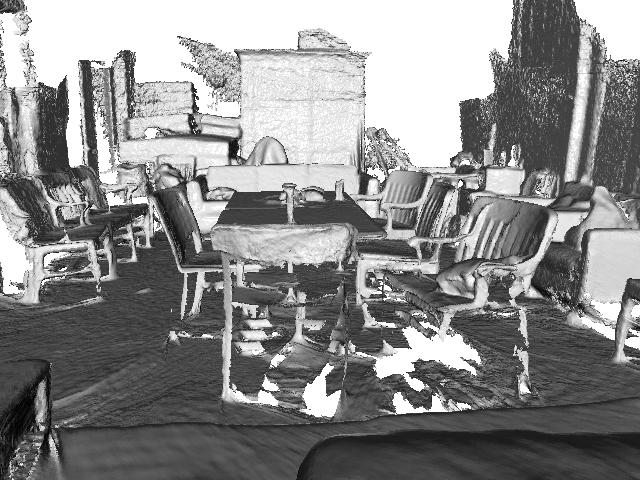} \\
    \includegraphics[width=\sz\linewidth]{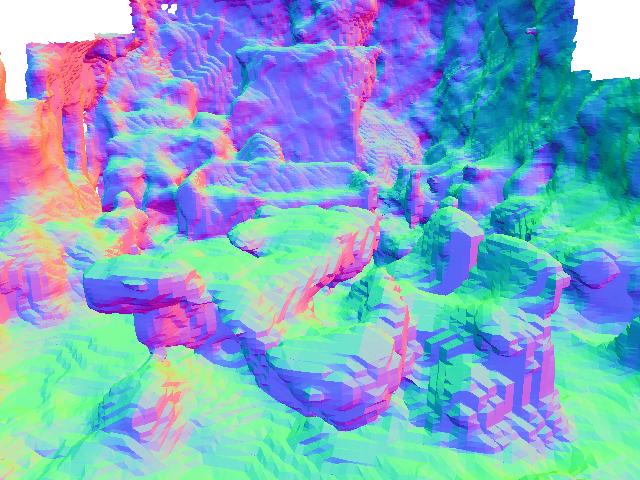} &
    \includegraphics[width=\sz\linewidth]{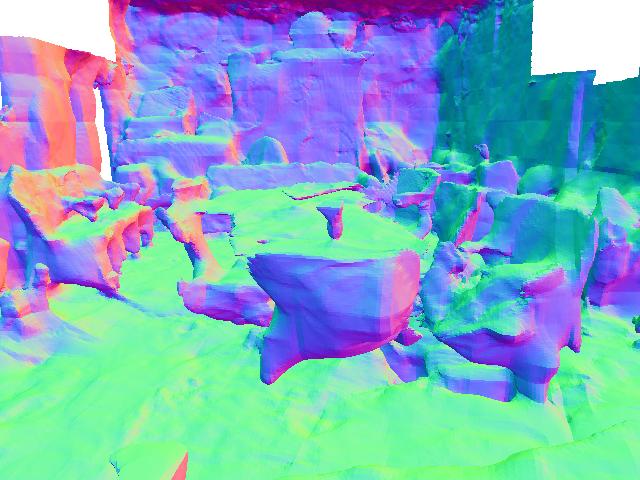} &
    \includegraphics[width=\sz\linewidth]{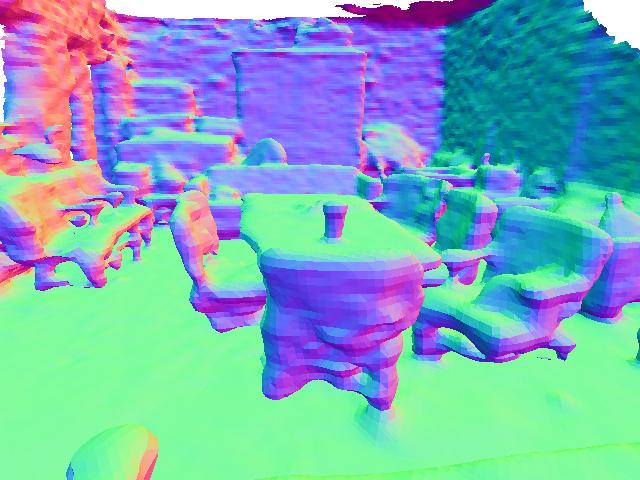} &
    \includegraphics[width=\sz\linewidth]{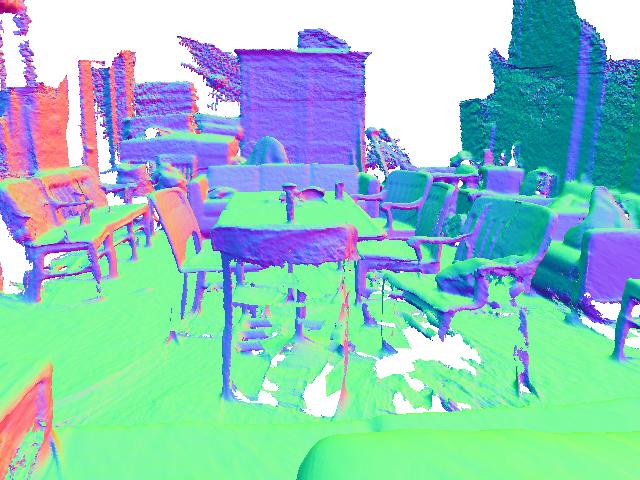} \\
  \end{tabular} 
  \caption{Qualitative comparison on ScanNet \texttt{scene0059} with different shading mode. Our method achieve smooth reconstruction of the floor while accurately recover the chairs. }
  \label{fig:scannet_scene0059}
\end{figure*}

\begin{figure*}[htbp]
  \centering
  \footnotesize
  \setlength{\tabcolsep}{1.5pt}
  \newcommand{\sz}{0.235}
  \begin{tabular}{cccc}
    iMAP$^*$~\cite{sucarIMAPImplicitMapping2021a} & NICE-SLAM~\cite{zhuNiceslamNeuralImplicit2022} & Ours & ScanNet Mesh \\
    \includegraphics[width=\sz\linewidth]{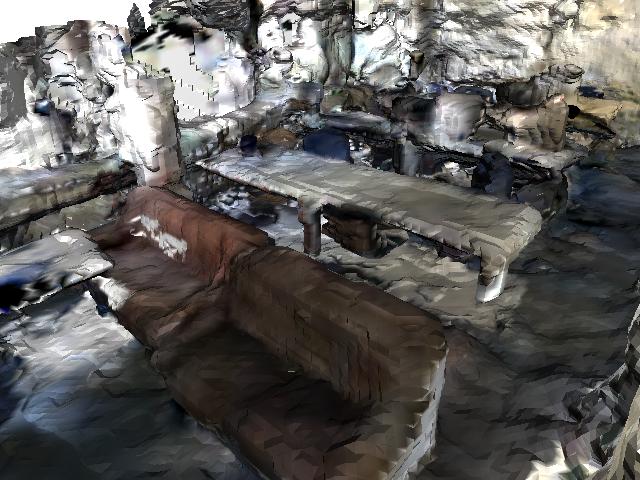} &
    \includegraphics[width=\sz\linewidth]{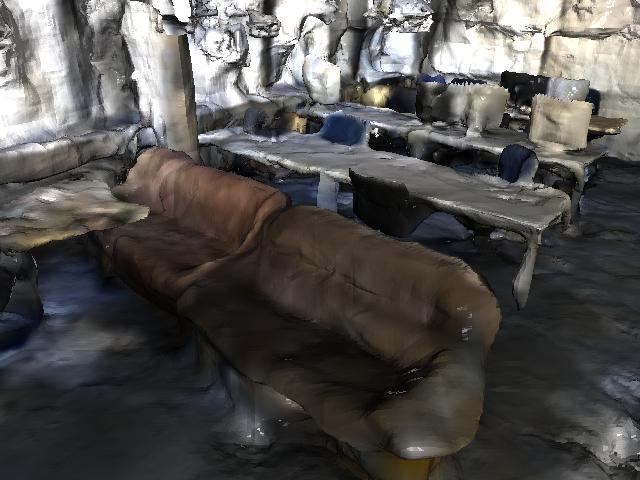} &
    \includegraphics[width=\sz\linewidth]{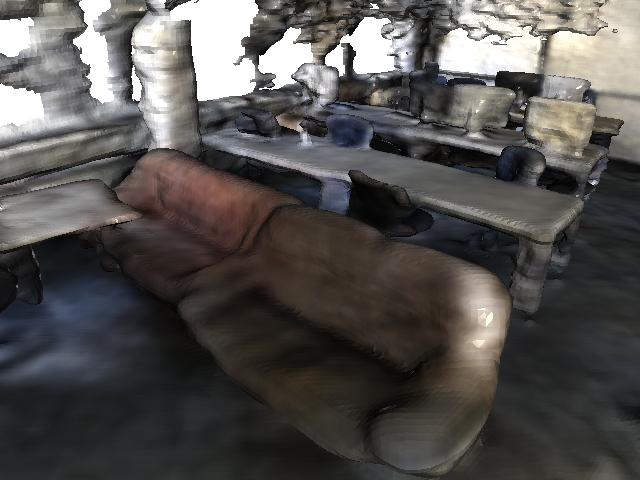} &
    \includegraphics[width=\sz\linewidth]{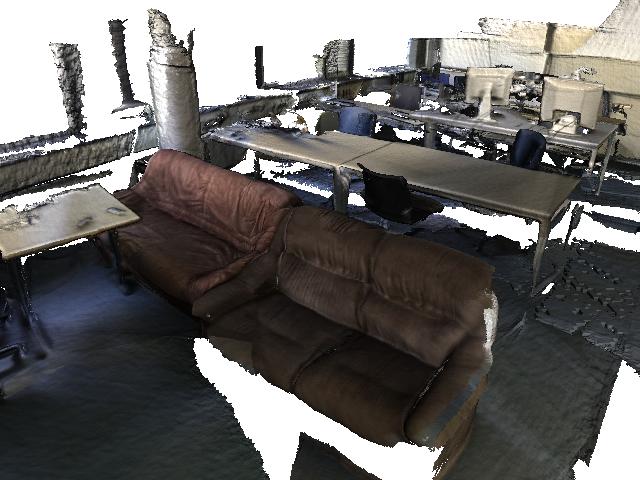} \\
    \includegraphics[width=\sz\linewidth]{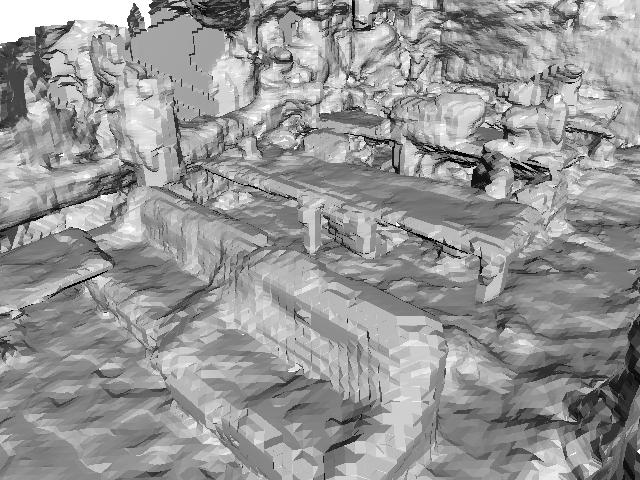} &
    \includegraphics[width=\sz\linewidth]{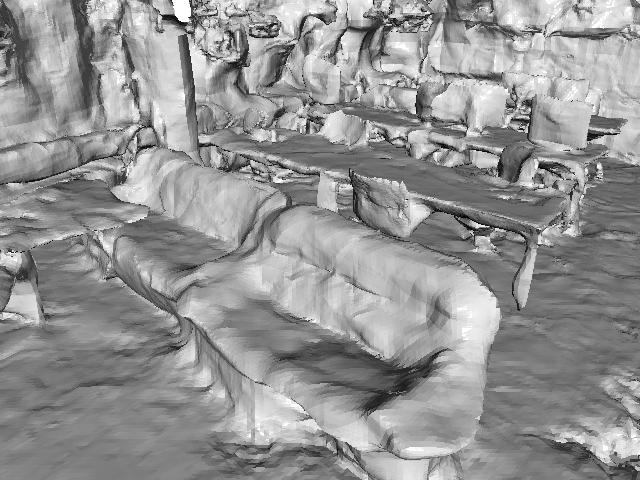} &
    \includegraphics[width=\sz\linewidth]{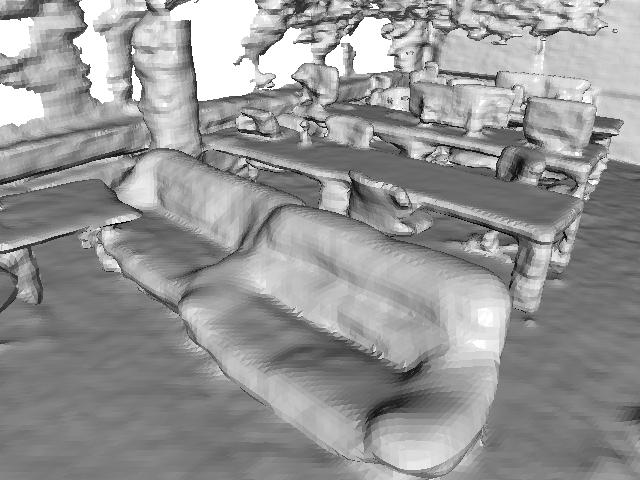} &
    \includegraphics[width=\sz\linewidth]{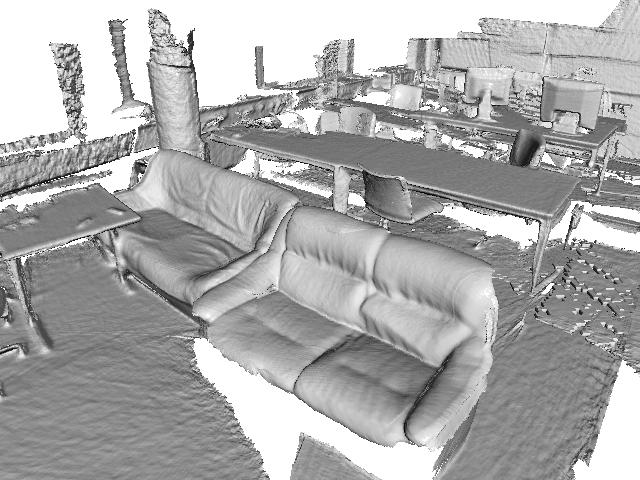} \\
    \includegraphics[width=\sz\linewidth]{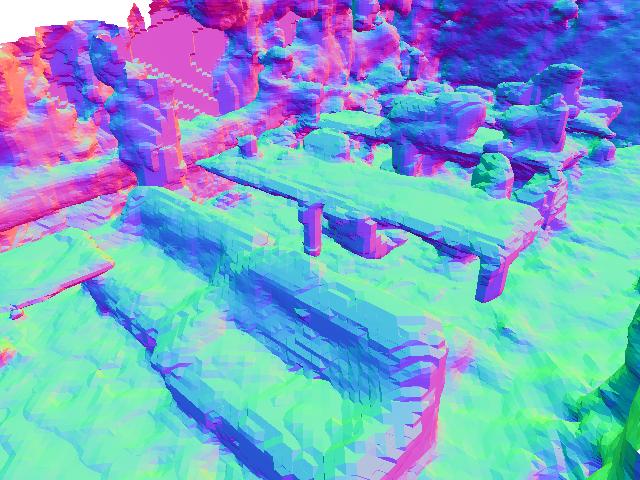} &
    \includegraphics[width=\sz\linewidth]{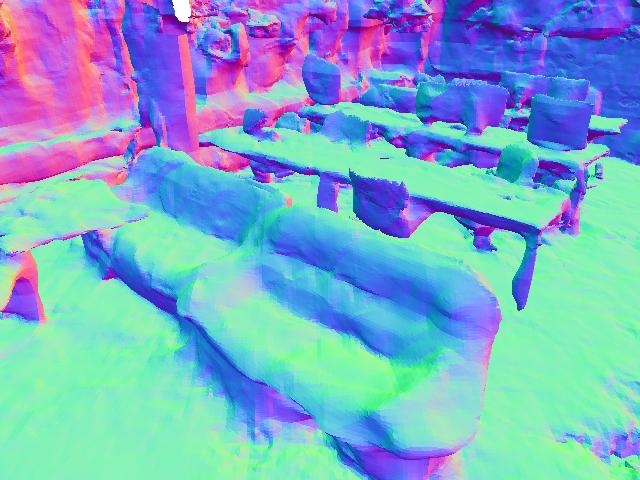} &
    \includegraphics[width=\sz\linewidth]{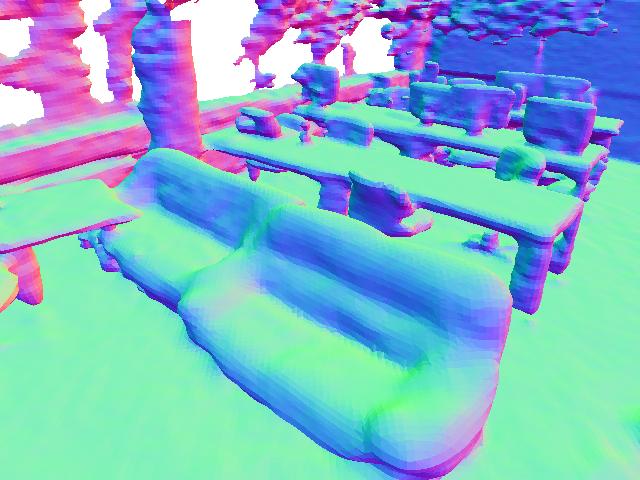} &
    \includegraphics[width=\sz\linewidth]{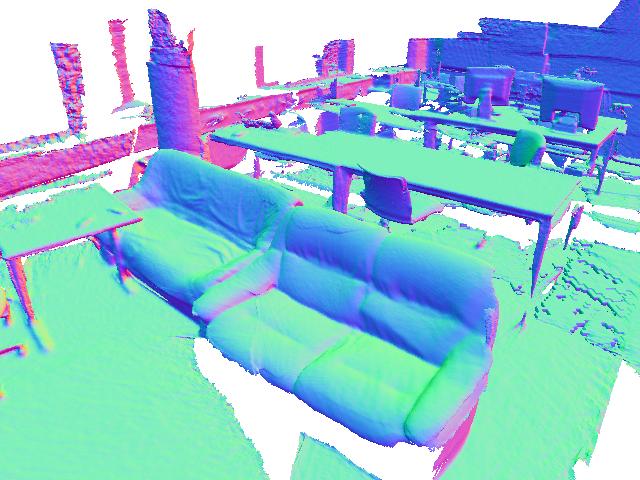} \\
  \end{tabular} 
  \caption{Qualitative comparison on ScanNet \texttt{scene0106} with different shading mode. Our reconstruction result is clearly less noisy in comparison to other two baseline models.}
  \label{fig:scannet_scene0106}
\end{figure*}

\noindent\textbf{ScanNet sequences~\cite{daiScannetRichlyannotated3d2017}.} In our main paper we showed top-down view comaparison on ScanNet sequence, which highlighted more on overall reconstruction quality and tracking accuracy. In this section we show more detailed zoom-in views in Fig.~\ref{fig:scannet_scene0000}-\ref{fig:scannet_scene0106} to better showcase the level of details and fidality that Co-SLAM can achieve on those challenging real-world sequences.

\begin{figure*}[htbp]
  \centering
  \footnotesize
  \setlength{\tabcolsep}{1.5pt}
  \newcommand{\sz}{0.24}
  \begin{tabular}{lcccc}
    & \texttt{top-down} & \texttt{view-1} & \texttt{view-2} & \texttt{view-3} \\
    \makecell{\rotatebox{90}{\tt Ours}} &
    \makecell{\includegraphics[width=\sz\linewidth]{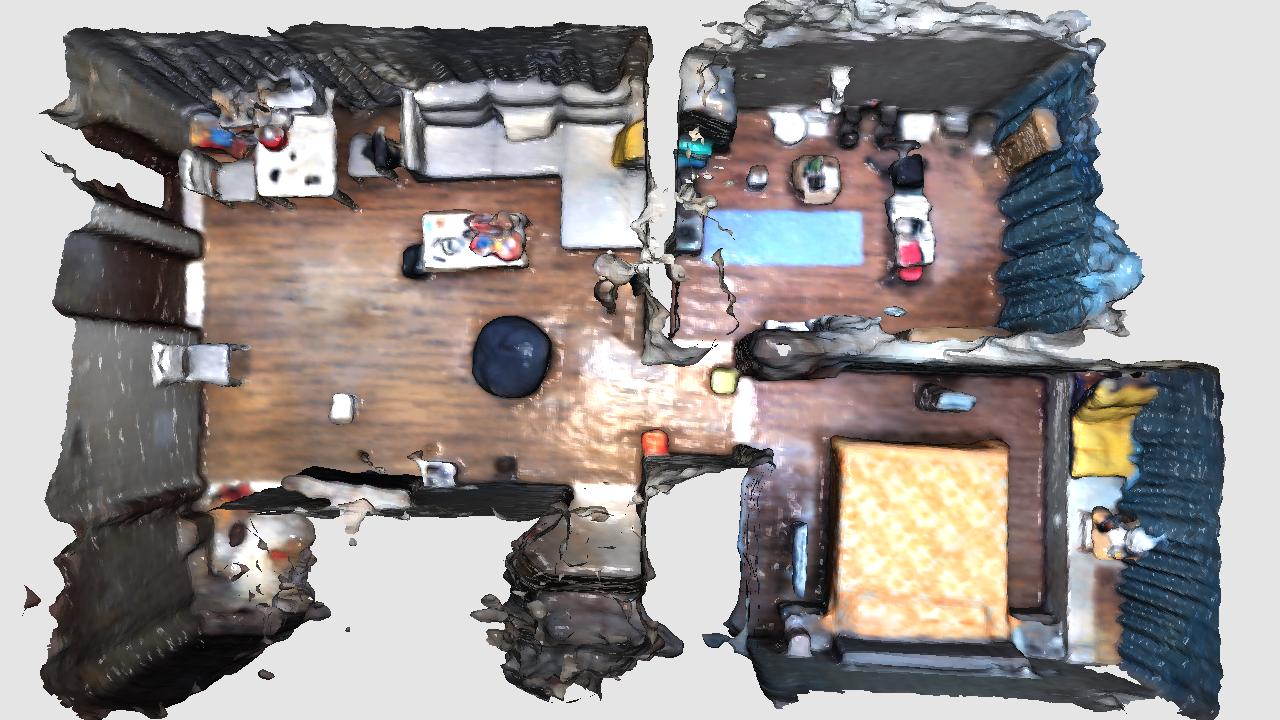}} &
    \makecell{\includegraphics[width=\sz\linewidth]{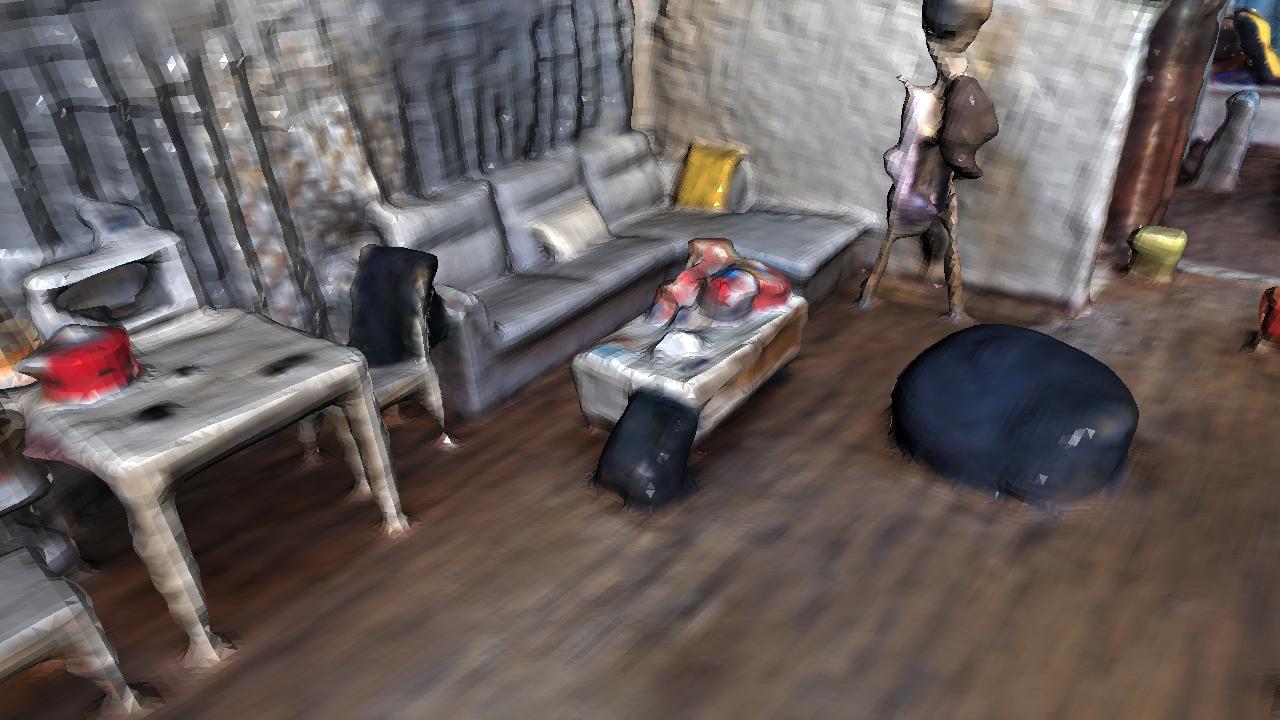}} &
    \makecell{\includegraphics[width=\sz\linewidth]{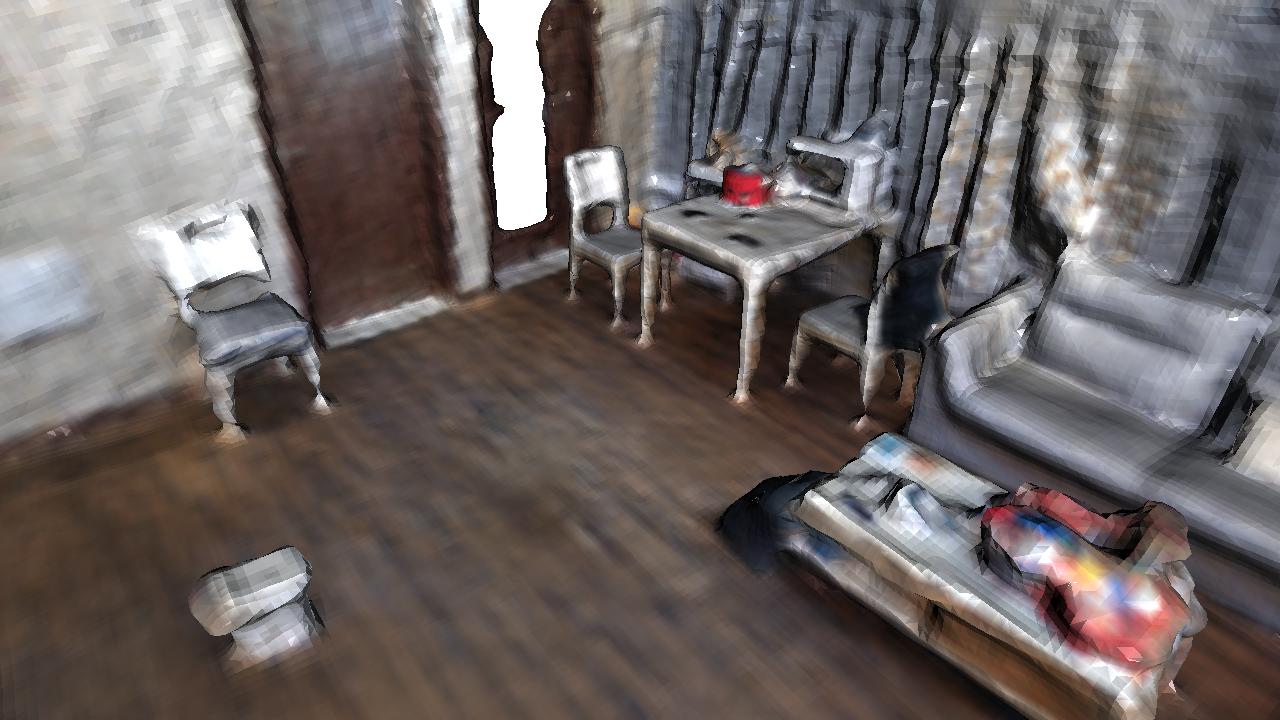}} &
    \makecell{\includegraphics[width=\sz\linewidth]{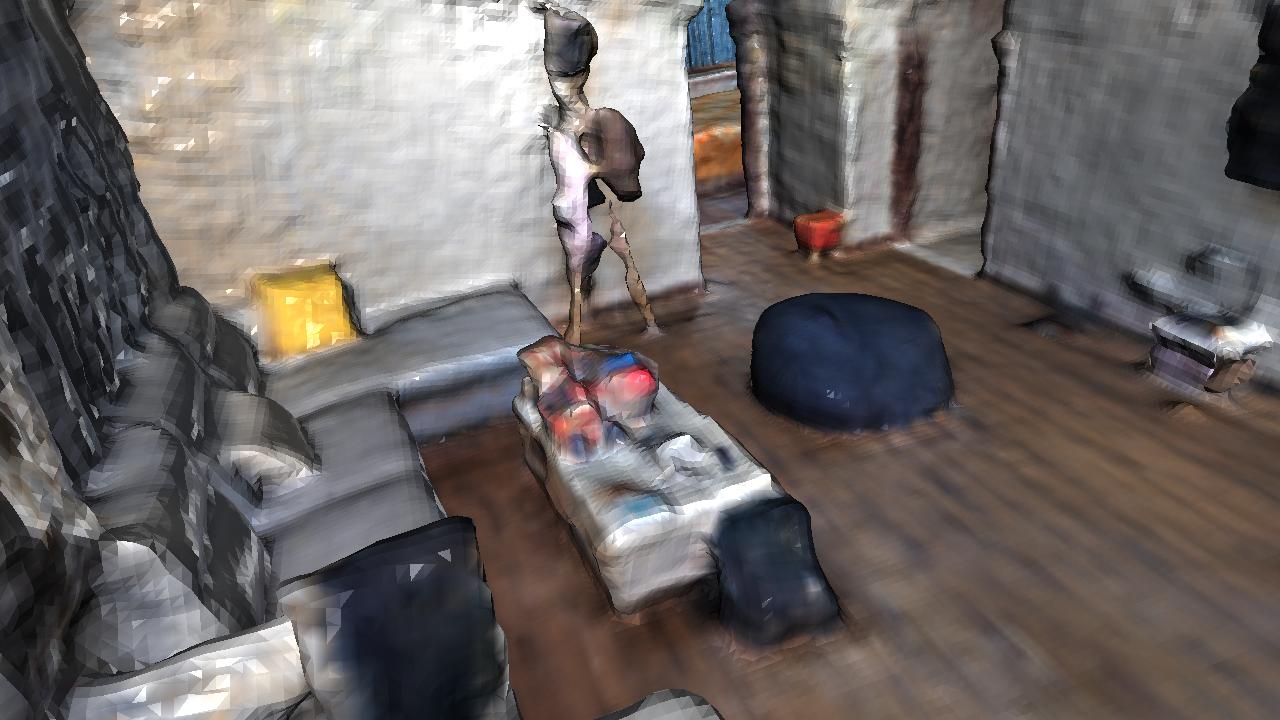}} \\
    \makecell{\rotatebox{90}{\tt NICE-SLAM}} &
    \makecell{\includegraphics[width=\sz\linewidth]{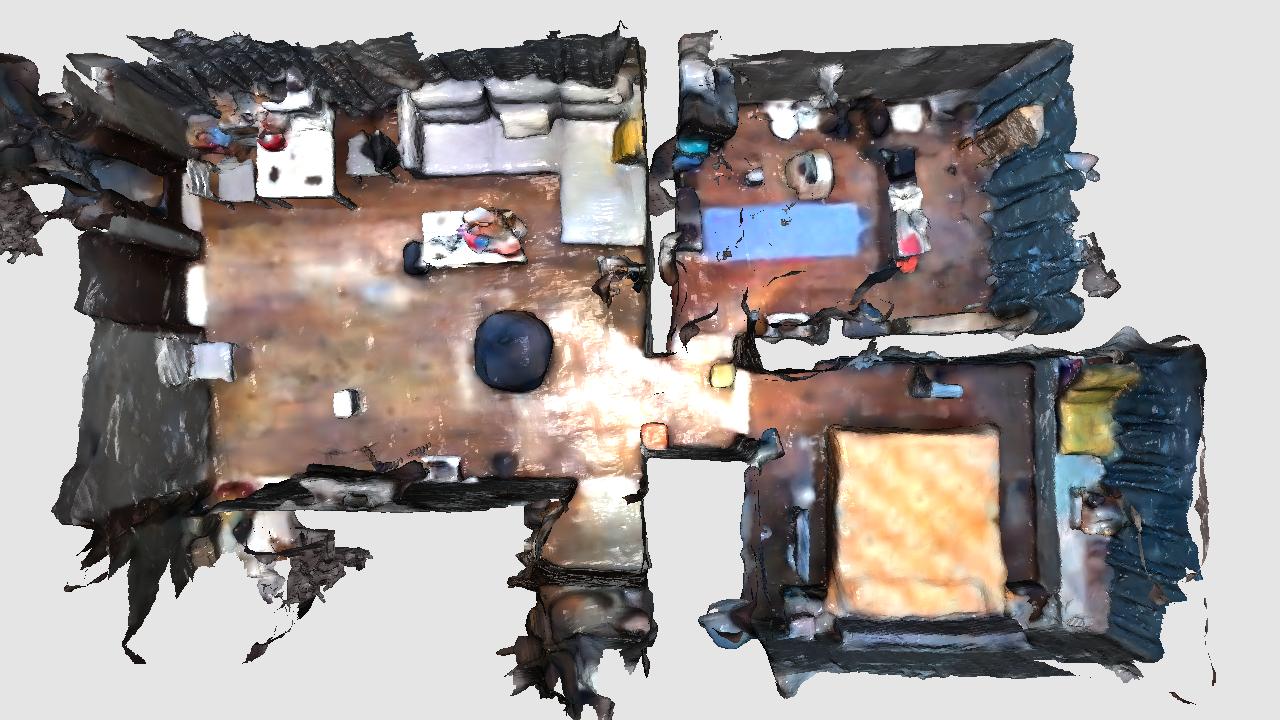}} &
    \makecell{\includegraphics[width=\sz\linewidth]{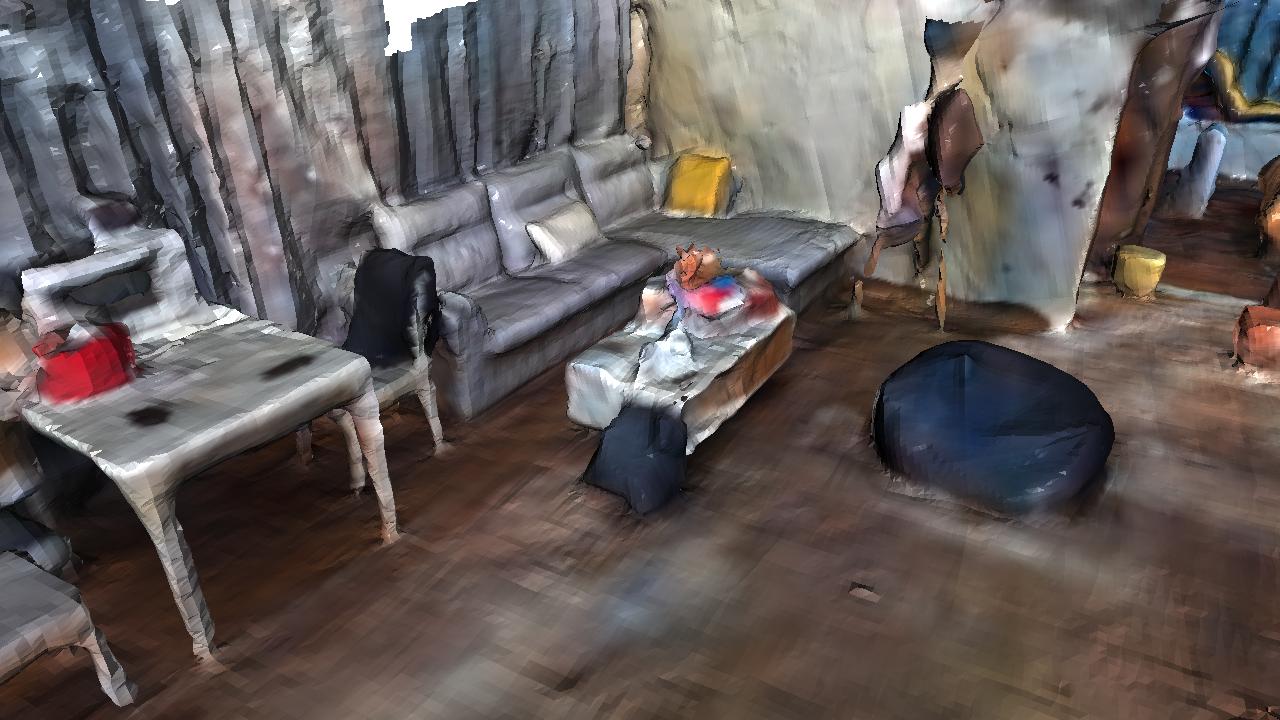}} &
    \makecell{\includegraphics[width=\sz\linewidth]{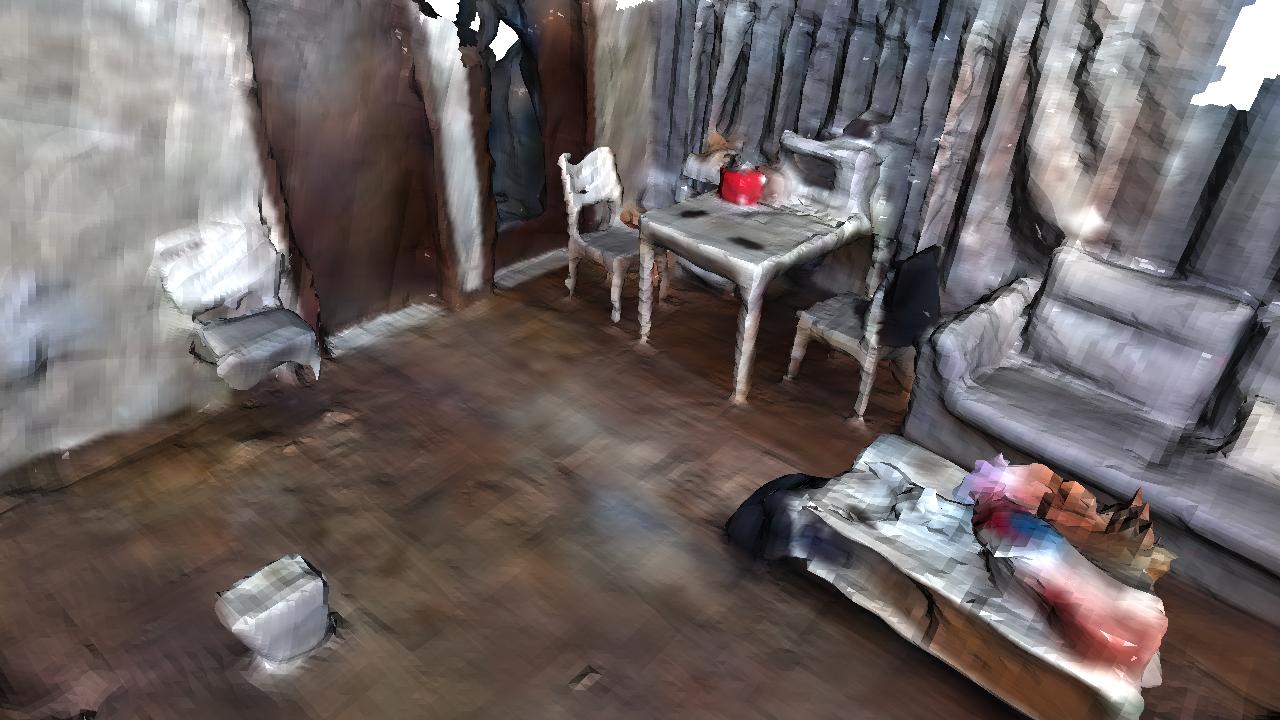}} &
    \makecell{\includegraphics[width=\sz\linewidth]{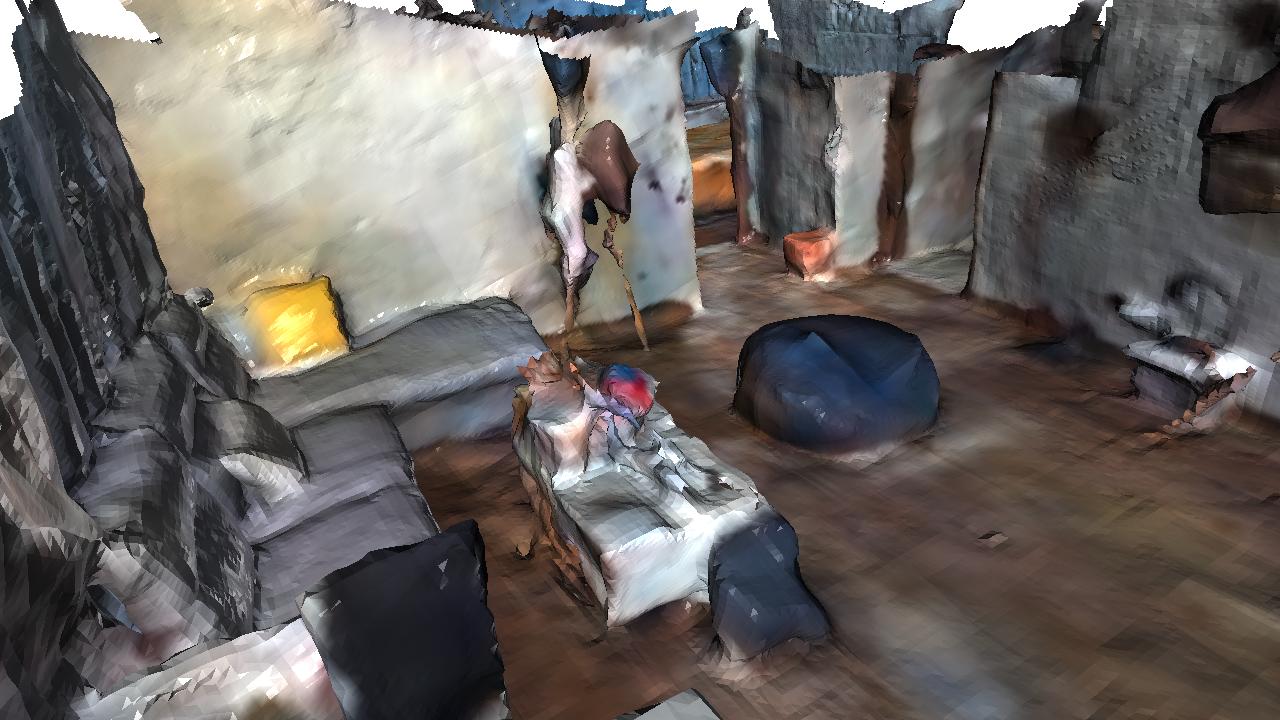}} \\
    \makecell{\rotatebox{90}{\tt Ours}} &
    \makecell{\includegraphics[width=\sz\linewidth]{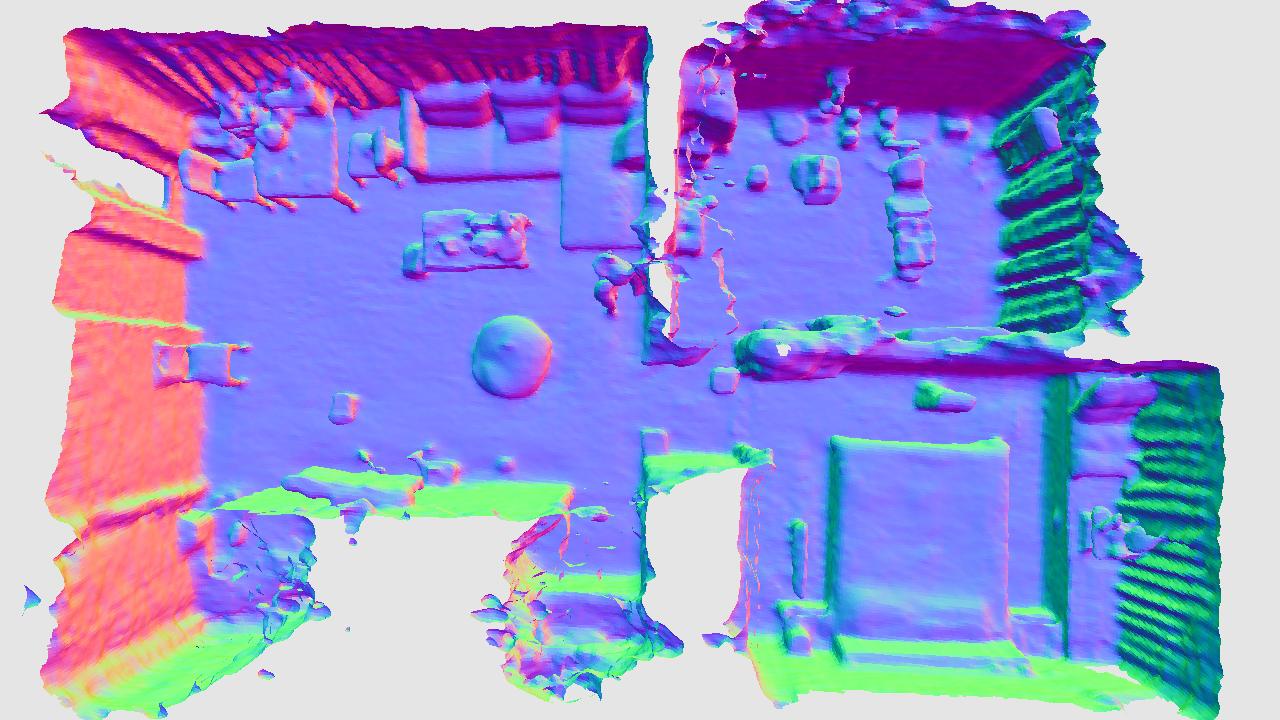}} &
    \makecell{\includegraphics[width=\sz\linewidth]{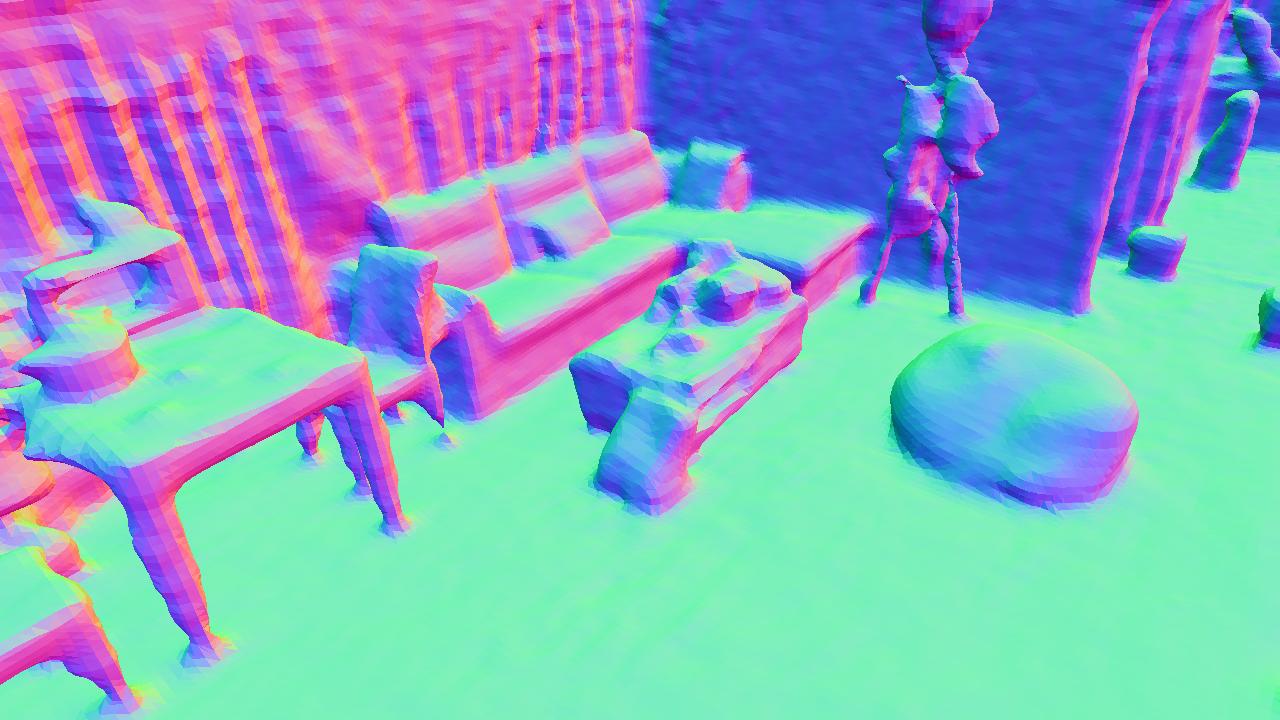}} &
    \makecell{\includegraphics[width=\sz\linewidth]{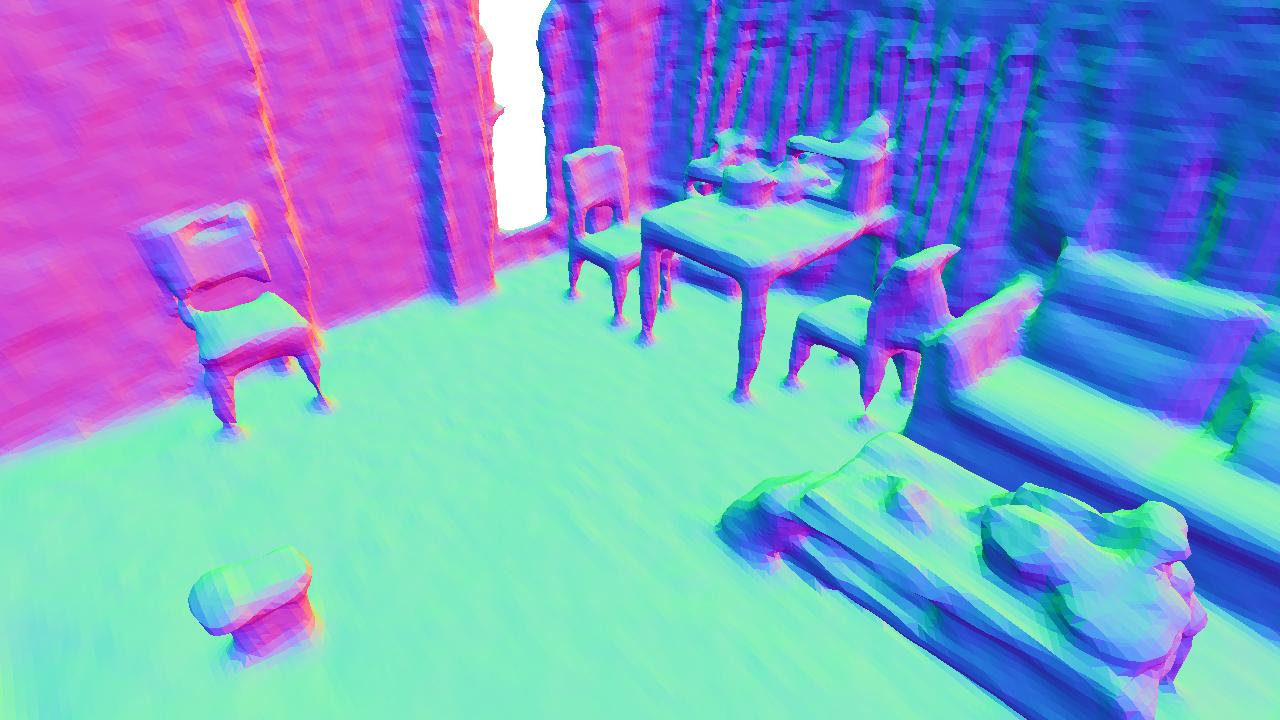}} &
    \makecell{\includegraphics[width=\sz\linewidth]{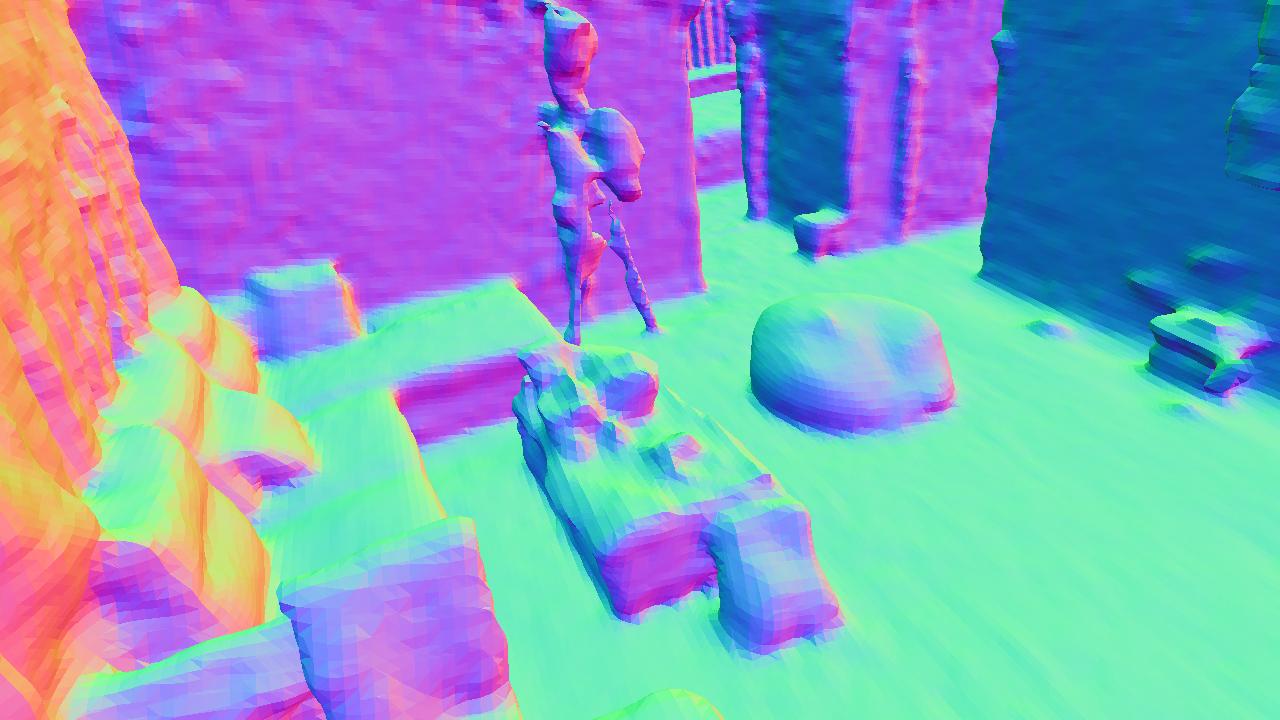}} \\
    \makecell{\rotatebox{90}{\tt NICE-SLAM}} &
    \makecell{\includegraphics[width=\sz\linewidth]{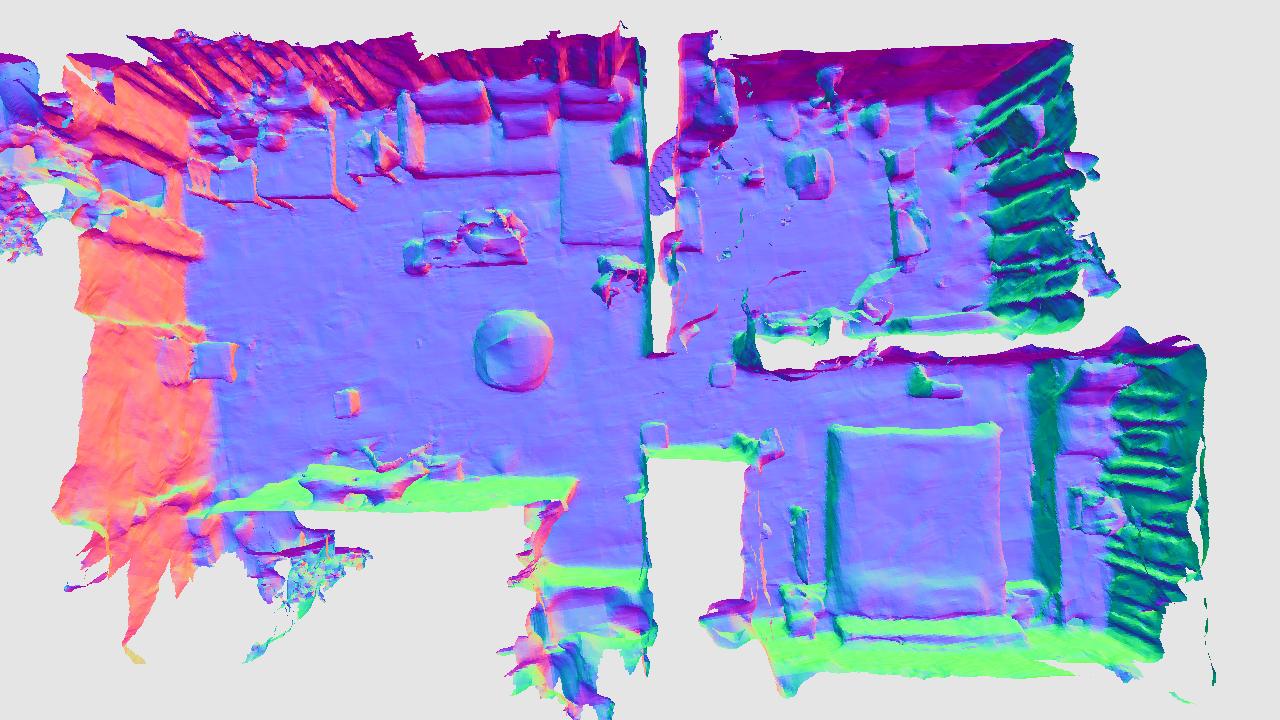}} &
    \makecell{\includegraphics[width=\sz\linewidth]{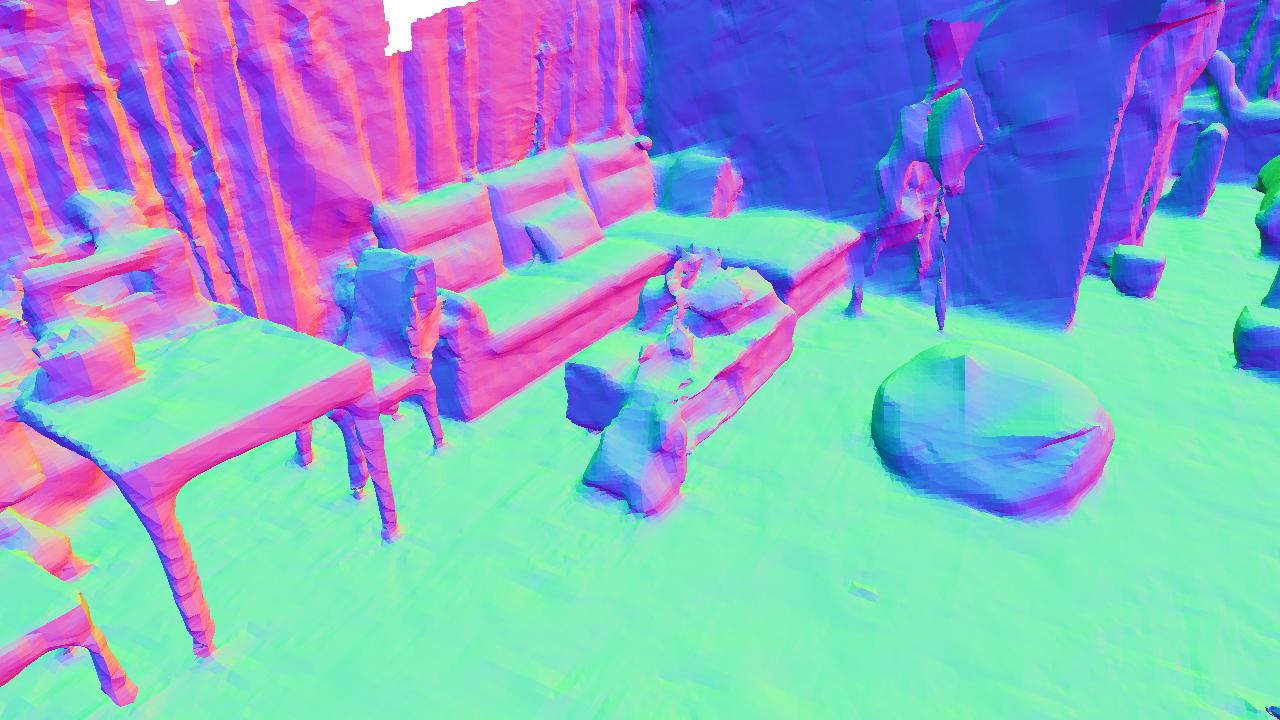}} &
    \makecell{\includegraphics[width=\sz\linewidth]{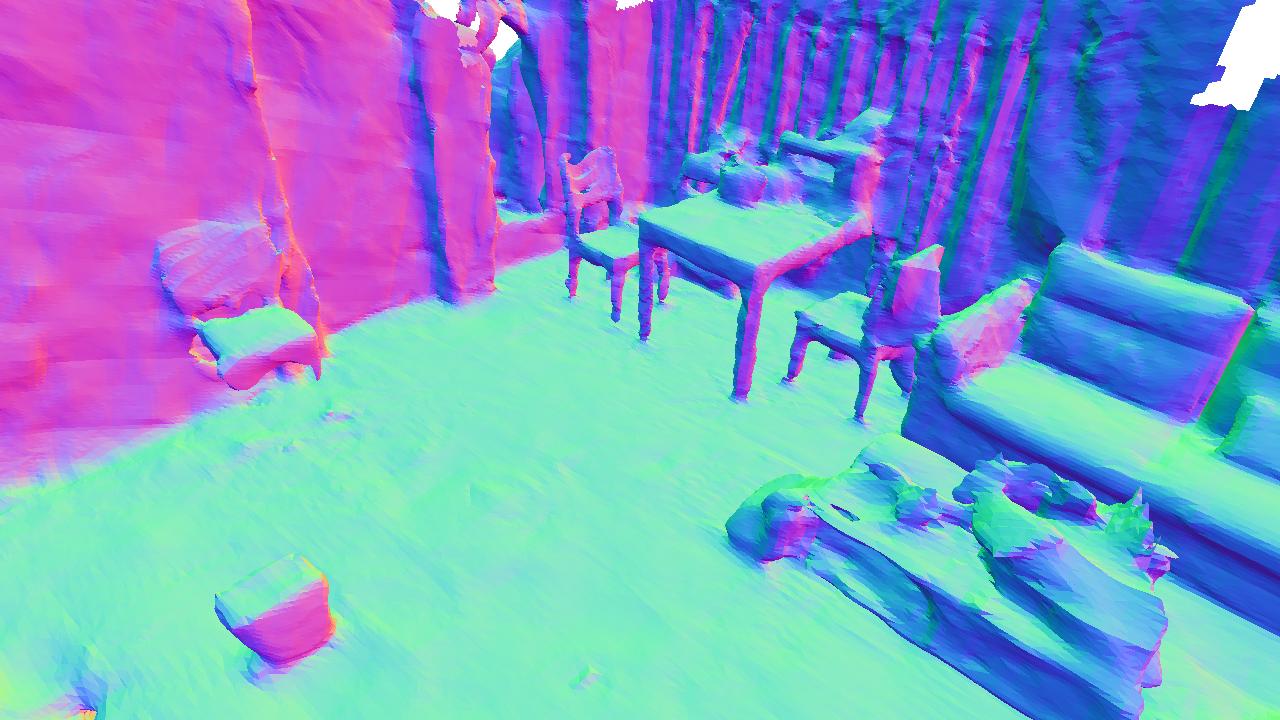}} &
    \makecell{\includegraphics[width=\sz\linewidth]{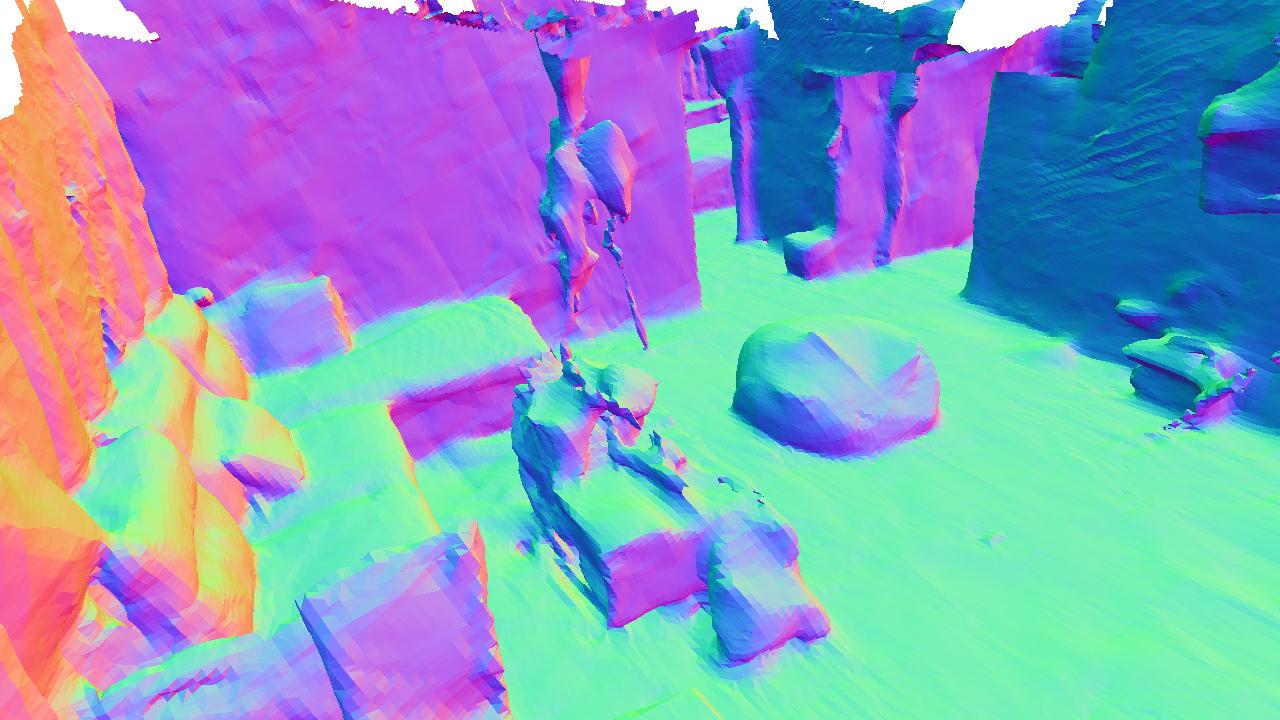}} \\
    \makecell{\rotatebox{90}{\tt Ours}} &
    \makecell{\includegraphics[width=\sz\linewidth]{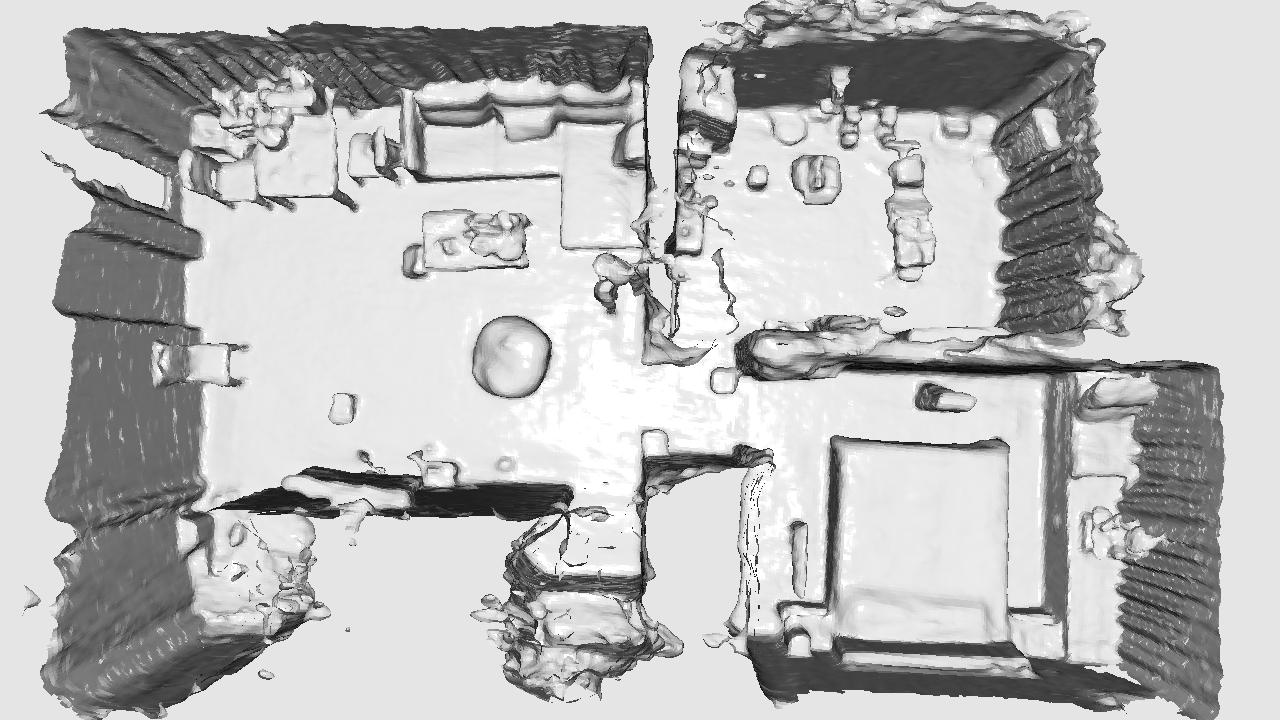}} &
    \makecell{\includegraphics[width=\sz\linewidth]{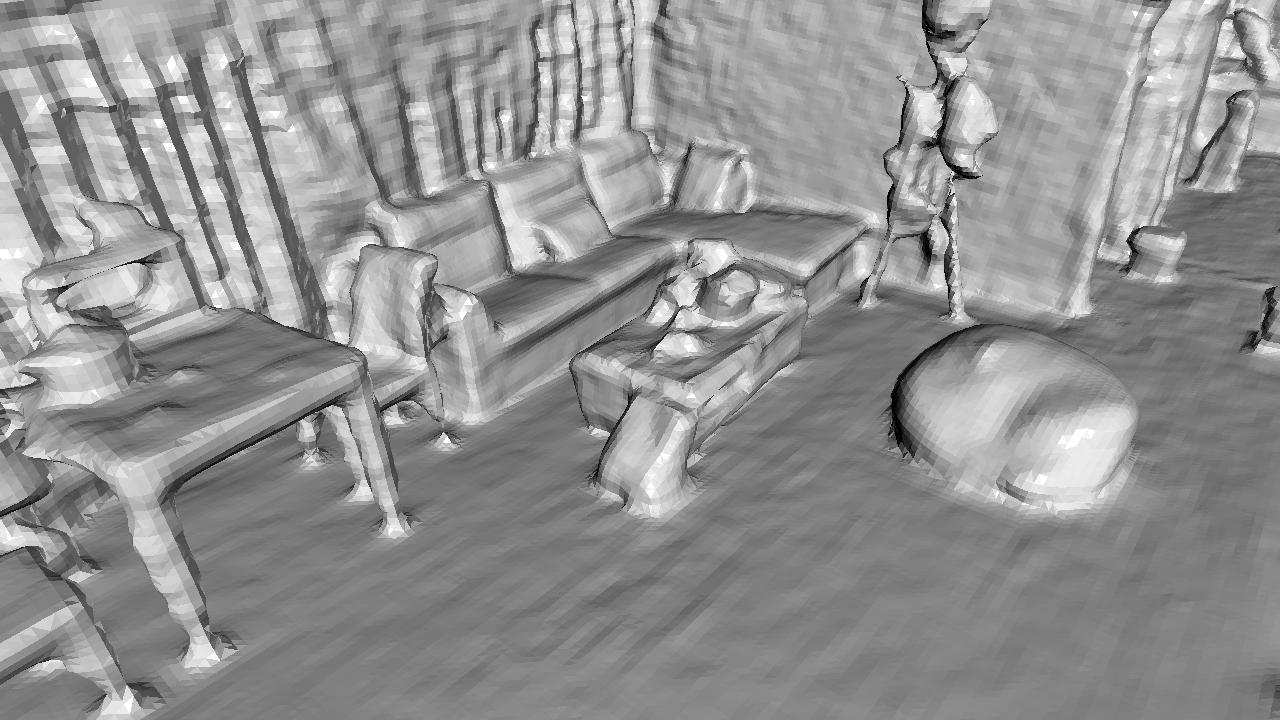}} &
    \makecell{\includegraphics[width=\sz\linewidth]{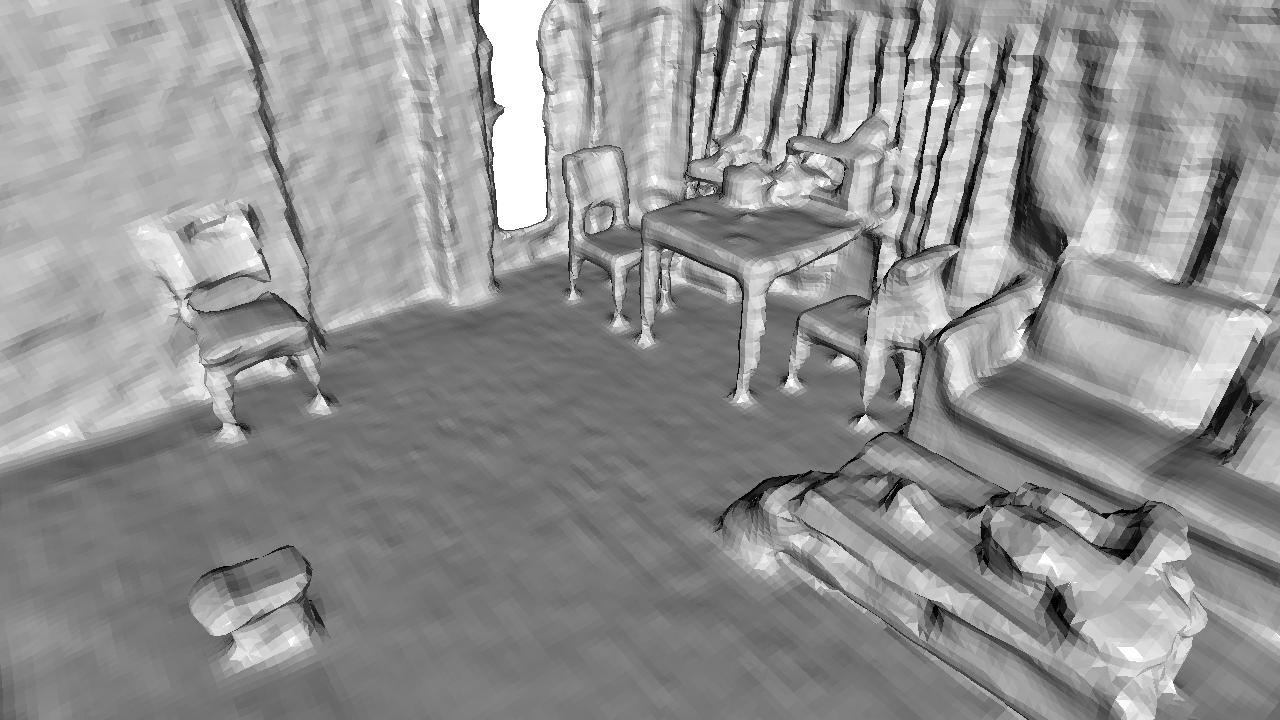}} &
    \makecell{\includegraphics[width=\sz\linewidth]{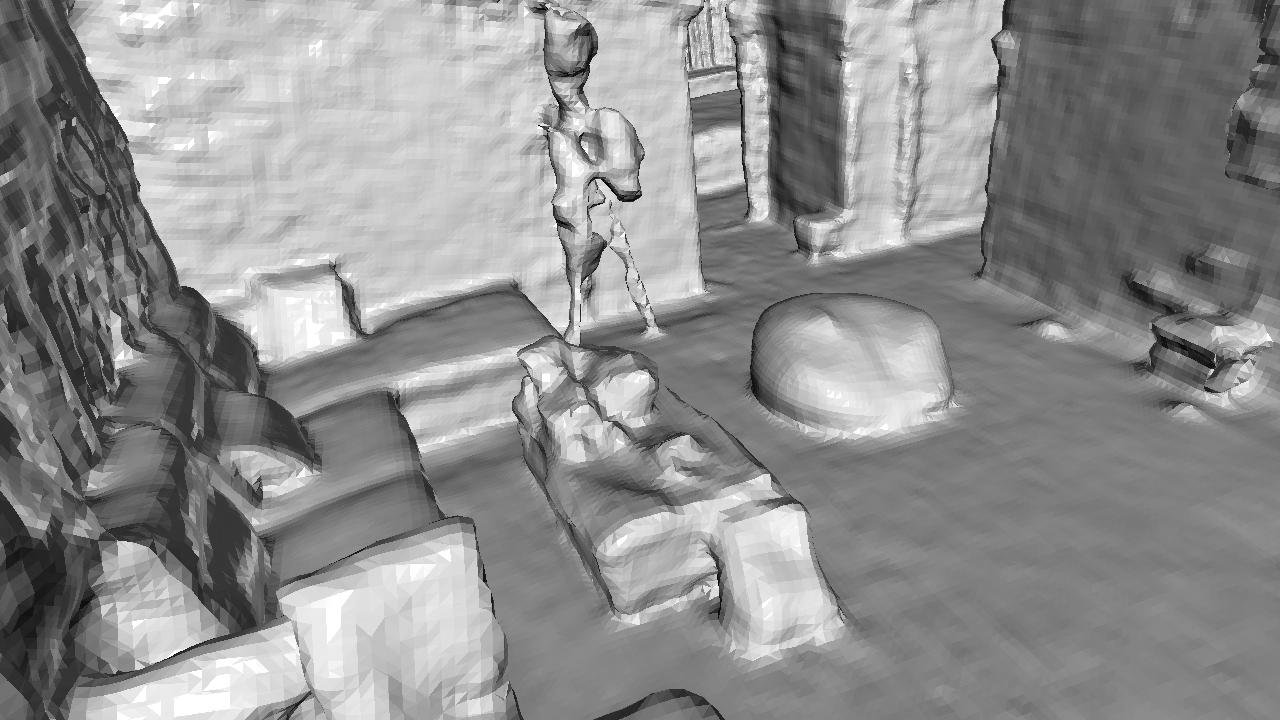}} \\
    \makecell{\rotatebox{90}{\tt NICE-SLAM}} &
    \makecell{\includegraphics[width=\sz\linewidth]{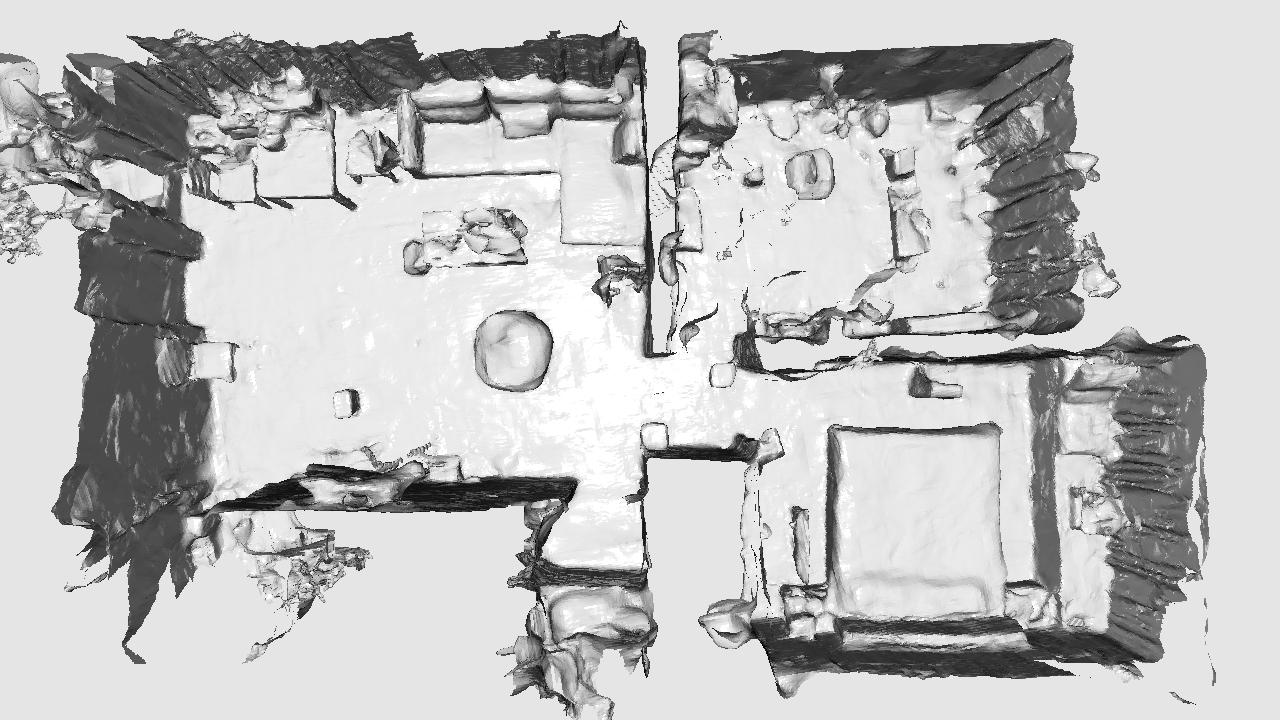}} &
    \makecell{\includegraphics[width=\sz\linewidth]{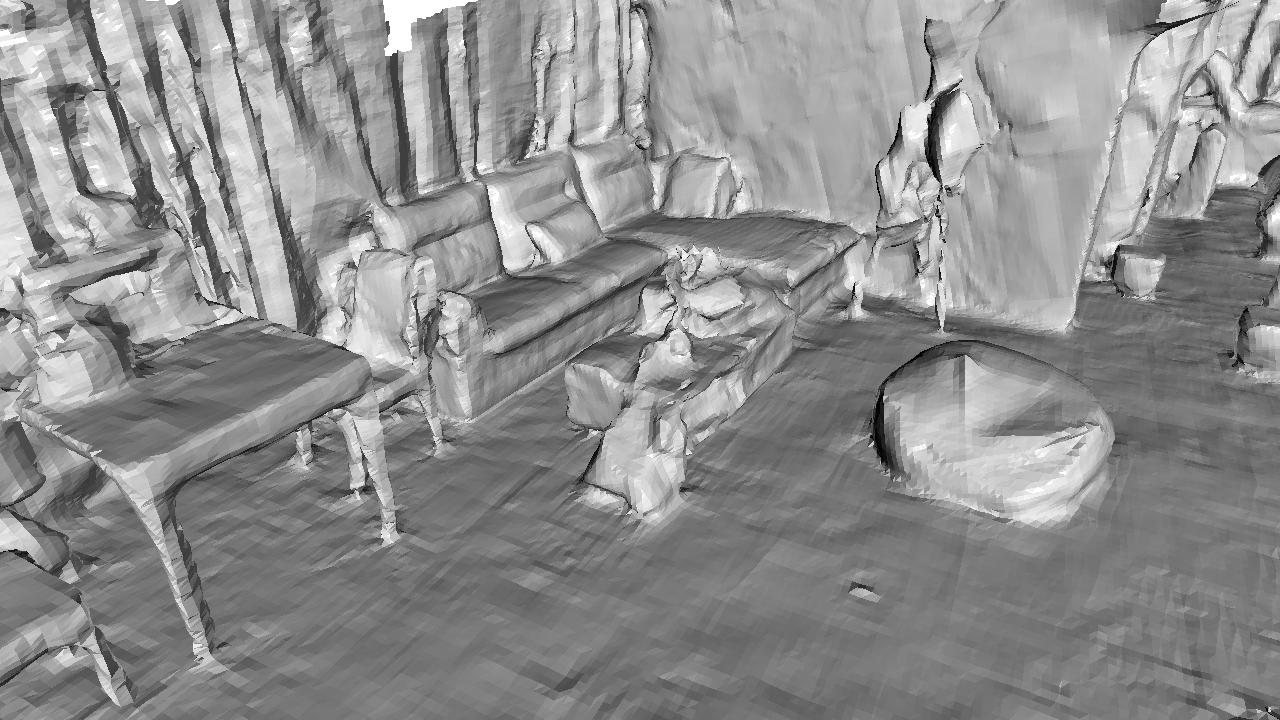}} &
    \makecell{\includegraphics[width=\sz\linewidth]{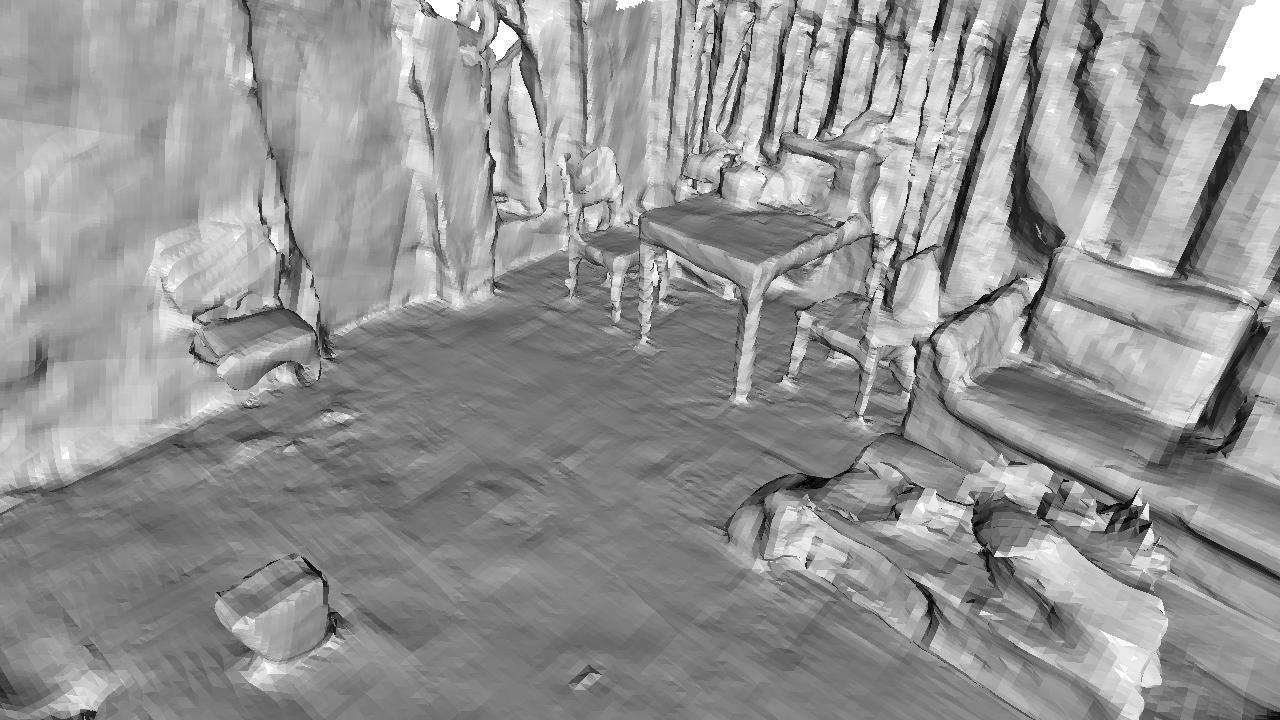}} &
    \makecell{\includegraphics[width=\sz\linewidth]{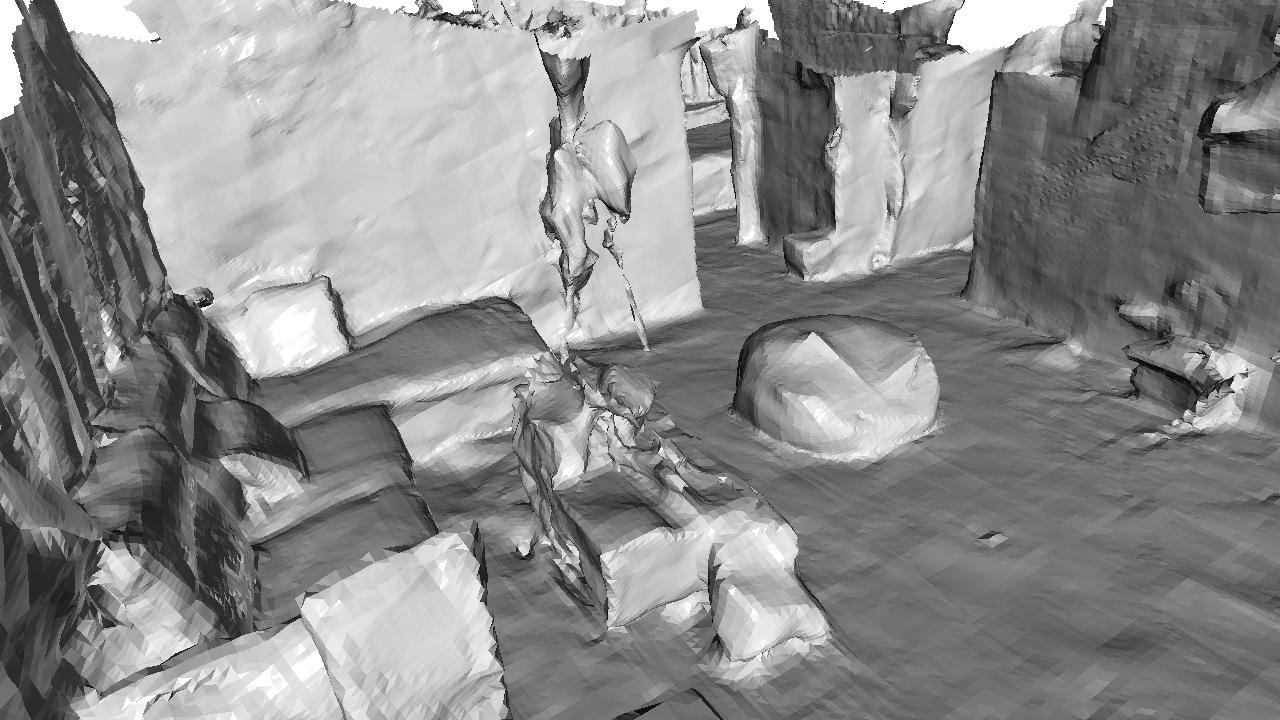}} \\
  \end{tabular} 
  \vspace{-5pt}
  \caption{
  Qualitative comparison on NICE-SLAM apartment sequence on different view-point with different shading mode. Co-SLAM achieves smooth, detailed and high-fidelity reconstruction while running $>10$ times faster.
  }
  \label{fig:apartment}
\end{figure*}

\noindent\textbf{NICE-SLAM apartment~\cite{zhuNiceslamNeuralImplicit2022}.} In addition to the 6 sequences from ScanNet, we also compared Co-SLAM and NICE-SLAM on the apartment sequence collected by the authors of NICE-SLAM using Azure Kinect depth camera. We run Co-SLAM with our ScanNet setting. We show qualitative comparison from different views in Fig.~\ref{fig:apartment}. As can be seen Co-SLAM achieves smoother and better quality reconstruction in much shorter time (40 minutes vs. 10 hours).

\begin{figure*}[htbp]
  \centering
  \footnotesize
  \setlength{\tabcolsep}{1.5pt}
  \newcommand{\sz}{0.24}
  \begin{tabular}{lcccc}
    & \texttt{top-down} & \texttt{view-1} & \texttt{view-2} & \texttt{view-3} \\
    \makecell{\rotatebox{90}{\tt Ours}} &
    \makecell{\includegraphics[width=\sz\linewidth]{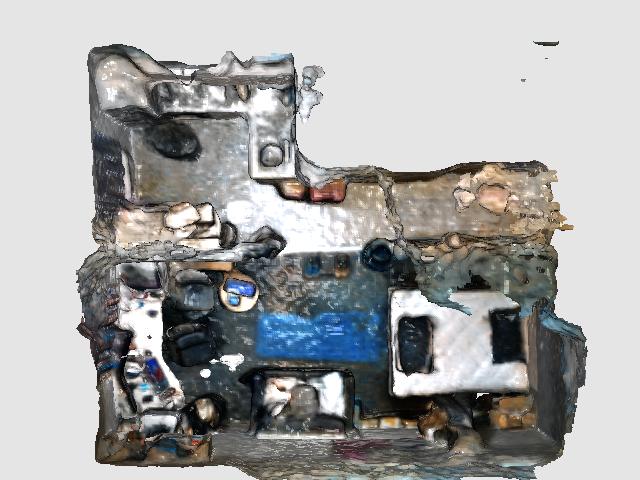}} &
    \makecell{\includegraphics[width=\sz\linewidth]{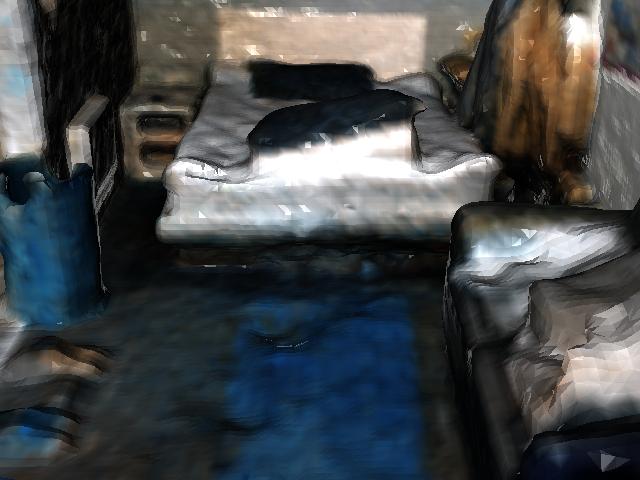}} &
    \makecell{\includegraphics[width=\sz\linewidth]{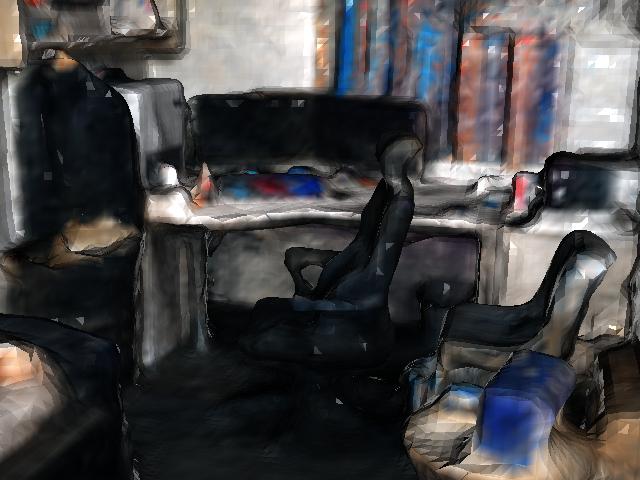}} &
    \makecell{\includegraphics[width=\sz\linewidth]{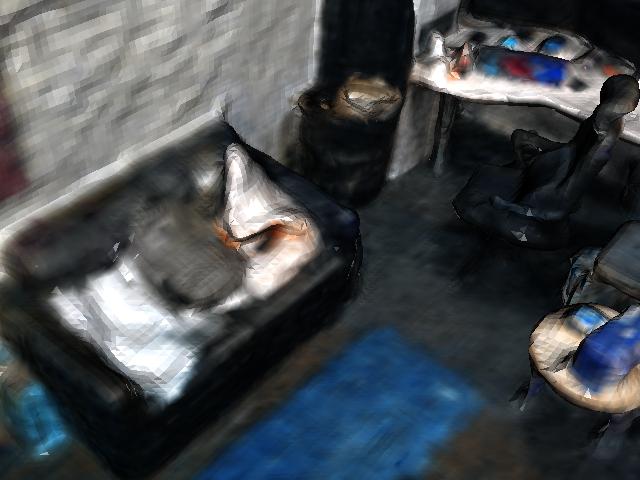}} \\
    \makecell{\rotatebox{90}{\tt NICE-SLAM}} &
    \makecell{\includegraphics[width=\sz\linewidth]{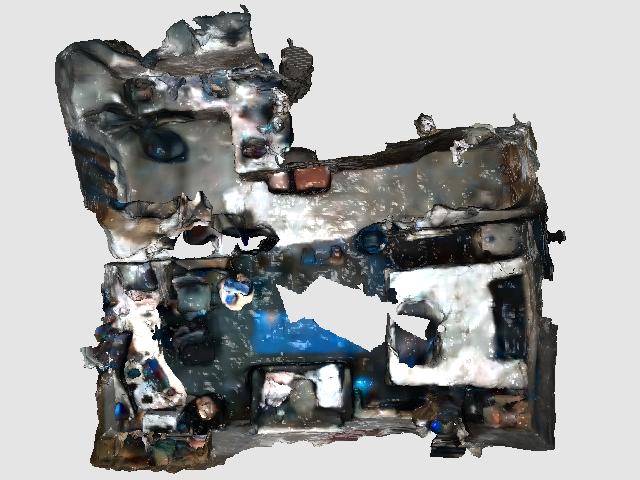}} &
    \makecell{\includegraphics[width=\sz\linewidth]{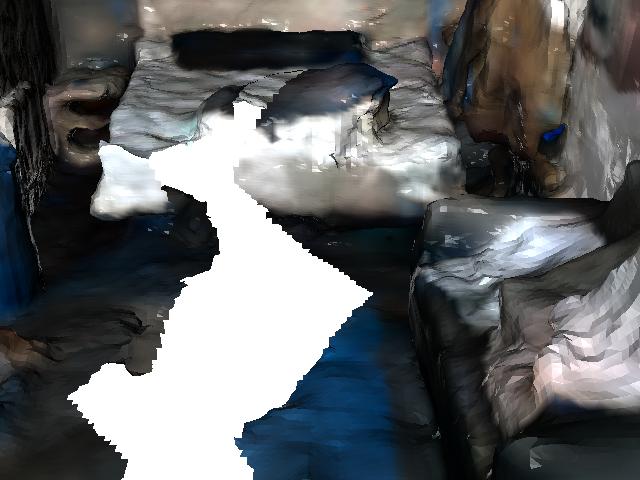}} &
    \makecell{\includegraphics[width=\sz\linewidth]{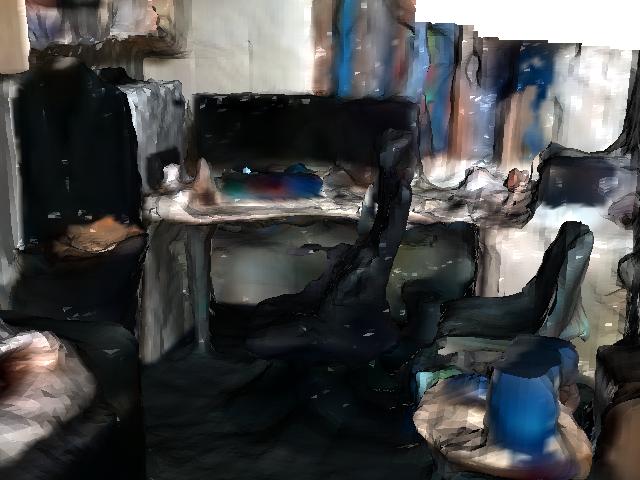}} &
    \makecell{\includegraphics[width=\sz\linewidth]{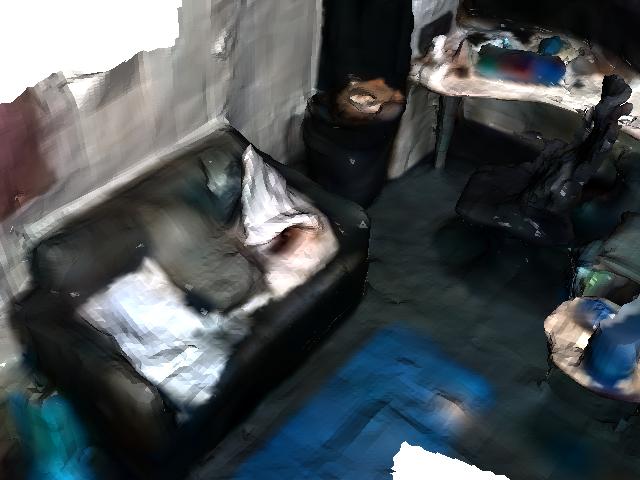}} \\
    \makecell{\rotatebox{90}{\tt Ours}} &
    \makecell{\includegraphics[width=\sz\linewidth]{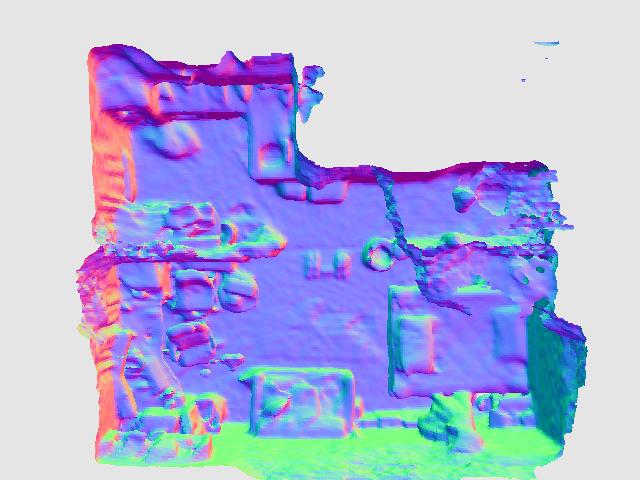}} &
    \makecell{\includegraphics[width=\sz\linewidth]{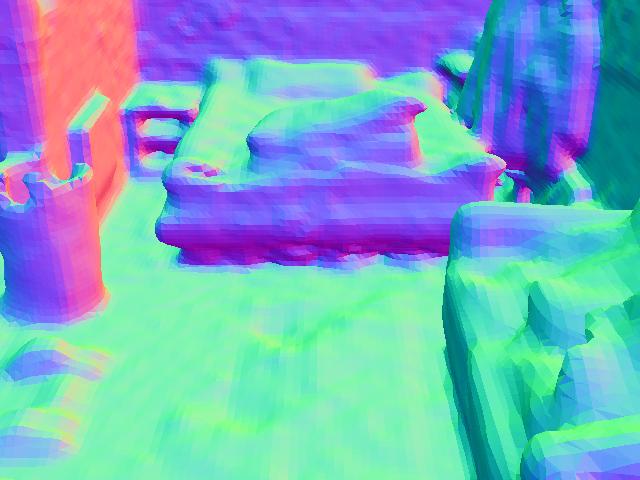}} &
    \makecell{\includegraphics[width=\sz\linewidth]{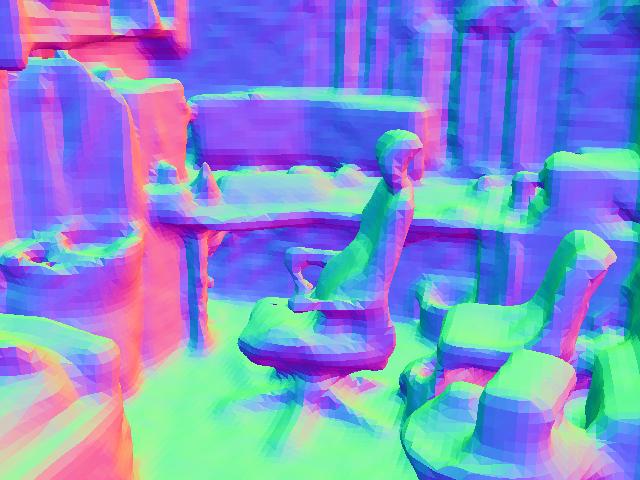}} &
    \makecell{\includegraphics[width=\sz\linewidth]{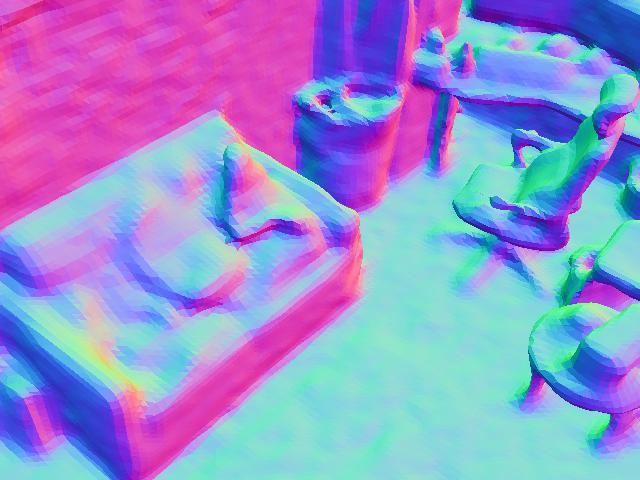}} \\
    \makecell{\rotatebox{90}{\tt NICE-SLAM}} &
    \makecell{\includegraphics[width=\sz\linewidth]{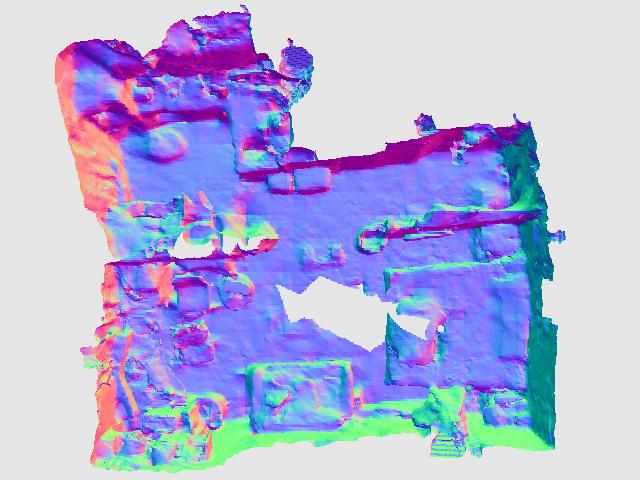}} &
    \makecell{\includegraphics[width=\sz\linewidth]{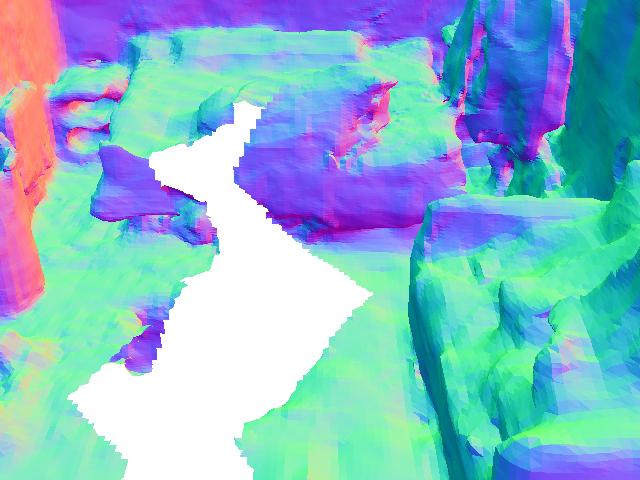}} &
    \makecell{\includegraphics[width=\sz\linewidth]{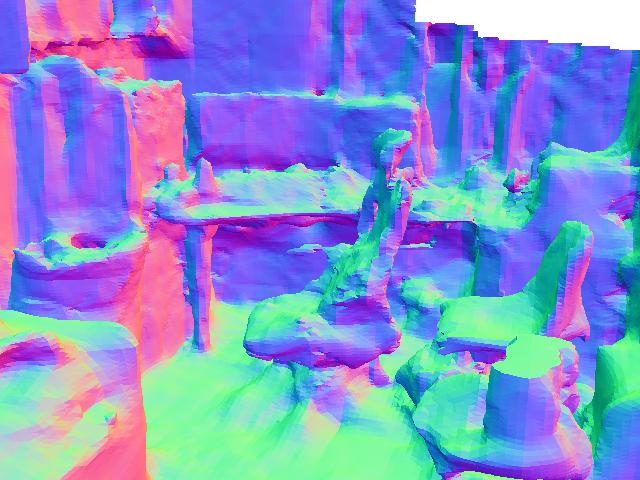}} &
    \makecell{\includegraphics[width=\sz\linewidth]{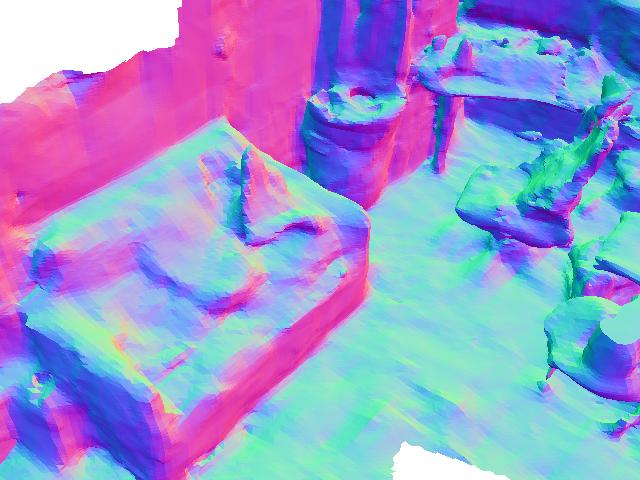}} \\
    \makecell{\rotatebox{90}{\tt Ours}} &
    \makecell{\includegraphics[width=\sz\linewidth]{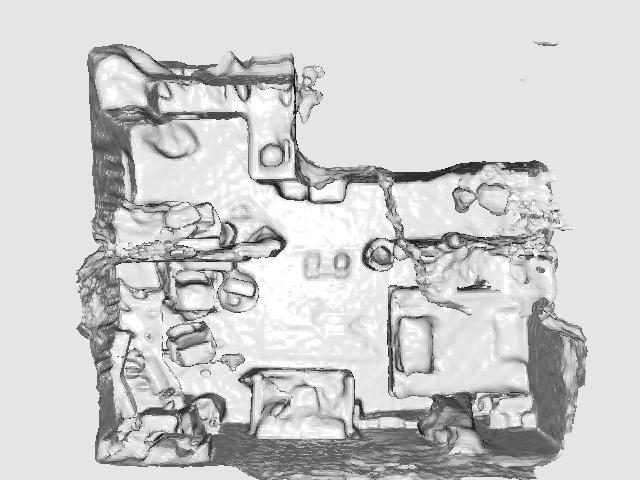}} &
    \makecell{\includegraphics[width=\sz\linewidth]{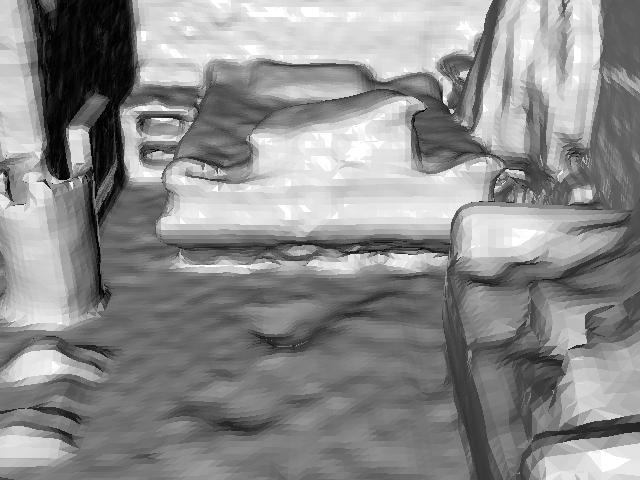}} &
    \makecell{\includegraphics[width=\sz\linewidth]{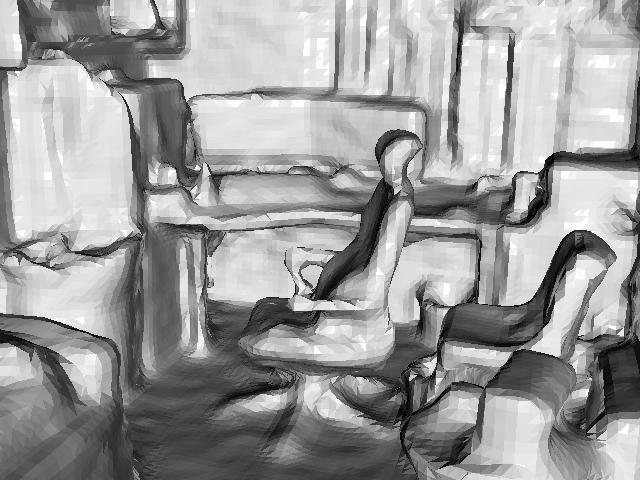}} &
    \makecell{\includegraphics[width=\sz\linewidth]{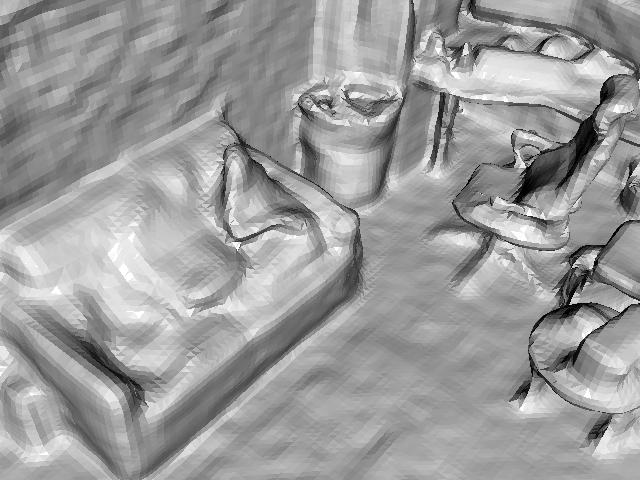}} \\
    \makecell{\rotatebox{90}{\tt NICE-SLAM}} &
    \makecell{\includegraphics[width=\sz\linewidth]{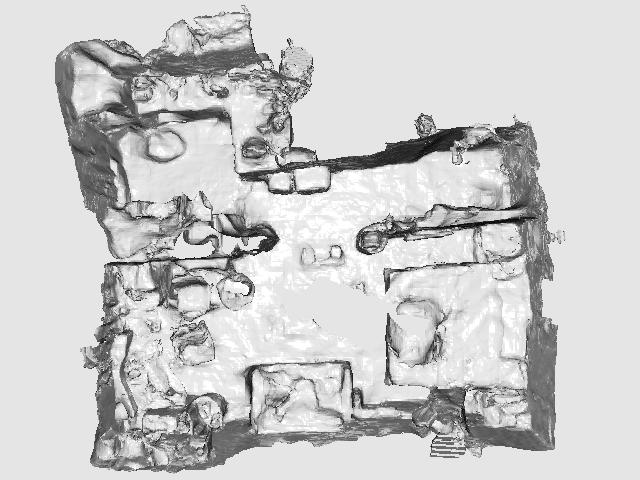}} &
    \makecell{\includegraphics[width=\sz\linewidth]{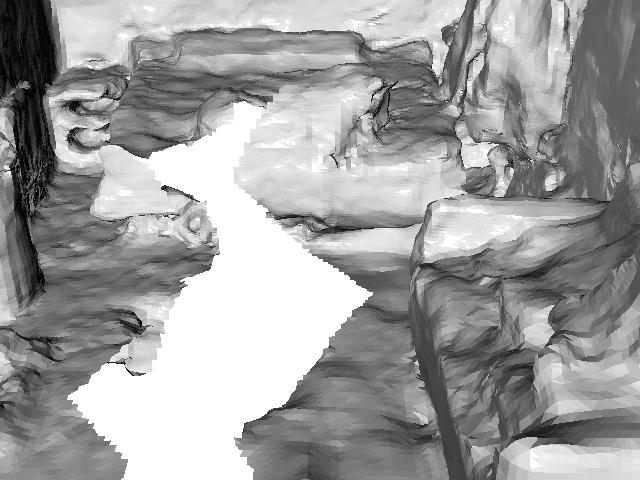}} &
    \makecell{\includegraphics[width=\sz\linewidth]{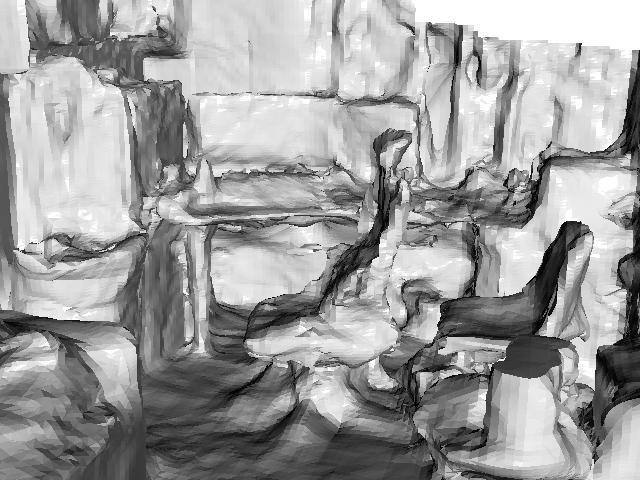}} &
    \makecell{\includegraphics[width=\sz\linewidth]{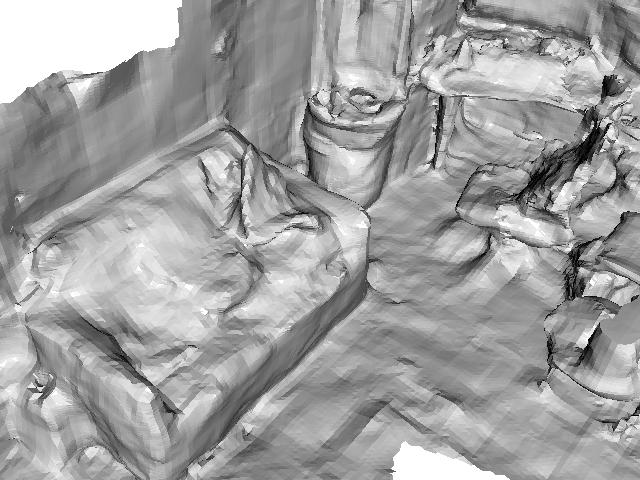}} \\
  \end{tabular} 
  \vspace{-5pt}
  \caption{Qualitative comparison on self-captured room sequence on different view-point with different shading mode. Overall Co-SLAM produces higher quality surface reconstruction with finer details (the desk chair, the objects on the desk and sofa, the curtain, etc). Also note that NICE-SLAM lost tracking slightly causing the reconstructed scene to be torn apart.}
  \label{fig:my_room07}
\end{figure*}

\begin{figure*}[htbp]
  \centering
  \footnotesize
  \setlength{\tabcolsep}{1.5pt}
  \newcommand{\sz}{0.24}
  \begin{tabular}{lcccc}
    & \texttt{top-down} & \texttt{view-1} & \texttt{view-2} & \texttt{view-3} \\
    \makecell{\rotatebox{90}{\tt Ours}} &
    \makecell{\includegraphics[width=\sz\linewidth]{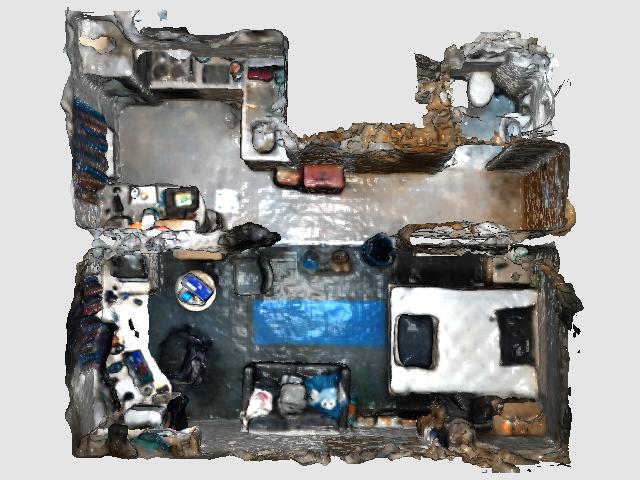}} &
    \makecell{\includegraphics[width=\sz\linewidth]{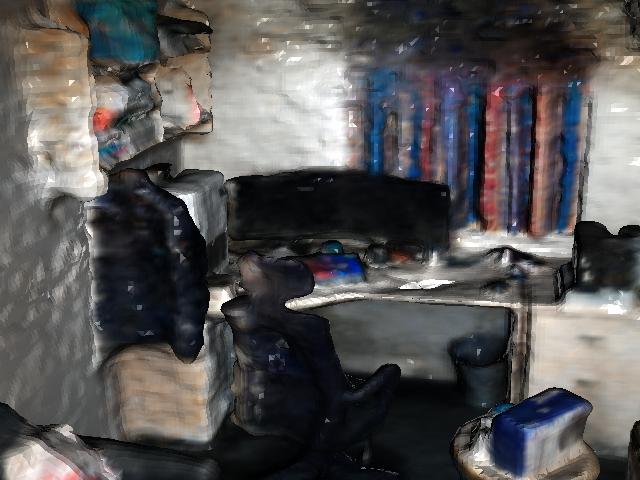}} &
    \makecell{\includegraphics[width=\sz\linewidth]{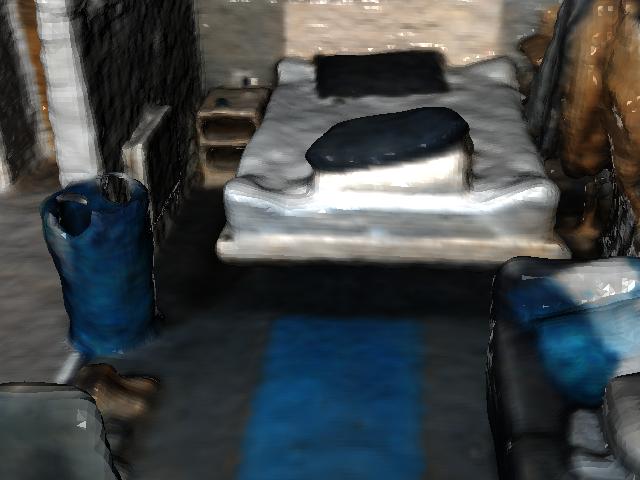}} &
    \makecell{\includegraphics[width=\sz\linewidth]{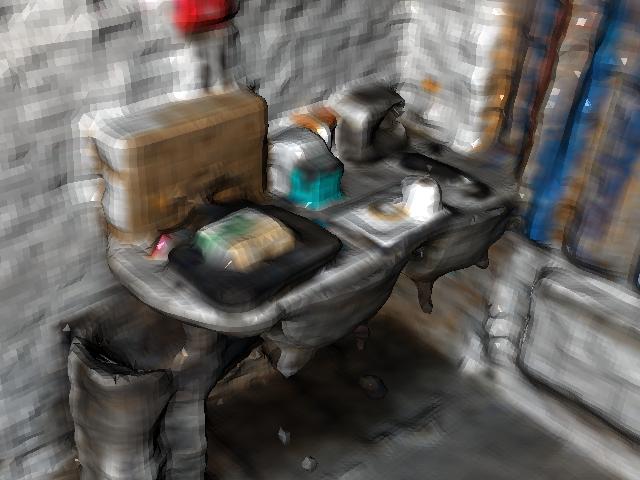}} \\
    \makecell{\rotatebox{90}{\tt NICE-SLAM}} &
    \makecell{\includegraphics[width=\sz\linewidth]{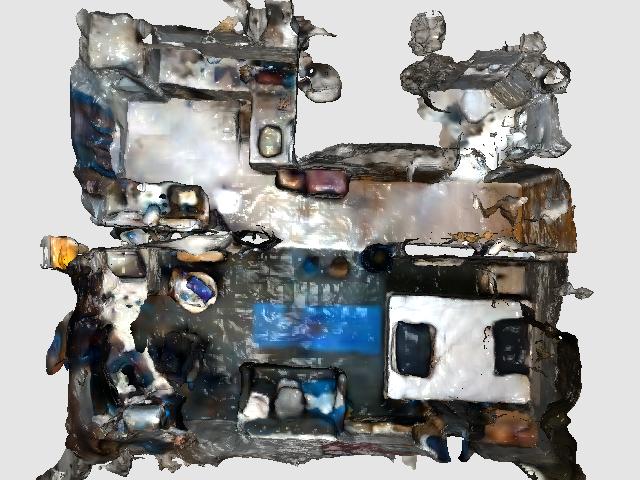}} &
    \makecell{\includegraphics[width=\sz\linewidth]{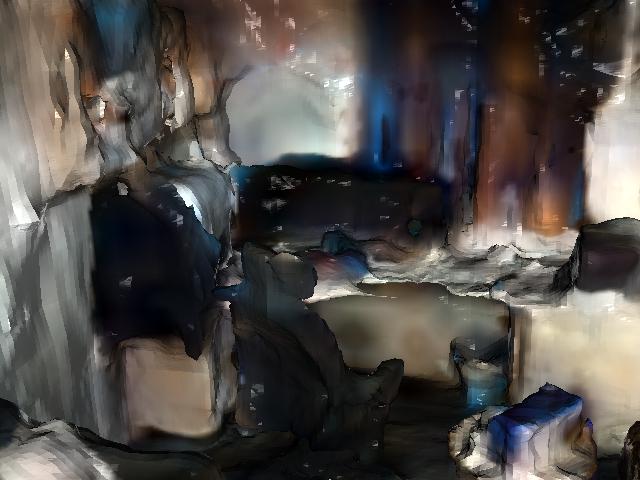}} &
    \makecell{\includegraphics[width=\sz\linewidth]{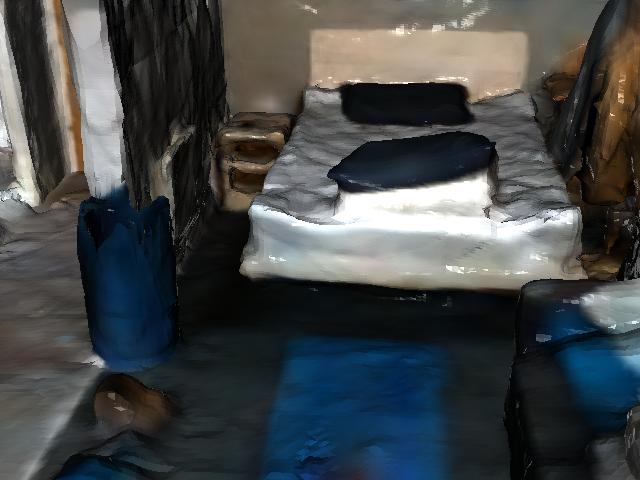}} &
    \makecell{\includegraphics[width=\sz\linewidth]{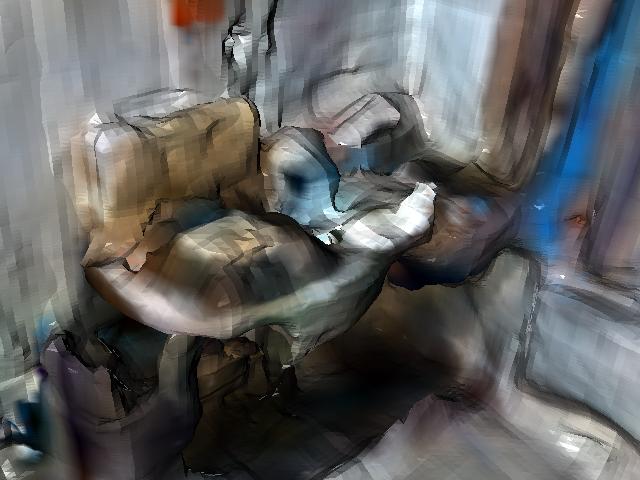}} \\
    \makecell{\rotatebox{90}{\tt Ours}} &
    \makecell{\includegraphics[width=\sz\linewidth]{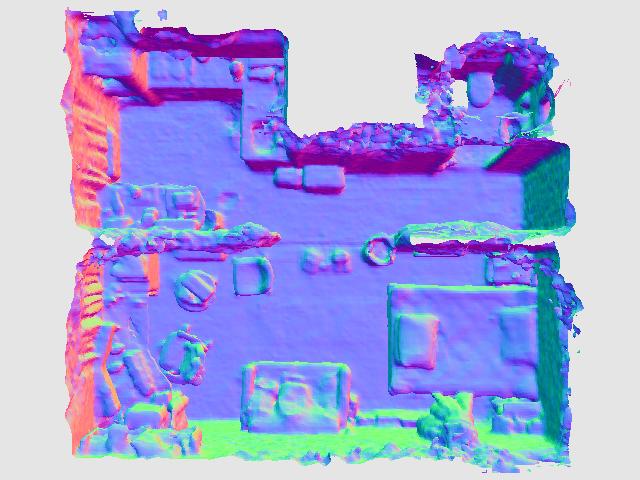}} &
    \makecell{\includegraphics[width=\sz\linewidth]{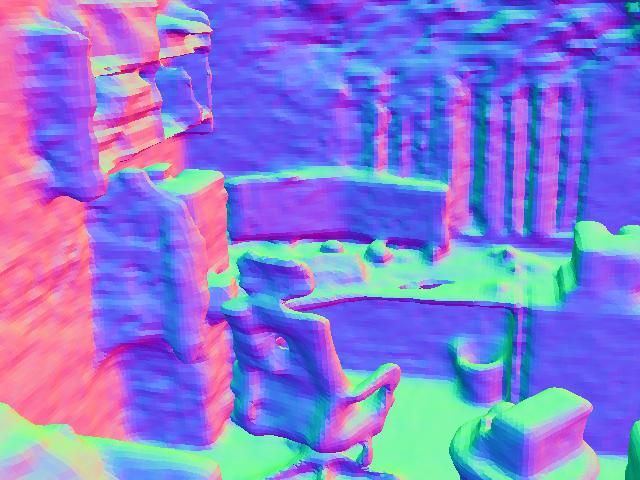}} &
    \makecell{\includegraphics[width=\sz\linewidth]{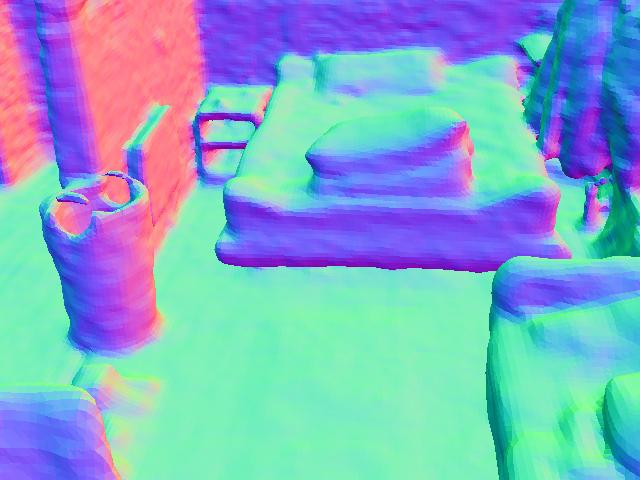}} &
    \makecell{\includegraphics[width=\sz\linewidth]{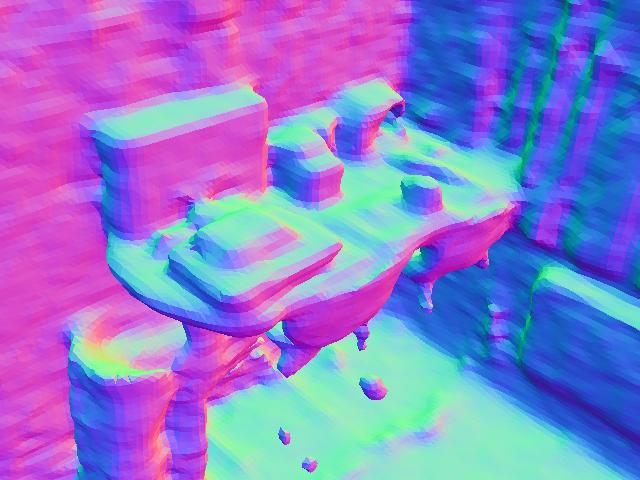}} \\
    \makecell{\rotatebox{90}{\tt NICE-SLAM}} &
    \makecell{\includegraphics[width=\sz\linewidth]{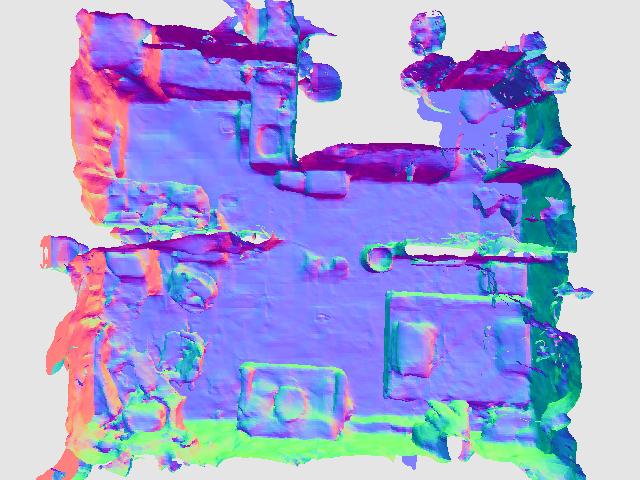}} &
    \makecell{\includegraphics[width=\sz\linewidth]{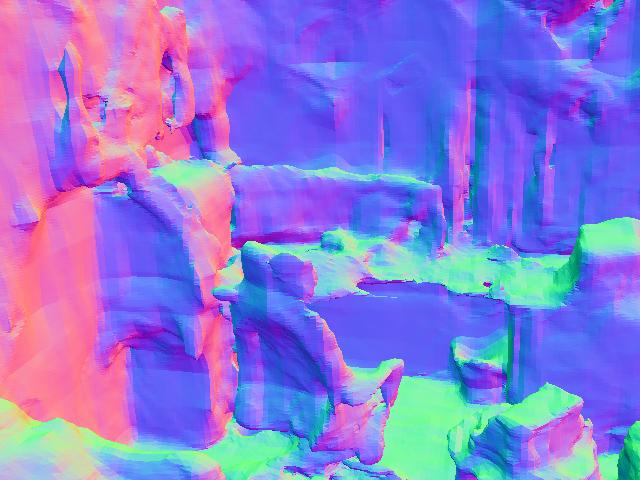}} &
    \makecell{\includegraphics[width=\sz\linewidth]{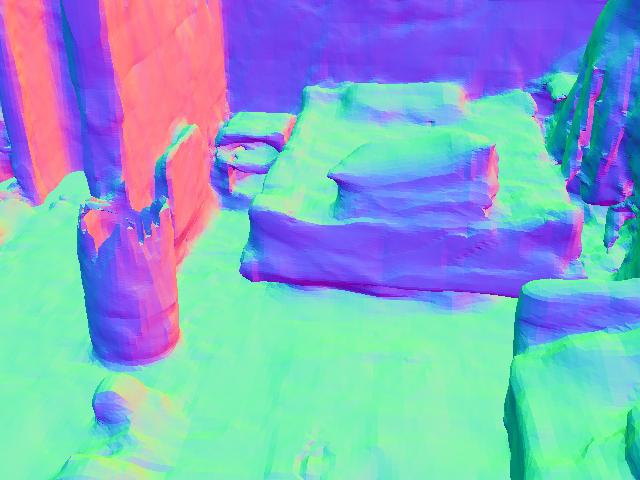}} &
    \makecell{\includegraphics[width=\sz\linewidth]{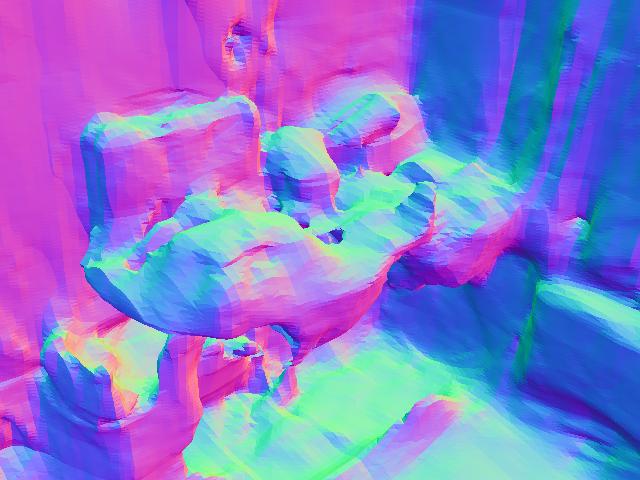}} \\
    \makecell{\rotatebox{90}{\tt Ours}} &
    \makecell{\includegraphics[width=\sz\linewidth]{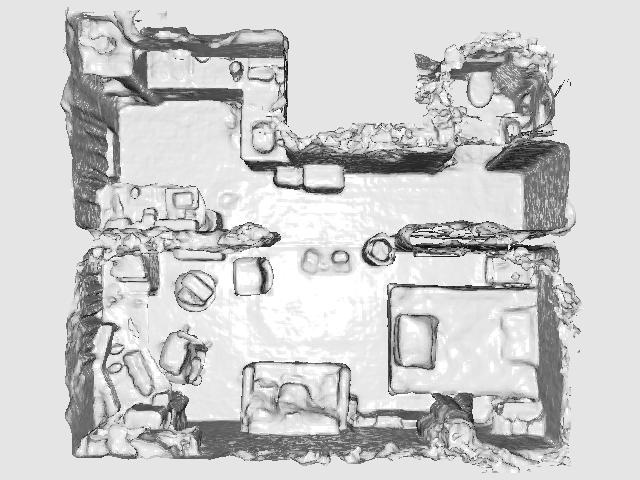}} &
    \makecell{\includegraphics[width=\sz\linewidth]{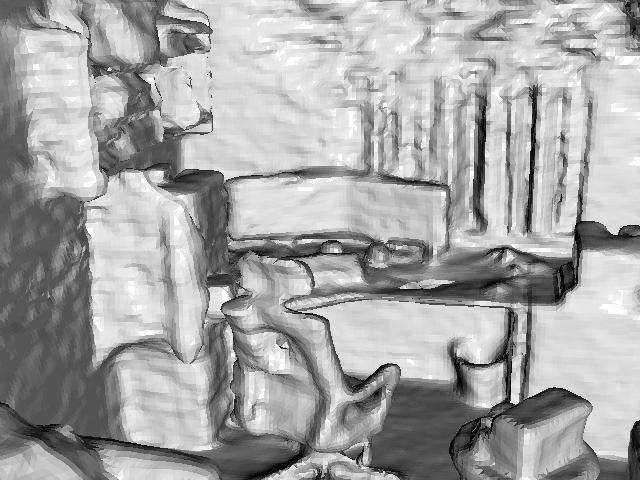}} &
    \makecell{\includegraphics[width=\sz\linewidth]{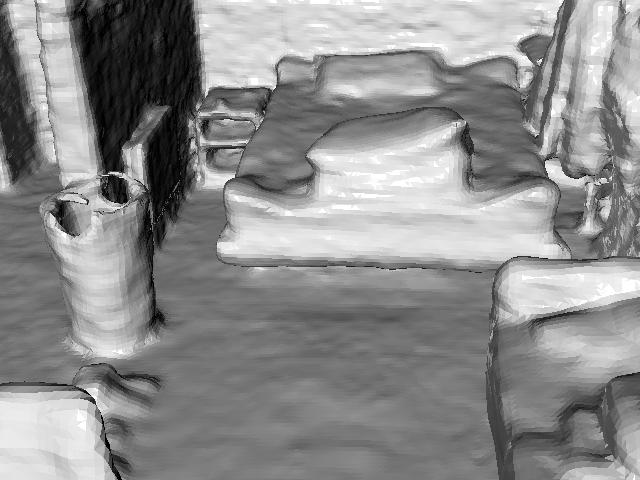}} &
    \makecell{\includegraphics[width=\sz\linewidth]{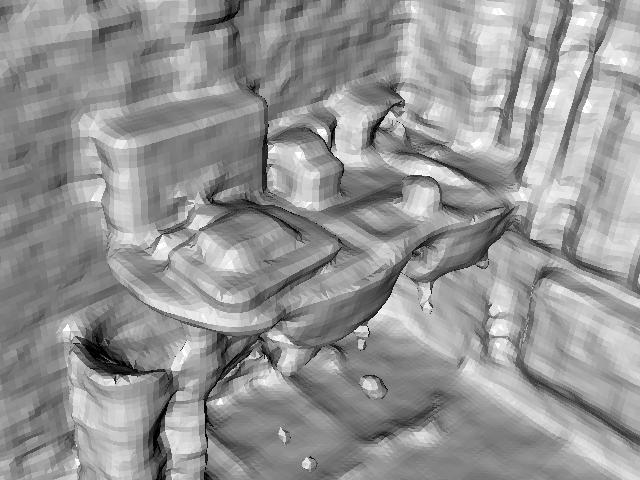}} \\
    \makecell{\rotatebox{90}{\tt NICE-SLAM}} &
    \makecell{\includegraphics[width=\sz\linewidth]{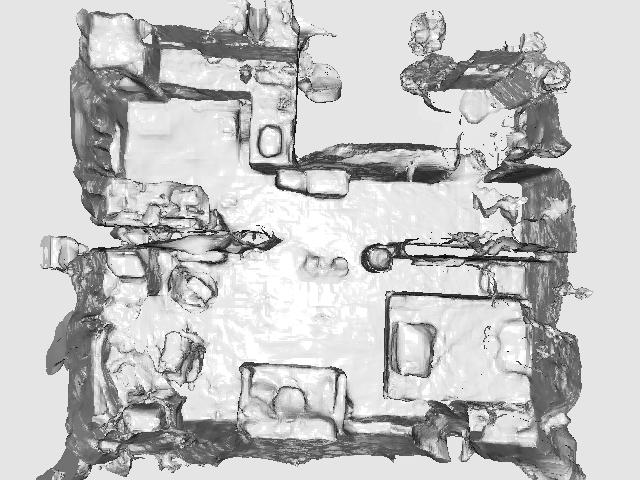}} &
    \makecell{\includegraphics[width=\sz\linewidth]{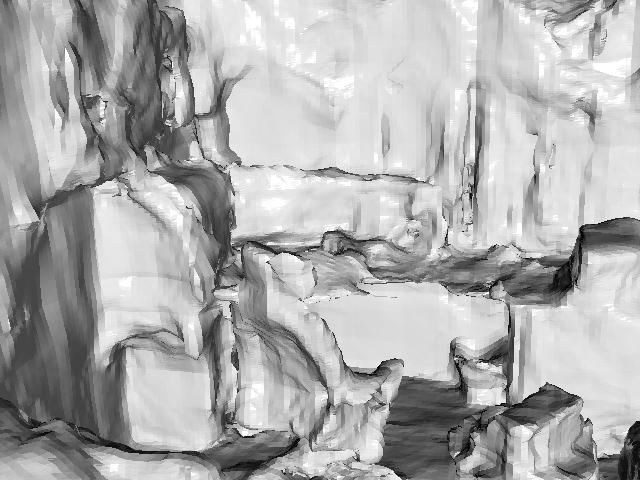}} &
    \makecell{\includegraphics[width=\sz\linewidth]{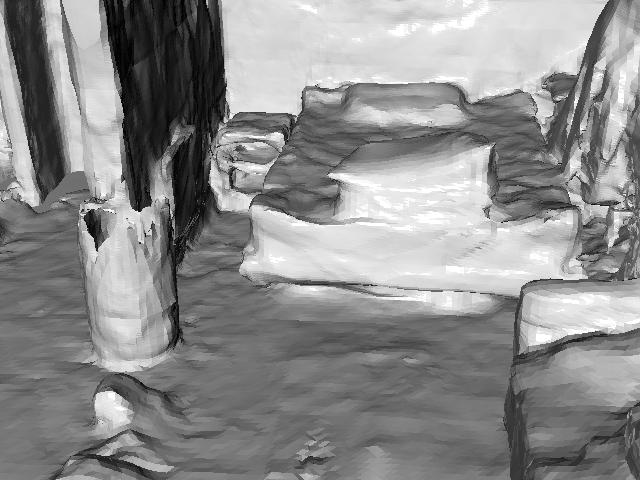}} &
    \makecell{\includegraphics[width=\sz\linewidth]{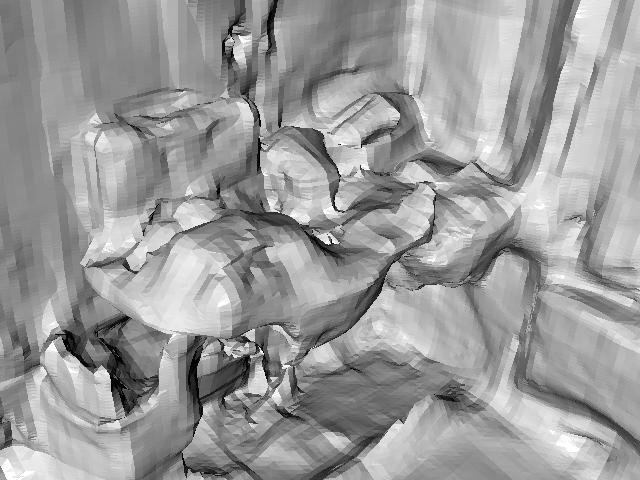}} \\
  \end{tabular} 
  \vspace{-5pt}
  \caption{Qualitative comparison on a different scan of the same room in Fig.~\ref{fig:my_room07}. Note how Co-SLAM produces better surface reconstruction while running $> 10$ time faster. }
  \label{fig:my_room10}
\end{figure*}

\noindent\textbf{Self-captured room.} In addition to ScanNet sequences and the apartment sequences captured by the authors of NICE-SLAM, we also collected our two real-world indoor sequences using RealSense D435i depth camera, whose depth quality is slightly worse than Azure Kinect. We show qualitative comparison in Fig.~\ref{fig:my_room07} and Fig.~\ref{fig:my_room10}.

\subsection{ScanNet Camera Tracking Results}

In this section, we provide a more comprehensive view of the camera tracking results on the ScanNet dataset. In evaluating the absolute trajectory error (ATE), a rigid transformation is estimated to align the estimated trajectory with the ground truth. While this protocol is widely used in traditional SLAM and also in NICE-SLAM~\cite{zhuNiceslamNeuralImplicit2022}, we observe that this does not always tell the whole story. For example, Fig.~\ref{fig:ATE} shows the reconstructed ScanNet scene with estimated camera trajectory under the same world coordinate, i.e. \textbf{without} doing the trajectory alignment. It can be seen that \methodname performs better in terms of camera tracking (Note the top-left and top-right corner of the trajectory) and leads to less distorted reconstruction. However, this is not reflected in Tab. 4 in our main paper as trajectory alignment. Therefore in Tab.~\ref{tab:ATE_no_align} we report the full camera tracking results both with and without trajectory alignment. As can be seen, \methodname achieves overall better and more robust tracking results.

\subsection{More Ablation Studies}

\noindent\textbf{Using separate color grid.} As dedcribed om our main paper, we adopted two separate MLP decoders for color and geometry but only use a single hash-grid. In Tab~\ref{tab:twogrid} we shows the comparison of using two separate hash grids for color and geometry, and using one hash grid (our default setting). We empirically discover that thanks to our joint coordinate and sparse parametric encoding, using a single hash-grid already achieves similar tracking accuracy and reconstruction quality while is more computational efficient and requires much less memory storage.

\begin{figure}[t]
  \centering
  \setlength{\tabcolsep}{1.5pt}
  \newcommand{\sz}{0.48}
  \begin{tabular}{cc}
 
    \includegraphics[width=\sz\columnwidth]{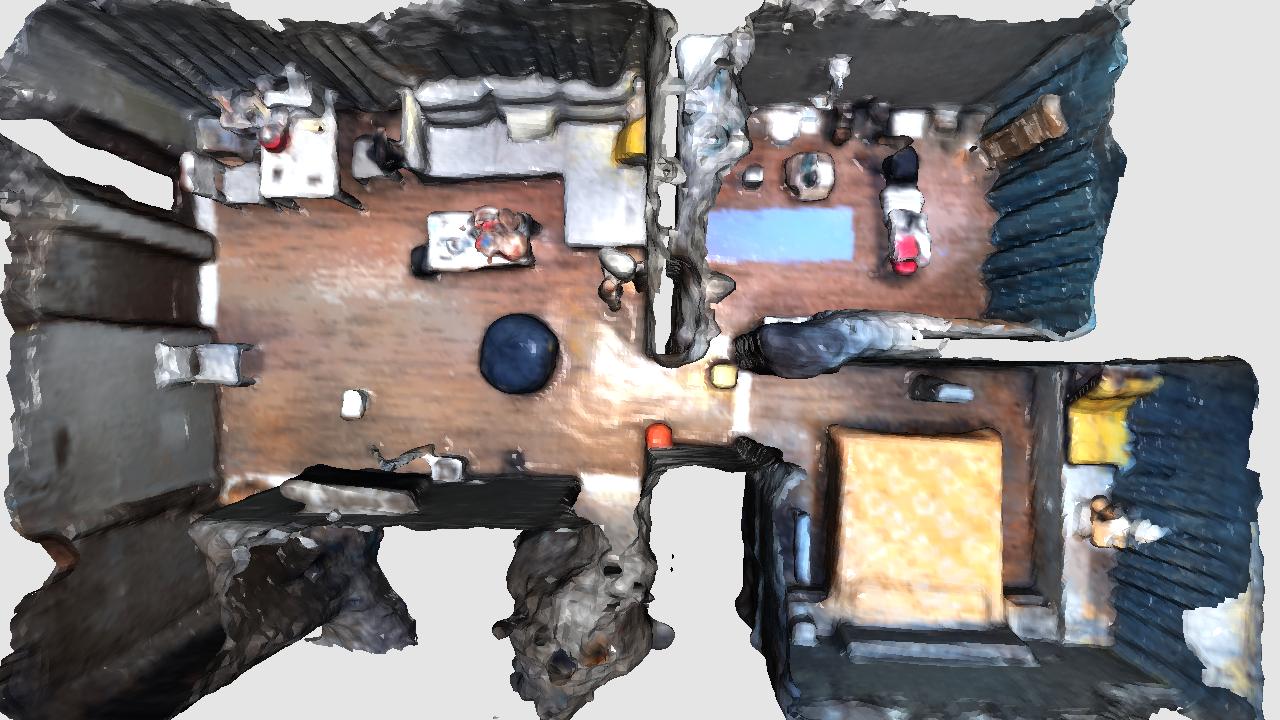} &
    \includegraphics[width=\sz\columnwidth]{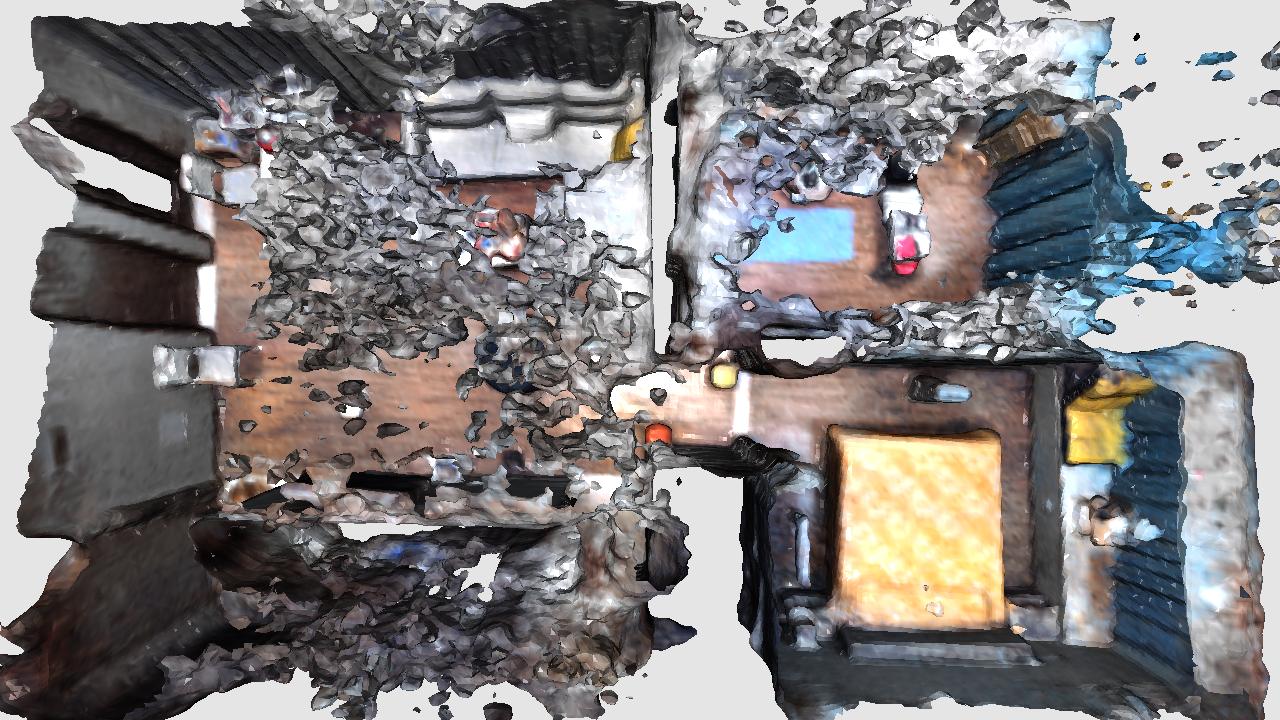} \\
    \includegraphics[width=\sz\columnwidth]{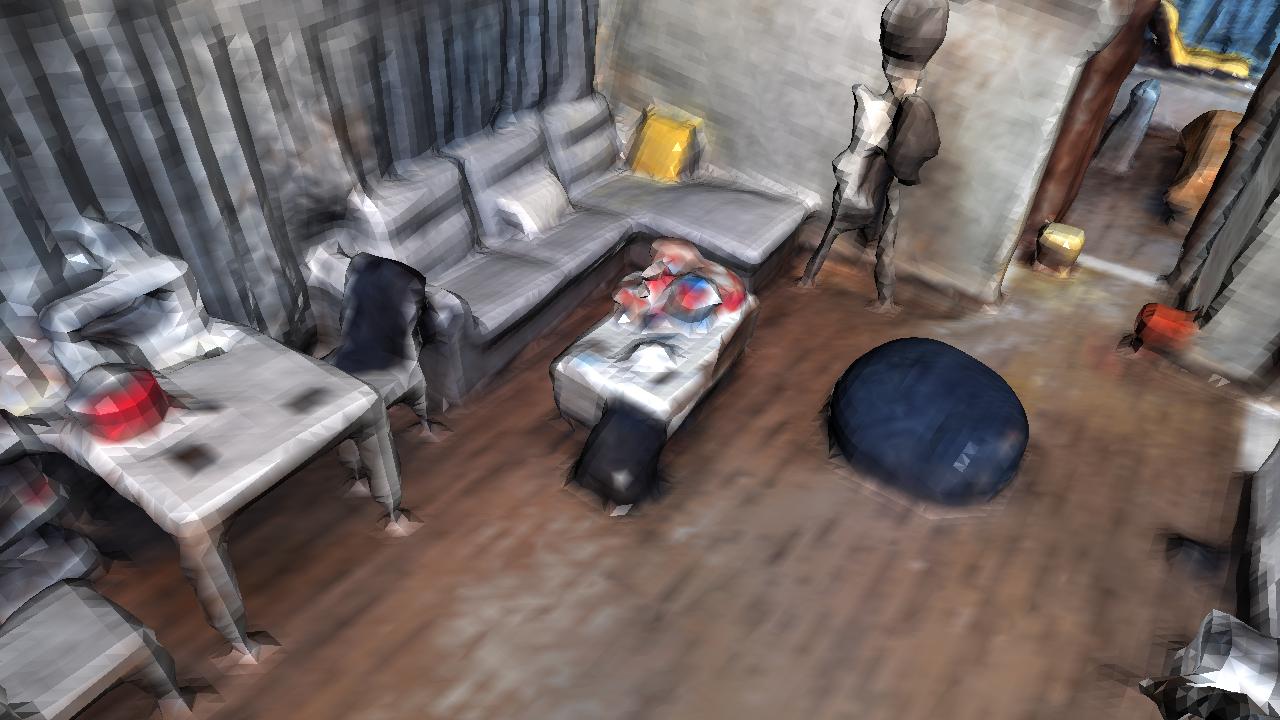} &
    \includegraphics[width=\sz\columnwidth]{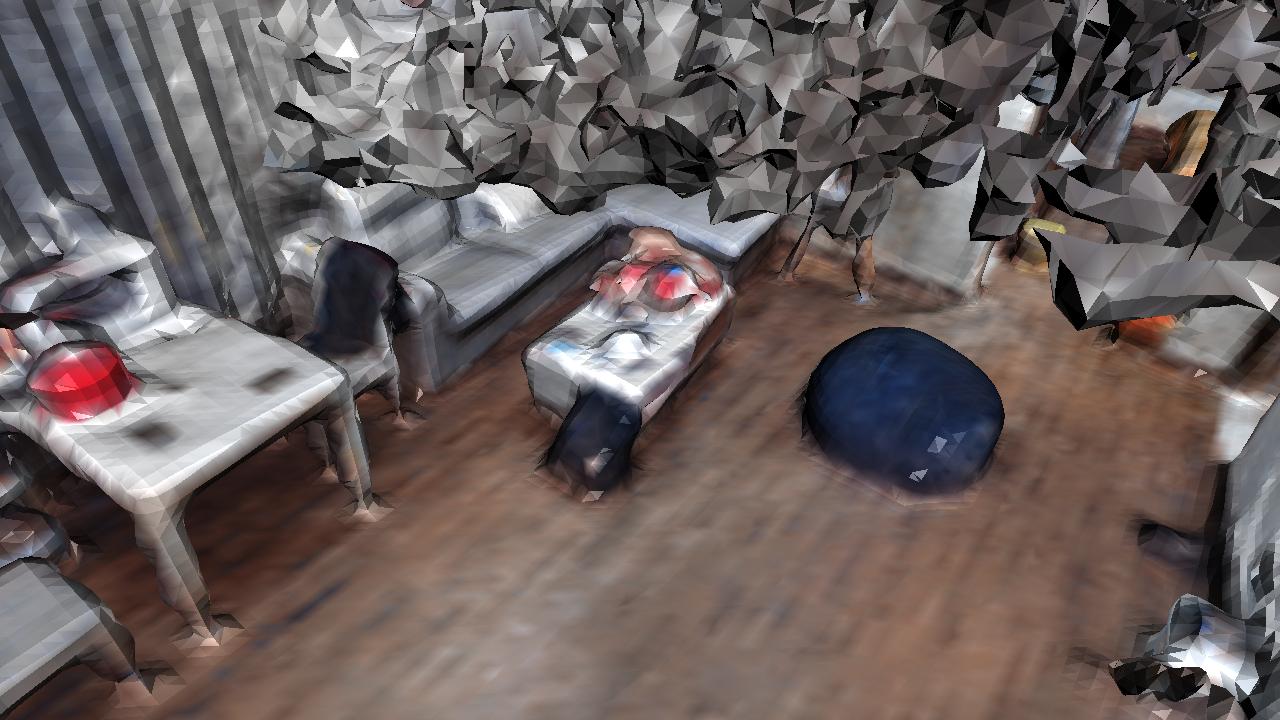} \\
    \texttt{w/ smooth term} & \texttt{w/o smooth term} \\
    
  \end{tabular} 
  \vspace{-5pt}
  \caption{Reconstruction result w/ and w/o our smoothness term. Smoothness term could effectively remove hash collision artefacts.}
  \label{fig:ab_smooth}
  \vspace{-12pt}
\end{figure}

\noindent\textbf{Effect of smoothness term.} We also conduct ablation study on the effectiveness of our smoothness term applied on the features. As shown in Fig.~\ref{fig:ab_smooth}, our smooth loss could provide a effective regularisation and remove artefacts caused by hash collisions in unobserved regions that do not have any supervision.

\noindent\textbf{Effect of pose optimization in global BA.} We perform an additional experiment on performing our GBA without any pose optimization (GBA$^{\ddagger}$) in Tab.~\ref{tab:ablation_gba_supp}. We show that even the sample points in each sampled batch (2048 rays) may not have large overlapping, optimizing camera pose with such sampling strategy can still significantly improve the performance and the robustness of the pose estimation. 

\begin{table}[t]
    \centering
    \resizebox{1.00\columnwidth}{!}
    {
    \begin{tabular}{lcccc}
    \toprule
    Method & Tracking (ms) & Mapping (ms) & Memory &
    ATE$\downarrow$\\
    \midrule
    Two grids & 9.9$\times$20 & 25.4$\times$10 &  1.7M &8.69\\
    One grid & 7.8$\times$20& 20.2$\times$10& 0.8M & 8.75\\
    
    \bottomrule
    \end{tabular}
    }
    \vspace{-4pt}
    \caption{Performance analysis of modeling the geometry and appearance using one/two grids. The ATE is similar while using one grid require less computational cost. Run-time is reported in \texttt{ms/iter $\times$ \#iter} format.}
    \label{tab:twogrid}
    \vspace{-8pt}
\end{table}
\begin{table}[tpb]
    \centering
	\resizebox{0.99\columnwidth}{!}{
    \begin{tabular}{lcccccc}
        \toprule
        \textbf{Method}  & \textbf{Acc.} $\downarrow$ & \textbf{Comp.} $\downarrow$ & \textbf{C-$\ell_1$} $\downarrow$ & \textbf{NC} $\uparrow$  & \textbf{F-score} $\uparrow$ & \textbf{Run time}
        \\
        \midrule
        Neural RGBD   &  0.0151  & 0.0197  &	0.0174 & 0.9316 & \textbf{0.9635} & 10-25h \\
        GO-Surf & 0.0158   & 0.0195  &  0.0177 & \textbf{0.9317} & 0.9591 &15-45min \\
        Ours  & \textbf{0.0149}   & \textbf{0.0179}  &  \textbf{0.0164} & 0.9292 & 0.9629 &100-500s \\
        \bottomrule
    \end{tabular}
    }
    \vspace{-4pt}
    \caption[Quantitative comparison]{Quantitative results of the reconstruction on 10 synthetic scenes~\cite{azinovicNeuralRGBDSurface2022c}. The evaluation metrics and protocol follow Neural RGBD~\cite{azinovicNeuralRGBDSurface2022c} and GO-Surf~\cite{wangGosurfNeuralFeature2022}. We achieve on-par performance but our training is significantly faster.
    }
    \vspace{-3mm}
    \label{tab:batchmode}
\end{table}
\begin{table}[t]
\centering
\resizebox{1.0\columnwidth}{!}{
\begin{tabular}{l cc ccc c c c}
\toprule
\multirow{2}{*}{Name} & \multicolumn{2}{c}{KF selection} & \multicolumn{3}{c}{\#KF} & \multirow{2}{*}{\begin{tabular}[c]{@{}c@{}}Pose \\ optim.\end{tabular}} & \multirow{2}{*}{ATE (cm)} & \multirow{2}{*}{Std. (cm)} \\
\cmidrule(lr){2-3} \cmidrule(lr){4-6}
& Local & Global & 0 & 10 & All & & & \\
\midrule
w/o BA & & & \checkmark & & & & 16.81 & 1.69\\
LBA&\checkmark& & & \checkmark & & \checkmark & 9.69 & 1.38\\

GBA-10&& \checkmark& & \checkmark & & \checkmark & 9.54 & 0.67\\
GBA$^\ddagger$&& \checkmark& &  & \checkmark& & 9.72 & 0.53\\
GBA&& \checkmark& &  & \checkmark& \checkmark& \textbf{8.75} & \textbf{0.33}\\
\bottomrule
\end{tabular}}
\vspace{-5pt}
\caption{Ablation using different BA strategies on Co-SLAM: (LBA) BA with rays from 10 local keyframes; (GBA-10) BA with rays from 10 randomly selected keyframes; (GBA$^{\ddagger}$): BA with rays from all keyframes w/o pose optimization; (GBA) BA with rays from all keyframes and pose optimization (our full method). All methods sample a total of 2048 rays per iteration. }
\vspace{-10pt}
\label{tab:ablation_gba_supp}
\end{table}

\subsection{Batch-mode Optimisation}

To validate the representation ability of the proposed joint coordinate and parametric encoding, we also perform experiments of batch mode optimisation, which is an offline approach and the pose initialised by BundleFusion~\cite{daiBundlefusionRealtimeGlobally2017a} is given. Tab.~\ref{tab:batchmode} shows the quantitative results of different methods. Neural RGBD~\cite{azinovicNeuralRGBDSurface2022c} is a coordinate encoding-based method while GO-Surf~\cite{wangGosurfNeuralFeature2022} is a parametric encoding-based method. By using the proposed joint coordinate and parametric encoding, we achieve competitive reconstruction performance with significantly faster training speed.

\begin{figure*}[htbp]
  \centering
  \footnotesize
  \setlength{\tabcolsep}{1.5pt}
  \newcommand{\sz}{0.235}
  \begin{tabular}{cccc}
    iMAP$^\dagger$~\cite{sucarIMAPImplicitMapping2021a} & NICE-SLAM~\cite{zhuNiceslamNeuralImplicit2022} & Ours$^\dagger$ & Ground Truth \\
    \includegraphics[width=\sz\linewidth]{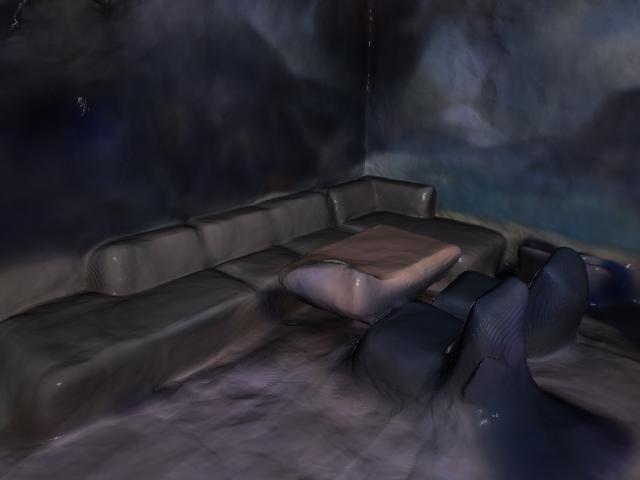} &
    \includegraphics[width=\sz\linewidth]{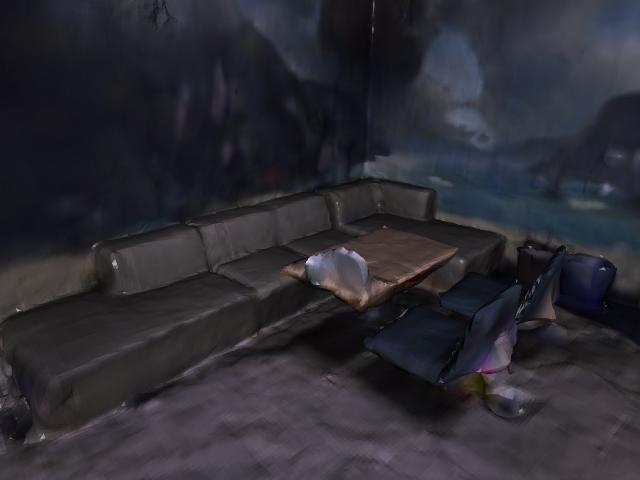} &
    \includegraphics[width=\sz\linewidth]{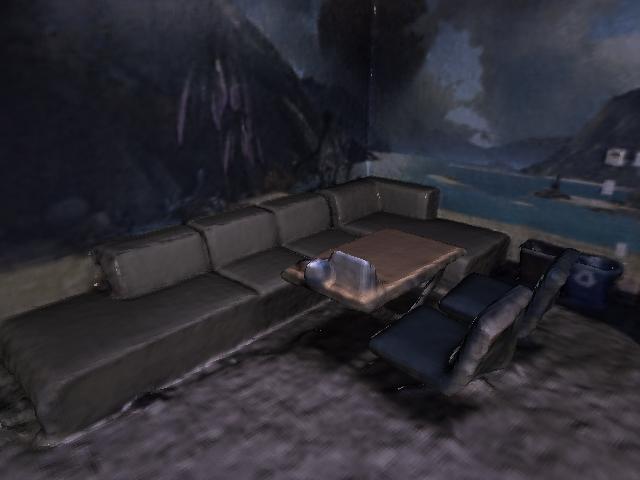} &
    \includegraphics[width=\sz\linewidth]{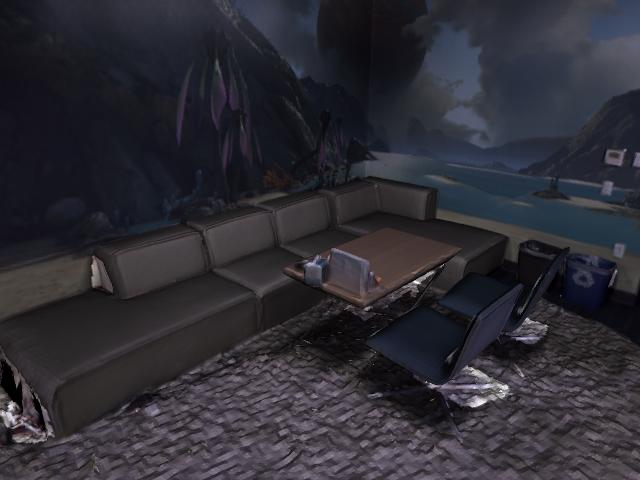} \\
    \includegraphics[width=\sz\linewidth]{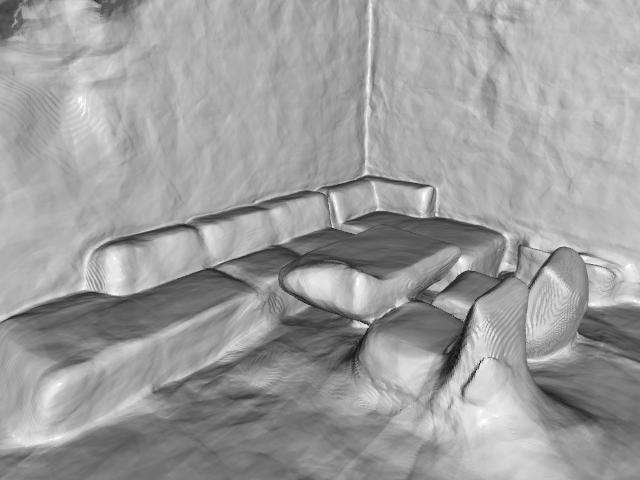} &
    \includegraphics[width=\sz\linewidth]{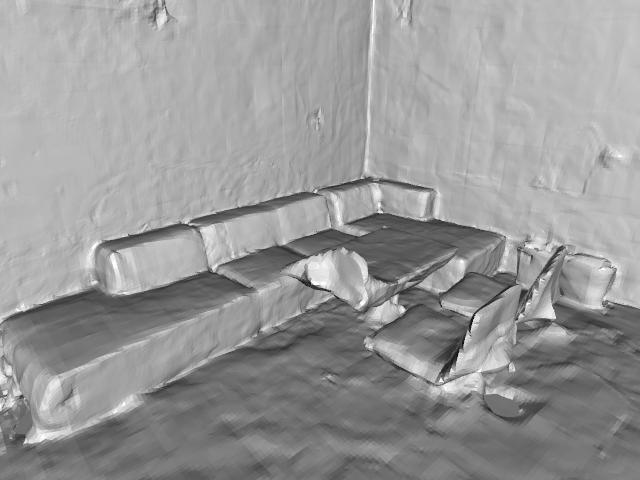} &
    \includegraphics[width=\sz\linewidth]{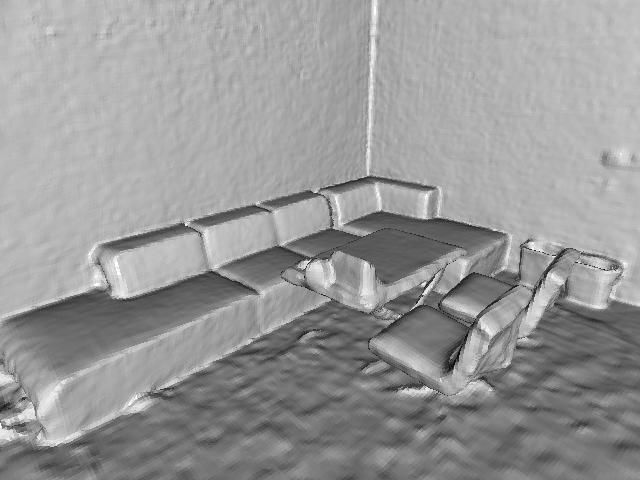} &
    \includegraphics[width=\sz\linewidth]{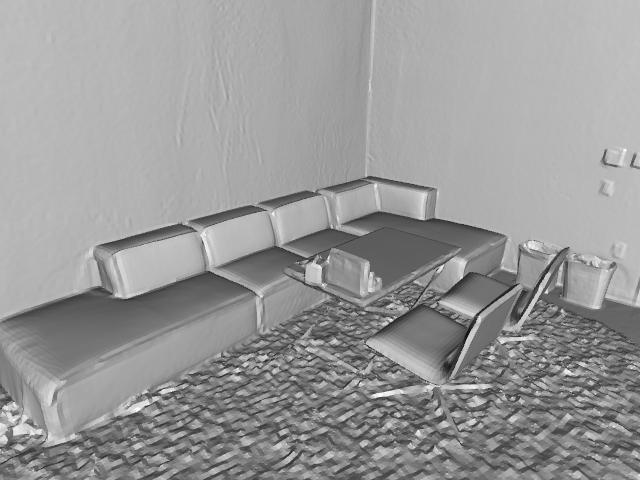} \\
    \includegraphics[width=\sz\linewidth]{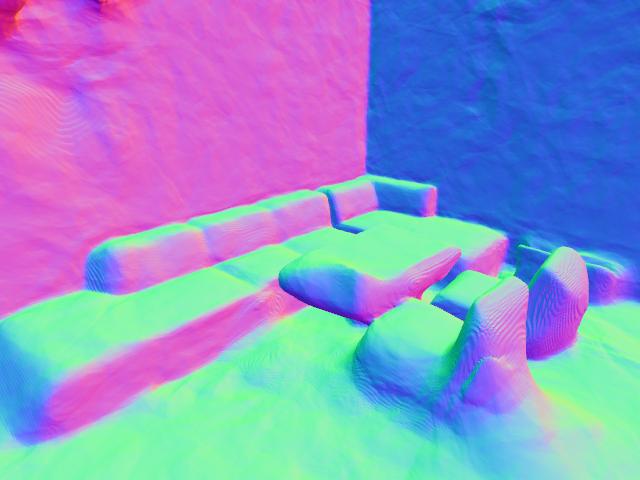} &
    \includegraphics[width=\sz\linewidth]{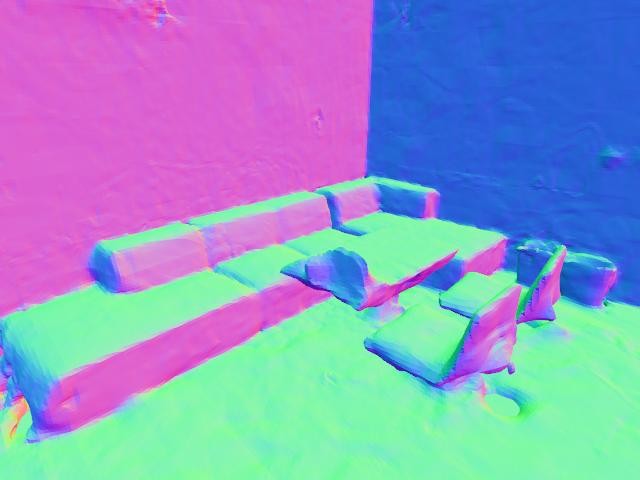} &
    \includegraphics[width=\sz\linewidth]{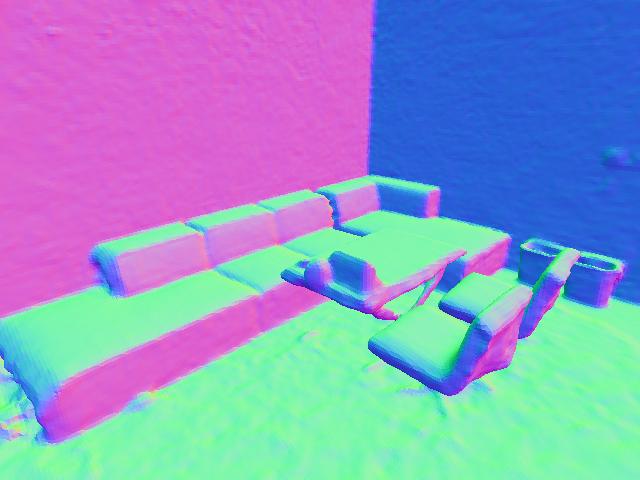} &
    \includegraphics[width=\sz\linewidth]{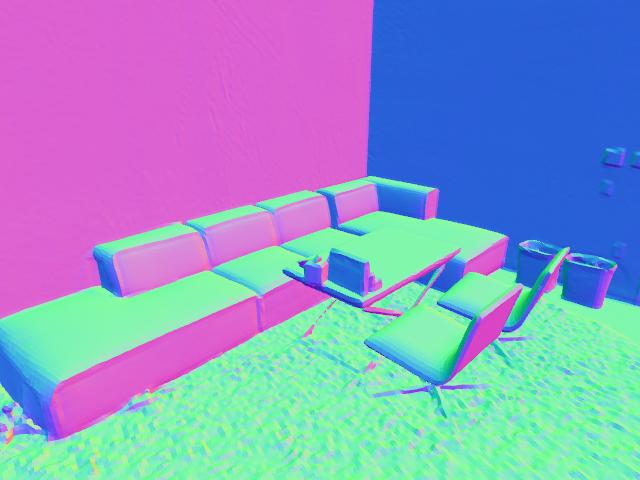} \\
  \end{tabular} 
  \vspace{-5pt}
  \caption{Qualitative comparison on Replica \texttt{office-0} with different shading mode. Our methods achieve accurate scene reconstruction with high frequence details. At the same time, our reconstruction is also sharper and smoother.   }
  \label{fig:Replica_office0}
\end{figure*}

\begin{figure*}[htbp]
  \centering
  \footnotesize
  \setlength{\tabcolsep}{1.5pt}
  \newcommand{\sz}{0.235}
  \begin{tabular}{cccc}
    iMAP$^\dagger$~\cite{sucarIMAPImplicitMapping2021a} & NICE-SLAM~\cite{zhuNiceslamNeuralImplicit2022} & Ours$^\dagger$ & Ground Truth \\
    \includegraphics[width=\sz\linewidth]{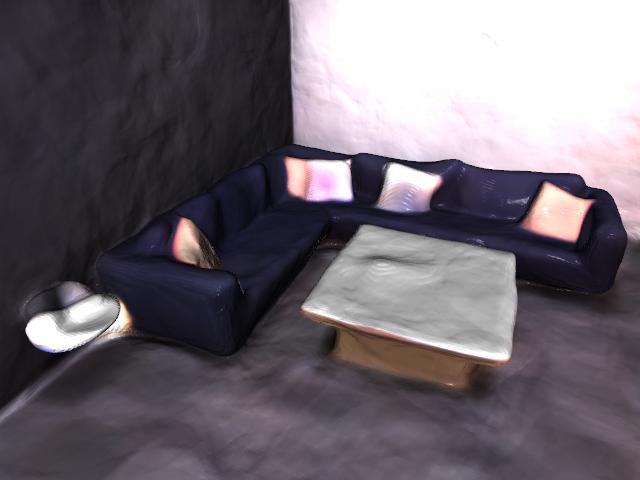} &
    \includegraphics[width=\sz\linewidth]{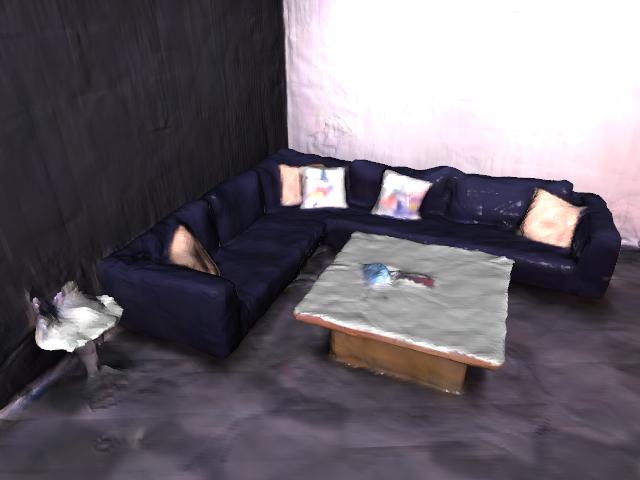} &
    \includegraphics[width=\sz\linewidth]{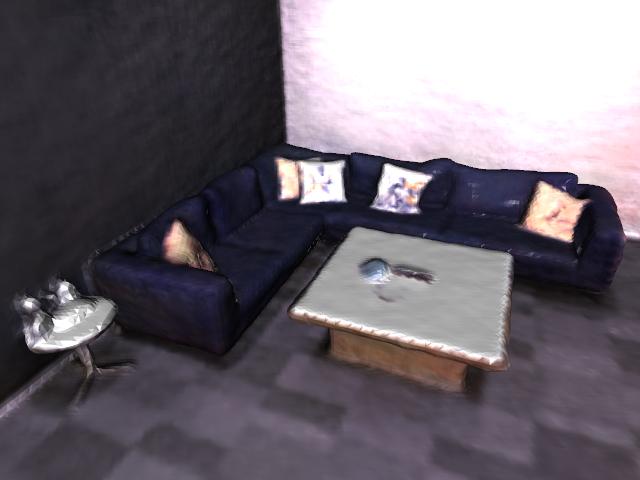} &
    \includegraphics[width=\sz\linewidth]{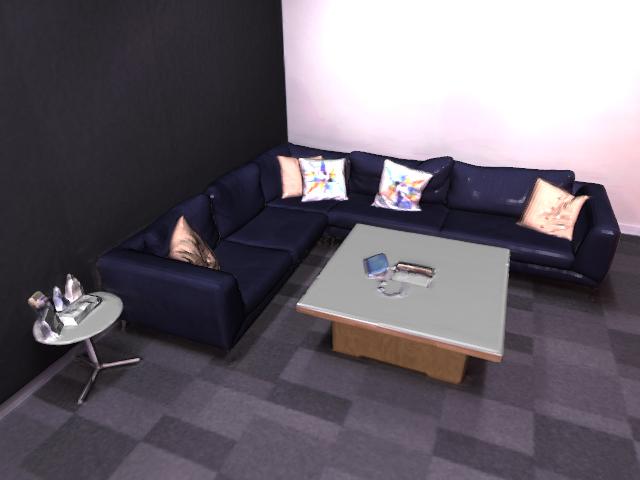} \\
    \includegraphics[width=\sz\linewidth]{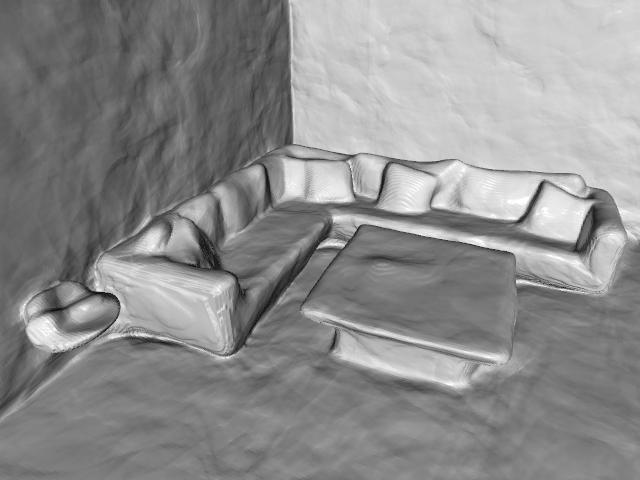} &
    \includegraphics[width=\sz\linewidth]{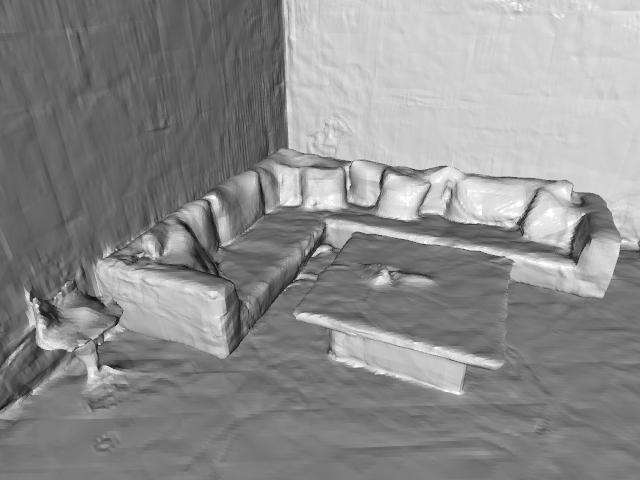} &
    \includegraphics[width=\sz\linewidth]{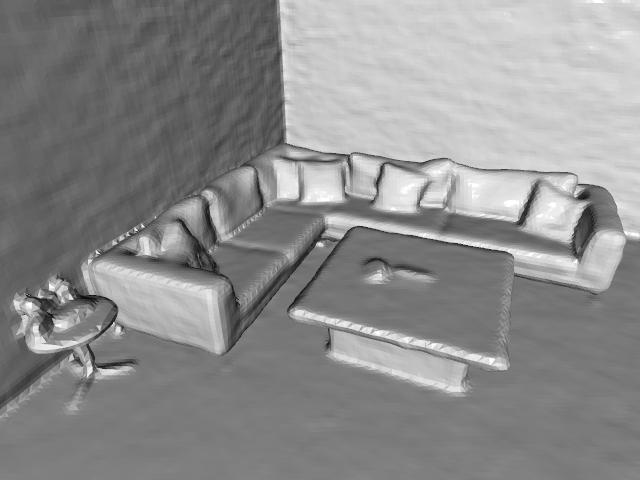} &
    \includegraphics[width=\sz\linewidth]{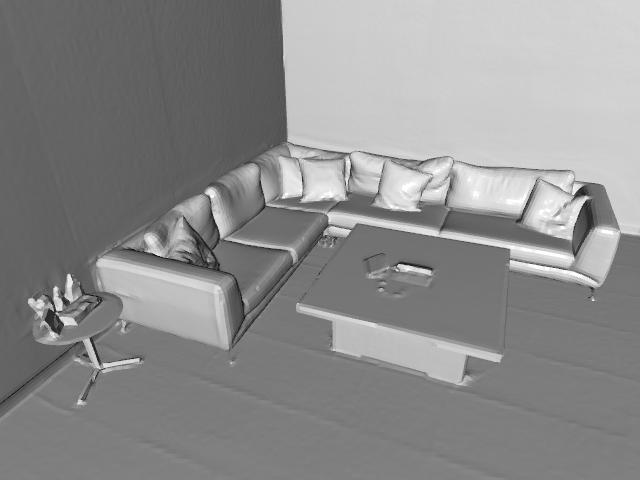} \\
    \includegraphics[width=\sz\linewidth]{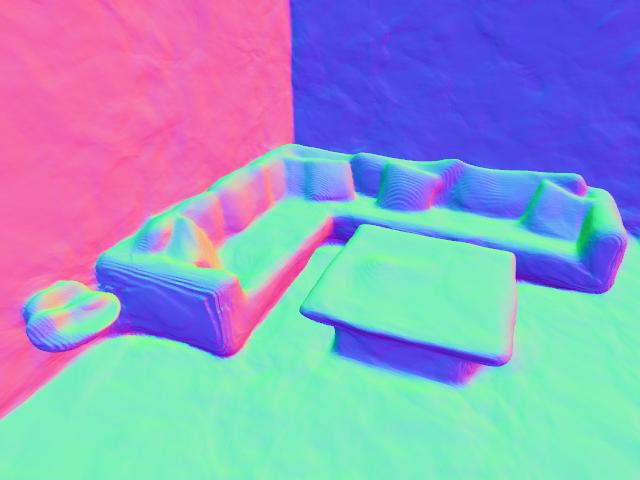} &
    \includegraphics[width=\sz\linewidth]{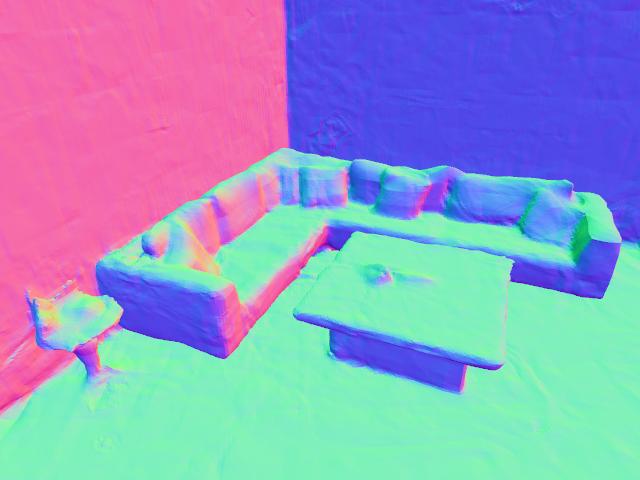} &
    \includegraphics[width=\sz\linewidth]{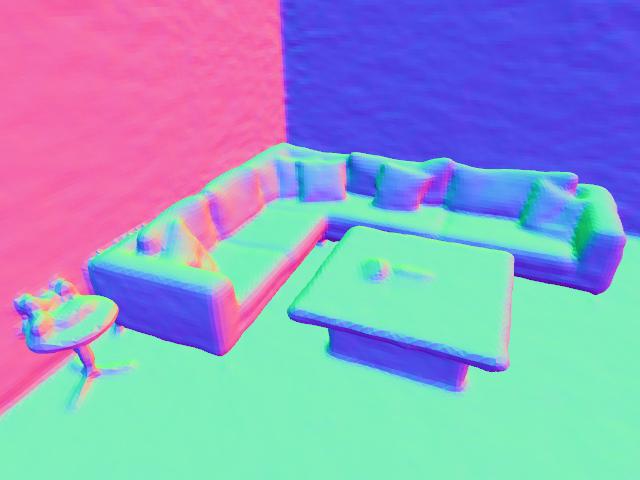} &
    \includegraphics[width=\sz\linewidth]{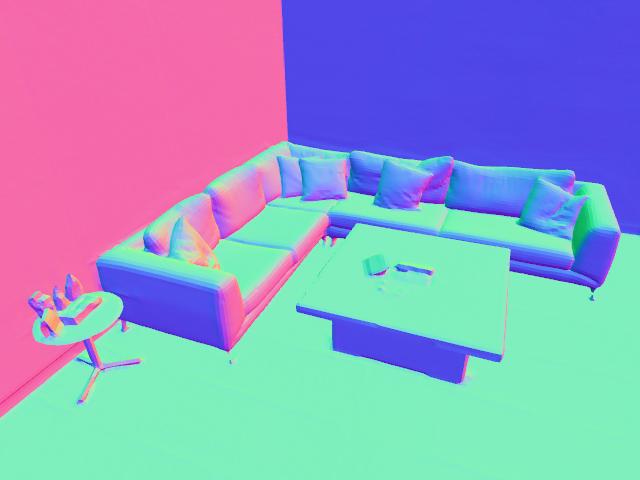} \\
  \end{tabular} 
  \caption{Qualitative comparison on Replica \texttt{office-2} with different shading mode. Note that regions with different color styles in the groundtruth color image indicate the unobserved region. Our method achieve better scene completion for unobserved regions.}
  \label{fig:Replica_office2}
\end{figure*}

\begin{figure*}[htbp]
  \centering
  \footnotesize
  \setlength{\tabcolsep}{1.5pt}
  \newcommand{\sz}{0.235}
  \begin{tabular}{cccc}
    iMAP$^\dagger$~\cite{sucarIMAPImplicitMapping2021a} & NICE-SLAM~\cite{zhuNiceslamNeuralImplicit2022} & Ours$^\dagger$ & Ground Truth \\
    \includegraphics[width=\sz\linewidth]{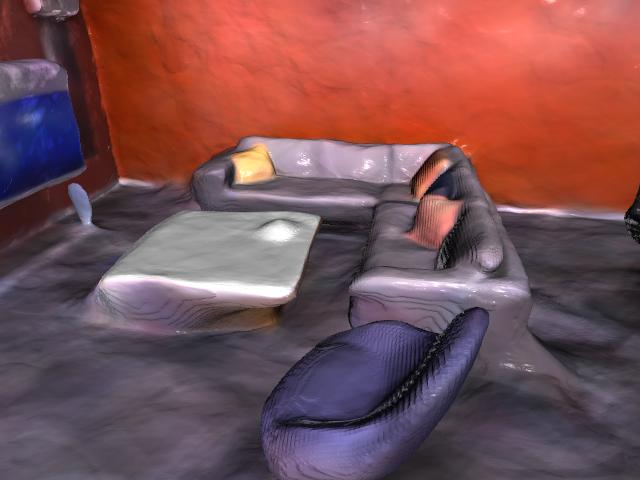} &
    \includegraphics[width=\sz\linewidth]{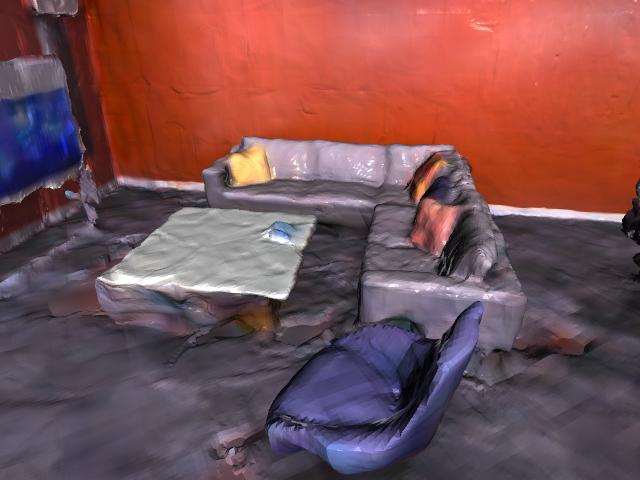} &
    \includegraphics[width=\sz\linewidth]{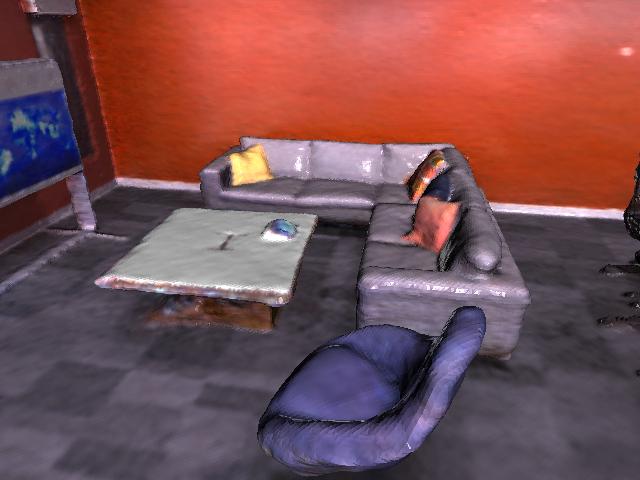} &
    \includegraphics[width=\sz\linewidth]{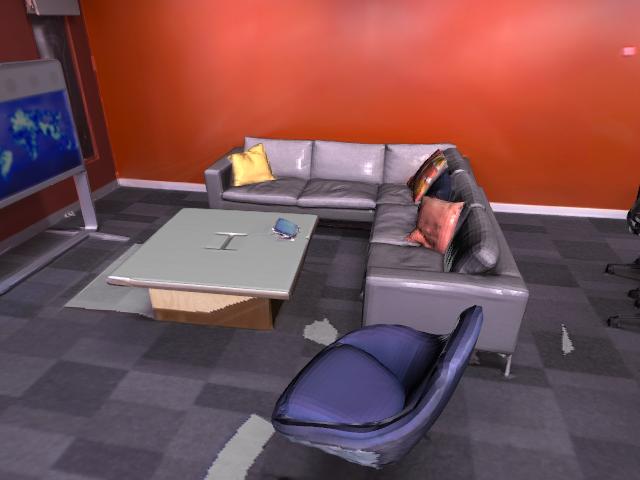} \\
    \includegraphics[width=\sz\linewidth]{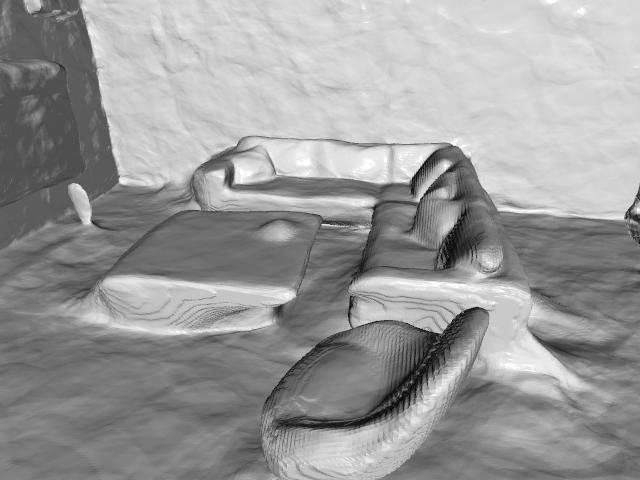} &
    \includegraphics[width=\sz\linewidth]{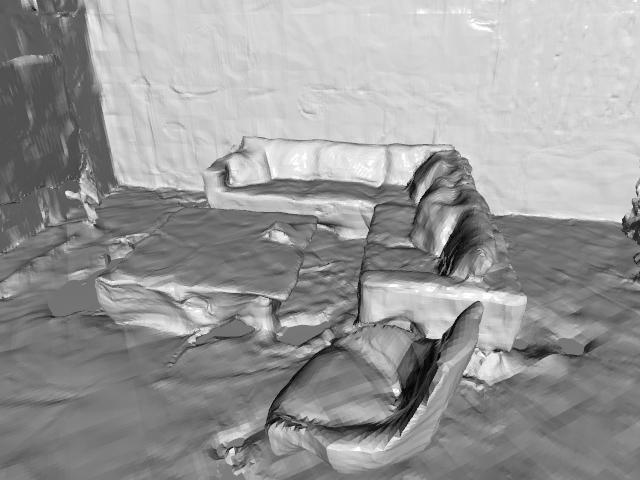} &
    \includegraphics[width=\sz\linewidth]{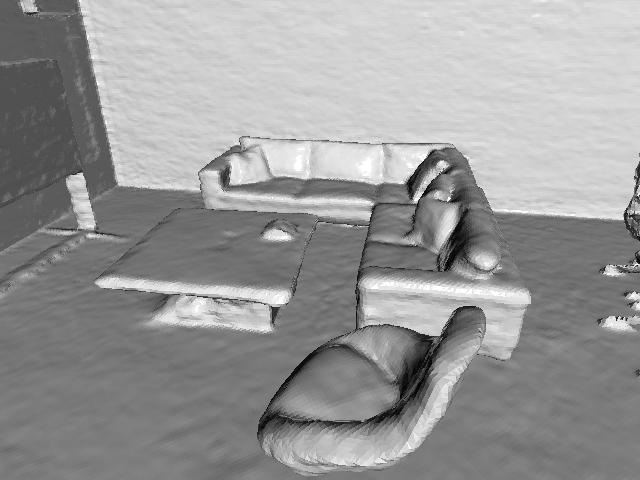} &
    \includegraphics[width=\sz\linewidth]{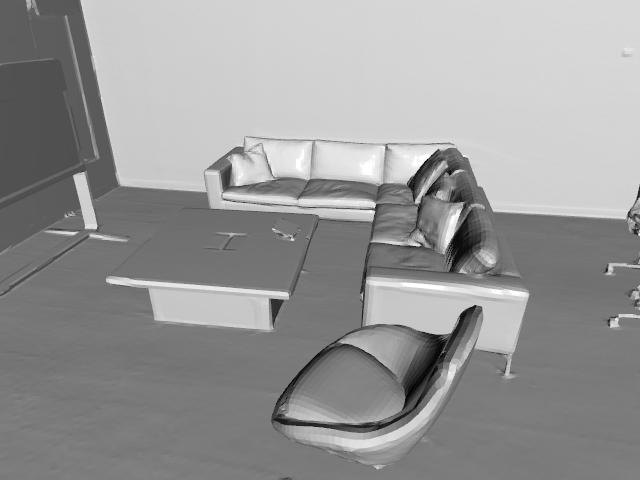} \\
    \includegraphics[width=\sz\linewidth]{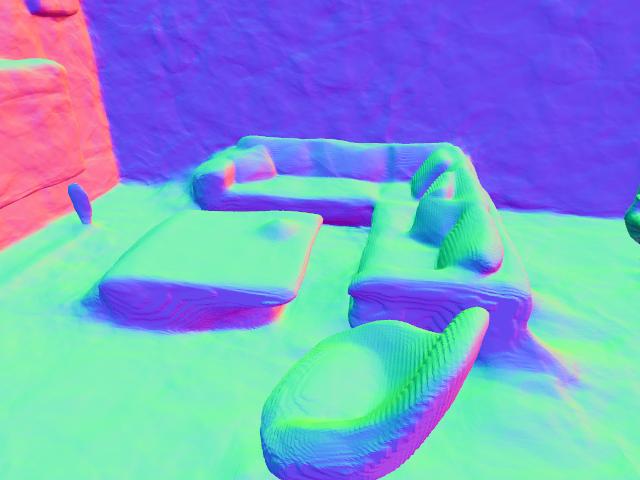} &
    \includegraphics[width=\sz\linewidth]{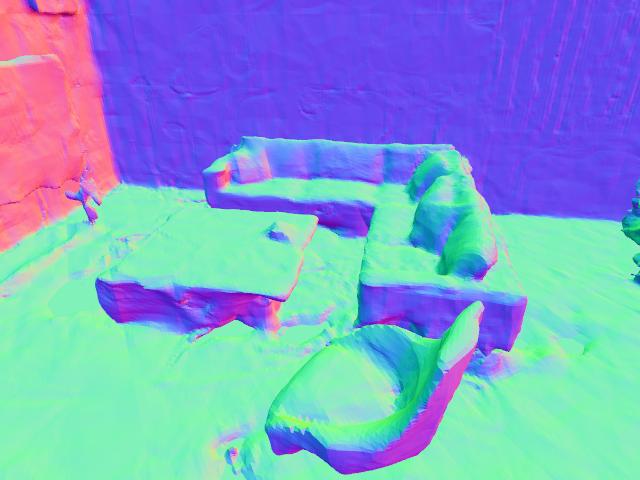} &
    \includegraphics[width=\sz\linewidth]{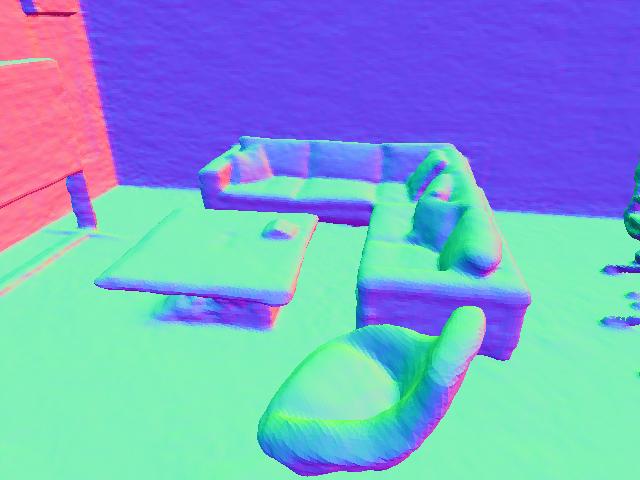} &
    \includegraphics[width=\sz\linewidth]{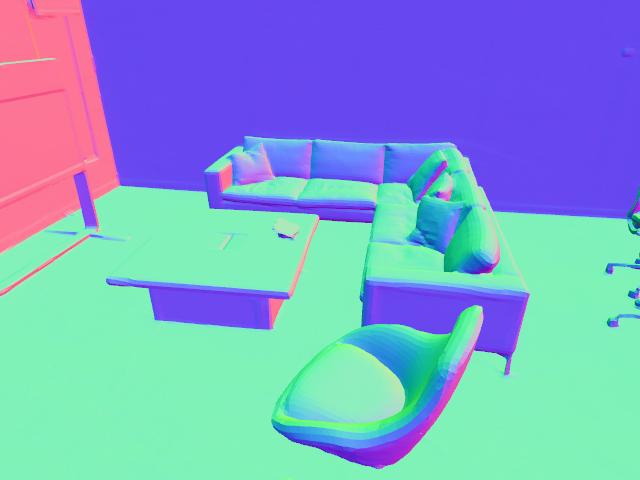} \\
  \end{tabular} 
  \vspace{-5pt}
  \caption{Qualitative comparison on Replica \texttt{office-3} with different shading mode. Our method achieves smooth reconstruction for regions that contain multiple objects while other methods contain some build-up effect.}
  \label{fig:Replica_office3}
\end{figure*}

\begin{figure*}[htbp]
  \centering
  \footnotesize
  \setlength{\tabcolsep}{1.5pt}
  \newcommand{\sz}{0.235}
  \begin{tabular}{cccc}
    iMAP$^\dagger$~\cite{sucarIMAPImplicitMapping2021a} & NICE-SLAM~\cite{zhuNiceslamNeuralImplicit2022} & Ours$^\dagger$ & Ground Truth \\
    \includegraphics[width=\sz\linewidth]{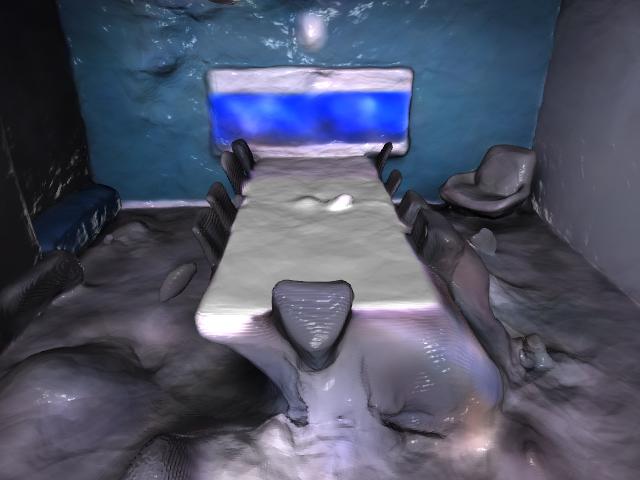} &
    \includegraphics[width=\sz\linewidth]{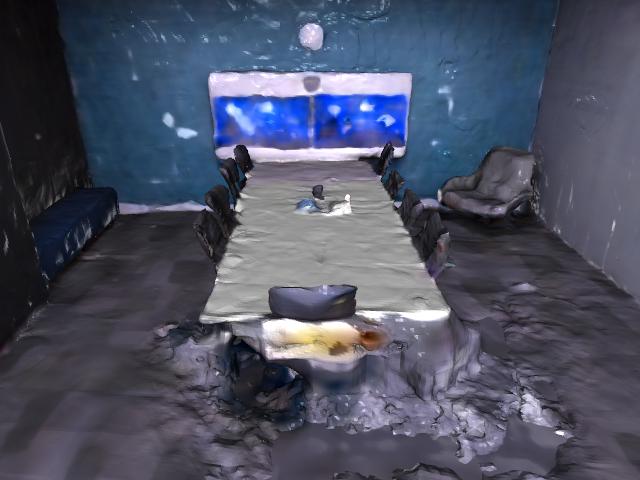} &
    \includegraphics[width=\sz\linewidth]{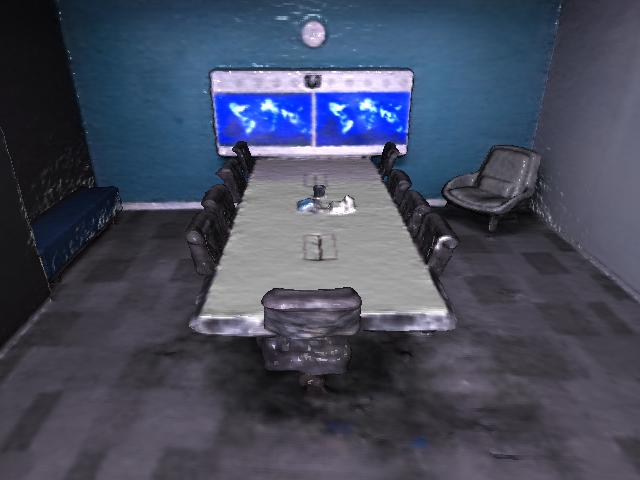} &
    \includegraphics[width=\sz\linewidth]{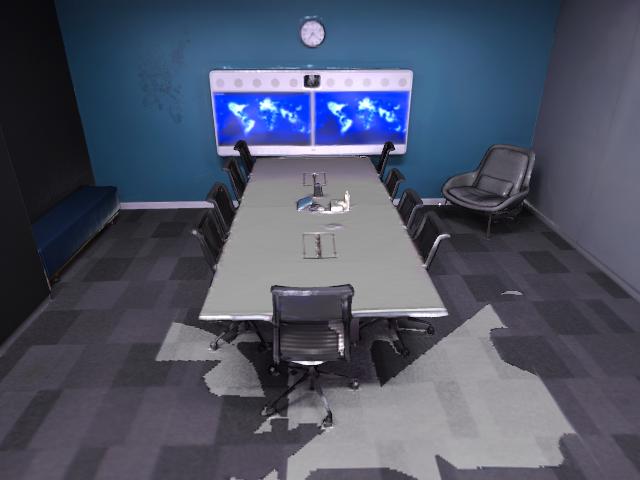} \\
    \includegraphics[width=\sz\linewidth]{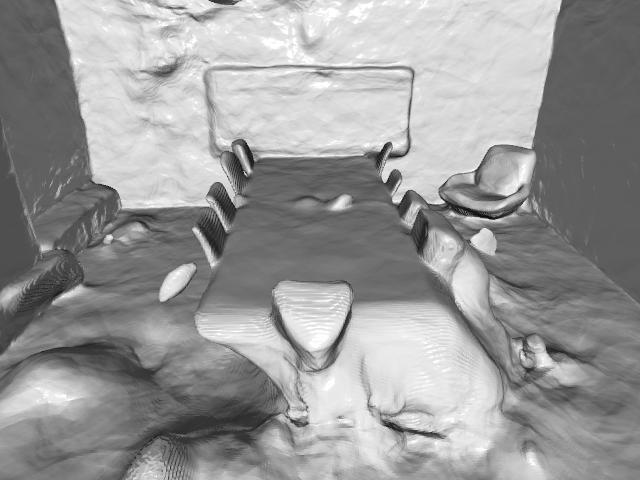} &
    \includegraphics[width=\sz\linewidth]{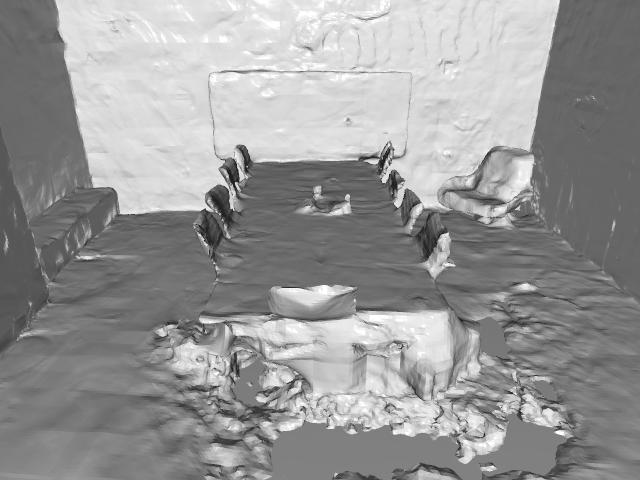} &
    \includegraphics[width=\sz\linewidth]{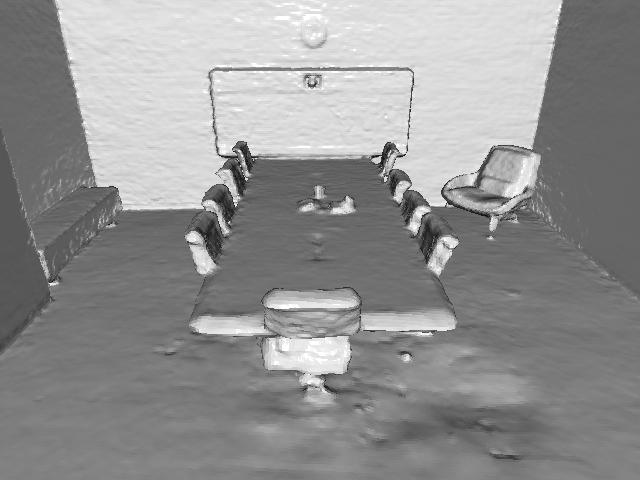} &
    \includegraphics[width=\sz\linewidth]{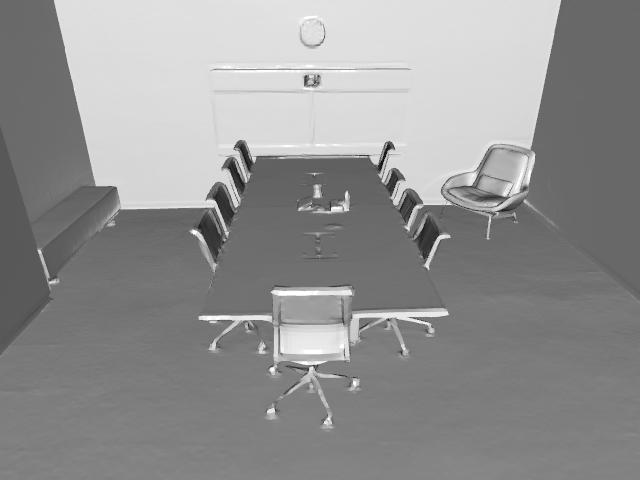} \\
    \includegraphics[width=\sz\linewidth]{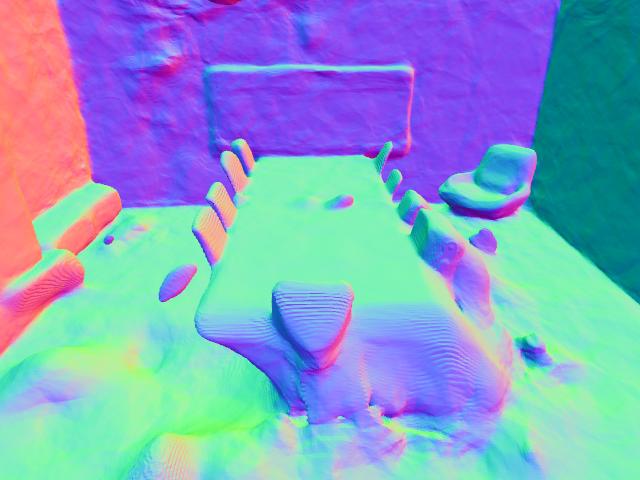} &
    \includegraphics[width=\sz\linewidth]{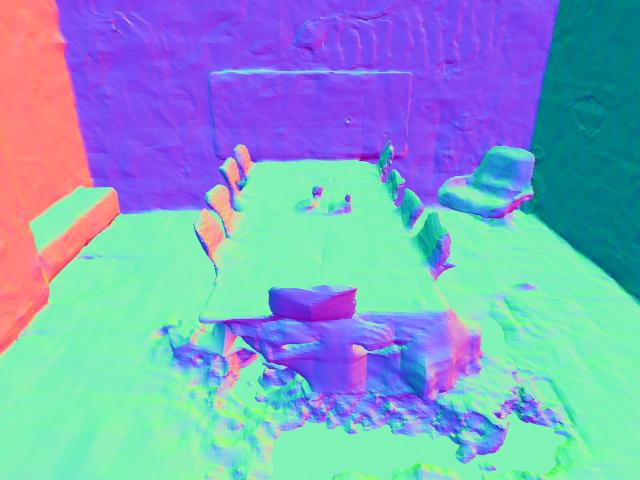} &
    \includegraphics[width=\sz\linewidth]{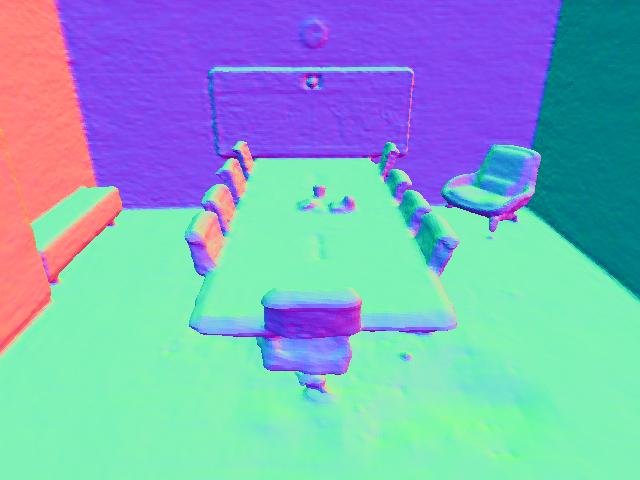} &
    \includegraphics[width=\sz\linewidth]{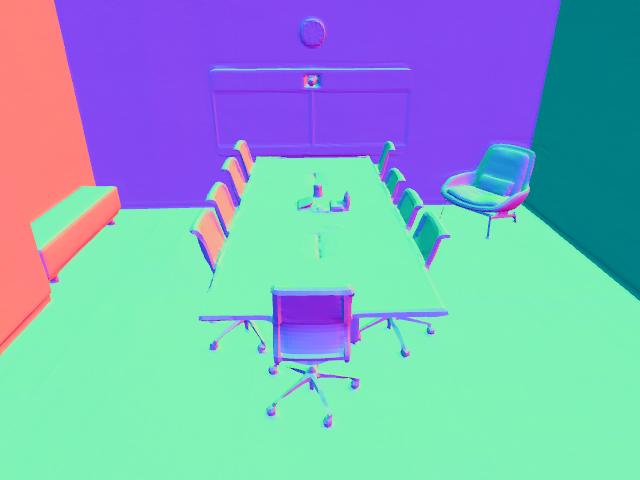} \\
  \end{tabular} 
  \vspace{-5pt}
  \caption{Qualitative comparison on Replica \texttt{office-4} with different shading mode. Note that regions with different color styles in the groundtruth color image indicate the unobserved region. Our method can accurately recover the thin structures while achieve smooth reconstruction around the flat regions that have not been observed.}
  \label{fig:Replica_office4}
\end{figure*}

{\small
\bibliographystyle{ieee_fullname}
\bibliography{egbib}
}